\documentclass{article}
\usepackage{microtype}
\usepackage{graphicx}
\usepackage{subfigure}
\usepackage{booktabs} 
\usepackage{multirow}
\usepackage{hyperref}
\usepackage{amsthm}

\usepackage{algorithmic}
\usepackage{bm}

\usepackage[accepted]{icml2024}

\usepackage{amsmath}
\usepackage{amssymb}
\usepackage{mathtools}
\usepackage{amsthm}
\usepackage{tikz}
\usepackage{afterpage}
\usepackage{bm}

\def\va{{\bm{a}}}
\def\vx{{\bm{x}}}

\def\vbeta{{\bm{\beta}}}
\def\vmu{{\bm{\mu}}}

\def\vtheta{{\bm{\theta}}}

\DeclareMathAlphabet{\mathsfit}{\encodingdefault}{\sfdefault}{m}{sl}
\SetMathAlphabet{\mathsfit}{bold}{\encodingdefault}{\sfdefault}{bx}{n}

\def\vdelta{\bm{\delta}}

\def\gF{\mathcal{F}}
\def\tgF{\tilde{\gF}}
\def\vb{\bm{b}}
\def\tvb{\tilde{\vb}}

\def\tva{\tilde{\va}}
\def\gN{\mathcal{N}}
\def\tmu{\tilde{\mu}}
\def\tsigma{\tilde{\sigma}}
\def\E{\mathbb{E}}
\def\argmax{\mathrm{argmax}}
\def\Std{\mathrm{Std}}
\def\gY{\mathcal{Y}}
\def\gX{\mathcal{X}}
\def\vdelta{\bm{t}}

\usepackage[capitalize,noabbrev]{cleveref}

\theoremstyle{plain}
\newtheorem{theorem}{Theorem}[section]

\theoremstyle{definition}

\theoremstyle{remark}
\newtheorem{remark}[theorem]{Remark}

\usepackage[textsize=tiny]{todonotes}
\usepackage{makecell}
 \graphicspath{{./}{figs/}}
 \newcommand{\etal}{\textit{et~al.}}
 
\newcommand{\ie}{\textit{i.e.}}
\newcommand{\eg}{\textit{e.g.}}
\usepackage{color}
\def\red#1{\textcolor{red}{#1}}
\def\tred#1{\textcolor{red}{#1}}

\icmltitlerunning{IBD-PSC: Input-level Backdoor Detection via Parameter-oriented Scaling Consistency}

\begin{document}

\twocolumn[
\icmltitle{IBD-PSC: Input-level Backdoor Detection via Parameter-oriented\\Scaling Consistency}



\icmlsetsymbol{equal}{*}

\begin{icmlauthorlist}
\icmlauthor{Linshan Hou}{yyy}
\icmlauthor{Ruili Feng}{yyy1,yyy11}
\icmlauthor{Zhongyun Hua}{yyy}
\icmlauthor{Wei Luo}{yyy2}
\icmlauthor{Leo Yu Zhang}{yyy3}
\icmlauthor{Yiming Li}{sch}
\end{icmlauthorlist}

\icmlaffiliation{yyy}{School of Computer Science and Technology, Harbin Institute of Technology, Shenzhen, China}
\icmlaffiliation{yyy1}{Alibaba Group, China}
\icmlaffiliation{yyy11}{University of Science and Technology of China, China}
\icmlaffiliation{yyy2}{School of Information Technology, Deakin University, Australia}
\icmlaffiliation{yyy3}{School of Information and Communication Technology, Griffith University, Australia}
\icmlaffiliation{sch}{Nanyang Technological University, Singapore}

\icmlcorrespondingauthor{Zhongyun Hua}{huazhongyun@hit.edu.cn}
\icmlcorrespondingauthor{Yiming Li}{liyiming.tech@gmail.com}

\icmlkeywords{Machine Learning, ICML}
\vskip 0.3in
]



\printAffiliationsAndNotice{}  

\begin{abstract}
Deep neural networks (DNNs) are vulnerable to backdoor attacks, where adversaries can maliciously trigger model misclassifications by implanting a hidden backdoor during model training. This paper proposes a simple yet effective input-level backdoor detection (dubbed IBD-PSC) as a `firewall' to filter out malicious testing images. Our method is motivated by an intriguing phenomenon, \ie, parameter-oriented scaling consistency (PSC), where the prediction confidences of poisoned samples are significantly more consistent than those of benign ones when amplifying model parameters. In particular, we provide theoretical analysis to safeguard the foundations of the PSC phenomenon. We also design an adaptive method to select BN layers to scale up for effective detection. Extensive experiments are conducted on benchmark datasets, verifying the effectiveness and efficiency of our IBD-PSC method and its resistance to adaptive attacks. Codes are available at \href{https://github.com/THUYimingLi/BackdoorBox}{BackdoorBox}.
\end{abstract}
\section{Introduction}
Backdoor attacks are an emerging training-phase threat to deep neural networks (DNNs)~\cite{li2022backdoor}. A backdoored model behaves normally on benign samples while misclassifying malicious samples containing adversary-specified patterns (\ie, triggers). This attack could happen whenever the training stage is not fully controlled. It poses a significant threat to the lifecycle and supply chain of DNNs.


Currently, there are five representative defense strategies to alleviate backdoor threats, including \textbf{(1)} data purification~\cite{tran2018spectral,li2021anti,jebreel2023defending}, \textbf{(2)} poison suppression~\cite{wang2022training,huang2022backdoor,tang2023setting}, \textbf{(3)} model-level backdoor detection~\cite{wang2019neural,xiang2023umd,wang2024mm}, \textbf{(4)} model-level backdoor mitigation~\cite{liu2018fine,zeng2022adversarial,guo2023policycleanse}, and \textbf{(5)} input-level backdoor detection (IBD)~\cite{gao2021design,liu2023detecting,guo2023scale}. In general, the first four strategies typically demand substantial computational resources since they usually require model training. However, these resources are unavailable for many researchers and developers, especially those using third-party models. In contrast, the last one is less resource-intensive and is, therefore, our main focus. It aims to detect and prevent malicious inputs and can serve as the firewall of deployed models.

 \begin{figure}[!t]
    \centering
    \includegraphics[width=0.473\textwidth]{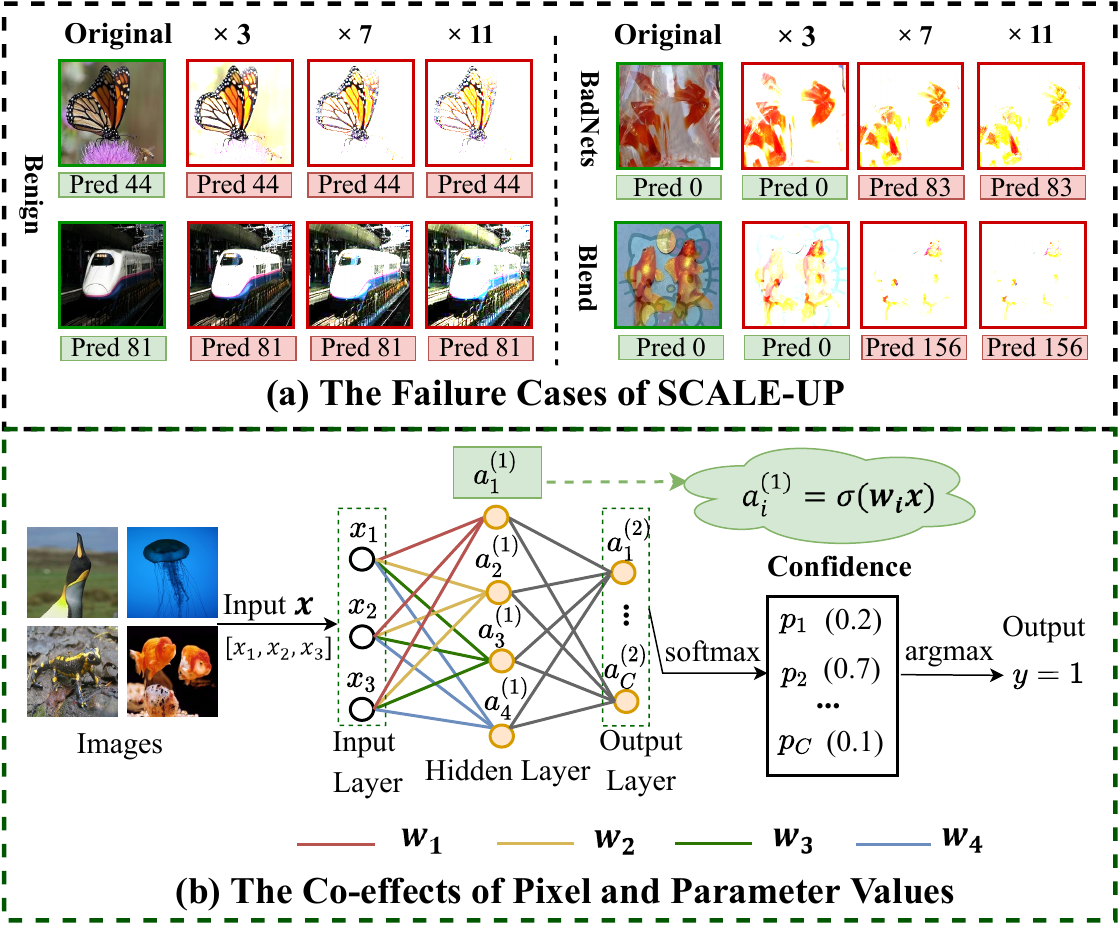} 
    \vspace{-0.5em}
    \caption{The limitation of SCALE-UP and the co-effects of pixel and parameter values. 
    \textbf{(a)} Failures in SCALE-UP due to bounded pixel value ($i.e.$, [0, 255]). Specifically, benign samples with black and white pixels are immune to amplification, preserving scaled prediction stability. Multiplying larger pixel values can easily turn them white, making the trigger disappear and become useless. \textbf{(b)} The prediction is the co-effects of the image and model parameters.}
    \vspace{-1em}
    \label{fig:intro}
\end{figure}

To the best of our knowledge, SCALE-UP~\cite{guo2023scale}  currently stands as the most advanced IBD. It observes that the predictions of poisoned samples (\ie, those containing triggers) exhibit more robustness to pixel-level amplification compared with those of benign samples and provides the theoretical foundations for this phenomenon. Employing this intriguing phenomenon, SCALE-UP directly enlarges all pixel values of the suspicious input sample with varying amplification intensities and assesses its prediction consistency for detection. However, SCALE-UP encounters some intrinsic limitations due to the restriction of pixel values (\ie, bounded in [0, 255]). For example, as shown in~\cref{fig:intro}(a), benign samples containing black and white pixels maintain their initial predictions during the amplification process. This stability is due to their extreme pixel values (0 or 255), which remain unaffected against amplification. Conversely, in poisoned samples, amplification often turns higher pixel values to the maximum (\ie, 255). It leads to large blank areas in the scaled poisoned images, masking the triggers and thus leading to changes in their predictions. Recognizing that prediction results are from the co-effects of pixel and parameter values, as shown in~\cref{fig:intro}(b), while parameter values are not bounded, an intriguing question arises: 


\emph{Shall the model's parameters expose backdoors with more grace than the humble pixel's tale?}


Fortunately, the answer is yes! In this paper, we reveal that the prediction confidences of poisoned samples have \emph{parameter-oriented scaling consistency (PSC)}. Specifically, we scale up the learned parameters of the batch normalization (BN) layers, which are widely exploited in advanced DNN structures. We demonstrate that the prediction confidences of poisoned samples are significantly more consistent than those of benign ones when the number of amplified BN layers increases. In particular, we show that this intriguing phenomenon is not accidental, where we prove that we can always find a scaling factor for BN parameters to expose latent backdoors for all attacked models (under some classical assumptions in learning theory). The scaled model can misclassify benign samples while maintaining the predictions of poisoned samples, leading to the PSC phenomenon.

Motivated by this finding, we propose a simple yet effective IBD method to identify and filter malicious testing samples, dubbed IBD-PSC. Specifically, for each suspicious testing image, our IBD-PSC measures its PSC value. This PSC value is defined as the average confidence generated over a range of parameter-scaled versions of the original model on the label, which is predicted by the original model. The larger the PSC value, the more likely the suspicious sample is poisoned. In particular, we start from the last layer of the deployed model and scale up different numbers of BN layers to obtain the scaled models. It is motivated by the previous findings~\cite{huang2022backdoor,jebreel2023defending} that trigger patterns often manifest as complicated features learned by the deeper layers of models, especially for those attacks with elaborate designs \cite{huang2022backdoor,jebreel2023defending}. To effectively determine the optimal number of layers for amplification, we design an adaptive algorithm by evaluating the scaling impact on the model's performance when processing benign samples.

In conclusion, our main contributions are four-fold. \textbf{(1)} We disclose an intriguing phenomenon, $i.e.$, parameter-oriented scaling consistency (PSC), where the prediction confidences of poisoned samples are more consistent than benign ones when scaling up BN parameters. \textbf{(2)} We provide theoretical insights to elucidate the PSC phenomenon. \textbf{(3)} We design a simple yet effective method (\ie, IBD-PSC) to filter out poisoned testing images based on our findings. \textbf{(4)} We conduct extensive experiments on benchmark datasets, verifying the effectiveness of our method against 13 representative attacks and its resistance to potential adaptive attacks.

\section{Related Work}
\subsection{Backdoor Attacks}

In general, existing backdoor attacks can be categorized into three types based on the adversaries' capabilities: \textbf{(1)} poison-only attacks, \textbf{(2)} training-controlled attacks, and \textbf{(3)} model-controlled attacks. These attacks could happen whenever the training stage is not fully controlled.

\noindent\textbf{Poison-only Backdoor Attacks.} In these attacks, the adversaries can only manipulate the training dataset. Gu~\etal~\cite{gu2017badnets} proposed the first poison-only attack (\ie, BadNets). BadNets poisoned a few training samples by patching a predefined trigger, \eg, a $3\times 3$ white square, onto the bottom right corner of these samples. It then altered the labels of the modified samples to an adversaries-specified target label. Models trained on such poisoned training sets create a relation between the trigger and the target label. Subsequent studies further developed more stealthy attack methods, including invisible and clean-label attacks. The former methods~\cite{chen2017targeted,li2021backdoor} typically used imperceptible triggers to bypass manual detection, while the latter ones ~\cite{turner2019label,zeng2023narcissus,gao2023not} maintained the ground-truth label of poisoned samples. Besides, there are also the physical attack~\cite{wenger2021backdoor,gong2023kaleidoscope,xu2023batt} that adopt physical objects or spatial transformations as triggers and the adaptive attack methods~\cite{tang2021demon,qi2023revisiting} that are specifically designed to evade defenses.

\noindent\textbf{Training-controlled Backdoor Attacks.} In these attacks, adversaries can modify both the training dataset and the training process. One line of work aimed to circumvent existing defenses and human detection. For instance, the adversaries may introduce a `noise mode'~\cite{nguyen2020input,nguyen2021wanet,mo2023robust,zhang2024detector} or incorporate well-designed regularization terms into training loss~\cite{li2020invisible,doan2021backdoor,xia2022enhancing}. Another line of work focused on augmenting the effectiveness of attacks. For instance, Wang~\etal~\cite{Wangbpp} exploited learning algorithms beyond supervised learning to ensure the correct injections of subtle triggers. Besides, Li~\etal~\cite{li2021backdoor} and Zhang~\etal~\cite{zhang2022poison} introduced spatial transformations to poisoned samples to hide the triggers more robustly, extending the threat of backdoor attacks to the real physical scenarios.

\noindent\textbf{Model-controlled Backdoor Attacks.} In model-controlled backdoor attacks, adversaries modify model architectures or parameters directly to inject backdoors. For example, Tang~\etal~\cite{tang2020embarrassingly} implanted hidden backdoors by inserting an additional malicious module into the benign victim model. Qi~\etal~\cite{qi2022towards} proposed to maliciously modify the parameters of a narrow subnet in the benign model instead of inserting an additional module. This approach was more stealthy and was highly effective in both digital and physical scenarios.

Recently, a few works exploit backdoor attacks for positive purposes~\citep{li2022untargeted,li2022defending1,li2023black,guo2023domain,tang2023setting,ya2024towards}, which are out of our scope.

\subsection{Backdoor Defenses}

Based on the stage of the model lifecycle where defense occurs, existing defenses can be mainly divided into five main categories: \textbf{(1)} data purification~\cite{tran2018spectral,li2021anti,jebreel2023defending}, \textbf{(2)} poison suppression~\cite{wang2022training,huang2022backdoor,tang2023setting}, \textbf{(3)} model-level backdoor detection~\cite{wang2019neural,xiang2023umd,wang2024mm,yao2024reverse,wang2022unicorn,wang2022rethinking}, \textbf{(4)} model-level backdoor mitigation~\cite{liu2018fine,zeng2022adversarial,guo2023policycleanse,li2024purifying,li2024nearest,xu2024towards}, and \textbf{(5)} input-level backdoor detection (IBD)~\cite{gao2021design,liu2023detecting,guo2023scale}. Specifically, data purification intends to filter out all poisoned samples in a given (third-party) dataset. It usually needs to train a model before identifying the influence of each training sample; Poison suppression aims to hinder the model's learning of the poisoned samples by modifying its training process to prevent backdoor creation; Model-level detection usually trains a meta-classifier or approximates trigger generation to determine whether a suspicious model contains hidden backdoors; IBD detects and prevents malicious inputs and acts as a `firewall' of deployed models. In general, the first four strategies demand substantial computational resources since they typically necessitate model training or fine-tuning. However, these resources are unavailable for many researchers and developers, especially those using third-party models. This paper primarily focuses on IBD, which is more computation-friendly.

Previous IBD methods~\cite{chou2018sentinet,gao2021design,liu2023detecting} are effective under implicit assumptions concerning the backdoor triggers. For example, STRIP~\cite{gao2021design} posited that trigger features play a dominant role, and the predictions of poisoned samples will not be affected even when benign features are overlaid. These assumptions can be easily circumvented by adaptive backdoor attacks~\cite{nguyen2020input,li2021invisible,duan2024conditional}. To the best of our knowledge, the most advanced IBD method is SCALE-UP~\cite{guo2023scale}. It amplified all pixel values of an input sample with varying intensities and treated it as poisoned if the predictions were consistent. However, SCALE-UP inherited some potential limitations due to pixel value constraints (bounded in [0, 255]). For example, these constraints may alter predictions of poisoned samples, as amplification can transform higher pixel values into the maximum value of 255, causing triggers (\eg, a white square) to disappear. How to design effective yet efficient IBD methods is still a critical open question. 

\begin{figure*}[!t]
\vspace{-1em}
\centering
\subfigure[Benign\label{fig:benign_conf}]{\includegraphics[width=0.245\textwidth]{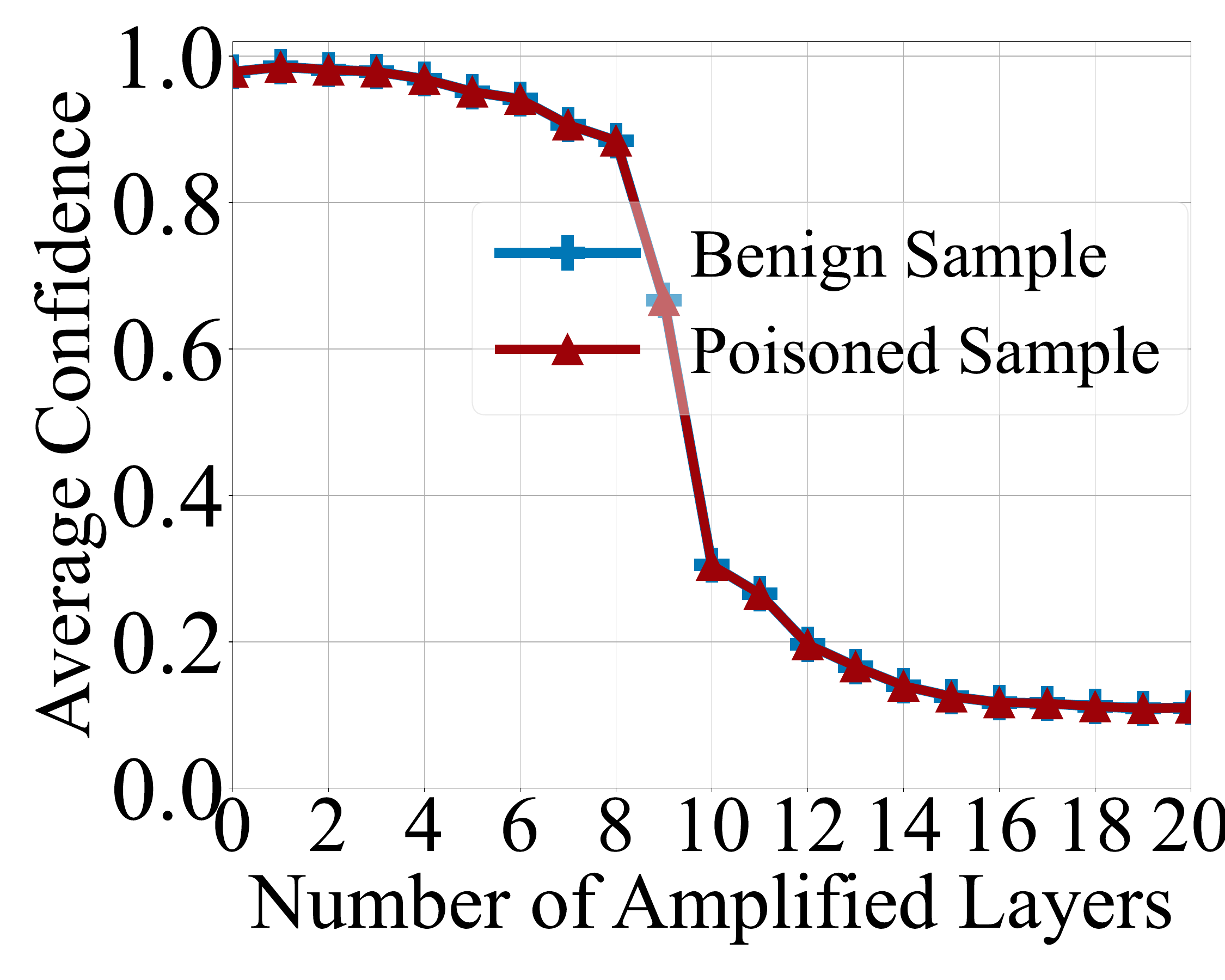}}
\subfigure[BadNets\label{fig:badnets_conf}]{\includegraphics[width=0.245\textwidth]{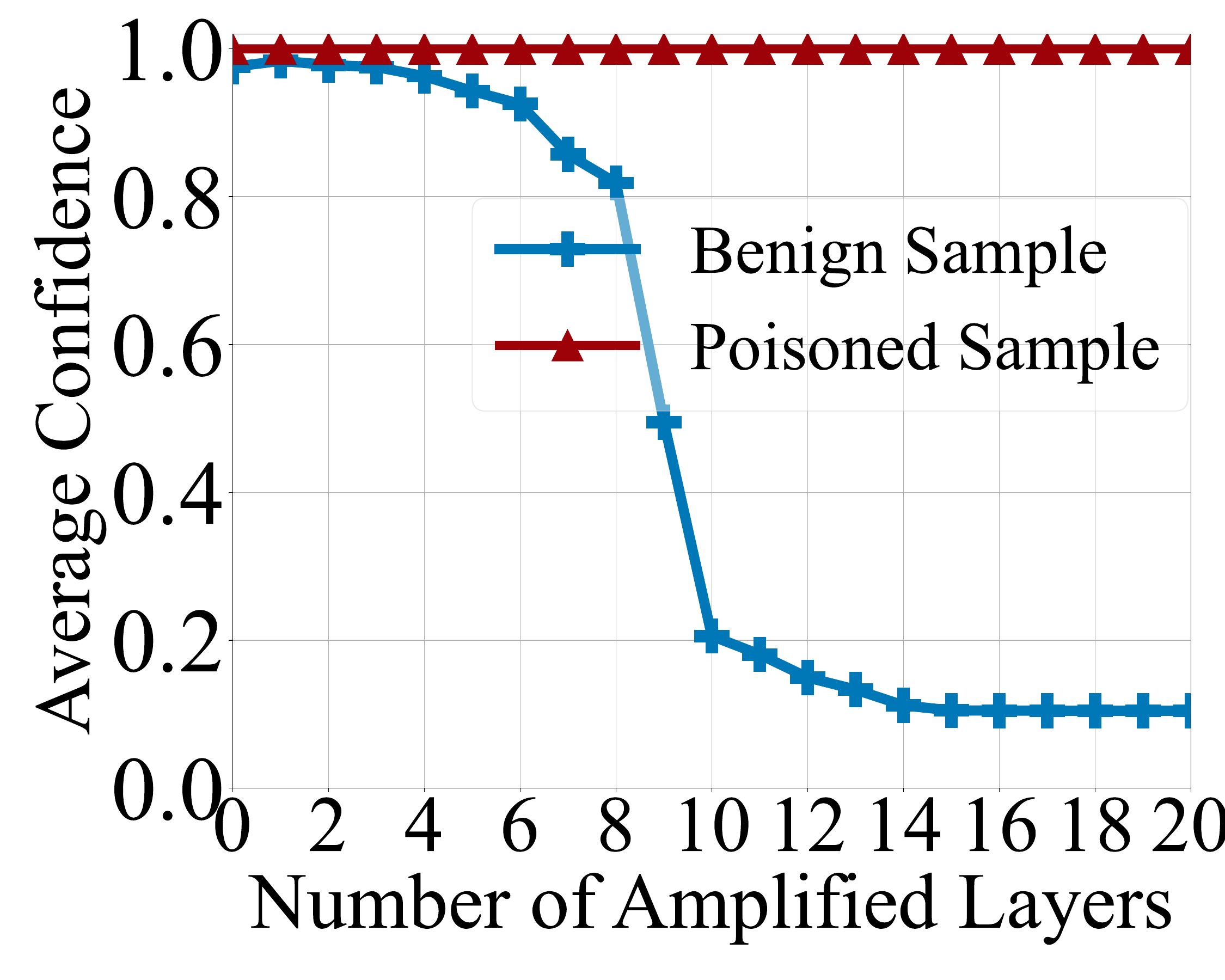}}
\subfigure[WaNet\label{fig:wanet_conf}]{\includegraphics[width=0.245\textwidth]{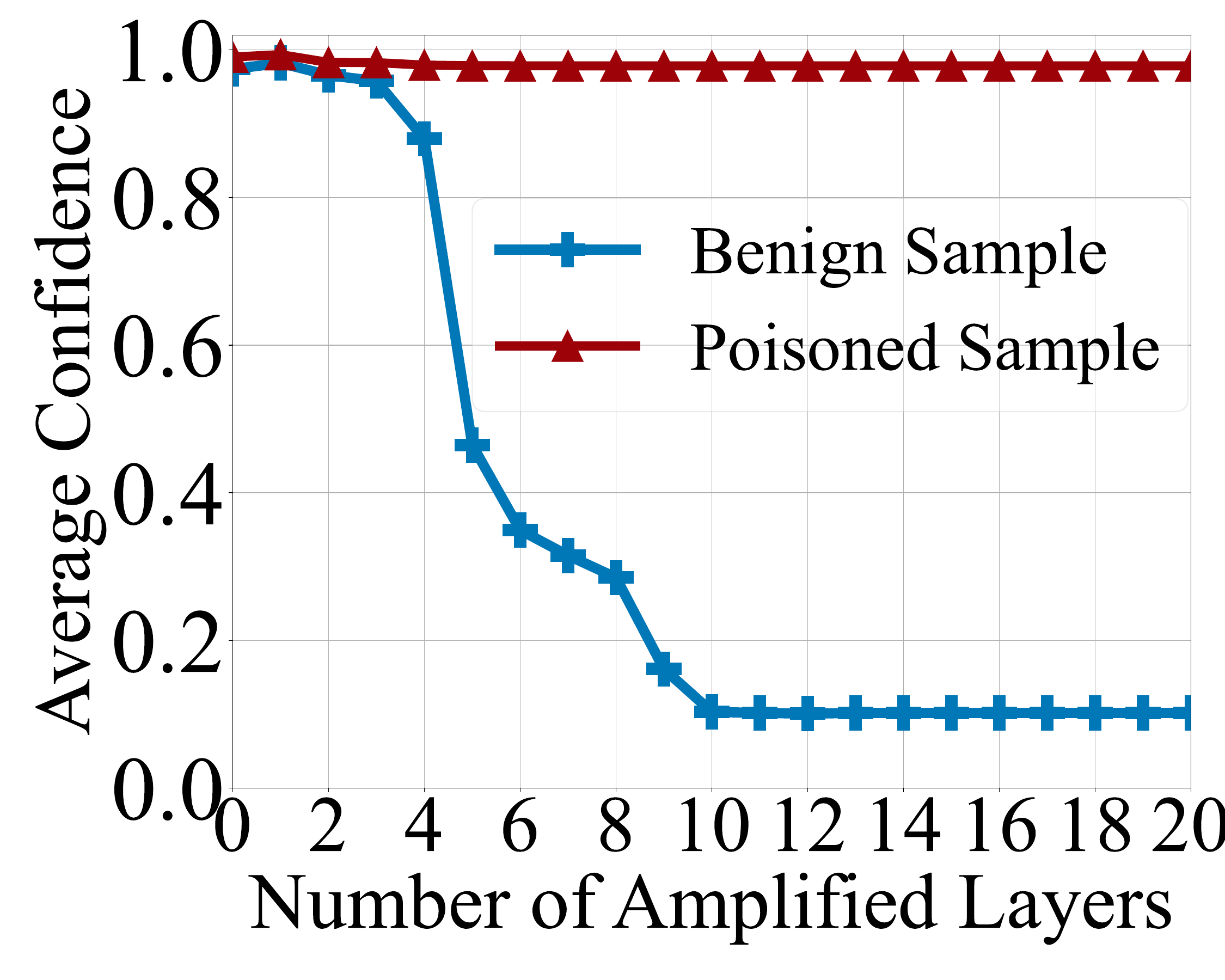}}
\subfigure[BATT\label{fig:batt_conf}]{\includegraphics[width=0.245\textwidth]{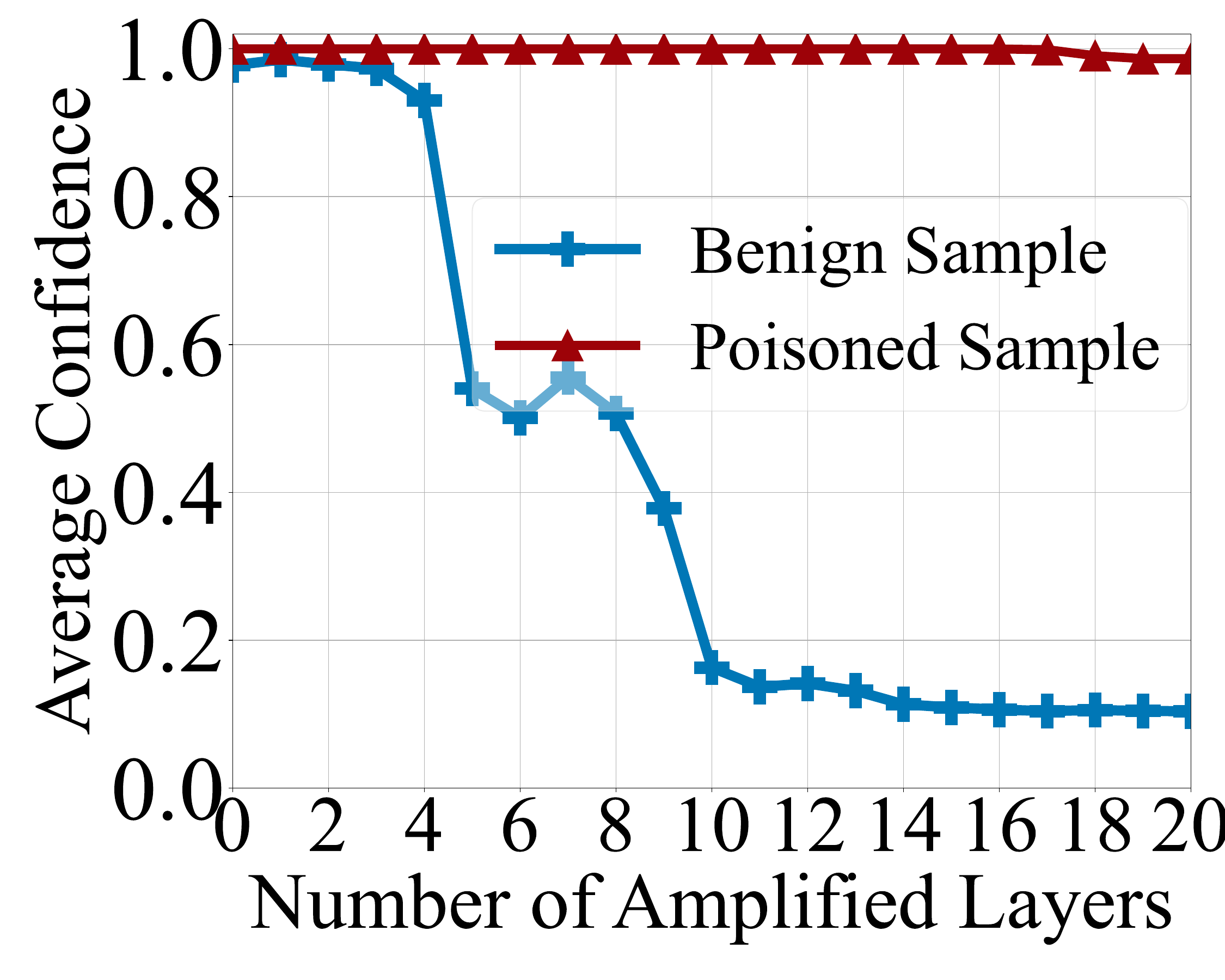}}
\vspace{-1em} 
\caption{The average confidence of benign and poisoned samples when amplifying different numbers of BN layers under benign and backdoored models (starting from the last layer).}
\vspace{-1em} 
\label{fig:intuition}
\end{figure*} 

\section{Parameter-oriented Scaling Consistency}
\label{sec:phe}

\begin{figure*}[!t]
\centering
\subfigure[Benign Model\label{fig:benign_l2}]{\includegraphics[width=0.24\textwidth]{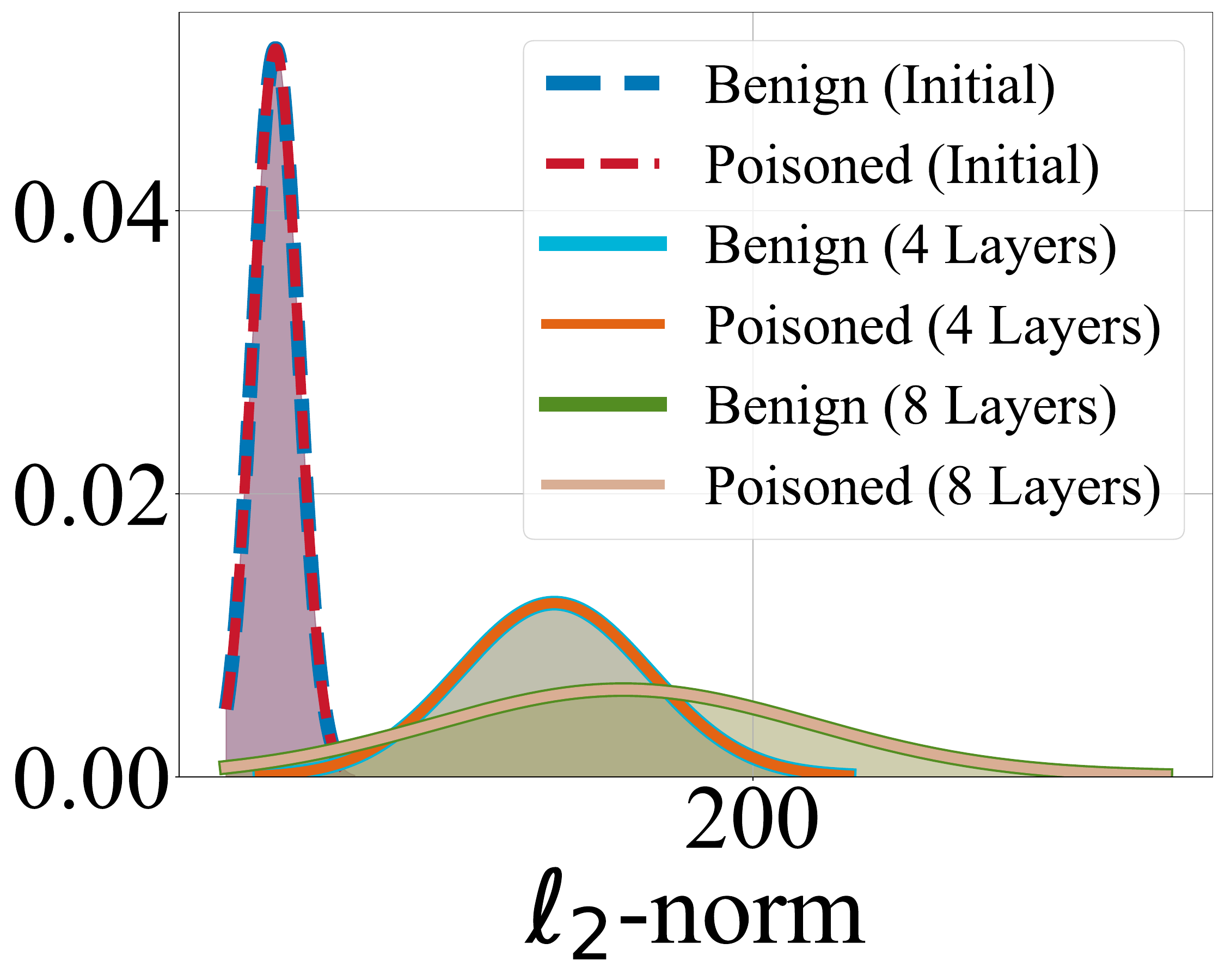}}
\subfigure[BadNets\label{fig:badnets_l2}]{\includegraphics[width=0.24\textwidth]{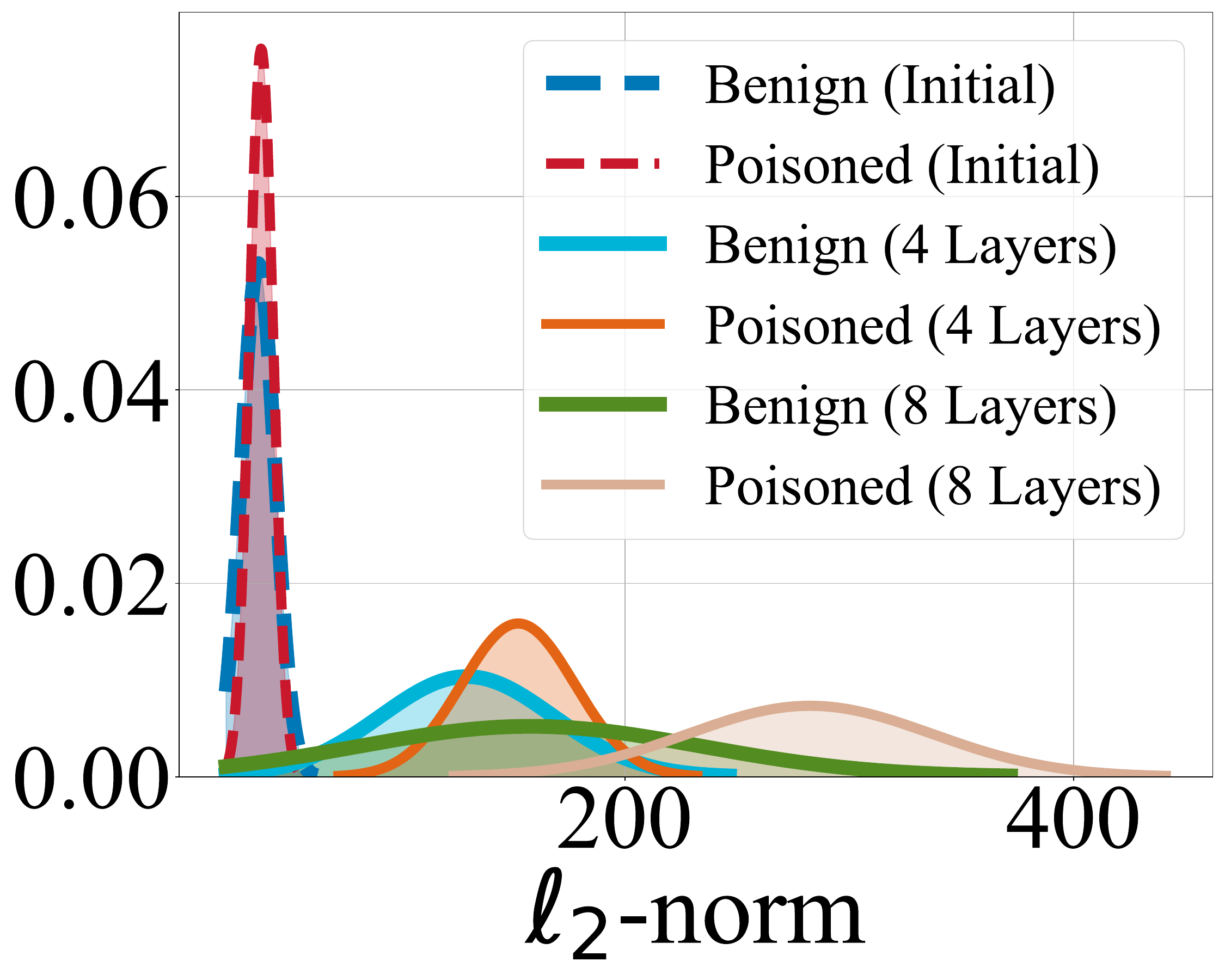}}
\subfigure[WaNet\label{fig:wanet_l2}]{\includegraphics[width=0.24\textwidth]{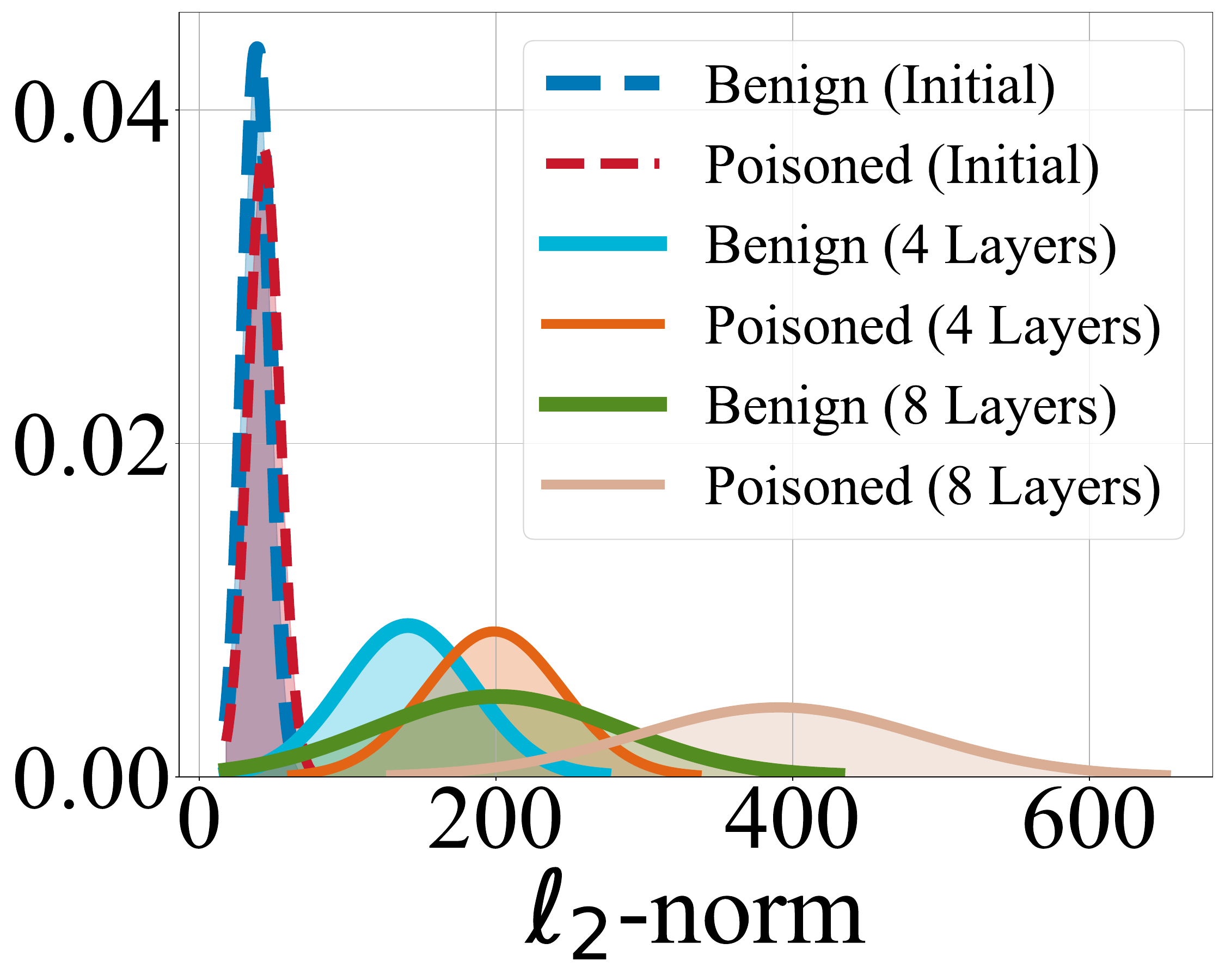}}
\subfigure[BATT\label{fig:batt_l2}]{\includegraphics[width=0.24\textwidth]{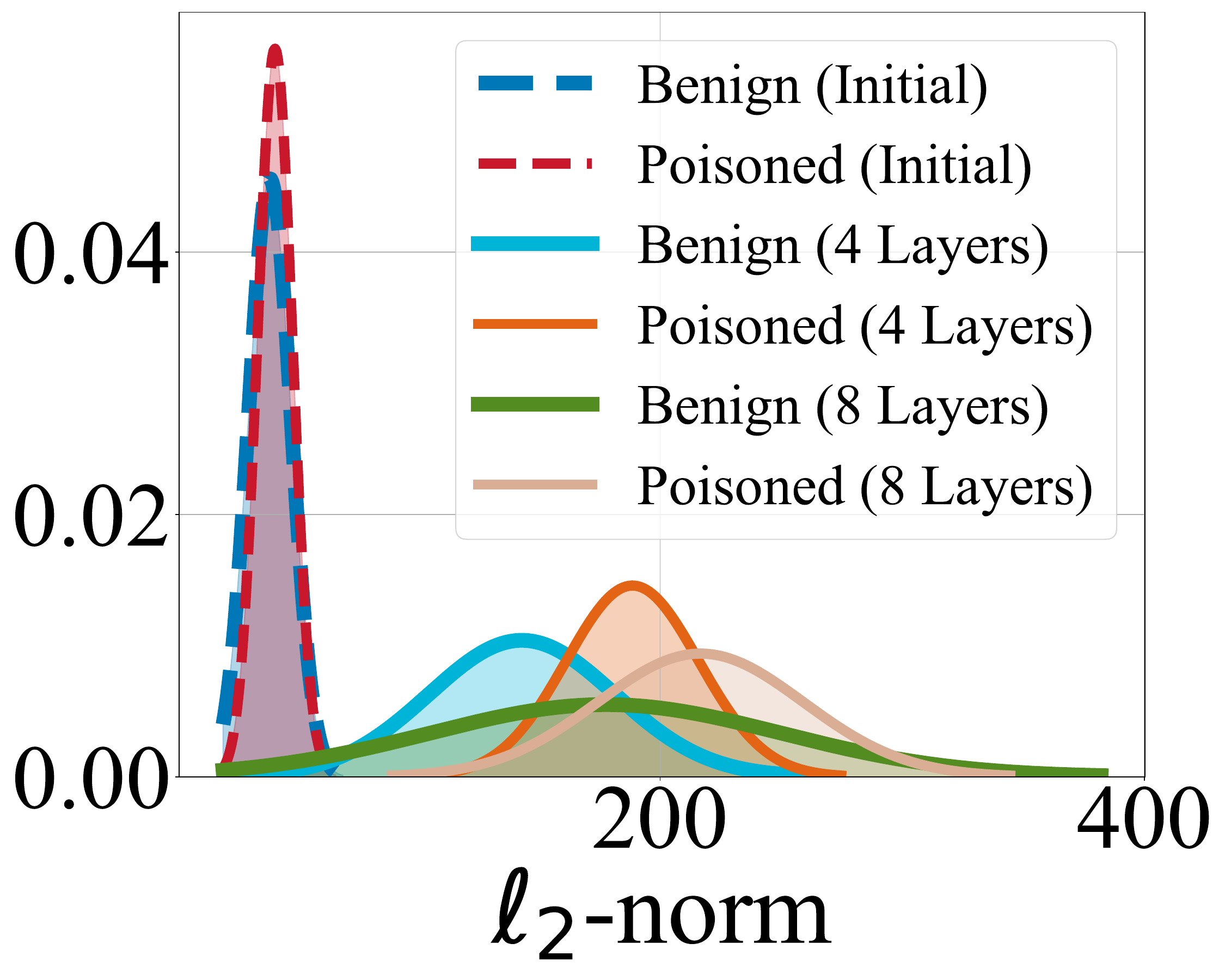}}
\vspace{-1em} 
\caption{The approximated distribution of the $\ell_2$-norm, fitted by Gaussian, of the final feature map of samples generated by models with different numbers of amplified BN layers. Increasing the number of amplified layers increases both value and variance of features.}
\vspace{-1.3em}
\label{fig:intuition_l2}
\end{figure*}

As demonstrated in \cite{guo2023scale}, the predictions of poisoned samples are significantly more consistent than benign ones when amplifying all pixel values. Motivated by the fact that model predictions result from the co-effects of samples and model parameters, in this section, we explore whether a similar intriguing phenomenon still exists if we scale up model parameters instead of pixel values.

For simplicity, we mainly focus on the learnable parameters of BN layers since they are used to transform features and are widely exploited in almost all advanced DNNs. Before illustrating our key observation and its theoretical support, we first briefly review the mechanism of BN.

\vspace{0.3em}
\textbf{Batch Normalization.} Let $\phi(\cdot;\gamma, \vbeta)$ denote the BN function, for a given batch feature maps $\va$, the BN operation transforms it into $\vb$: $\vb=\phi(\va; \gamma, \vbeta)$. This transformation is expressed as $\phi(\va; \gamma, \vbeta) = \gamma \left(\frac{\va - \vmu_a}{\sqrt{\sigma_a^2 + \epsilon}}\right) + \vbeta$, where $\epsilon$ is a small value, $\vmu_a$ and $\sigma_a$ are the mean and standard deviation of $\va$, respectively. The $\gamma$ and $\vbeta$ are learnable parameters, designed to scale and shift normalized features, and learned during the training process.

\vspace{-0.3em}
\noindent\textbf{Settings.} In this section, we adopt BadNets~\cite{gu2017badnets}, WaNet~\cite{nguyen2021wanet}, and BATT~\cite{xu2023batt} on CIFAR-10 \cite{krizhevsky2009learning} as examples for analyses. They are the representative of \textbf{(1)} patch-based attack, \textbf{(2)} sample-specific attack, and \textbf{(3)} physical attack, respectively. We exploit a standard ResNet-18 \cite{he2016deep} as our model structure. It contains twenty BN layers. For all attacks, we set the poisoning rate as 10\%. 
Specifically, for each benign and poisoned image, we scale up on the BN parameters (\ie, $\gamma$ and $\vbeta$) with $\omega=1.5$ times starting from the last layer and gradually moving forward to more layers. Similar to \cite{guo2023scale}, we also calculate the \emph{average confidence} defined as the average probability of samples on the label predicted by the original unamplified model. More details are in our~\cref{appendix:empirical}.

\noindent\textbf{Results.} 
As shown in~\cref{fig:benign_conf}, the average prediction confidence of the poisoned and benign samples decreases at almost the same rate as the number of amplified BN layers increases under the benign model. In contrast, as shown in~\cref{fig:badnets_conf}-\cref{fig:batt_conf}, the average prediction confidence of the poisoned samples remains nearly unchanged, whereas that of the benign samples also decreases during the parameter-amplified process under all three attacked models. In other words, benign and poisoned samples enjoy different BN-amplified prediction behaviors under attacked models. We call this intriguing phenomenon (of poisoned samples) as \emph{parameter-oriented scaling consistency (PSC)}.

To verify that the PSC phenomenon is not accidental, we provide the following theoretical and empirical analyses.

\begin{theorem}
\label{theorem1}
Let $F=FC\circ f_L\circ\dots\circ f_1$ be a backdoored DNN with $L$ hidden layers and FC denotes the fully connected layers. Let $\bm{x}$ be an input, $\vb=f_l\circ\cdots\circ f_1(\bm{x})$ be its batch-normalized feature after the $l$-th layer ($1\leq l\leq L$), and $t$ represent the attacker-specified target class. Assume that $\vb$ follows a mixture of Gaussian distributions. Then the following two statements hold: (1) Amplifying the $\vbeta$ and $\gamma$ parameters of the $l$-th BN layer can make $\Vert \tvb \Vert_2$ ($\tvb$ is the amplified version of $\vb$) arbitrarily large, and (2) There exists a positive constant $M$ that is independent of $\tvb$, such that whenever $\Vert \tvb \Vert_2 > M$, then $\arg \max FC\circ f_L\circ\dots\circ f_{l+1}(\tvb)=t$, even when $\arg \max FC\circ f_L\circ\dots\circ f_{l+1}(\vb)\not=t$.
\end{theorem}

\cref{theorem1} indicates that larger enough feature norms can induce decreasing confidence in the original predicted class if the inputs are benign samples (under certain classical assumptions in learning theory). Poisoned samples, instead, will stay fine. 
Its proof is in~\cref{th:proof}.

In practice, we find amplifying only a single BN layer may require an unreasonably large amplification factor and is unstable among different attacks or even BN layers, as demonstrated in~\cref{appendix:onebn}. Fortunately, as shown in~\cref{fig:intuition_l2}, amplifying multiple BN layers with a small factor (\eg, 1.5) can also significantly increase the feature norm in the last pre-FC layer and is more stable across different settings. As such, we amplify multiple layers throughout this work.

\begin{figure*}[!t]
    \centering
    \includegraphics[width=\textwidth]{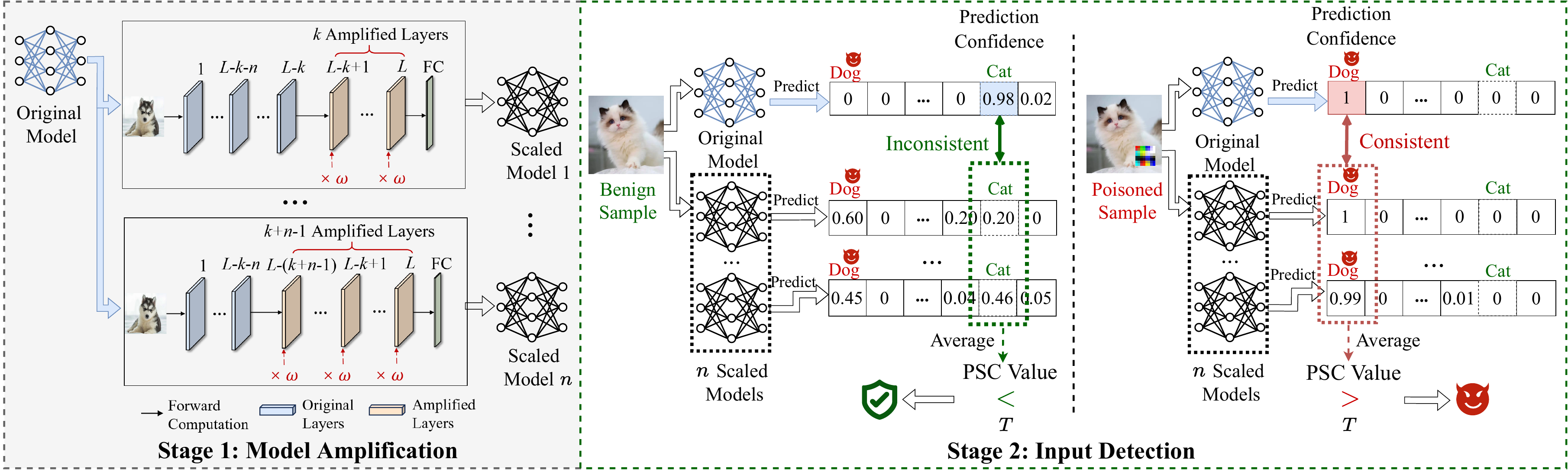} 
    \vspace{-1.5em}
    \caption{The main pipeline of our IBD-PSC. \textbf{Stage 1. Model Amplification}: Starting from the penultimate $k$-th layer of the original model, IBD-PSC gradually forward amplifies the parameters of more BN layers simultaneously to obtain $n$ different parameter-amplified models. \textbf{Stage 2. Input Detection}: For each suspicious image, IBD-PSC will first calculate the prediction confidence of the obtained $n$ parameter-amplified models on the label predicted by the original model. After that, IBD-PSC determines whether it is a poisoned sample by whether the average of obtained prediction confidences (defined as PSC value) is greater than a given threshold $T$.}
    \label{fig:work_flow}
    \vspace{-1em}
\end{figure*}

\section{The Proposed Method}

\subsection{Preliminaries}
\noindent\textbf{Threat Model.} This work focuses on input-level backdoor detection under the white-box setting with limited computational capacities. Defenders have full access to the suspicious model downloaded from a third party, but they lack the resources to remove potential backdoors (via backdoor mitigation). Similar to prior works~\cite{gao2021design,guo2023scale}, we assume that defenders have access to a limited number of local benign samples.

\noindent\textbf{Defenders' Goals.} An ideal IBD solution aims to precisely identify and eliminate all poisoned input samples while preserving the inference efficiency of the deployed model. Consequently, defenders have two main goals: \textbf{(1)} Effectiveness: The defense should accurately identify whether a given suspicious image is malicious. \textbf{(2)} Efficiency: The defense must operate in real-time and integrate seamlessly as a plug-and-play module, ensuring minimal impact on the model's inference time.

\noindent\textbf{The Overview of DNNs.} Consider a DNN model $\mathcal{F}: \mathcal{X} \rightarrow [0,1]^C$ consisting $L$ hidden layers, where $\mathcal{X}$ is the input space and $C$ is the number of classes. We can specify it as  
 \begin{equation}
     \mathcal{F}= \mathrm{FC}\circ f_L \circ f_{L-1} \circ \cdots \circ f_2 \circ f_1,
 \end{equation}
where $\mathrm{FC}$ denotes the fully-connected layers and $f_{i}$ represents $i$-th hidden layer consisting of one convolutional, batch normalization, and activation layer.

\noindent\textbf{The Main Pipeline of Backdoor Attack.} Let $ \mathcal{D} = \{ (\vx_i, y_i) \}_{i=1}^{N} $ denote a training set, consisting of $ N $ \textit{i.i.d.} samples. For each sample $(\vx, y)$, $ \vx \in \mathcal{X} = [0,1]^{d_c \times d_w \times d_h} $ and $ y \in \mathcal{Y} = \{1,2, \cdots, C\}$, an adversary creates a poisoned training set $\hat{\mathcal{D}}$ by injecting a pre-defined trigger $\vdelta$ into a subset of benign samples (\ie, $\mathcal{D}_s$). The trigger is procured through a designated trigger generating function, symbolized as $\vdelta = \tau(\vx)$, where $\tau: \mathcal{X} \rightarrow \mathcal{X}$. The generated poisoned samples are represented as $\mathcal{D}_p =\{\left( \hat{\vx}, t \right) \mid \hat{\vx}=\vx+\vdelta,(\vx,y)\in \mathcal{D}_s \}$. The final poisoned training set $\hat{\mathcal{D}}$ is formed by combining $\mathcal{D}_p$ with the remaining benign samples $\mathcal{D}_b$, \ie,  $\hat{\mathcal{D}} = \mathcal{D}_p \cup \mathcal{D}_b$. The poisoning rate is $\rho = |\mathcal{D}_p| / |\hat{\mathcal{D}}|$. The backdoor will be created for DNNs trained on the poisoned dataset $\hat{\mathcal{D}}$. 

\subsection{The Overview of IBD-PSC}

As demonstrated in~\cref{sec:phe}, the prediction confidences of poisoned samples exhibit greater consistency than those of benign ones when scaling up BN parameters of attacked DNNs. As such, we can detect whether a suspicious image is malicious by examining its parameter-oriented scaled consistency (PSC), a method we refer to as IBD-PSC. 

In general, as shown in~\cref{fig:work_flow}, our IBD-PSC has two main stages, including \textbf{(1)} model amplification and \textbf{(2)} input detection. In the first stage, we amplify the BN parameters of different layers in the original model to obtain \emph{a series of} parameter-amplified models. In the second stage, we calculate the PSC value of the suspicious image based on the obtained models and the original one. A larger PSC value indicates a higher likelihood that the suspicious image is malicious. The technical details are as follows.

\subsection{Model Amplification}
\noindent\textbf{Overview.} In this stage, we intend to obtain $n$ different parameter-amplified versions of the original model by amplifying the parameters (\ie, $\gamma$ and $\vbeta$) of its different BN layers. In particular, we amplify the later parts of the original model. It is motivated by the previous findings that trigger patterns often manifest as complicated features learned by the deeper (convolutional) layers of DNNs, especially for those attacks with elaborate designs \cite{huang2022backdoor,jebreel2023defending}. This finding is consistent with our observations in~\cref{fig:intuition}. Specifically, let $k$ denote the penultimate BN layers in which we scale up in the first parameter-amplified model. For the $i$-th amplified model, we scale up the parameters in the last $(k+i-1)$ BN layers with the same scaling factor $\omega$. Let $\mathcal{F}$ denotes the original model, its parameter-amplified version containing $k$ amplified BN layers with scaling factor $\omega$ (\ie, $\hat{\mathcal{F}}_{k}^{\omega}$) can be defined as 
\begin{equation}
\label{eq:hatF}
    \hat{\mathcal{F}}_{k}^{\omega} = \mathrm{FC} \circ \hat{f}_L^{\omega} \circ \hat{f}_{L-1}^{\omega} \circ ... \circ \hat{f}_{L-k+1}^{\omega} \circ ... \circ f_2 \circ f_1,
\end{equation}
where $\hat{f}_{i}^{\omega}$ represents the BN layer of the $i$-th hidden layer undergoing an amplification process. It scales the original BN layer's parameters $\gamma$ and $\vbeta$ by a scaling factor ${\omega}$, \ie, $\hat{\gamma}=\omega\cdot \gamma$ and $\hat{\vbeta}=\omega\cdot \vbeta$. We also conduct ablation studies in~\cref{appendix:forward} and~\cref{appendix:allbnlayers} to assess the impact of amplifying BN layers in a forward sequential manner and that of amplifying all BN layers, respectively.

We exploit $n$ instead of one parameter-amplified model (with many amplified BN layers) to balance the performance on benign and poisoned samples. In practice, $n$ is a defender-assigned hyper-parameter. More details and its impact are included in~\cref{appendix:timesn}. Accordingly, the last remaining question for model amplification is selecting a suitable starting point $k$. Its technical details are as follows.

\noindent\textbf{Layer Selection.} To optimally determine the number of amplified BN layers, we design an adaptive algorithm to dynamically select a suitable $k$. Motivated by our PSC phenomenon (see \cref{fig:intuition}), we intend to \emph{find the point where the prediction accuracy for benign samples begins to decline significantly}. Specifically, we incrementally increase $k$ from $1$ to $L$ and monitor the error rate $\eta$. Let $\mathcal{D}_{r}$ denote the set of remaining benign samples. We can then compute the error rate $\eta$ as the proportion of samples within $\mathcal{D}_{r}$ that are misclassified by the parameter-amplified model $\hat{\mathcal{F}}_{k}^{\omega}$, \ie, 
\begin{equation}
\label{eq:errorrate}
\eta = \frac{1}{|\mathcal{D}_r|} \sum_{(\vx,y)\in \mathcal{D}_r} \mathbb{I}\left({\argmax}\left(\hat{\mathcal{F}}_{k}^{\omega}\left(\vx\right)\right)\neq y\right),    
\end{equation} 
where $\mathbb{I}$ denotes the indicator function. Once $\eta$ exceeds a predefined threshold $\xi$ (\eg, 60\%), the BN layers from the $(L-k+1)$-th to the $L$-th layer are determined as the target layers for amplification. The details of the adaptive algorithm are outlined in~\cref{alg:identifylayers}.
\begin{algorithm}[!t]
   \caption{Adaptive layer selection.}
  \label{alg:identifylayers}
\begin{algorithmic}
   \STATE {\bfseries Input:} original model $\mathcal{F}$, scaling factor $\omega$, error rate threshold $\xi$, local benign dataset $\mathcal{D}_r$
   \STATE {\bfseries Output:} optimal number of amplified BN layers (\ie, $k$) for the first parameter-amplified model 
   \FOR{$i \gets 1$ \textbf{to} $L$}
        \STATE $k = i$
       \STATE Generate the parameter-amplified model $\hat{\mathcal{F}}_{k}^{\omega}$ using~\cref{eq:hatF}
       \STATE Calculate the error rate $\eta$ using~\cref{eq:errorrate}
   \IF{$\eta>\xi$}
        \STATE \textbf{break} 
   \ENDIF
   \ENDFOR
   \STATE \textbf{return} $k$
\end{algorithmic}
\end{algorithm}
\subsection{Input Detection} 
Once we obtain $n$ parameter-amplified versions of the original model $\mathcal{F}$ with the starting amplified point $k$ (\ie, $\{\hat{\mathcal{F}}_{k+i-1}^{\omega}\}_{i=1}^n$), for each suspicious image, our IBD-PSC can examine it by calculating its PSC value based on their predictions. Specifically, we define the PSC value as the average confidence generated over parameter-amplified models on the label predicted by the original model, \ie, 
\begin{equation}
\textrm{PSC}(\vx) = \frac{1}{n} \sum_{i=k}^{k+n-1} \hat{\mathcal{F}}_{i}^\omega(\vx)_{y'},
\end{equation}
where $y' = \argmax \left(\mathcal{F}(\vx)\right)$.
After obtaining the PSC value, IBD-PSC assesses whether the input sample is malicious by comparing it to a predefined threshold $T$. If $\textrm{PSC} > T$, it is marked as a poisoned image.

\section{Experiments}
\label{sec:exp}
\subsection{Experiment Settings}

\begin{table*}[!t]
\centering
\vspace{-0.8em}
\caption{ The performance (AUROC, F1) on the CIFAR-10 dataset. We mark the best result in boldface and failed cases ($<0.7$) in red.}
\label{tab:main-cifar10}
\small
\setlength{\tabcolsep}{1pt}
\begin{tabular}{lccccccccccccccccc} 
\toprule
Attacks$\rightarrow$  &\multicolumn{2}{c}{BadNets} &\multicolumn{2}{c}{Blend} & \multicolumn{2}{c}{PhysicalBA} & \multicolumn{2}{c}{IAD} & \multicolumn{2}{c}{WaNet} & \multicolumn{2}{c}{ISSBA} & \multicolumn{2}{c}{BATT} & \multicolumn{2}{c}{Avg.}  \\
Defenses$\downarrow$ & AUROC & F1 &  AUROC & F1& AUROC & F1&   AUROC & F1&   AUROC  & F1  &  AUROC & F1  &  AUROC & F1 &  AUROC & F1\\
\midrule
 STRIP & 0.931 &0.842 &\tred{0.453} & \tred{0.114} & 0.884 &  0.882& 0.962 & 0.907& \tred{0.469} & \tred{0.125} & \tred{0.364} & \tred{0.526} &\tred{0.449} & \tred{0.258} & \tred{0.663} & \tred{0.494}\\ 
  TeCo & 0.998 & 0.970 & \tred{0.675}& \tred{0.678} & 0.748 & \tred{0.689} & 0.909 & 0.920 & 0.923 & 0.915 & 0.901 & 0.942 &  0.914 & \tred{0.673} & 0.858 & 0.834\\
  SCALE-UP& 0.962 & 0.913 & \tred{0.644}& \tred{0.453}& 0.969  & 0.715 & 0.967 & 0.869& \tred{0.672}& \tred{0.529}  & 0.942  &  0.894 & 0.959 & 0.911 & 0.731 & 0.757\\
 IBD-PSC  & \bf{1.000} & \bf{0.967} & \bf{0.998} & \bf{0.960} & \bf{0.972} &  \bf{0.942} & \bf{0.983} & \bf{0.952} & \bf{0.984} & \bf{0.956} & \bf{1.000} & \bf{0.986}& \bf{0.999} & \bf{0.966} & \bf{0.992} & \bf{0.961}\\
\bottomrule
\end{tabular}
\vspace{-1.5em}
\end{table*}

\begin{table*}[!t]
\centering
\caption{The performance (AUROC, F1) on the GTSRB dataset. We mark the best result in boldface and failed cases ($<0.7$) in red.}
\label{tab:main-gtsrb}
\small
\setlength{\tabcolsep}{1pt}
\begin{tabular}{lccccccccccccccccc} 
\toprule
Attacks$\rightarrow$  &\multicolumn{2}{c}{BadNets} &\multicolumn{2}{c}{Blend} & \multicolumn{2}{c}{PhysicalBA} & \multicolumn{2}{c}{IAD} & \multicolumn{2}{c}{WaNet} & \multicolumn{2}{c}{ISSBA} & \multicolumn{2}{c}{BATT} & \multicolumn{2}{c}{Avg.}  \\
 Defenses$\downarrow$ & AUROC & F1 &  AUROC & F1& AUROC & F1&   AUROC & F1&   AUROC  & F1  &  AUROC & F1  &  AUROC & F1 &  AUROC & F1\\
\midrule
 STRIP  & 0.962 & 0.915 &\tred{0.426} & \tred{0.088}& 0.700  & \tred{0.479} & 0.855 & 0.890& \tred{0.356} &\tred{0.201} & \tred{0.640}& \tred{0.625} & \tred{0.648} &\tred{0.368} & \tred{0.657} & \tred{0.588} \\
 TeCo& 0.879 &  0.905 & 0.917& 0.913& 0.860 & \tred{0.673} & 0.955 & 0.962 & 0.954 & 0.935 & 0.941 & 0.947 & 0.829 & \tred{0.673} & 0.907 & 0.858\\
 SCALE-UP & 0.913 & 0.858 & \tred{0.579}&\tred{0.421}  &0.762 & 0.709& 0.885 & 0.860 & \tred{0.309} & \tred{0.149} & 0.733 & \tred{0.691} & 0.902 & 0.876 & 0.700 & \tred{0.669}\\
  IBD-PSC & \bf{0.968} &  \bf{0.965}& \bf{0.953} &\bf{0.928} & \bf{0.940} & \bf{0.946} & \bf{0.970} & \bf{0.971} & \bf{0.986} & \bf{0.973} & \bf{0.972} & \bf{0.971} & \bf{0.969} & \bf{0.968} &\bf{0.969} & \bf{0.962}\\
\bottomrule
\end{tabular}
\vspace{-1.3em}
\end{table*}

\begin{table*}[!t]
\centering
\caption{The performance (AUROC, F1) on SubImageNet-200. We mark the best result in boldface and failed cases ($<0.7$) in red.}
\label{tab:main-imagenet}
\small
\setlength{\tabcolsep}{1pt}
\begin{tabular}{lccccccccccccccccc} 
\toprule
Attacks$\rightarrow$  &\multicolumn{2}{c}{BadNets} &\multicolumn{2}{c}{Blend} & \multicolumn{2}{c}{PhysicalBA} & \multicolumn{2}{c}{IAD} & \multicolumn{2}{c}{WaNet} & \multicolumn{2}{c}{ISSBA} & \multicolumn{2}{c}{BATT} & \multicolumn{2}{c}{Avg.}  \\
 Defenses$\downarrow$ & AUROC & F1 &  AUROC & F1& AUROC & F1&   AUROC & F1&   AUROC  & F1  &  AUROC & F1  &  AUROC & F1 &  AUROC & F1\\
\midrule
 STRIP  &0.840 &0.828 & 0.799& 0.772&\tred{0.618} &\tred{0.468} & \tred{0.528}& \tred{0.419}& \tred{0.563} & \tred{0.356} &  0.768 & 0.765 & \tred{0.554} & \tred{0.361} & \tred{0.681} & \tred{0.596}\\
TeCo& 0.978 & 0.880 & 0.958& 0.849& 0.926 & 0.842& 0.927 & 0.920 & 0.903 & 0.747 & 0.945 & 0.921 & \tred{0.690}& \tred{0.692} & 0.908 & 0.846\\
 SCALE-UP & 0.967 & 0.895 &\tred{0.531} &\tred{0.356} & 0.932 & 0.876 & \tred{0.322} & \tred{0.030} & \tred{0.563} & \tred{0.356} & 0.945 & 0.912&  0.967 & 0.921 & 0.725 & \tred{0.651}\\
 IBD-PSC & \bf{1.000} & \bf{0.992} & \bf{0.989} &\bf{0.833}&\bf{0.994}&\bf{0.988}&\bf{0.994}& \bf{0.996}&\bf{0.967} & \bf{0.981} &\bf{0.989} & \bf{0.987} &\bf{0.998} &\bf{0.998} & \bf{0.990} & \bf{0.974}\\
\bottomrule
\end{tabular}
\vspace{-1em}
\end{table*}

\noindent\textbf{Datasets and Models.} We follow the settings in existing backdoor defenses and conduct experiments on CIFAR-10~\cite{krizhevsky2009learning}, GTSRB~\cite{stallkamp2012man} and a subset of ImageNet dataset with 200 classes (dubbed `SubImageNet-200')~\cite{deng2009imagenet} using the ResNet18 architecture ~\cite{he2016deep}. More detailed settings are presented in~\cref{appendix:dataset}.


\noindent\textbf{Attack Baselines.} We evaluate the effectiveness of IBD-PSC against thirteen representative backdoor attacks, including \textbf{1)} BadNets~\cite{gu2017badnets}, \textbf{2)} Blend~\cite{chen2017targeted}, \textbf{3)} LC~\cite{turner2019label}, \textbf{4)} ISSBA~\cite{li2021invisible}, \textbf{5)} TaCT~\cite{tang2021demon}, \textbf{6)} NARCISSUS~\cite{zeng2023narcissus}, \textbf{7)} Adap-Patch~\cite{qi2023revisiting}, \textbf{8)} BATT~\cite{xu2023batt}, \textbf{9)} PhysicalBA~\cite{li2021backdoor}, \textbf{10)} IAD~\cite{nguyen2020input}, \textbf{11)} WaNet~\cite{nguyen2021wanet}, \textbf{12)} BPP~\cite{Wangbpp}, and \textbf{13)} SRA~\cite{qi2022towards}. The first eight attacks are representative of poison-only attacks, while the last one is a model-controlled attack. The remaining four are training-controlled attacks. More details about the attack baselines are in the~\cref{appendix:baselineattacks}. 

\noindent\textbf{Defense Settings.}
We compare our defense with classical and advanced input-level backdoor defenses, including STRIP~\cite{gao2021design}, TeCo~\cite{liu2023detecting} and SCALE-UP~\cite{guo2023scale}. We implement these defenses using their official codes with default settings. Our IBD-PSC defense maintains a consistent hyper-parameter setting across various attacks and datasets. Specifically, we set $\omega=1.5$, $n=5$, $\epsilon=60\%$, and $T=0.9$. Defenders can only access 100 benign samples as their local samples. More setting details about baseline methods are in~\cref{appendix:baselinedefenses}.


\noindent\textbf{Evaluation Metrics.}
We employ two common metrics in our evaluation: \textbf{1)} the area under the receiver operating curve (AUROC) measures the overall performance of detection methods across different thresholds, and \textbf{2)} the F1 score measures both detection precision and recall.

\subsection{Main Results}
\label{sec:mainresult}
As shown in~\cref{tab:main-cifar10}-\ref{tab:main-imagenet}, IBD-PSC consistently achieves promising performance in all cases across various datasets. For instance, it achieves AUROC and F1 scores approaching 1.0, indicating its effectiveness in various attack scenarios. The results also demonstrate that IBD-PSC achieves a substantial improvement in detection performance compared to the defense baselines.  
In contrast, all baseline defenses fail in some cases (marked in red), especially under attacks involving subtle alterations across multiple pixels (\eg, Blend, WaNet) or physical attacks. This failure is primarily caused by their implicit assumptions about backdoors, such as sample-agnostic triggers and robustness against image preprocessing. We also provide the results with PreActResNet18~\cite{he2016identity} and MobileNet~\cite{krizhevsky2009learning} architectures in~\cref{appendix:moremodels}. We also provide the ROC curves of defenses against four representative attacks in~\cref{appendix_auccurve}. Besides, for more experimental results under other attack baselines in~\cref{appendix:moreattacks}.

We also calculated the inference time of all methods under identical and ideal conditions for evaluating efficiency. For example, we assume that defenders will load all required models and images simultaneously (with more memory requirements compared to the vanilla model inference). Arguably, this comparison is fair and reasonable since different defenses differ greatly in their mechanisms and requirements. Detailed settings can be found in~\cref{appendix:time}. As shown in ~\cref{fig:time}, the efficiency of our IBD-PSC is on par with or even better than all baseline defenses. The extra time is negligible compared to no defense, although IBD-PSC may increase some storage or computational consumptions. More detailed discussions about our running efficiency and storage requirements are in \cref{appendix:limitations}. 

\begin{figure}[!t]
    \centering
    \begin{minipage}{0.7\linewidth}
        \includegraphics[width=\linewidth]{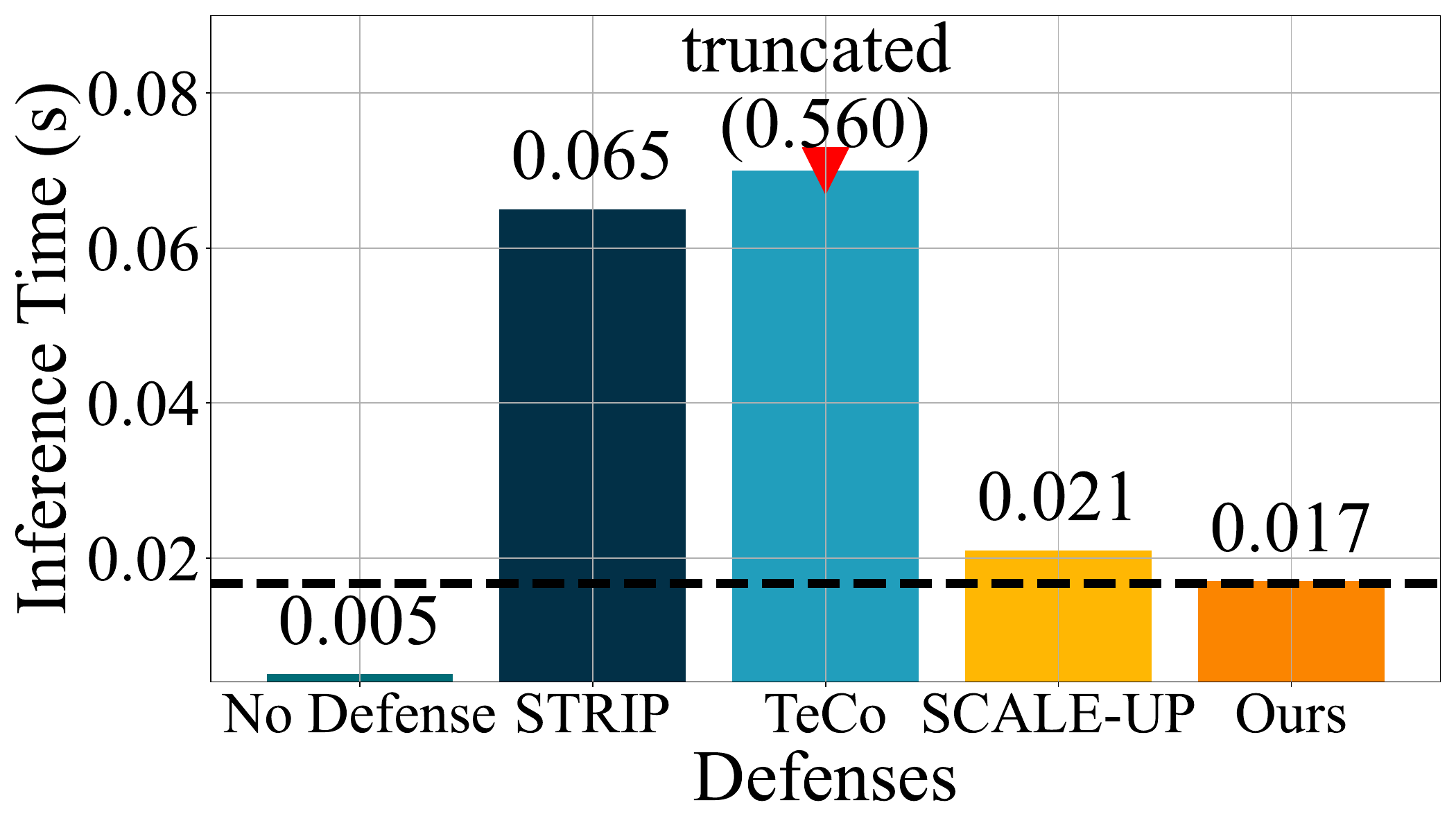}
    \end{minipage}
    \vspace{-1em}
 \caption{The inference time on the CIFAR-10 dataset.} 
 \label{fig:time}
\vspace{-2em}
\end{figure}
\subsection{Ablation Study}

\begin{figure}[!t]
	\centering
 \begin{minipage}{0.99\linewidth}
    \begin{minipage}{0.325\linewidth}
            \includegraphics[width=1\linewidth]{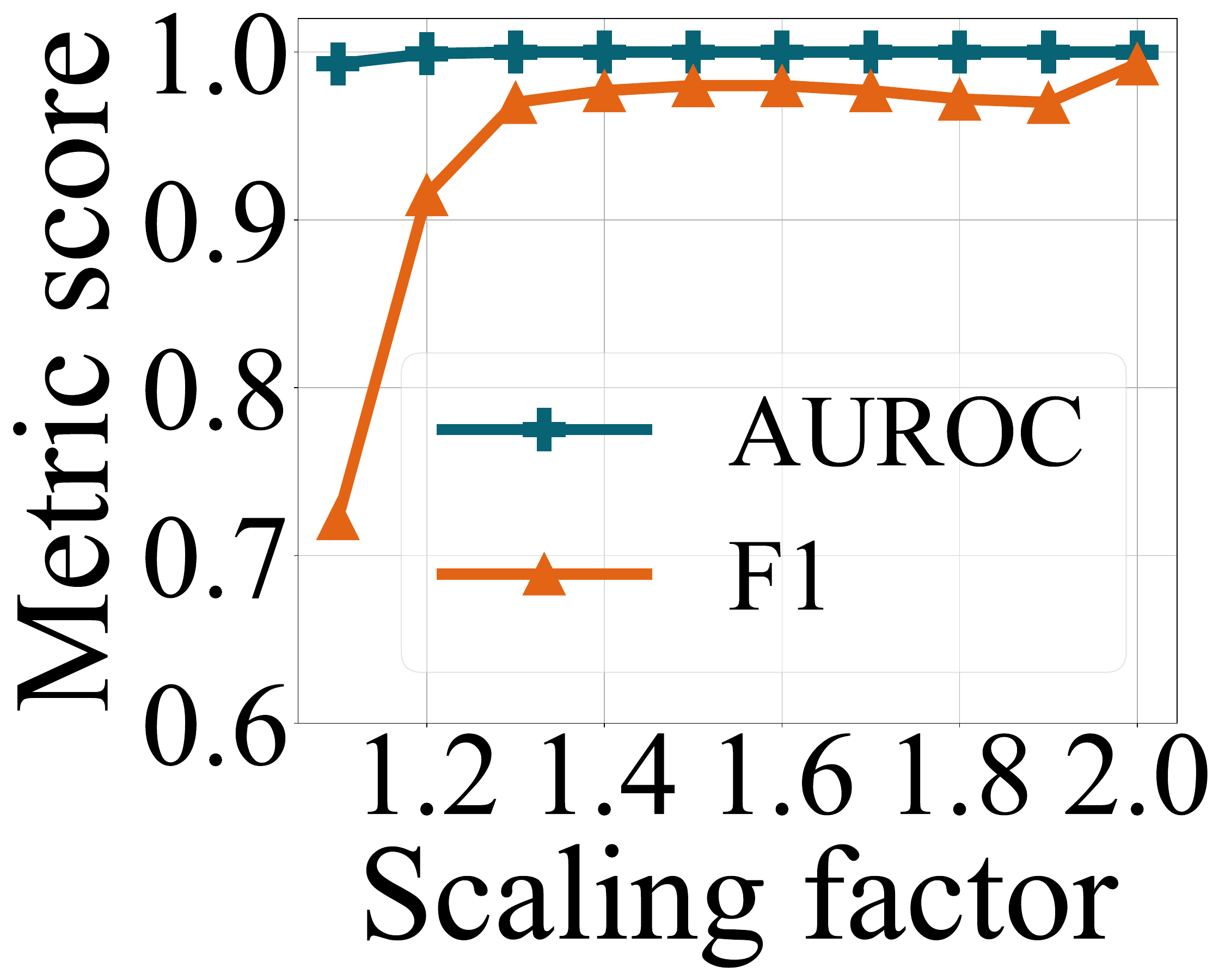}
            \centerline{(a) BadNets}
    \end{minipage}
    \begin{minipage}{0.325\linewidth}
            \includegraphics[width=1\linewidth]{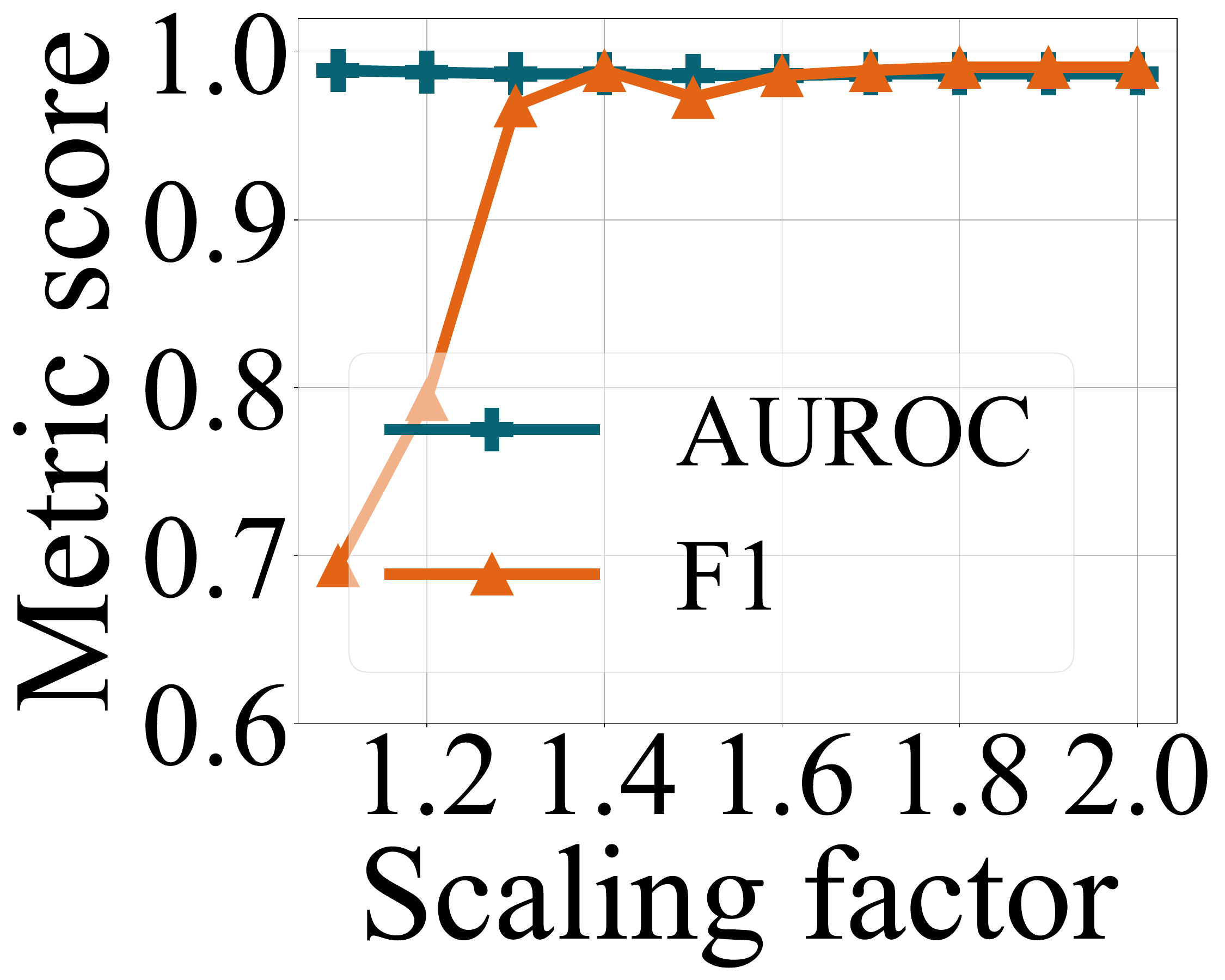}
            \centerline{(b) WaNet}
    \end{minipage}
    \begin{minipage}{0.325\linewidth}
            \includegraphics[width=1\linewidth]{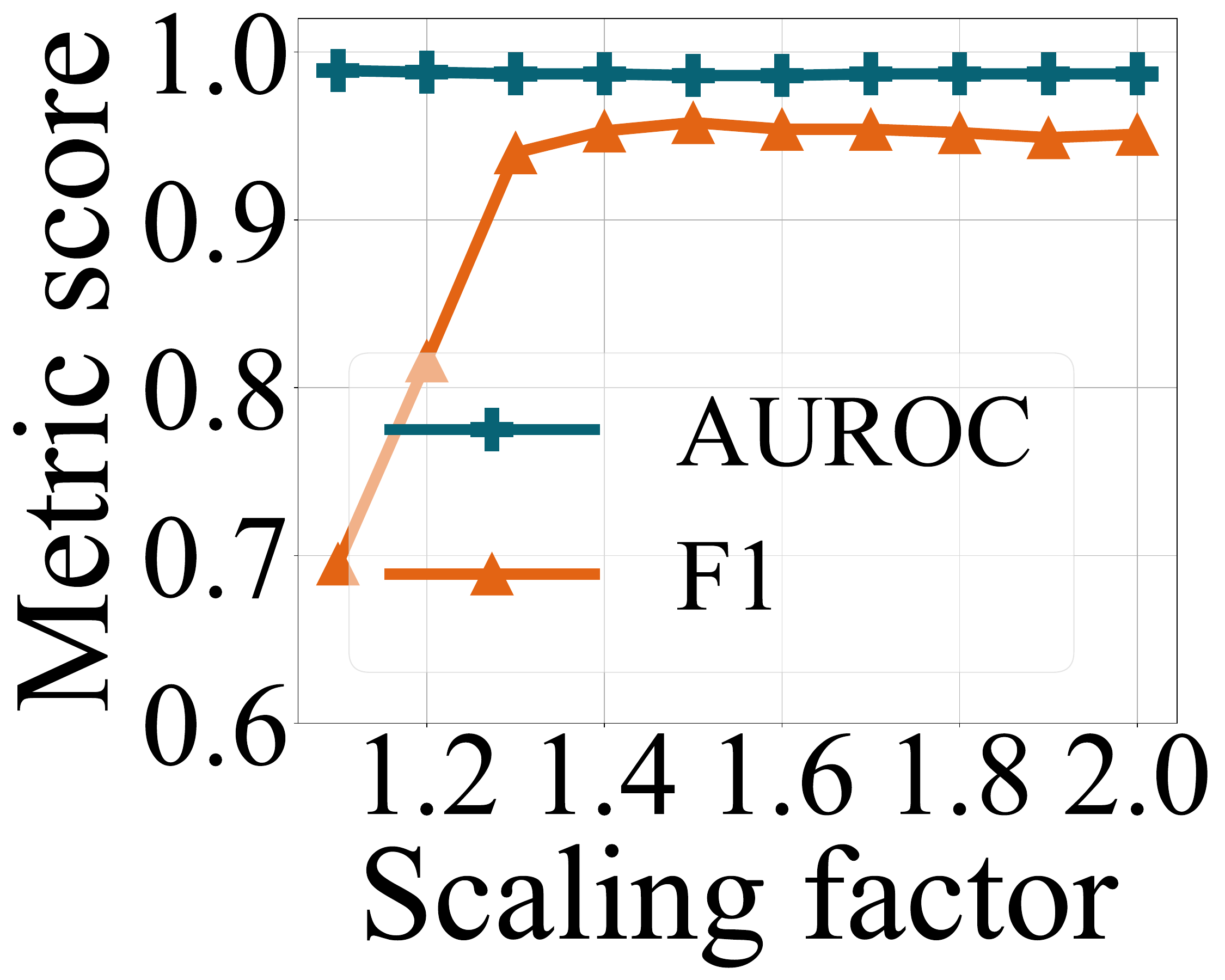}
            \centerline{(c) BATT}
    \end{minipage}
\end{minipage}
\vspace{-0.5em}
 \caption{The impact of scaling factors on CIFAR-10.}
 \label{fig:magnification}
\vspace{-1em}
\end{figure}

\noindent\textbf{Impact of Scaling Factor $\omega$.} IBD-PSC generates scaled models by amplifying the learnable parameter values of the selected BN layers with a fixed scalar $\omega$. We hereby explore its effects on our method. Specifically, we vary $\omega$ from 1 to 2 and calculate the AUROC and F1 scores of IBD-PSC against three representative backdoor attacks (\ie, BadNets, WaNet, and BATT) on CIFAR-10. As shown in~\cref{fig:magnification}, in the initial phase, increasing $\omega$ can significantly improve both the AUROC and F1 scores against different backdoor attacks. Furthermore, the AUROC and F1 scores converge to nearly one and stabilize at approximately one for $\omega$ values of 1.5 or higher, \ie, the scaling factor has a relatively minor influence when it is sufficiently large. Besides, we conduct further ablation studies on other hyper-parameters of our method, as detailed in~\cref{appendix:ablation}.

\noindent\textbf{Impact of Confidence Consistency.} 
IBD-PSC leverages the consistency of confidence for detection, in contrast to the SCALE-UP method, which relies on the consistency of the predicted label. SCALE-UP is designed for black-box scenarios where defenders only have access to predicted labels, while our IBD-PSC focuses on white-box settings where predicted confidences are naturally available. To validate the effectiveness of IBD-PSC, we develop a variant that uses label consistency (dubbed `Ours-L'). We then calculate the False Positive Rate (FPR) (\%) for both target and benign classes on the CIFAR-10 dataset across various backdoor attacks. As shown in~\cref{tab:FPR_target}, our method significantly reduces false positives in both the target and benign classes, outperforming both the Ours-L and SCALE-UP.

\begin{figure}[!t]
	\centering
 \begin{minipage}{0.99\linewidth}
    \begin{minipage}{0.325\linewidth}
            \includegraphics[width=1\linewidth]{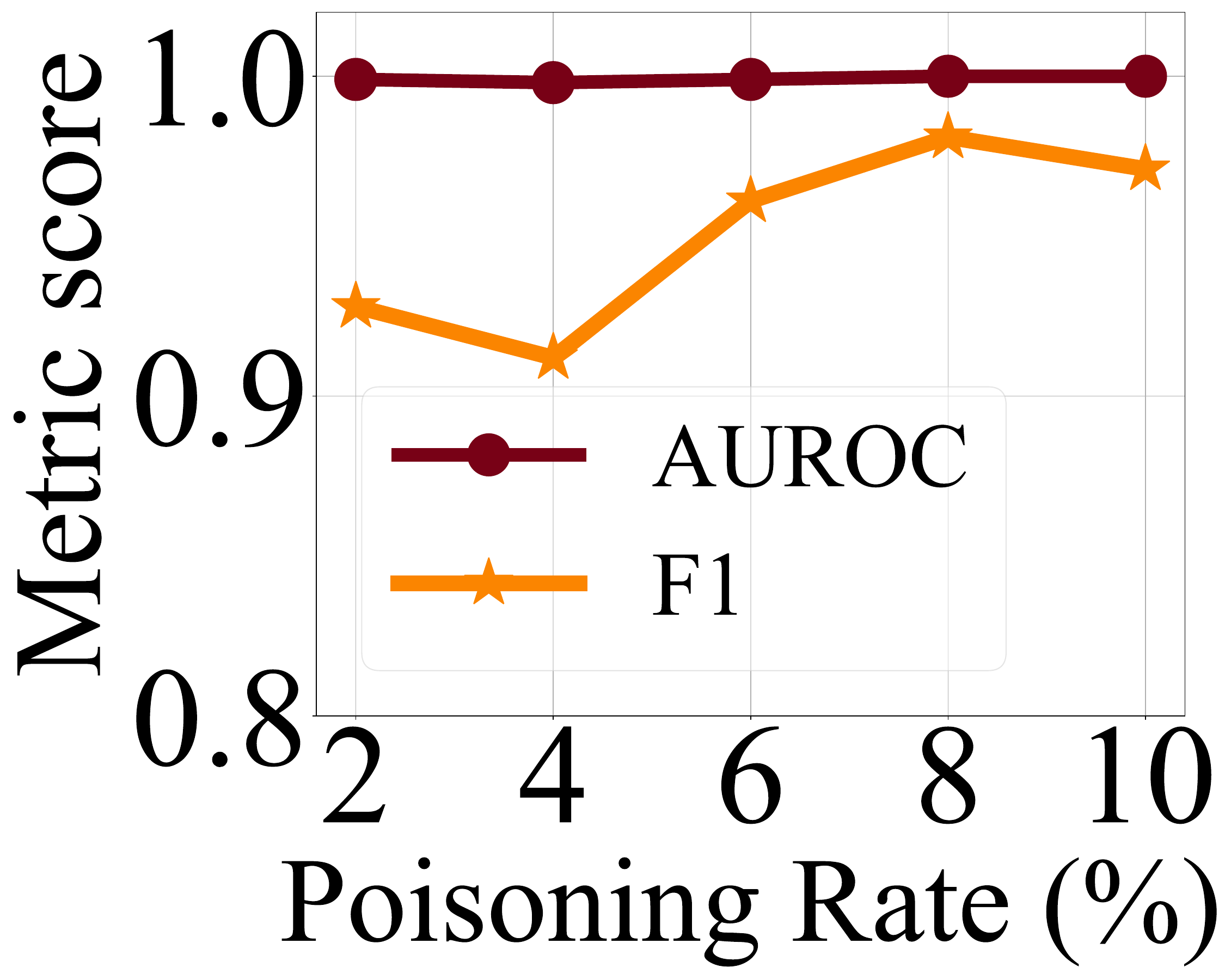}
            \centerline{(a) BadNets}
    \end{minipage}
    \begin{minipage}{0.325\linewidth}
            \includegraphics[width=1\linewidth]{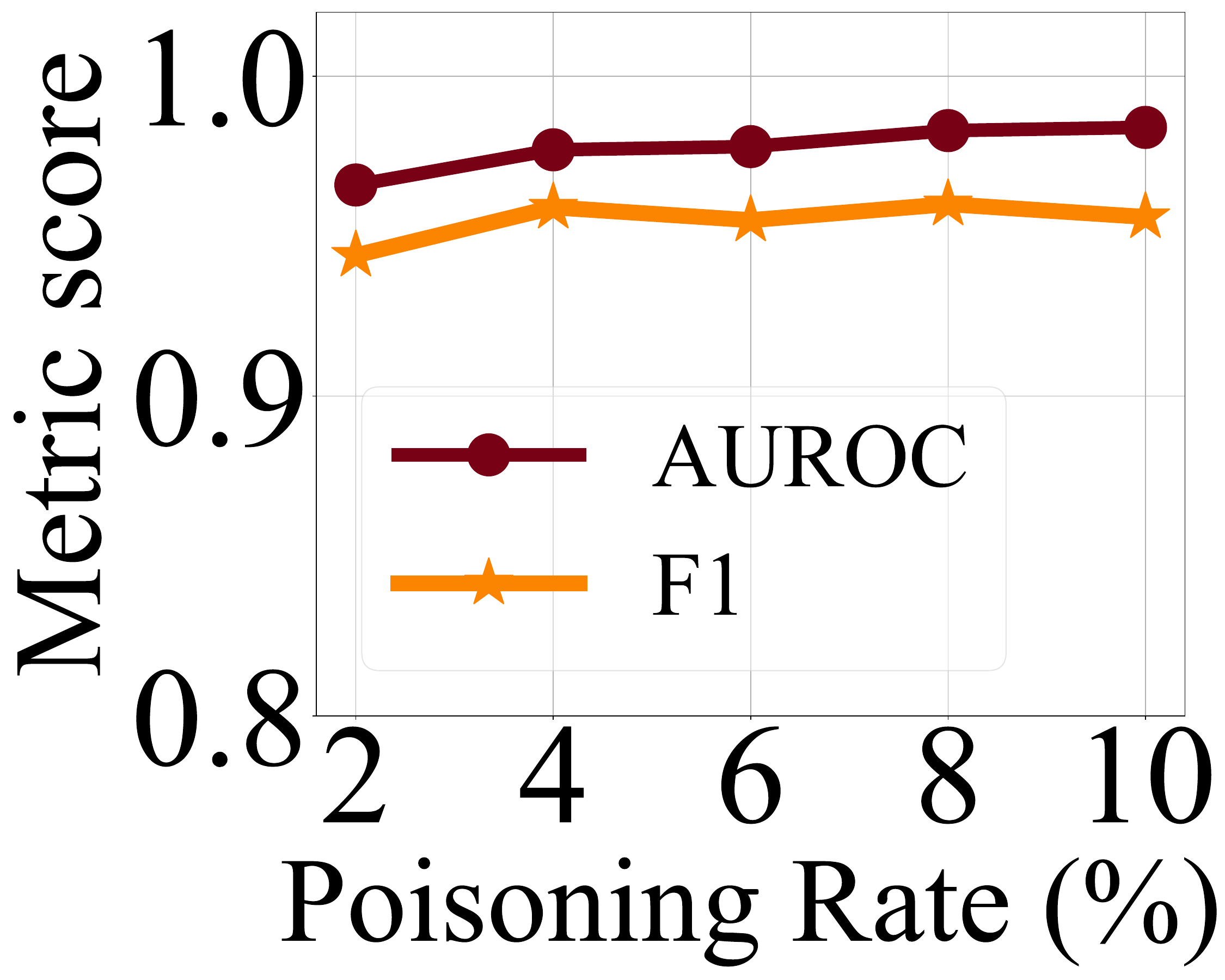}
            \centerline{(b) WaNet}
    \end{minipage}
    \begin{minipage}{0.325\linewidth}
            \includegraphics[width=1\linewidth]{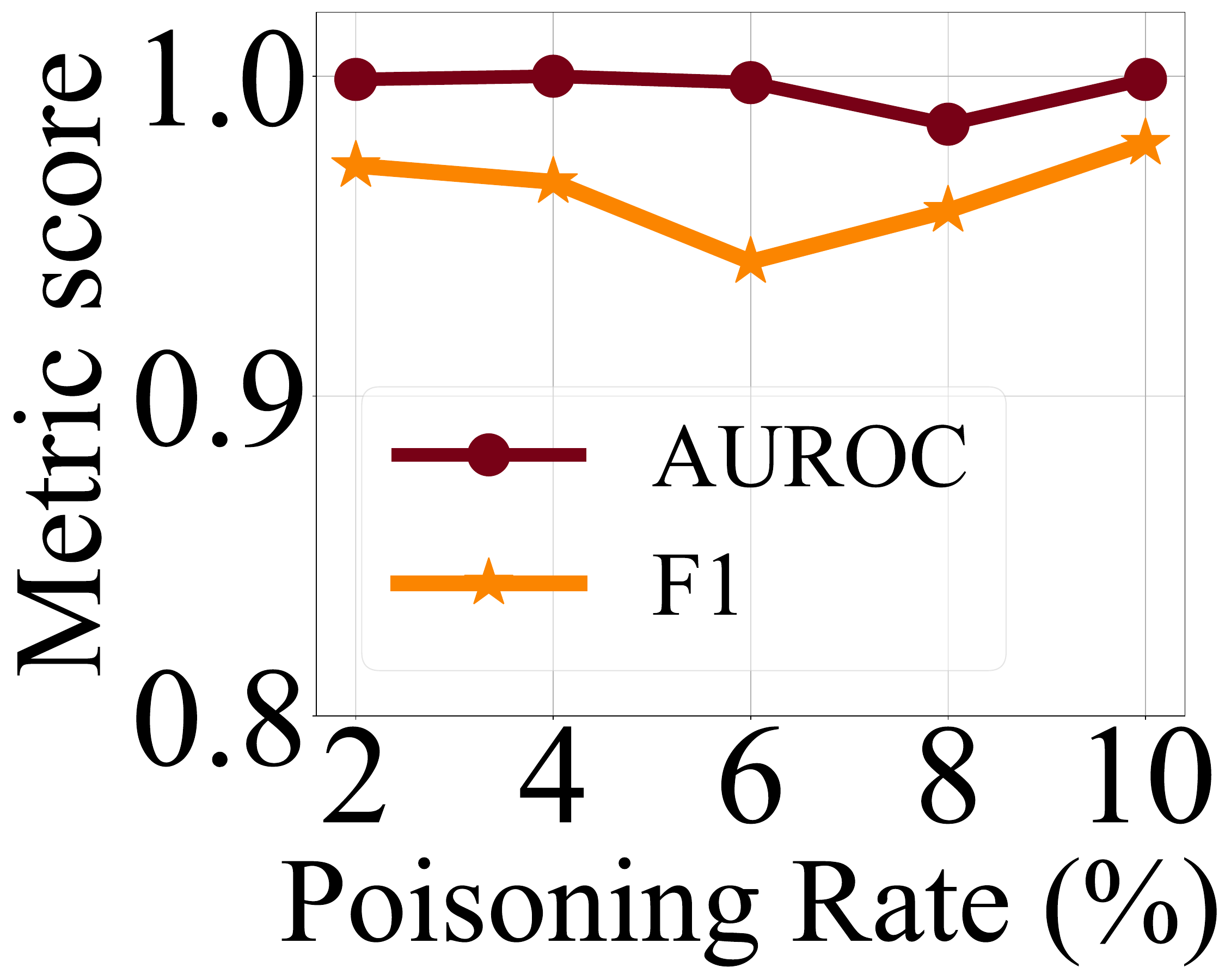}
            \centerline{(c) BATT}
    \end{minipage}
\end{minipage}
\vspace{-0.5em}
 \caption{The impact of poisoning rate on CIFAR-10.} 
 \label{fig:poisoningratio}
 \vspace{-1.5em}
\end{figure}

\subsection{Resistance to Potential Adaptive Attacks} 
\label{sec: ada}
We initially assess the performance of IBD-PSC against attacks with low poisoning rates. This is because a small poisoning rate $\rho$ can prevent models from over-fitting triggers, thus weakening the association between triggers and target labels, as demonstrated in previous studies~\cite{guo2023scale,qi2023revisiting}. Specifically, we conduct attacks (BadNets, WaNet, and BATT) on the CIFAR-10 dataset with $\rho$ ranging from 0.02 to 0.1, ensuring the attack success rates exceed 80\%. The results in~\cref{fig:poisoningratio} consistently demonstrate the effectiveness of IBD-PSC, with AUROC and F1 scores consistently above 0.98 and 0.95, respectively. Results on SubImageNet-200 are shown in~\cref{appendix:small_rate_defense}.

\begin{table}[!t]
\centering
\vspace{-0.8em}
\small
\caption{The False Positive Rate (FPR) (\%) of SCALE-UP and our defense on target and benign classes on the CIFAR-10 dataset.}
\label{tab:FPR_target}
\setlength{\tabcolsep}{2pt}
\scalebox{0.95}{
\begin{tabular}{lcccccccccccccc} 
\toprule
 Defenses$\rightarrow$ & \multicolumn{2}{c}{SCALE-UP} & \multicolumn{2}{c}{Ours-L} & \multicolumn{2}{c}{Ours} \\
 \cmidrule(lr){2-3}  \cmidrule(lr){4-5} \cmidrule(lr){6-7}
 AttackS$\downarrow$ & Target & Benign & Target & Benign & Target & Benign \\ 
\midrule
BadNets &72.74	&29.00	&0.40	&9.76&	0.20	&1.88\\
Blend&	54.28&	19.80&	22.55	&3.39&	18.34&	2.64\\
PhysicalBA	&90.58	&23.98	&4.60&	5.42&	4.10&	1.50\\
WaNet&	76.70&	28.11	&81.41&	10.05&	69.20&	8.16\\
ISSBA	&93.93	&20.70&	20.94	&3.00&	17.22&	0.61\\
BATT	&57.74	&18.78&	2.35	&9.72&	0.87&	6.90\\
SRA&	65.55	&29.33	&0.62	&10.48	&0.50	&10.13\\
Ada-Patch	&93.80	&25.77	&8.67&	4.78&	4.34&	3.00\\
\bottomrule
\end{tabular}
}
\vspace{-1em}
\end{table}

\begin{table}[!t]
\centering
\caption{Performance of IBD-PSC under adaptive attacks. }
\label{tab:ada}
\tiny
\setlength{\tabcolsep}{4pt}
\begin{tabular}{lcccccccccccccc} 
\toprule
 $\alpha$$\rightarrow$ & \multicolumn{2}{c}{0.2} & \multicolumn{2}{c}{0.5} & \multicolumn{2}{c}{0.9} & \multicolumn{2}{c}{0.99} \\
  \cmidrule{2-9}
 Attacks$\downarrow$ & AUROC & F1 & AUROC & F1 & AUROC & F1 & AUROC & F1  \\ 
\midrule
BadNets & 0.992& 0.978& 0.986& 0.964& 0.995& 0.962& 0.996&0.951 \\
WaNet &0.947 & 0.949&0.956 &0.942 & 0.931&  0.927& 0.819& 0.862\\ 
BATT & 0.986& 0.968& 0.994& 0.956& 0.982& 0.975 & 0.979& 0.959\\
\bottomrule
\end{tabular}
\vspace{-1.3em}
\end{table}

\begin{figure*}[!t]
	\centering
 \begin{minipage}{0.99\linewidth}
    \begin{minipage}{0.245\linewidth}
            \includegraphics[width=1\linewidth]{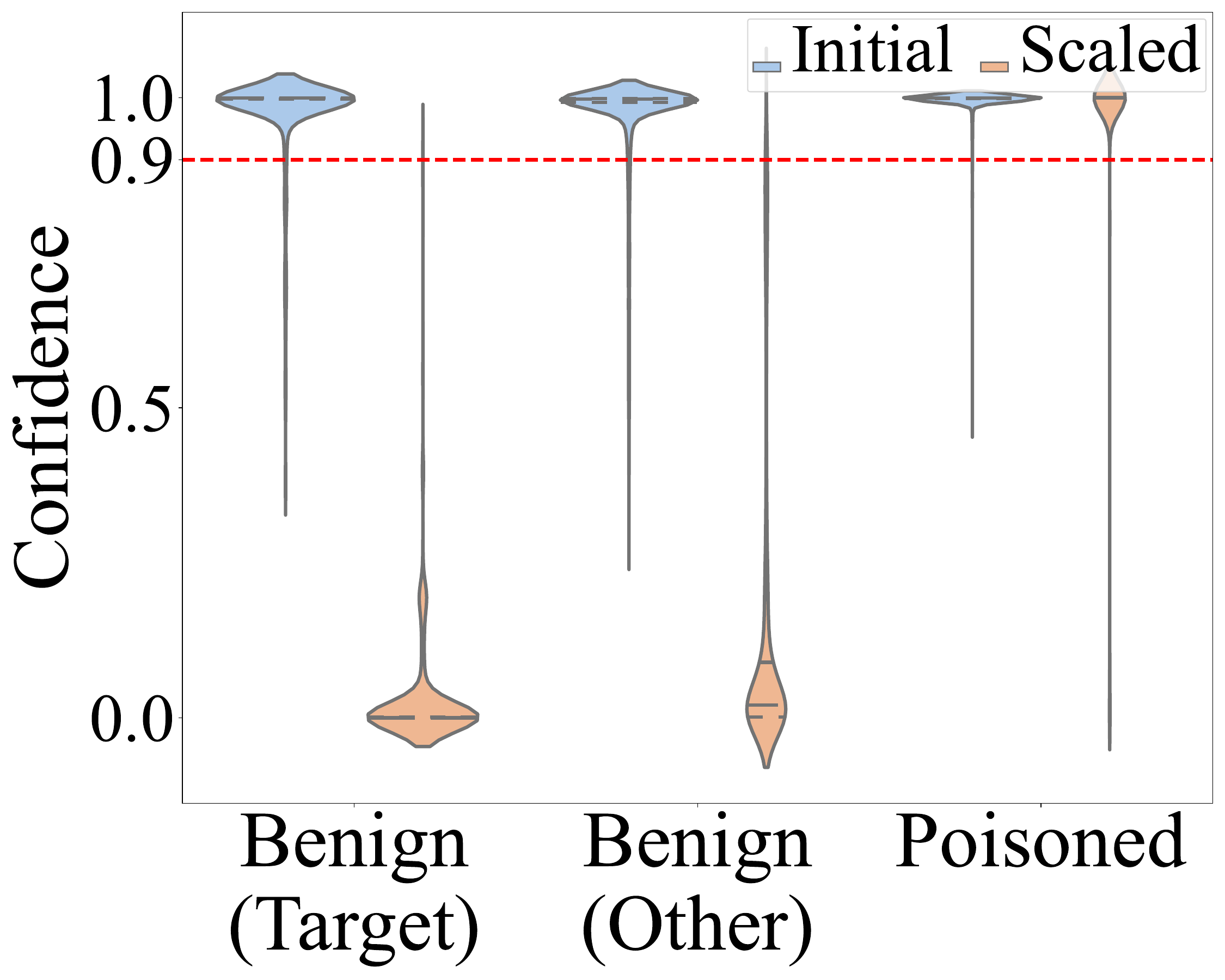}
            \centerline{(a) BadNets}
    \end{minipage}
    \begin{minipage}{0.245\linewidth}
            \includegraphics[width=1\linewidth]{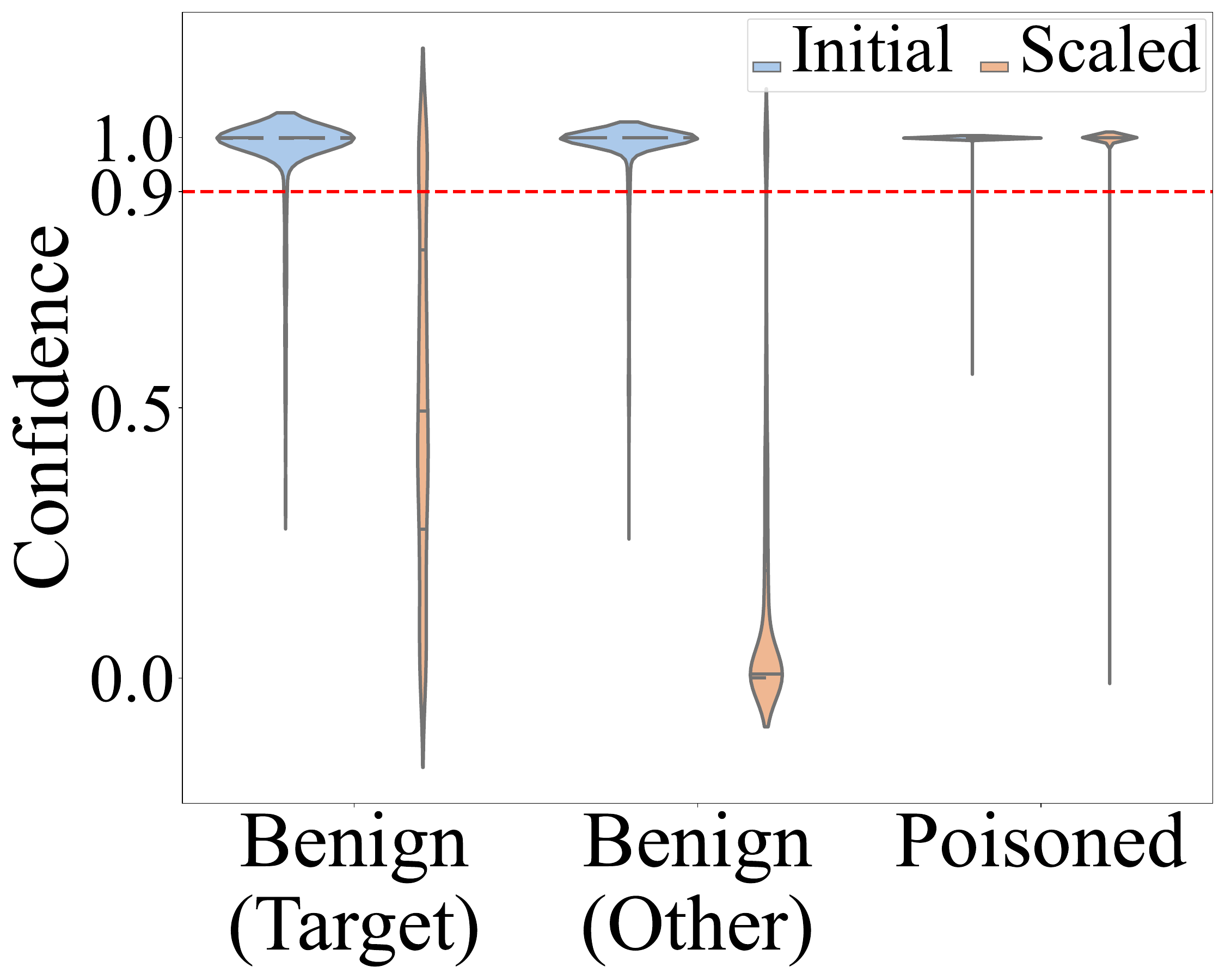}
            \centerline{(b) Blend}
    \end{minipage}
        \begin{minipage}{0.245\linewidth}
            \includegraphics[width=1\linewidth]{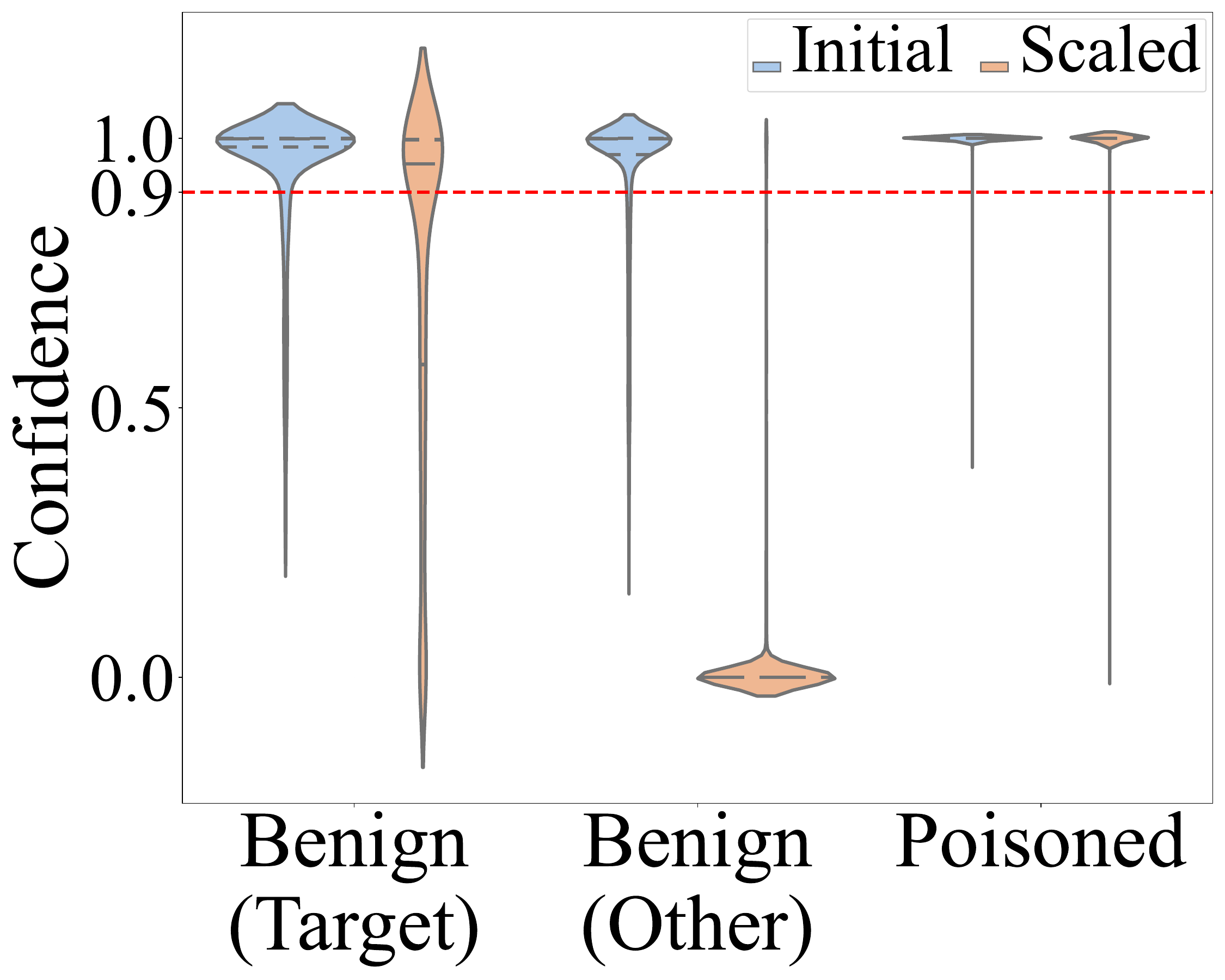}
            \centerline{(b) BATT}
    \end{minipage}
    \begin{minipage}{0.245\linewidth}
            \includegraphics[width=1\linewidth]{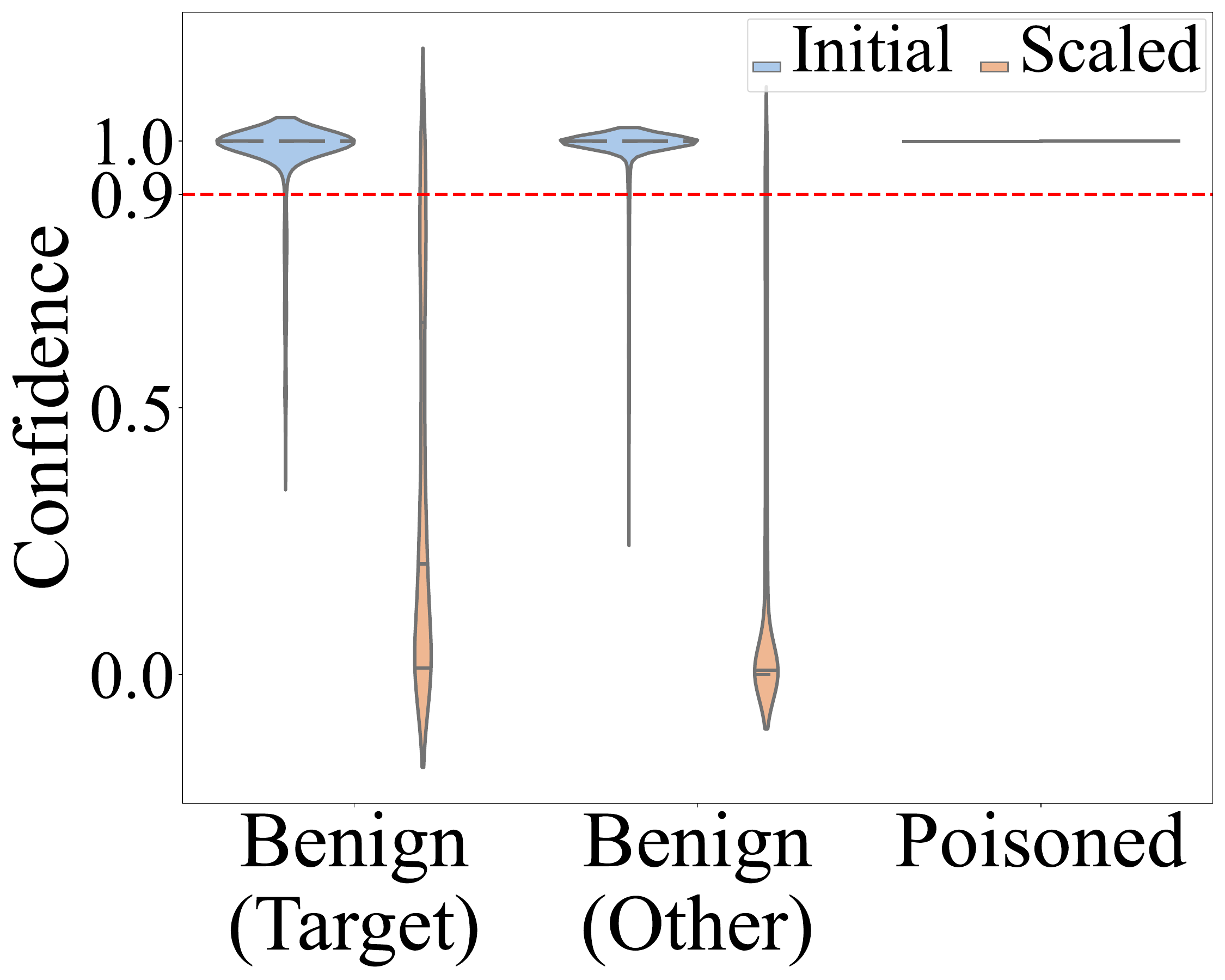}
            \centerline{(f) Ada-patch}
    \end{minipage}
    \end{minipage}
\vspace{-0.5em}
 \caption{The violin plots of the prediction confidences for benign samples in the target and other classes, as well as for poisoned samples, as predicted by the initial and scaled models on CIFAR-10. The threshold is 0.9.}
 \label{fig:violin-main}
 \vspace{-1.2em}
\end{figure*}

We further evaluate the robustness of IBD-PSC against potential adaptive attacks in the worst-case scenario, where adversaries possess complete knowledge of our defense. Typically, a vanilla backdoored model functions normally with benign samples but yields adversary-specific predictions when exposed to poisoned samples. The loss function for training such backdoored models is defined as follows:
\begin{equation}
\vspace{-0.5em}
\label{eq:bd}
      \mathcal{L}_{\textrm{bd}} = \sum_{i=1}^{|\mathcal{D}_b|} \mathcal{L}(\mathcal{F}(\vx_i), y_i) +\sum_{j=1}^{|\mathcal{D}_p|}  \mathcal{L}(\mathcal{F}(\vx_j), y_t),
\end{equation}
where $\mathcal{L}(\cdot)$ represents the cross entropy loss function.

We design an adaptive loss term $\mathcal{L}_{\textrm{ada}}$ to ensure benign samples are correctly predicted under parameter amplification:
\begin{equation}
\label{eq:adapt}
 \vspace{-0.5em}
       \mathcal{L}_{\textrm{ada}} = \sum_{i=1}^{|\mathcal{D}_b|} \mathcal{L}(\hat{\mathcal{F}_k^{\omega}}(\vx_{i};\hat{\vtheta}),y_i).
\end{equation}
Subsequently, we integrate this adaptive loss $\mathcal{L}_{\textrm{ada}}$ with the vanilla loss $\mathcal{L}_{\textrm{bd}}$ to formulate the overall loss function as $\mathcal{L} = \alpha \mathcal{L}_{\textrm{bd}} + (1-\alpha) \mathcal{L}_{\textrm{ada}}$, where $\alpha$ is a weighting factor. We then optimize the original model's parameters $\vtheta$ by minimizing $\mathcal{L}$ during the training phase.

Similar to previous experiments, we also employ the three representative backdoor attacks to develop adaptive attacks on the CIFAR-10 dataset.~\cref{tab:ada} demonstrates the sustained robustness of our IBD-PSC across all cases. The effectiveness primarily originates from our adaptive layer selection strategy, which dynamically identifies BN layers for amplification, regardless of whether it is a vanilla or an adaptive backdoored model. The layers selected during the inference stage typically differ from those used in the training phase, enabling the IBD-PSC to effectively detect poisoned samples. More results and the resistance to another adaptive attack can be found in~\cref{appendix:ada}.

\subsection{Performance on Benign Samples from Target Class}
In this section, we evaluate the effectiveness of our defense on benign samples from the target class. We conduct experiments on the CIFAR-10 dataset against four different attacks and present the confidences of both the initial backdoored model and the scaled models in~\cref{fig:violin-main}. As shown, the confidences of benign samples from both the target class and other classes decrease due to parameter amplification, falling below the threshold. In contrast, the confidence values for poisoned samples mostly remain above this threshold. These results demonstrate that our defense effectively distinguishes between benign and poisoned samples, regardless of whether the benign sample originates from the target class. In particular, we observe an interesting phenomenon that scaled models tend to cluster the confidences for benign samples from the target class in the more difficult-to-learn class(es), rather than in the easier ones, which is unexpected. We will further explore its intrinsic mechanism in our future work. Additional analysis can be found in~\cref{appendix:target_benign}.

\subsection{A Closer Look to the Effectiveness of our Method}
To gain deeper insights, we delve into the mechanisms of both SCALE-UP and our IBD-PSC. We utilize t-SNE~\cite{van2008visualizing} for visualizing the features of benign and poisoned samples in the last hidden layer. We adopt the representative BadNets attack method on the CIFAR-10 dataset as an example for our discussions. More results about other attack methods can be found in~\cref{appendix:closer}. The results in~\cref{fig:cluster} demonstrate that both SCALE-UP and our IBD-PSC induce more significant shifts in the feature space for benign samples compared to the poisoned samples. These larger shifts result in changes in the predictions for benign samples. These results provide clear evidence of the effectiveness of the two defense methods. Furthermore, in contrast to SCALE-UP, our IBD-PSC method induces more significant shifts in benign samples. This disparity in shift magnitude may stem from the constrained pixel value range of [0, 255], potentially mitigating the impact of amplification. However, the values of model parameters do not have such bounded constraints. Consequently, the larger shifts contribute to a more distinct separation between benign and poisoned samples, significantly augmenting the effectiveness of IBD-PSC.

\subsection{The Extension to Training Set Purification}
Although our method is initially and primarily designed to filter malicious testing samples, it can also be used to detect potentially poisoned samples within a compromised training set. Specifically, users can first train a model on this dataset with a standard process and then exploit our detection method. To verify our effectiveness, we conduct experiments on the CIFAR-10 dataset against three representative attacks. The results show a 100\% TPR and nearly 100\% AUROC scores, with FPR scores close to 0\%. We compare the detection performance of our method with the most advanced defenses, \ie, CD~\cite{huang2023distilling} and MSPC~\cite{pal2024backdoor}, and the results show that our method achieves the best detection performance. Subsequently, we retrain a model on this purified dataset to evaluate both its BA and the ASR. The ASR scores of these retrained models are less than 0.5\%, rendering the attacks ineffective. More settings and results can be found in~\cref{appendix:training_set}. 

\begin{figure}[!t]
	\centering
 \begin{minipage}{0.44\linewidth}
    \includegraphics[width=1\linewidth]{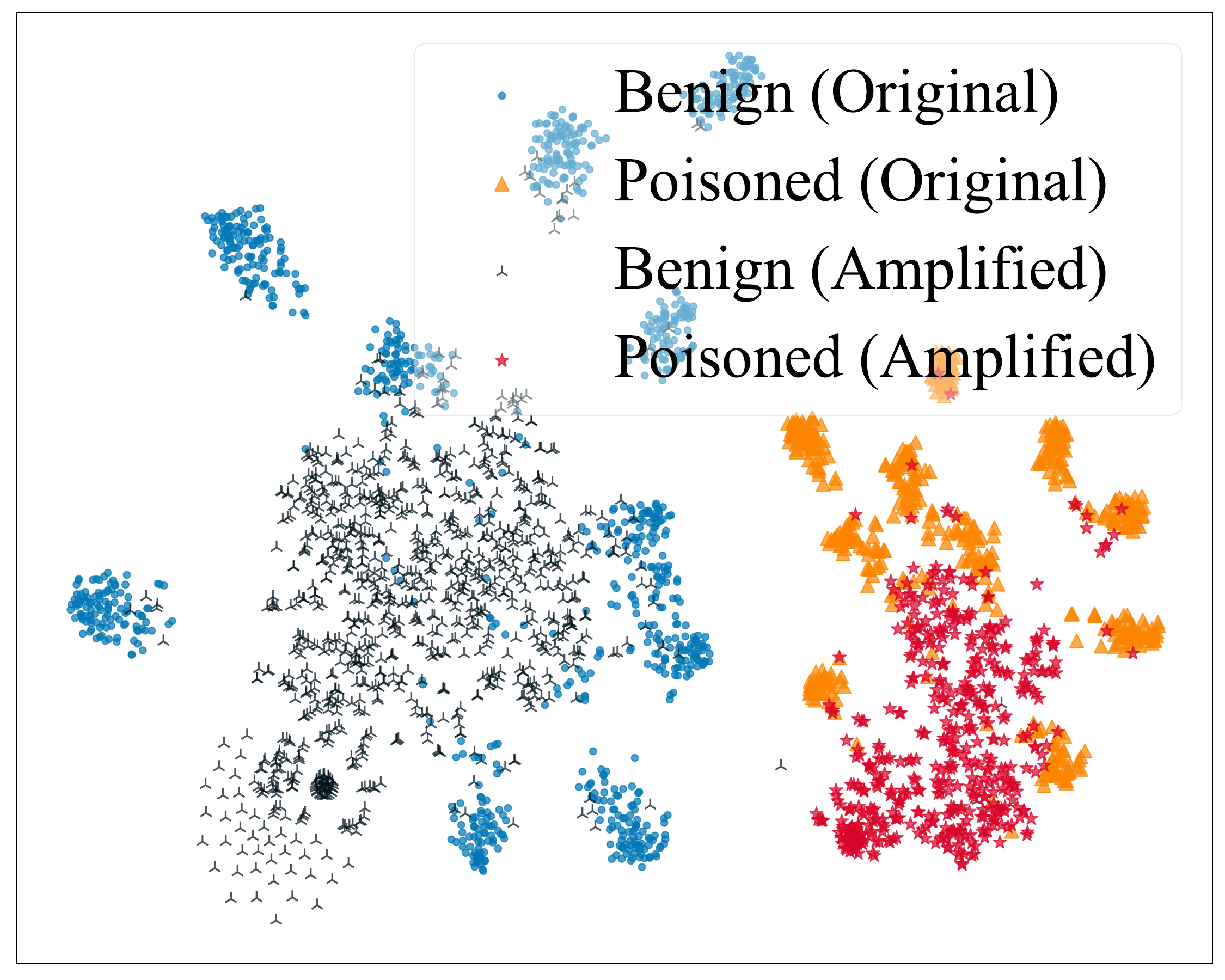}
     \centerline{(a) SCALE-UP}
\end{minipage}\hspace{0.5em}
 \begin{minipage}{0.44\linewidth}
    \includegraphics[width=1\linewidth]{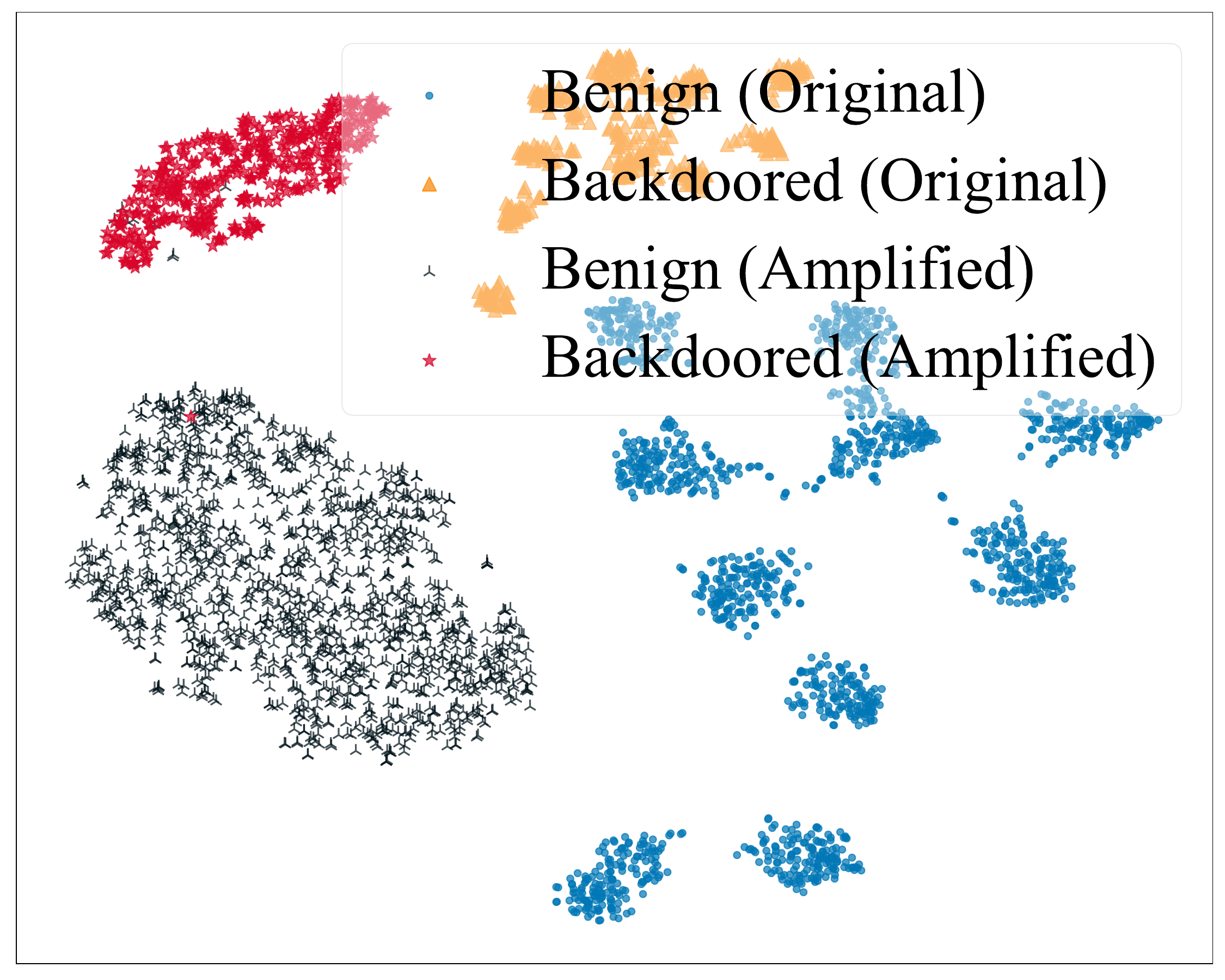}
    \centerline{(b) Ours}
\end{minipage}
\vspace{-0.3em}
 \caption{t-SNE of feature representations of benign and poisoned samples on the CIFAR-10 dataset against BadNets attack.} 
 \label{fig:cluster}
  \vspace{-1em}
 \end{figure}
\section{Conclusion}
In this paper, we proposed a simple yet effective method (dubbed IBD-PSC) for determining whether a suspicious image is poisoned. The IBD-PSC was inspired by our discovery of an intriguing phenomenon, named parameter-oriented scaled consistency (PSC). This phenomenon manifests through a significant uniformity of prediction confidences for poisoned samples, in contrast to benign ones, when the parameters of selected BN layers undergo amplification. We provided the theoretical and empirical foundations to support this phenomenon. To enhance the detection performance, we also designed an adaptive algorithm to dynamically select the number of BN layers for amplification. We conducted thirteen backdoor attack methods on benchmark datasets to comprehensively verify the effectiveness of our IBD-PSC. We also demonstrated that our IBD-PSC is highly efficient and resistant to potential adaptive attacks.

\newpage

\section*{Acknowledgement}
This work was supported in part by the National Natural Science Foundation of China under Grants 62071142 and by the Guangdong Basic and Applied Basic Research Foundation under Grant 2024A1515012299.

\section*{Impact Statement}

Backdoor attacks have posed severe threats in DNNs since developers often rely on external untrustworthy training resources (\eg, datasets and model backbones). This paper proposes a simple yet effective input-level backdoor detection to identify and filter malicious testing samples. It generally has no ethical issues since it does not expose new vulnerabilities within DNNs and is purely defensive. However, we need to notice that our work can only filter out poisoned input images but cannot repair potential backdoors in the deployed model. Besides, it cannot recover trigger patterns or the ground-truth class of the poisoned samples. People should not be too optimistic about eliminating backdoor threats. Moreover, the adversaries may design more advanced backdoor attacks against our defense, although we have demonstrated that it is challenging. People should use only trusted training resources and models to eliminate and prevent backdoor attacks at the source.

\bibliography{example_paper}

\begin{thebibliography}{74}
\providecommand{\natexlab}[1]{#1}
\providecommand{\url}[1]{\texttt{#1}}
\expandafter\ifx\csname urlstyle\endcsname\relax
  \providecommand{\doi}[1]{doi: #1}\else
  \providecommand{\doi}{doi: \begingroup \urlstyle{rm}\Url}\fi

\bibitem[Bai et~al.(2021)Bai, Wu, Zhang, Li, Li, and Xia]{bai2021targeted}
Bai, J., Wu, B., Zhang, Y., Li, Y., Li, Z., and Xia, S.-T.
\newblock {Targeted Attack against Deep Neural Networks via Flipping Limited Weight Bits}.
\newblock In \emph{ICLR}, 2021.

\bibitem[Chen et~al.(2018)Chen, Carvalho, Baracaldo, Ludwig, Edwards, Lee, Molloy, and Srivastava]{chen2018detecting}
Chen, B., Carvalho, W., Baracaldo, N., Ludwig, H., Edwards, B., Lee, T., Molloy, I., and Srivastava, B.
\newblock Detecting backdoor attacks on deep neural networks by activation clustering.
\newblock In \emph{CEUR Workshop}, 2018.

\bibitem[Chen et~al.(2022)Chen, Wu, and Wang]{chen2022effective}
Chen, W., Wu, B., and Wang, H.
\newblock Effective backdoor defense by exploiting sensitivity of poisoned samples.
\newblock In \emph{NeurIPS}, 2022.

\bibitem[Chen et~al.(2017)Chen, Liu, Li, Lu, and Song]{chen2017targeted}
Chen, X., Liu, C., Li, B., Lu, K., and Song, D.
\newblock {Targeted Backdoor Attacks on Deep Learning Systems Using Data Poisoning}.
\newblock \emph{arXiv}, 2017.

\bibitem[Chou et~al.(2020)Chou, Tram{\`e}r, Pellegrino, and Boneh]{chou2018sentinet}
Chou, E., Tram{\`e}r, F., Pellegrino, G., and Boneh, D.
\newblock Senti{N}et: Detecting physical attacks against deep learning systems.
\newblock In \emph{IEEE S\&P Workshop}, 2020.

\bibitem[Deng et~al.(2009)Deng, Dong, Socher, Li, Li, and Fei-Fei]{deng2009imagenet}
Deng, J., Dong, W., Socher, R., Li, L.-J., Li, K., and Fei-Fei, L.
\newblock {ImageNet: A large-scale hierarchical image database}.
\newblock In \emph{CVPR}, 2009.

\bibitem[Doan et~al.(2021)Doan, Lao, and Li]{doan2021backdoor}
Doan, K., Lao, Y., and Li, P.
\newblock {Backdoor Attack with Imperceptible Input and Latent Modification}.
\newblock In \emph{NeurIPS}, 2021.

\bibitem[Duan et~al.(2024)Duan, Hua, Liao, Zhang, and Zhang]{duan2024conditional}
Duan, Q., Hua, Z., Liao, Q., Zhang, Y., and Zhang, L.~Y.
\newblock Conditional backdoor attack via jpeg compression.
\newblock In \emph{AAAI}, 2024.

\bibitem[Gao et~al.(2021)Gao, Kim, Doan, Zhang, Zhang, Nepal, Ranasinghe, and Kim]{gao2021design}
Gao, Y., Kim, Y., Doan, B.~G., Zhang, Z., Zhang, G., Nepal, S., Ranasinghe, D.~C., and Kim, H.
\newblock {Design and Evaluation of a Multi-Domain Trojan Detection Method on Deep Neural Networks}.
\newblock \emph{IEEE Transactions on Dependable and Secure Computing}, 2021.

\bibitem[Gao et~al.(2023)Gao, Li, Zhu, Wu, Jiang, and Xia]{gao2023not}
Gao, Y., Li, Y., Zhu, L., Wu, D., Jiang, Y., and Xia, S.-T.
\newblock {Not all samples are born equal: Towards effective clean-label backdoor attacks}.
\newblock \emph{PR}, 2023.

\bibitem[Gong et~al.(2023)Gong, Wang, Chen, Xue, Wang, and Shen]{gong2023kaleidoscope}
Gong, X., Wang, Z., Chen, Y., Xue, M., Wang, Q., and Shen, C.
\newblock {Kaleidoscope: Physical Backdoor Attacks against Deep Neural Networks with RGB Filters}.
\newblock \emph{IEEE Transactions on Dependable and Secure Computing}, 2023.

\bibitem[Gu et~al.(2017)Gu, Dolan{-}Gavitt, and Garg]{gu2017badnets}
Gu, T., Dolan{-}Gavitt, B., and Garg, S.
\newblock {BadNets: Identifying Vulnerabilities in the Machine Learning Model Supply Chain}.
\newblock \emph{IEEE Access}, 2017.

\bibitem[Guo et~al.(2024)Guo, Lu, Bao, Pang, Yan, Du, and Li]{guo2024gaussian}
Guo, H., Lu, C., Bao, F., Pang, T., Yan, S., Du, C., and Li, C.
\newblock Gaussian mixture solvers for diffusion models.
\newblock In \emph{NeurIPS}, 2024.

\bibitem[Guo et~al.(2023{\natexlab{a}})Guo, Li, Wang, and Liu]{guo2023policycleanse}
Guo, J., Li, A., Wang, L., and Liu, C.
\newblock {PolicyCleanse: Backdoor Detection and Mitigation in Reinforcement Learning}.
\newblock In \emph{ICCV}, 2023{\natexlab{a}}.

\bibitem[Guo et~al.(2023{\natexlab{b}})Guo, Li, Chen, Guo, Sun, and Liu]{guo2023scale}
Guo, J., Li, Y., Chen, X., Guo, H., Sun, L., and Liu, C.
\newblock {SCALE-UP: An efficient black-box input-level backdoor detection via analyzing scaled prediction consistency}.
\newblock In \emph{ICLR}, 2023{\natexlab{b}}.

\bibitem[Guo et~al.(2023{\natexlab{c}})Guo, Li, Wang, Xia, Huang, Liu, and Li]{guo2023domain}
Guo, J., Li, Y., Wang, L., Xia, S.-T., Huang, H., Liu, C., and Li, B.
\newblock Domain watermark: Effective and harmless dataset copyright protection is closed at hand.
\newblock In \emph{NeurIPS}, 2023{\natexlab{c}}.

\bibitem[Hayase et~al.(2021)Hayase, Kong, Somani, and Oh]{hayase2021spectre}
Hayase, J., Kong, W., Somani, R., and Oh, S.
\newblock Spectre: Defending against backdoor attacks using robust statistics.
\newblock In \emph{ICML}, 2021.

\bibitem[He et~al.(2016{\natexlab{a}})He, Zhang, Ren, and Sun]{he2016deep}
He, K., Zhang, X., Ren, S., and Sun, J.
\newblock {Deep Residual Learning for Image Recognition}.
\newblock In \emph{CVPR}, 2016{\natexlab{a}}.

\bibitem[He et~al.(2016{\natexlab{b}})He, Zhang, Ren, and Sun]{he2016identity}
He, K., Zhang, X., Ren, S., and Sun, J.
\newblock {Identity Mappings in Deep Residual Networks}.
\newblock In \emph{ICCV}, 2016{\natexlab{b}}.

\bibitem[Huang et~al.(2023)Huang, Ma, Erfani, and Bailey]{huang2023distilling}
Huang, H., Ma, X., Erfani, S., and Bailey, J.
\newblock Distilling cognitive backdoor patterns within an image.
\newblock In \emph{ICLR}, 2023.

\bibitem[Huang et~al.(2022)Huang, Li, Wu, Qin, and Ren]{huang2022backdoor}
Huang, K., Li, Y., Wu, B., Qin, Z., and Ren, K.
\newblock {Backdoor Defense via Decoupling the Training Process}.
\newblock In \emph{ICLR}, 2022.

\bibitem[Jebreel et~al.(2023)Jebreel, Domingo-Ferrer, and Li]{jebreel2023defending}
Jebreel, N.~M., Domingo-Ferrer, J., and Li, Y.
\newblock {Defending Against Backdoor Attacks by Layer-wise Feature Analysis}.
\newblock In \emph{SIGKDD}, 2023.

\bibitem[Krizhevsky et~al.(2009)Krizhevsky, Hinton, et~al.]{krizhevsky2009learning}
Krizhevsky, A., Hinton, G., et~al.
\newblock {Learning Multiple Layers of Features from Tiny Images}.
\newblock \emph{Technical report}, 2009.

\bibitem[Li et~al.(2024{\natexlab{a}})Li, Cai, Cai, Li, Qiu, Wang, and Zhang]{li2024purifying}
Li, B., Cai, Y., Cai, J., Li, Y., Qiu, H., Wang, R., and Zhang, T.
\newblock Purifying quantization-conditioned backdoors via layer-wise activation correction with distribution approximation.
\newblock In \emph{ICML}, 2024{\natexlab{a}}.

\bibitem[Li et~al.(2024{\natexlab{b}})Li, Cai, Li, Xue, Li, and Li]{li2024nearest}
Li, B., Cai, Y., Li, H., Xue, F., Li, Z., and Li, Y.
\newblock Nearest is not dearest: Towards practical defense against quantization-conditioned backdoor attacks.
\newblock In \emph{CVPR}, 2024{\natexlab{b}}.

\bibitem[Li et~al.(2020)Li, Xue, Zhao, Zhu, and Zhang]{li2020invisible}
Li, S., Xue, M., Zhao, B. Z.~H., Zhu, H., and Zhang, X.
\newblock {Invisible Backdoor Attacks on Deep Neural Networks via Steganography and Regularization}.
\newblock \emph{IEEE Transactions on Dependable and Secure Computing}, 2020.

\bibitem[Li et~al.(2021{\natexlab{a}})Li, Li, Wu, Li, He, and Lyu]{li2021invisible}
Li, Y., Li, Y., Wu, B., Li, L., He, R., and Lyu, S.
\newblock {Invisible Backdoor Attack with Sample-Specific Triggers}.
\newblock In \emph{ICCV}, 2021{\natexlab{a}}.

\bibitem[Li et~al.(2021{\natexlab{b}})Li, Lyu, Koren, Lyu, Li, and Ma]{li2021anti}
Li, Y., Lyu, X., Koren, N., Lyu, L., Li, B., and Ma, X.
\newblock {Anti-Backdoor Learning: Training Clean Models on Poisoned Data}.
\newblock In \emph{NeurIPS}, 2021{\natexlab{b}}.

\bibitem[Li et~al.(2021{\natexlab{c}})Li, Zhai, Jiang, Li, and Xia]{li2021backdoor}
Li, Y., Zhai, T., Jiang, Y., Li, Z., and Xia, S.-T.
\newblock {Backdoor Attack in the Physical World}.
\newblock In \emph{ICLR Workshop}, 2021{\natexlab{c}}.

\bibitem[Li et~al.(2022{\natexlab{a}})Li, Bai, Jiang, Yang, Xia, and Li]{li2022untargeted}
Li, Y., Bai, Y., Jiang, Y., Yang, Y., Xia, S.-T., and Li, B.
\newblock Untargeted backdoor watermark: Towards harmless and stealthy dataset copyright protection.
\newblock In \emph{NeurIPS}, 2022{\natexlab{a}}.

\bibitem[Li et~al.(2022{\natexlab{b}})Li, Jiang, Li, and Xia]{li2022backdoor}
Li, Y., Jiang, Y., Li, Z., and Xia, S.-T.
\newblock {Backdoor Learning: A Survey}.
\newblock \emph{IEEE Transactions on Neural Networks and learning systems}, 2022{\natexlab{b}}.

\bibitem[Li et~al.(2022{\natexlab{c}})Li, Zhu, Jia, Jiang, Xia, and Cao]{li2022defending1}
Li, Y., Zhu, L., Jia, X., Jiang, Y., Xia, S.-T., and Cao, X.
\newblock Defending against model stealing via verifying embedded external features.
\newblock In \emph{AAAI}, 2022{\natexlab{c}}.

\bibitem[Li et~al.(2023{\natexlab{a}})Li, Mengxi, Yang, Yong, and Shu-Tao]{backdoorbox}
Li, Y., Mengxi, Y., Yang, B., Yong, J., and Shu-Tao, X.
\newblock {BackdoorBox: A Python Toolbox for Backdoor Learning}.
\newblock In \emph{ICLR Workshop}, 2023{\natexlab{a}}.

\bibitem[Li et~al.(2023{\natexlab{b}})Li, Zhu, Yang, Jiang, Wei, and Xia]{li2023black}
Li, Y., Zhu, M., Yang, X., Jiang, Y., Wei, T., and Xia, S.-T.
\newblock Black-box dataset ownership verification via backdoor watermarking.
\newblock \emph{IEEE Transactions on Information Forensics and Security}, 2023{\natexlab{b}}.

\bibitem[Liu et~al.(2018)Liu, Dolan-Gavitt, and Garg]{liu2018fine}
Liu, K., Dolan-Gavitt, B., and Garg, S.
\newblock {Fine-Pruning: Defending Against Backdooring Attacks on Deep Neural Networks}.
\newblock In \emph{RAID}, 2018.

\bibitem[Liu et~al.(2023)Liu, Li, Wang, Hu, Ye, Jin, Wu, and Xiao]{liu2023detecting}
Liu, X., Li, M., Wang, H., Hu, S., Ye, D., Jin, H., Wu, L., and Xiao, C.
\newblock {Detecting Backdoors During the Inference Stage Based on Corruption Robustness Consistency}.
\newblock In \emph{CVPR}, 2023.

\bibitem[Loureiro et~al.(2021)Loureiro, Sicuro, Gerbelot, Pacco, Krzakala, and Zdeborov{\'a}]{loureiro2021learning}
Loureiro, B., Sicuro, G., Gerbelot, C., Pacco, A., Krzakala, F., and Zdeborov{\'a}, L.
\newblock Learning gaussian mixtures with generalized linear models: Precise asymptotics in high-dimensions.
\newblock In \emph{NeurIPS}, 2021.

\bibitem[Ma et~al.(2022)Ma, Wang, Sun, Xue, Wen, and Xiang]{ma2022beatrix}
Ma, W., Wang, D., Sun, R., Xue, M., Wen, S., and Xiang, Y.
\newblock The" beatrix''resurrections: Robust backdoor detection via gram matrices.
\newblock In \emph{NDSS}, 2022.

\bibitem[Mo et~al.(2024)Mo, Zhang, Zhang, Luo, Sun, Hu, Gao, and Xiang]{mo2023robust}
Mo, X., Zhang, Y., Zhang, L.~Y., Luo, W., Sun, N., Hu, S., Gao, S., and Xiang, Y.
\newblock Robust backdoor detection for deep learning via topological evolution dynamics.
\newblock \emph{IEEE S\&P}, 2024.

\bibitem[Nguyen \& Tran(2020)Nguyen and Tran]{nguyen2020input}
Nguyen, T.~A. and Tran, A.
\newblock {Input-Aware Dynamic Backdoor Attack}.
\newblock In \emph{NeurIPS}, 2020.

\bibitem[Nguyen \& Tran(2021)Nguyen and Tran]{nguyen2021wanet}
Nguyen, T.~A. and Tran, A.~T.
\newblock {WaNet -- Imperceptible Warping-based Backdoor Attack}.
\newblock In \emph{ICLR}, 2021.

\bibitem[Pal et~al.(2024)Pal, Yao, Wang, Shen, and Liu]{pal2024backdoor}
Pal, S., Yao, Y., Wang, R., Shen, B., and Liu, S.
\newblock Backdoor secrets unveiled: Identifying backdoor data with optimized scaled prediction consistency.
\newblock In \emph{ICLR}, 2024.

\bibitem[Pan et~al.(2023)Pan, Zeng, Lyu, Lin, and Jia]{pan2023asset}
Pan, M., Zeng, Y., Lyu, L., Lin, X., and Jia, R.
\newblock $\{$ASSET$\}$: Robust backdoor data detection across a multiplicity of deep learning paradigms.
\newblock In \emph{USENIX Security}, 2023.

\bibitem[Papyan et~al.(2020)Papyan, Han, and Donoho]{papyan2020prevalence}
Papyan, V., Han, X., and Donoho, D.~L.
\newblock Prevalence of neural collapse during the terminal phase of deep learning training.
\newblock \emph{PNAS}, 2020.

\bibitem[Peri et~al.(2020)Peri, Gupta, Huang, Fowl, Zhu, Feizi, Goldstein, and Dickerson]{peri2020deep}
Peri, N., Gupta, N., Huang, W.~R., Fowl, L., Zhu, C., Feizi, S., Goldstein, T., and Dickerson, J.~P.
\newblock Deep k-nn defense against clean-label data poisoning attacks.
\newblock In \emph{ECCV}, 2020.

\bibitem[Qi et~al.(2022)Qi, Xie, Pan, Zhu, Yang, and Bu]{qi2022towards}
Qi, X., Xie, T., Pan, R., Zhu, J., Yang, Y., and Bu, K.
\newblock {Towards Practical Deployment-Stage Backdoor Attack on Deep Neural Networks}.
\newblock In \emph{CVPR}, 2022.

\bibitem[Qi et~al.(2023)Qi, Xie, Li, Mahloujifar, and Mittal]{qi2023revisiting}
Qi, X., Xie, T., Li, Y., Mahloujifar, S., and Mittal, P.
\newblock {Revisiting the Assumption of Latent Separability for Backdoor Defenses}.
\newblock In \emph{ICLR}, 2023.

\bibitem[Stallkamp et~al.(2012)Stallkamp, Schlipsing, Salmen, and Igel]{stallkamp2012man}
Stallkamp, J., Schlipsing, M., Salmen, J., and Igel, C.
\newblock {Man vs. computer: Benchmarking machine learning algorithms for traffic sign recognition}.
\newblock \emph{Neural Networks}, 2012.

\bibitem[Tang et~al.(2021)Tang, Wang, Tang, and Zhang]{tang2021demon}
Tang, D., Wang, X., Tang, H., and Zhang, K.
\newblock {Demon in the Variant: Statistical Analysis of DNNs for Robust Backdoor Contamination Detection}.
\newblock In \emph{USENIX Security}, 2021.

\bibitem[Tang et~al.(2020)Tang, Du, Liu, Yang, and Hu]{tang2020embarrassingly}
Tang, R., Du, M., Liu, N., Yang, F., and Hu, X.
\newblock {An Embarrassingly Simple Approach for Trojan Attack in Deep Neural Networks}.
\newblock In \emph{SIGKDD}, 2020.

\bibitem[Tang et~al.(2023)Tang, Yuan, Li, Liu, Chen, and Hu]{tang2023setting}
Tang, R., Yuan, J., Li, Y., Liu, Z., Chen, R., and Hu, X.
\newblock {Setting the Trap: Capturing and Defeating Backdoor Threats in PLMs through Honeypots}.
\newblock In \emph{NeurIPS}, 2023.

\bibitem[Tishby \& Zaslavsky(2015)Tishby and Zaslavsky]{tishby2015deep}
Tishby, N. and Zaslavsky, N.
\newblock {Deep Learning and the Information Bottleneck Principle}.
\newblock In \emph{ITW}, 2015.

\bibitem[Tran et~al.(2018)Tran, Li, and Madry]{tran2018spectral}
Tran, B., Li, J., and Madry, A.
\newblock {Spectral Signatures in Backdoor Attacks}.
\newblock In \emph{NeurIPS}, 2018.

\bibitem[Turner et~al.(2019)Turner, Tsipras, and Madry]{turner2019label}
Turner, A., Tsipras, D., and Madry, A.
\newblock {Label-Consistent Backdoor Attacks}.
\newblock \emph{arXiv}, 2019.

\bibitem[Van~der Maaten \& Hinton(2008)Van~der Maaten and Hinton]{van2008visualizing}
Van~der Maaten, L. and Hinton, G.
\newblock {Visualizing data using t-SNE}.
\newblock \emph{JMLR}, 2008.

\bibitem[Wang et~al.(2019)Wang, Yao, Shan, Li, Viswanath, Zheng, and Zhao]{wang2019neural}
Wang, B., Yao, Y., Shan, S., Li, H., Viswanath, B., Zheng, H., and Zhao, B.~Y.
\newblock {Neural Cleanse: Identifying and Mitigating Backdoor Attacks in Neural Networks}.
\newblock In \emph{IEEE S\&P}, 2019.

\bibitem[Wang et~al.(2024)Wang, Xiang, Miller, and Kesidis]{wang2024mm}
Wang, H., Xiang, Z., Miller, D.~J., and Kesidis, G.
\newblock {MM-BD: Post-Training Detection of Backdoor Attacks with Arbitrary Backdoor Pattern Types Using a Maximum Margin Statistic}.
\newblock In \emph{IEEE S\&P}, 2024.

\bibitem[Wang et~al.(2022{\natexlab{a}})Wang, Ding, Zhai, and Ma]{wang2022training}
Wang, Z., Ding, H., Zhai, J., and Ma, S.
\newblock {Training with More Confidence: Mitigating Injected and Natural Backdoors During Training}.
\newblock In \emph{NeurIPS}, 2022{\natexlab{a}}.

\bibitem[Wang et~al.(2022{\natexlab{b}})Wang, Mei, Ding, Zhai, and Ma]{wang2022rethinking}
Wang, Z., Mei, K., Ding, H., Zhai, J., and Ma, S.
\newblock Rethinking the reverse-engineering of trojan triggers.
\newblock In \emph{NeurIPS}, 2022{\natexlab{b}}.

\bibitem[Wang et~al.(2022{\natexlab{c}})Wang, Zhai, and Ma]{Wangbpp}
Wang, Z., Zhai, J., and Ma, S.
\newblock {BppAttack: Stealthy and Efficient Trojan Attacks against Deep Neural Networks via Image Quantization and Contrastive Adversarial Learning}.
\newblock In \emph{CVPR}, 2022{\natexlab{c}}.

\bibitem[Wang et~al.(2023)Wang, Mei, Zhai, and Ma]{wang2022unicorn}
Wang, Z., Mei, K., Zhai, J., and Ma, S.
\newblock Unicorn: A unified backdoor trigger inversion framework.
\newblock In \emph{ICLR}, 2023.

\bibitem[Wenger et~al.(2021)Wenger, Passananti, Bhagoji, Yao, Zheng, and Zhao]{wenger2021backdoor}
Wenger, E., Passananti, J., Bhagoji, A.~N., Yao, Y., Zheng, H., and Zhao, B.~Y.
\newblock {Backdoor Attacks Against Deep Learning Systems in the Physical World}.
\newblock In \emph{CVPR}, 2021.

\bibitem[Xia et~al.(2022)Xia, Niu, Li, and Li]{xia2022enhancing}
Xia, P., Niu, H., Li, Z., and Li, B.
\newblock Enhancing backdoor attacks with multi-level mmd regularization.
\newblock \emph{IEEE Transactions on Dependable and Secure Computing}, 2022.

\bibitem[Xiang et~al.(2023)Xiang, Xiong, and Li]{xiang2023umd}
Xiang, Z., Xiong, Z., and Li, B.
\newblock Umd: Unsupervised model detection for x2x backdoor attacks.
\newblock In \emph{ICML}, 2023.

\bibitem[Xu et~al.(2023)Xu, Li, Jiang, and Xia]{xu2023batt}
Xu, T., Li, Y., Jiang, Y., and Xia, S.-T.
\newblock Batt: Backdoor attack with transformation-based triggers.
\newblock In \emph{ICASSP}, 2023.

\bibitem[Xu et~al.(2024)Xu, Huang, Li, Qin, and Ren]{xu2024towards}
Xu, X., Huang, K., Li, Y., Qin, Z., and Ren, K.
\newblock Towards reliable and efficient backdoor trigger inversion via decoupling benign features.
\newblock In \emph{ICLR}, 2024.

\bibitem[Ya et~al.(2024)Ya, Li, Dai, Wang, Jiang, and Xia]{ya2024towards}
Ya, M., Li, Y., Dai, T., Wang, B., Jiang, Y., and Xia, S.-T.
\newblock Towards faithful xai evaluation via generalization-limited backdoor watermark.
\newblock In \emph{ICLR}, 2024.

\bibitem[Yao et~al.(2024)Yao, Zhang, Guo, Tian, Peng, Zou, Zhang, and Chen]{yao2024reverse}
Yao, Z., Zhang, H., Guo, Y., Tian, X., Peng, W., Zou, Y., Zhang, L.~Y., and Chen, C.
\newblock Reverse backdoor distillation: Towards online backdoor attack detection for deep neural network models.
\newblock \emph{IEEE Transactions on Dependable and Secure Computing}, 2024.

\bibitem[Zeng et~al.(2021)Zeng, Park, Mao, and Jia]{zeng2021rethinking}
Zeng, Y., Park, W., Mao, Z.~M., and Jia, R.
\newblock Rethinking the backdoor attacks' triggers: A frequency perspective.
\newblock In \emph{ICCV}, 2021.

\bibitem[Zeng et~al.(2022)Zeng, Chen, Park, Mao, Jin, and Jia]{zeng2022adversarial}
Zeng, Y., Chen, S., Park, W., Mao, Z.~M., Jin, M., and Jia, R.
\newblock {Adversarial Unlearning of Backdoors via Implicit Hypergradient}.
\newblock In \emph{ICLR}, 2022.

\bibitem[Zeng et~al.(2023)Zeng, Pan, Just, Lyu, Qiu, and Jia]{zeng2023narcissus}
Zeng, Y., Pan, M., Just, H.~A., Lyu, L., Qiu, M., and Jia, R.
\newblock {Narcissus: A Practical Clean-Label Backdoor Attack with Limited Information}.
\newblock In \emph{CCS}, 2023.

\bibitem[Zhang et~al.(2024)Zhang, Hu, Wang, Zhang, Zhou, Wang, Zhang, and Chen]{zhang2024detector}
Zhang, H., Hu, S., Wang, Y., Zhang, L.~Y., Zhou, Z., Wang, X., Zhang, Y., and Chen, C.
\newblock Detector collapse: Backdooring object detection to catastrophic overload or blindness.
\newblock In \emph{IJCAI}, 2024.

\bibitem[Zhang et~al.(2022)Zhang, Dongdong, Huang, Liao, Zhang, Feng, Hua, and Yu]{zhang2022poison}
Zhang, J., Dongdong, C., Huang, Q., Liao, J., Zhang, W., Feng, H., Hua, G., and Yu, N.
\newblock Poison ink: Robust and invisible backdoor attack.
\newblock \emph{IEEE Transactions on Image Processing}, 2022.

\bibitem[Zoran \& Weiss(2012)Zoran and Weiss]{zoran2012natural}
Zoran, D. and Weiss, Y.
\newblock Natural images, gaussian mixtures and dead leaves.
\newblock In \emph{NeurIPS}, 2012.

\end{thebibliography}
\bibliographystyle{icml2024}

\clearpage
\appendix
\onecolumn
\newcommand{\AppendixPrefix}{A}
\setcounter{section}{0}
\renewcommand{\thefigure}{\AppendixPrefix\arabic{figure}}
\setcounter{figure}{0}
\renewcommand{\thetable}{\AppendixPrefix\arabic{table}} 
\setcounter{table}{0}
\renewcommand{\theequation}{\AppendixPrefix\arabic{equation}} 
\setcounter{equation}{0}
\setcounter{theorem}{0}

\section*{Appendix}
\section{The Omitted Proof of Theorem 3.1}\label{th:proof}
\noindent\textbf{Theorem 3.1.} \emph{Let $F=FC\circ f_L\circ\dots\circ f_1$ be a backdoored DNN with $L$ hidden layers and FC denotes the fully-connected layers. Let $x$ be an input, $\vb=f_l\circ\cdots\circ f_1(x)$ be its batch-normalized feature after the $l$-th layer ($1\leq l\leq L$), and $t$ represent the attacker-specified target class. Assume that $\vb$ follows a mixture of Gaussian distribution. Then the following two statements hold: (1) Amplifying the $\vbeta$ and $\gamma$ parameters of the $l$-th BN layer can make $\Vert \tvb \Vert_2$ ($\tvb$ is the amplified version of $\vb$) arbitrarily large, and (2) There exists a positive constant $M$ that is independent of $\tvb$, such that whenever $\Vert \tvb \Vert_2 > M$, then $\arg \max FC\circ f_L\circ\dots\circ f_{l+1}(\tvb)=t$, even when $\arg \max FC\circ f_L\circ\dots\circ f_{l+1}(\vb)\not=t$}

\noindent\textbf{Proof of Theorem 3.1:}
For simplicity, let $\gF$ denote the benign model and $\tgF$ denote the backdoored model. We look at the $l$-th (pre-batch-norm) feature layer $\va_l$ such that
\begin{gather}
    \va = \va_l(\vx), \vb=\phi(\va;\gamma_{\mathrm{Benign}},\vbeta_{\mathrm{Benign}}),\gF(\vx)=\mathrm{FC}\circ f_L\circ\cdots \circ f_{l+1}(\vb),\\
  \tva = \tva_l(\vx), \tvb=\phi(\tva;\gamma, \vbeta),\tgF(\vx)=\mathrm{FC}\circ f_L\circ\cdots \circ f_{l+1}(\tvb).
\end{gather}
We assume all features follow the mixture of Gaussians, an assumption commonly used in many deep learning theory papers~\cite{guo2024gaussian,zoran2012natural,loureiro2021learning} as it simplifies analysis and provides a tractable framework for modeling complex data distributions. Consequently, $\va$ and $\tva$ follow:
\begin{gather}
\label{eq:3}
    \va\sim \frac{1}{C}\sum_{c=1}^C Z_c\exp{\frac{-\Vert\va-\mu_c\Vert_2^2}{2\sigma_c^2}},
    \va^{c}|c\sim Z_c\exp{\frac{-\Vert\va-\mu_c\Vert_2^2}{2\sigma_c^2}}, c\in\gY,\\
    \label{eq:4}
    \vb^{c}|c\sim B_c \exp{-\frac{\Vert\vb^{c}-(\vbeta_{\mathrm{Benign}}-\gamma_{\mathrm{Benign}}\frac{\mu_{\va}-\mu_c}{\sqrt{\sigma^2_{\va}+\epsilon}})\Vert_2^2}{2\frac{\gamma^2\sigma_c^2}{\sigma_{\va}^2}}}=\gN(\va;\vbeta_{\mathrm{Benign}}-\gamma_{\mathrm{Benign}}\frac{\mu_{\va}-\mu_c}{\sqrt{\sigma^2_{\va}+\epsilon}}, \frac{\gamma_{\mathrm{Benign}}\sigma_c}{\sigma_{\va}}), c\in\gY,
\end{gather}
and
\begin{gather}
        \tva\sim \frac{1}{C}\sum_{c=1}^C Z_c\exp{\frac{-\Vert\tva-\tmu_c\Vert_2^2}{2\tsigma_c^2}},
    \tva^{c}|c\sim Z_c\exp{\frac{-\Vert\tva-\tmu_c\Vert_2^2}{2\tsigma_c^2}}, c\in\ \gY,\\
    \tvb^{c}|c\sim B_c \exp{-\frac{\Vert\tvb^{c}-(\vbeta-\gamma\frac{\tmu_{\tva}-\tmu_c}{\sqrt{\tsigma^2_{\tva}+\epsilon}})\Vert_2^2}{2\frac{\gamma^2\tsigma_c^2}{\tsigma_{\tva}^2}}}=\gN(\tva;\vbeta-\gamma\frac{\tmu_{\tva}-\tmu_c}{\sqrt{\tsigma^2_{\tva}+\epsilon}}, \frac{\gamma\tsigma_c}{\tsigma_{\tva}}), c\in\gY,
\end{gather}
where
\begin{gather}
    \mu_c = \E_{\vx\sim p(\vx|\argmax\gF(\vx)=c)}[\va], \tmu_c=\E_{\vx\sim p(\vx|\argmax\tgF(\vx)=c)}[\tva],\\
    \sigma_c=\Std_{\vx\sim p(\vx|\argmax\gF(\vx)=c)}(\va), \tilde{\sigma}_c=\Std_{\vx\sim p(\vx|\argmax\tgF(\vx)=c)}(\tva), c\in\gY,\\
    \mu_{\va}=\E_{\vx\sim p(\vx)}[\va], \tmu_{\tva}=\E_{\vx\sim p(\vx)}[\tva],\sigma_{\va}=\Std_{\vx\sim p(\vx)}(\va), \tsigma_{\tva}=\E_{\vx\sim p(\tva)}[\tva].
\end{gather}\
For a sufficiently trained network, it is well-known that, with the neural collapse~\cite{papyan2020prevalence},  $\mu_c$ and $\sigma_c, c\in\gY$ form a simplex and are uniformly distributed. Specifically, in neural collapse scenarios, the features of each class form a simplex equiangular tight frame. This means that all features share (nearly) the same within-class variance and exhibit uniform mean values. Below, we try to find out the characteristics of a backdoored model.

\subsection{Characterize the Backdoored Model}
We denote the poisoned sample as \( x(t) = x + G(x) \), where $x$ is a benign input and \( G(x) \), the trigger, is very small (\ie, $\Vert G(x)\Vert_2 \ll 1$). The trigger $G(x)$ can be static or vary with different inputs, which fools the backdoored model $\tgF$ into recognizing the poisoned samples as the attacked target class $t$ instead of its true class $c$. For clarity, we simplify the trigger $G(x)$ as $\vdelta$. In this paper, we assume that all images have been normalized, \ie, $x \in [0, 1]$. Accordingly, $||t||_2 \ll 1$ holds in practice since the triggers are either very sparse (\eg, BadNets) or have a small overall magnitude (\eg, WaNet). 
So the feature distribution of $\vx(\vdelta)=\vx+\vdelta$ may be approximated by
\def\tvv{\tilde{\bm{v}}}
\begin{gather}
\tva(\vdelta)\approx\tva_l(\vx+\vdelta)=\tva_l(\vx)+\nabla\tva_l(\vx)^T\vdelta=\tva+\nabla\tva^T\vdelta,\\
    \tvb(\vdelta)=\gamma\left(\frac{\tva(\vdelta)-\tmu_{\tva}}{\sqrt{\tsigma_{\tva}^2+\epsilon}}\right)+\vbeta\approx \tvb+\frac{\nabla\tva^T\vdelta}{\sqrt{\tsigma_{\tva}^2+\epsilon}}\equiv \tvb+\tvv^T\vdelta.
\end{gather}
As $\vx(\vdelta)$ should be recognized as category $t$, the conditional probability of $\tvb(\vdelta)$ being sampled from $\tvb|c$ should be smaller than from $\tvb|t$ for all $c\in\gY$. The assumption holds, particularly for the deeper hidden layers, under the Gaussian mixture distribution and a well-trained network. Specifically, in~\cref{eq:3,eq:4}, we assume the conditional distribution $a_l(x) | \arg\max f(x) = c$ to be Gaussian. This conditional distribution is derived after completing the forward pass and examining the previous layers, and it remains unchanged, \ie, $p(a_l(x)\in a_l(A)| \arg\max f(x) = c )$$=p(a_{l+1}(x)\in a_{l+1}(A) | \arg\max f(x) = c), A={x: \arg\max f(x)=s}$. Clearly, in the last layer, the probability of $a_l(x)$ belonging to class $c$ will always be larger than other classes. Therefore, the assumption holds for the conditional distribution of $a_l(x) | \arg\max f(x) = c$. Thus, we can get, $\forall \vx,\vdelta\in\gX$,
\begin{gather}\label{eq:density_geq}
    B_t\exp{-\frac{\Vert\tvb(\vdelta)-(\vbeta-\gamma\frac{\tmu_{\tva}-\tmu_t}{\sqrt{\tsigma^2_{\tva}+\epsilon}})\Vert_2^2}{2\frac{\gamma^2\tsigma_t^2}{\tsigma_{\tva}^2}}}> B_c\exp{-\frac{\Vert\tvb(\vdelta)-(\vbeta-\gamma\frac{\tmu_{\tva}-\tmu_c}{\sqrt{\tsigma^2_{\tva}+\epsilon}})\Vert_2^2}{2\frac{\gamma^2\tsigma_c^2}{\tsigma_{\tva}^2}}},\forall c\in\gY,c\neq t,\\
    \Leftrightarrow \log \frac{B_t}{B_c}+\frac{1}{2\gamma^2\tsigma_t^2\tsigma_c^2}\left(\tsigma_t^2\Vert\tvb(\vdelta)-(\vbeta-\gamma\frac{\tmu_{\tva}-\tmu_c}{\sqrt{\tsigma^2_{\tva}+\epsilon}}-\tsigma_c^2)\Vert_2^2-\tsigma_c^2\Vert\tvb(\vdelta)-(\vbeta-\gamma\frac{\tmu_{\tva}-\tmu_t}{\sqrt{\tsigma^2_{\tva}+\epsilon}})\Vert_2^2\right)> 0,\\
    \forall\tvb(\vdelta)=\tvb+\tvv^T\vdelta.
\end{gather}
Note that this is actually a \textit{quadratic form} (the form of $ax^2+bx+c>0, \forall x$) of $\tvb$, to make sure the above inequality holds for all $\tvb$ (or at least most of $\tvb$ in the feature space), it is obvious that the quadratic coefficient $(\tsigma_t^2-\tsigma_c^2)$ must be positive, so we should have
\begin{equation}
    \tsigma_t>\tsigma_c,\forall c\in\gY, c\neq t.
\end{equation}
So we can confirm a key characteristic of the backdoored model, that the variance of the attacked target class $t$ is larger than any of the others.

\subsection{Parameter-oriented Scaling Consistency of Backdoored Models}
After obtaining the above characteristic of the backdoored model, we can then prove the parameter-oriented scaling consistency of it. 

Let 
\begin{gather}
    \Gamma_c=\vbeta-\gamma\frac{\tmu_{\tva}-\tmu_c}{\sqrt{\tsigma^2_{\tva}+\epsilon}}, c=1,\cdots,C.
\end{gather}
Considering the above mixture of the Gaussian model, a sample $\vx$ will be classified into class $t$ if and only if
\begin{gather}\label{eq:density_final}
     B_t\exp{-\frac{\Vert\tvb-(\vbeta-\gamma\frac{\tmu_{\tva}-\tmu_t}{\sqrt{\tsigma^2_{\tva}+\epsilon}})\Vert_2^2}{2\frac{\gamma^2\tsigma_t^2}{\tsigma_{\tva}^2}}}> B_c\exp{-\frac{\Vert\tvb-(\vbeta-\gamma\frac{\tmu_{\tva}-\tmu_c}{\sqrt{\tsigma^2_{\tva}+\epsilon}})\Vert_2^2}{2\frac{\gamma^2\tsigma_c^2}{\tsigma_{\tva}^2}}},\forall c\in\gY, c\neq d,\\
     \Leftrightarrow \log \frac{B_t}{B_c}+ \frac{\tsigma_{\tva}^2}{2\gamma^2\tsigma_t^2\tsigma_c^2}\left(\tsigma_t^2\Vert \tvb-\Gamma_c\Vert_2^2-\tsigma_c^2\Vert\tvb-\Gamma_t\Vert_2^2    \right)\geq 0.
\end{gather}
The above can stand if
\begin{gather}
     \log \frac{B_t}{B_c}+ \frac{\tsigma_{\tva}^2}{2\gamma^2\tsigma_t^2\tsigma_c^2}\left( (\tsigma_t^2-\tsigma_c^2)\Vert\tvb\Vert_2^2-2\Vert\tsigma_t^2\Gamma_c-\tsigma_c^2\Gamma_t\Vert_2\Vert\tvb\Vert_2+\frac{\Vert\tsigma_t^2\Gamma_c-\tsigma_c^2\Gamma_t\Vert_2^2}{\tsigma_t^2-\tsigma_c^2}   \right)\\
     +\frac{\tsigma_{\tva}^2}{2\gamma^2\tsigma_t^2\tsigma_c^2}\left(\tsigma_t^2\Vert\Gamma_c\Vert_2^2-\tsigma_c^2\Vert\Gamma_t\Vert_2^2-\Vert\tsigma_t^2\Gamma_c-\tsigma_c^2\Gamma_t\Vert_2^2\right)\geq 0\\
     \Leftrightarrow \Vert\tvb\Vert_2 \geq \max\{\frac{1}{\sqrt{\tsigma_t^2-\tsigma_c^2}}\sqrt{-\tsigma_t^2\Vert\Gamma_c\Vert_2^2+\tsigma_c^2\Vert\Gamma_t\Vert_2^2+\Vert\tsigma_t^2\Gamma_c-\tsigma_c^2\Gamma_t\Vert_2^2-\log \frac{B_t}{B_c}\frac{2\gamma^2\tsigma_t^2\tsigma_c^2}{\tsigma_{\tva}^2}}\\\label{eq:qua}
     +\frac{\Vert\tsigma_t^2\Gamma_c-\tsigma_c^2\Gamma_t\Vert_2}{\tsigma_t^2-\tsigma_c^2},0\}.
\end{gather}
So just like~\cref{eq:density_geq}, the above is also a quadratic form for $\Vert\tvb\Vert_2$ with positive quadratic coefficient $(\tsigma_t^2-\tsigma_c^2)>0$. So when $\Vert\tvb\Vert_2$ is large enough (\cref{eq:qua}), we will always have $\vx$ is more likely to be identified into category $t$ than all the others.

\begin{remark}
    Note that scale the parameter $\vbeta_{\mathrm{Inf}},\gamma_{\mathrm{Inf}}$ when inference will not influence the value of $\vbeta,\gamma$ in~\cref{eq:density_final}. The $\vbeta, \gamma$ in~\cref{eq:density_final} is used to describe the underlying feature distributions (which are assumed to be the mixture of Gaussians). They will not change upon training finished.
\end{remark}

As a result, when we scale the Batch Norm parameter $\vbeta_{\mathrm{Inf}}, \gamma_{\mathrm{Inf}}$, we will get a $\tvb_{\mathrm{Scale}}$ with larger norm propositional to $\gamma_{\mathrm{Inf}}$ and linearly increasing with respect to $\vbeta_{\mathrm{Inf}}$. When $\vbeta_{\mathrm{Inf}}, \gamma_{\mathrm{Inf}}$ are larger enough, the scaled feature $\tvb_{\mathrm{Scale}}$ will make~\cref{eq:density_final} always positive. 

\begin{remark}
    The above proof can be intuitively understood as follows: if we sample from a mixture of Gaussian distribution, then all remote points will be sampled from the Gaussian with the largest variance.
\end{remark}

\section{Detailed Configurations of the Empirical Study in~\cref{sec:phe}}
\label{appendix:empirical}
In this section, we adopt BadNets~\cite{gu2017badnets}, WaNet~\cite{nguyen2021wanet}, and BATT~\cite{xu2023batt} as examples for our analysis. These attacks epitomize static, dynamic, and physical backdoor attacks, respectively. Our experiments are conducted on the CIFAR-10 dataset~\cite{krizhevsky2009learning}, using the ResNet18 model~\cite{he2016deep}. For each attack, we set the poisoning rate ($\rho$) to 0.1, achieving ASRs over 99\%. In particular, we implement the backdoor attacks using their official codes with default settings. Specifically, the backdoor trigger for BadNets is represented as a $3\times 3$ grid in black-and-white and is added to the lower-right corner of the poisoned images. For WaNet, the trigger is applied to the original images through elastic image warping transformation. In the case of BATT, the poisoned samples are obtained by rotating the original images by sixteen degrees. These attacks are implemented using the BackdoorBox toolkit~\cite{backdoorbox}\footnote{\url{https://github.com/THUYimingLi/BackdoorBox}}.


Regarding the scaling procedure, we adopt a layer-wise weight scaling operation to generate the parameter-amplified models. we scale up on the BN parameters (\ie, $\gamma$ and $\vbeta$) with $\omega=1.5$ times starting from the last layer and gradually moving forward to more layers. For example, in a 20-layer model, the first iteration involves scaling the weights of the 20th layer, and the next iteration extends the scaling to the 20th and the 19th layers, and so on. We then calculate the \emph{average confidence} of 2000 testing samples for each parameter-scaled model. In this paper, \emph{confidence} refers to the predicted probability assigned to an input sample for a specified label. For instance, if an image of a \emph{cat} is predicted as the \emph{cat} label with a probability of 0.9, then the \emph{confidence} of the input under the \emph{cat} label is 0.9. The average confidence is defined as the average probability of samples on the label predicted by the original unamplified model. 

\section{Detailed Exploration of amplifying a single BN layers in~\cref{sec:phe}}
\label{appendix:onebn}
As described in~\cref{sec:phe}, we find amplifying only a single BN layer
may require an unreasonably large amplification factor, and due to the nonlinearity of neural network layers, often leads to unstable defense performance across different attacks. To further explain the phenomenon, we conduct an empirical investigation aimed at investigating the percentage of benign samples to be predicted as the target class when amplifying the learnable parameters of individual BN layers with scale $S$. The results are displayed in Table~\cref{tab:prop_onbn}, and we have three primary observations: 

(1) The amplification factor for achieving effective defense varies considerably from layer to layer. (2) Some attacks (\eg, WaNet and BATT) require an unreasonably large amplification factor to achieve a substantial misclassification rate. (3) Amplifying only a single BN layer may not be adequate to misclassify the majority of benign samples in some cases. For instance, amplifying the first BN layer alone cannot misclassify benign samples from the Ada-patch attack into the intended target class.

\begin{table}[!t]
\centering
\setlength{\tabcolsep}{2pt}
\small
\caption{The proportion (\%) of benign samples in CIFAR-10 predicted to the target class when amplifying only a single BN layer.}
\label{tab:prop_onbn}
\begin{tabular}{lccccccccccccccccc} 
\toprule
Index $\rightarrow$ &\multicolumn{4}{c}{1}	& \multicolumn{4}{c}{5} &	\multicolumn{4}{c}{15}\\
\cmidrule(lr){2-5} \cmidrule(lr){6-9} \cmidrule(lr){10-13}
Scales $S$ $\downarrow$	&BadNets	& WaNet	&BATT	&Ada-patch	&BadNets	& WaNet	& BATT	& Ada-patch & 	BadNets	& WaNet	&BATT	& Ada-patch\\
\midrule
5	&96.75	&10.50	&62.86	&0.00	&92.43&	93.25	&5.04&	12.85	&11.37	&99.32&	99.13	&76.81\\
10	&100.00	&53.53&	38.81	&0.00	&100.00&	100.00&	2.19&	27.40&	16.33	&100.00&	100.00&	89.66\\
100	&100.00&	100.00&	100.00	&0.15	&100.00&	100.00	&99.96&	91.56&	27.40&	100.00	&100.00&	96.10\\
1000	&100.00&	100.00&	100.00&	0.43&	100.00&	100.00&	100.00&	93.99	&28.89&	100.00&	100.00	&96.45\\
100000	&100.00	&100.00&	100.00&	0.44	&100.00	&100.00&	100.00&	94.18&	29.01	&100.00&	100.00	&96.49\\
\bottomrule
\end{tabular}
\end{table}

\begin{figure*}[!t]
\centering
\subfigure[Benign Model\label{fig:l2real_normL2real}]{\includegraphics[width=0.24\textwidth]{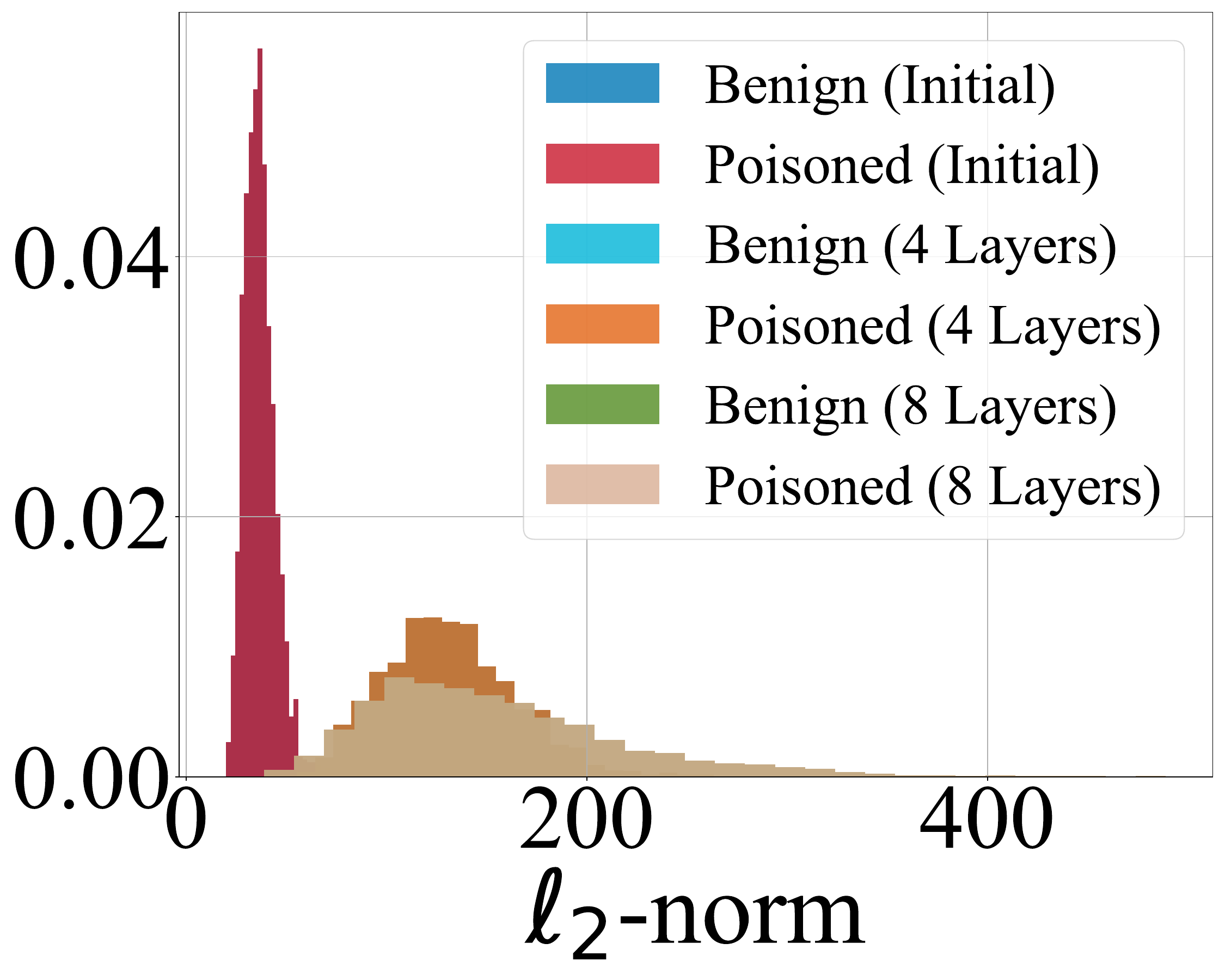}}
\subfigure[BadNets\label{fig:l2real_badnets_l2}]{\includegraphics[width=0.24\textwidth]{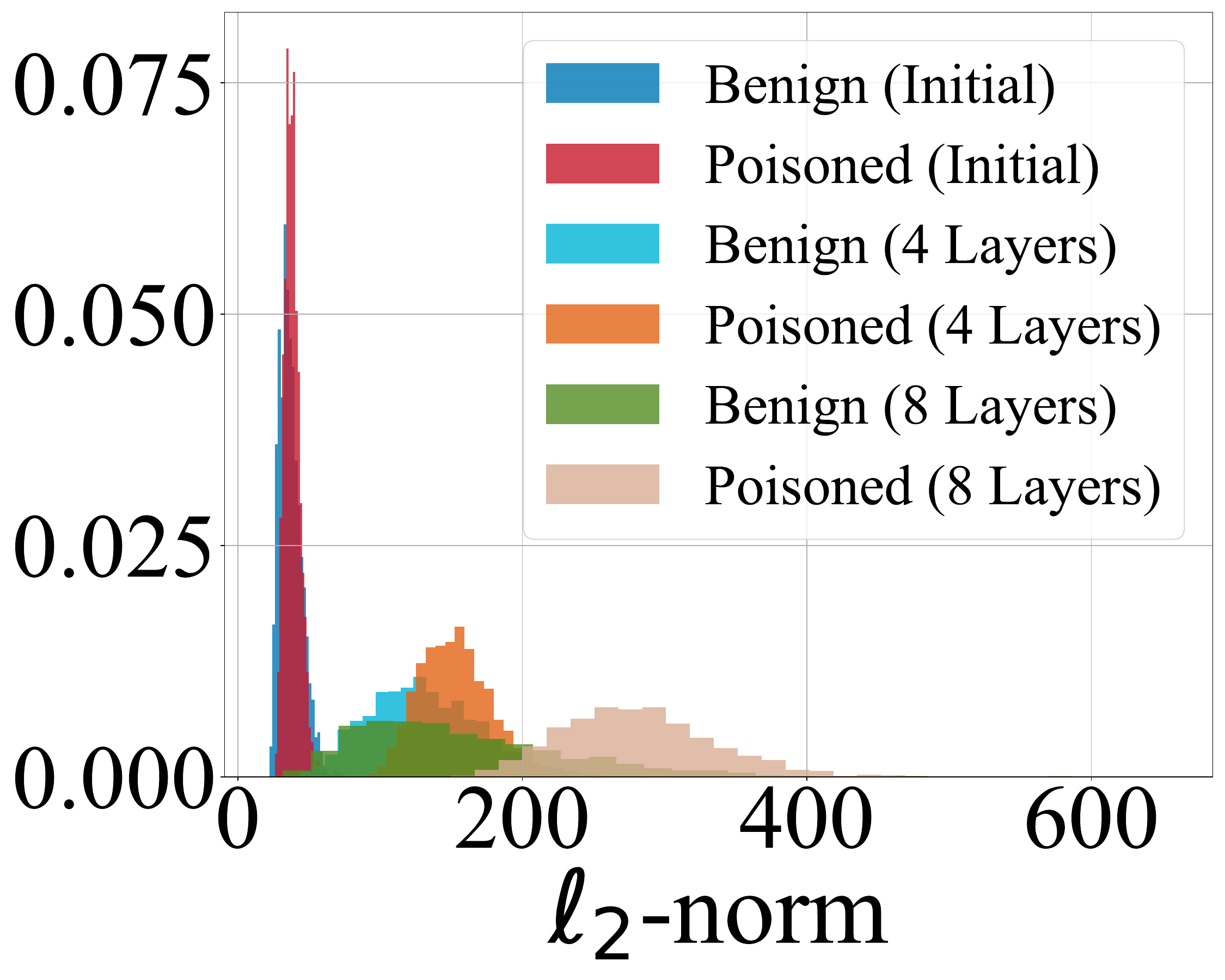}}
\subfigure[WaNet\label{fig:l2real_wanet_l2}]{\includegraphics[width=0.24\textwidth]{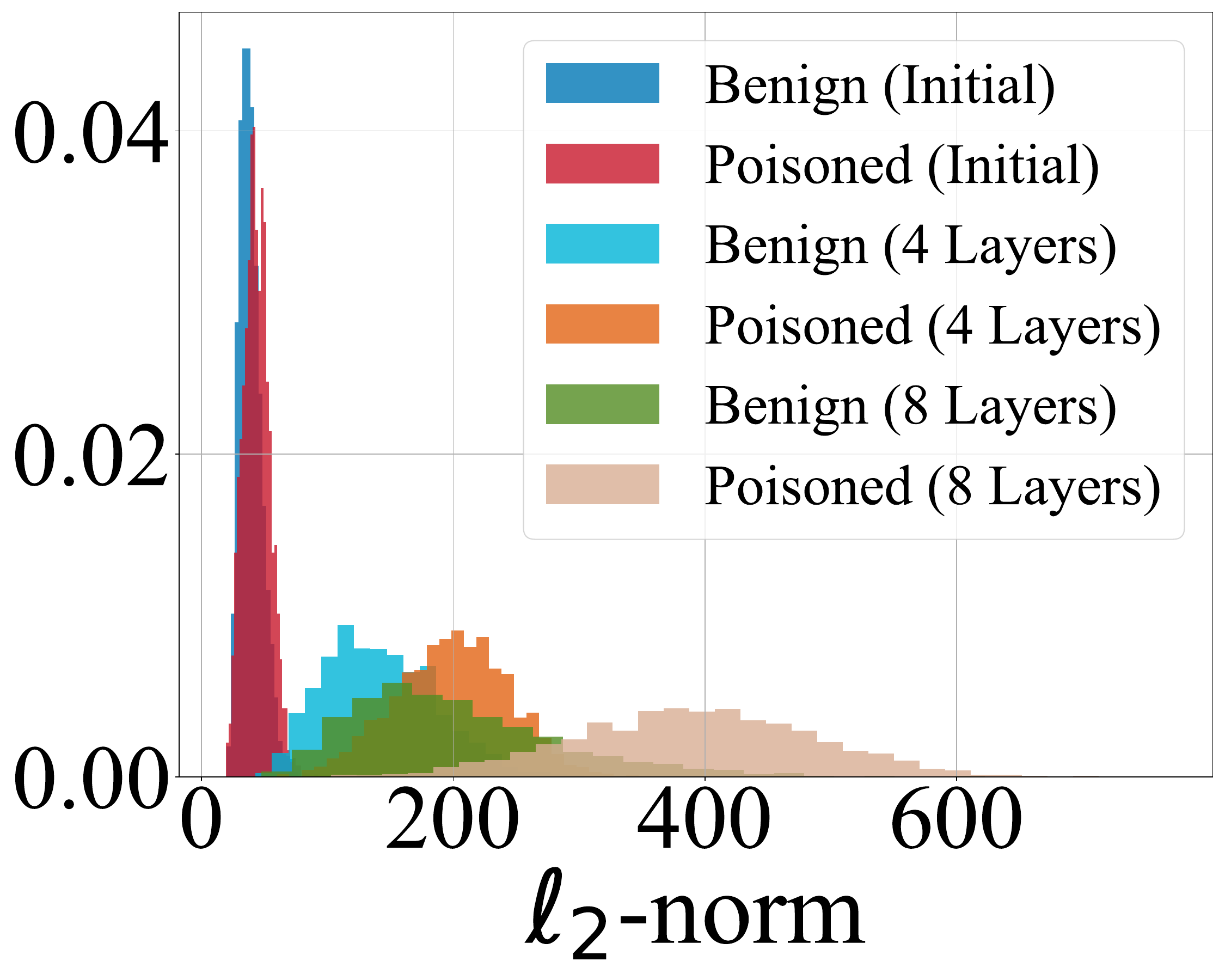}}
\subfigure[BATT\label{fig:l2real_batt_l2}]{\includegraphics[width=0.24\textwidth]{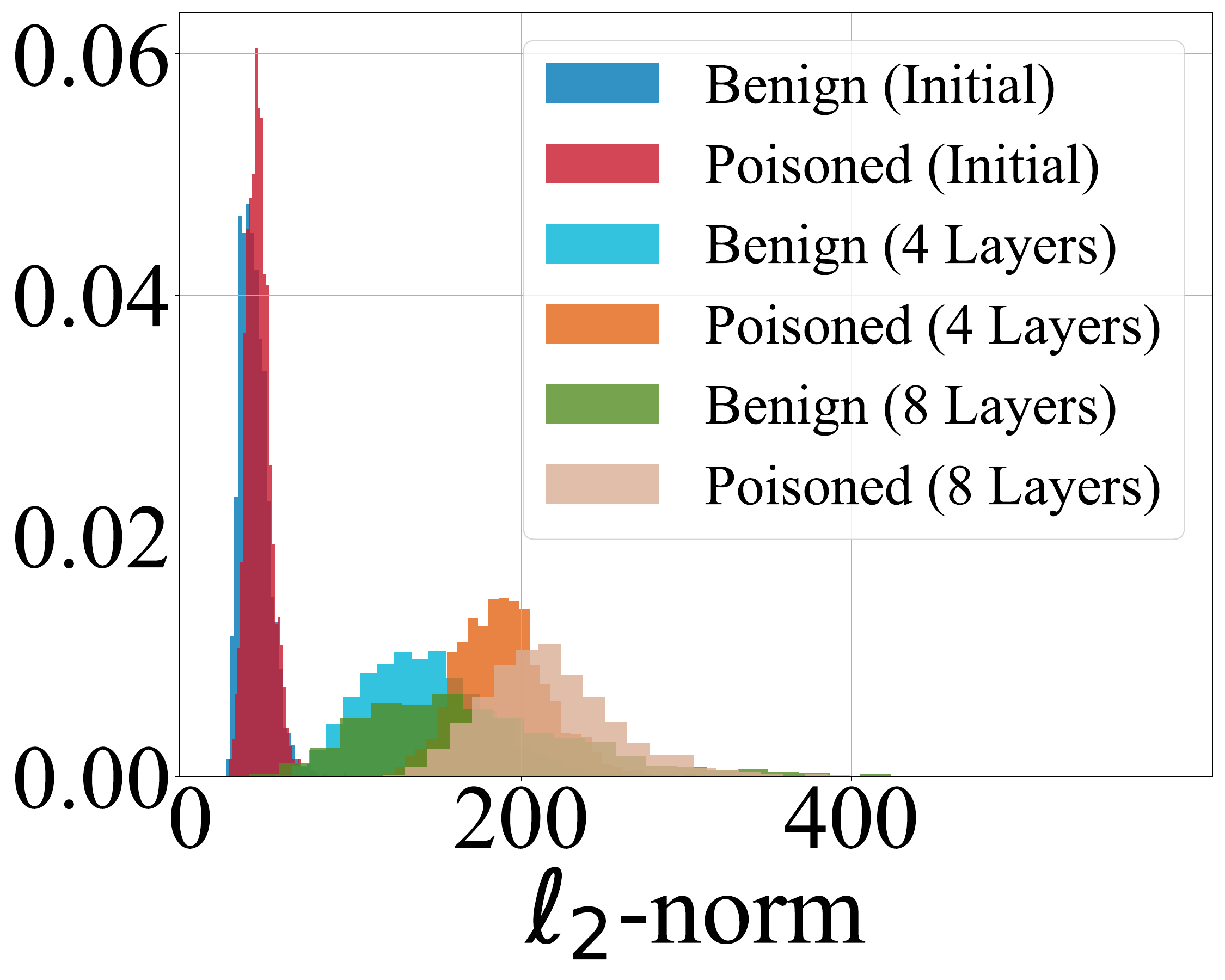}
}
\vspace{-1.2em}
\caption{Histograms of the $\ell_2$-norm of the final feature map of samples generated by models with different numbers of amplified BN layers. Increasing the number of amplified layers increases both value and variance of features.}
\vspace{-0.8em}
\label{fig:intuition_l2real}
\end{figure*}

To address this, we spread the amplification across multiple consecutive BN layers, using a small factor (\eg, 1.5) on each layer.
Instead of controlling the layer-wise amplification factor, we vary the number of amplified layers to achieve different levels of accumulated amplification.
This relation is demonstrated in~\cref{fig:intuition_l2real} (see~\cref{fig:intuition_l2} for the density plot), where we see amplifying more layers induces higher last-layer activations, and increases the room to differentiate poisoned samples from the benign ones.

\section{Why Scale the Later Layers?}
\label{appendix:forward}
Our defense relies on building a profile of how the target model behavior changes under progressive modifications to the model.
Motivated by a widely accepted hypothesis (\eg, \cite{tishby2015deep,huang2022backdoor,jebreel2023defending}) that layers situated towards the later stages exert a more direct influence on the ultimate model output, 
we designed our defense by
 amplifying the model parameters in stages, starting from the last hidden layer and progressively moving backward through the preceding layers.
Here, we examine the alternative of a forward model scaling approach,
which
scales
model parameters starting from the initial layers of the model and then progressing forward to the latter layers. 

The results in~\cref{tab:forward_scaling} demonstrate that while this defense strategy proves to be effective against most backdoor attacks, such as BadNets, and ISSBA, it exhibits poor performance against others like Blend, BATT, and LC attacks. This discrepancy may be attributed to the fact that in those attacks, the trigger features closely resemble benign features in the model's shallow layers, making it challenging for the amplification operation to sufficiently separate these two types of features. 

\begin{table}[!t]
\centering
\small
\caption{The performance (AUROC, F1) of our defense with forward model scaling process on the CIFAR-10 dataset. We mark the best result in boldface and failed cases ($<0.7$) in red.}
\label{tab:forward_scaling}
\setlength{\tabcolsep}{3pt}
\begin{tabular}{lcccccccccccccc} 
\toprule
 Metrics & BadNets & Blend & PhysicalBA & IAD &WaNet & ISSBA& BATT & SRA & LC & NARCISSUS &Adap-Patch\\
\midrule
AUROC & 0.997 &  \red{0.678} & 0.964 & 0.999 & 0.910& 0.998 &\red{0.635} & 0.952 & \red{0.450} &0.941& 0.960 \\
 F1 &  0.964 & \red{0.002} & 0.908 & 0.966 &\red{0.639}  & 0.970 &\red{0.052} & 0.904  & \red{0} & 0.922& 0.831\\ 
\bottomrule
\vspace{-1em}
\end{tabular}
\end{table}

\begin{table}[!t]
\centering
\caption{The performance (AUROC, F1) of our defense with amplifying all of the BN layers on the CIFAR-10 dataset. We mark the best result in boldface and failed cases ($<0.7$) in red.}
\label{tab:all_bn}
\small
\setlength{\tabcolsep}{3pt}
\begin{tabular}{lcccccccccccccc} 
\toprule
 Metrics & BadNets & Blend & PhysicalBA & IAD &WaNet & ISSBA& BATT & SRA & LC & NARCISSUS &Adap-Patch\\
\midrule
AUROC & 0.961 & \red{0.664} &0.947 & 0.949 & 0.938 & 0.949 & 0.947 & 0.942 &\red{0.224} &0.992 & \red{0.679}\\
 F1 & 0.949 & \red{0.060} & 0.926 & 0.952 &0.941 &0.951 &0.940 & 0.943 &\red{0} &0.938 &\red{0}\\ 
\bottomrule
\end{tabular}
\vspace{-1em}
\end{table}

\section{Why not Amplifying All BN Layers?}
\label{appendix:allbnlayers}
In our defense, we amplify the later parts of the original model. It is motivated by the previous findings that trigger patterns often manifest as complicated features learned by the deeper (convolutional) layers of DNNs, especially for those attacks with elaborate designs \cite{huang2022backdoor,jebreel2023defending}. It is also consistent with our observations in~\cref{fig:intuition}. 

We investigate the performance of our defense by amplifying all BN layers within a model. As shown in~\cref{tab:all_bn}, amplifying all layers leads to defense failure against Blend, LC, and WaNet attacks. In particular, its F1 score drops to 0, suggesting that amplifying all layers in the defense fails to detect any poisoned samples.

\section{Detailed Settings for Experimental Datasets and Configurations}
\label{appendix:dataset}
In line with the existing backdoor defense methods~\cite{guo2023scale,liu2023detecting,gao2021design}, we select the most commonly used benchmark datasets and model architectures for our experiments. The datasets and models used are outlined in~\cref{table:datasets}.

\begin{table}[ht]
  \caption{The overview of the image datasets and the related classifiers used in our experiments. }
  \label{table:datasets}
  \centering
  \small
  \begin{tabular}{lcccc}
    \toprule
    Datasets & \#Classes &  Input Sizes &\#Train. \& Test. Images  & Classifiers\\
    \midrule
    CIFAR-10 &  10 & 32 $\times$ 32 $\times$ 3 & 50,000, 10,000 & \makecell[c]{ResNet18, \\PreactResNet18, MobileNet}\\
    GTSRB & 43 & 32 $\times$ 32 $\times$ 3 & 39,200, 12,600  &  \makecell[c]{ResNet18, \\PreactResNet18, MobileNet}\\
    SubImageNet-200 & 200 & 224 $\times$ 224 $\times$ 3 & 100,000, 10,000 & ResNet18\\
  \bottomrule
\end{tabular}
\end{table}

\noindent\textbf{CIFAR-10} is a benchmark dataset consisting of 3 $\times$ 32 $\times$ 32 color images representing ten different object categories~\cite{krizhevsky2009learning}. The training set comprises 50,000 images, while the test set contains 10,000 images, with an equal distribution across the ten classes.

\noindent\textbf{GTSRB} is a benchmark dataset consisting of images of German traffic signs, categorized into 43 classes~\cite{stallkamp2012man}. The training set consists of 39,209 images, while the test set contains 12,630 images. Given the considerable variation in image sizes within this dataset, we resize all images to a uniform size of 3 $\times$ 32 $\times$ 32  for our experiments, ensuring consistency and convenience in handling.

\noindent\textbf{SubImageNet-200.} We adopt a subset of the ImageNet benchmark dataset~\cite{deng2009imagenet} by randomly selecting 200 categories from the most common categories in the original ImageNet. Specifically, the subset includes 100,000 images from the original ImageNet for training (500 images per class) and 10,000 images for testing (50 images per class). For simplicity, all images are resized to a uniform dimension of 3 $\times$ 224 $\times$ 224.

\section{Details of Training Backdoored Models}
\label{appendix:baselineattacks}

\subsection{Backdoor Attacks}
In~\cref{sec:exp}, we assess the effectiveness of our defense against thirteen backdoor attacks. These attacks are categorized into three types: \textbf{1)} poisoning-only attacks, \textbf{2)} training-controlled, \textbf{3)} and model-controlled attacks.

\vspace{-0.5em}
\begin{itemize}
    \item \textbf{Poison-only Backdoor Attacks:} For the most commonly studied poisoning-only attacks, we consider various forms. This includes classic static attacks like \textbf{(1)} BadNet~\cite{gu2017badnets} and \textbf{(2)} Blend~\cite{bai2021targeted}, sample-specific attack such as \textbf{(3)} ISSBA~\cite{li2021invisible}, clean-label attacks represented by \textbf{(4)} Label-Consistent (LC)~\cite{turner2019label} and \textbf{(5)} NARCISSUS~\cite{zeng2023narcissus}. 
    In addition, we also consider adaptive attacks like \textbf{(6)} TaCT~\cite{tang2021demon} and \textbf{(7)} Adap-Patch~\cite{qi2023revisiting}, which are designed to slip past existing defenses.
    
    \item \textbf{Training-controlled Backdoor Attacks:} The training-controlled attacks include the \textbf{(8)} Dynamic~\cite{nguyen2020input}, \textbf{(9)} WaNet~\cite{nguyen2021wanet}, \textbf{(10)} BPP~\cite{Wangbpp} attacks, and physical backdoor attacks, including \textbf{(11)} PhysicalBA~\cite{li2021backdoor} and \textbf{(12)} BATT~\cite{xu2023batt}.
    \item\textbf{Model-controlled Backdoor Attacks:} we assess attacks involving direct modification of model parameters, such as \textbf{(13)} subnet replacement attack (SRA)~\cite{qi2022towards}.
\end{itemize}

\begin{figure}[!t]
    \centering
    \includegraphics[width=0.89\textwidth]{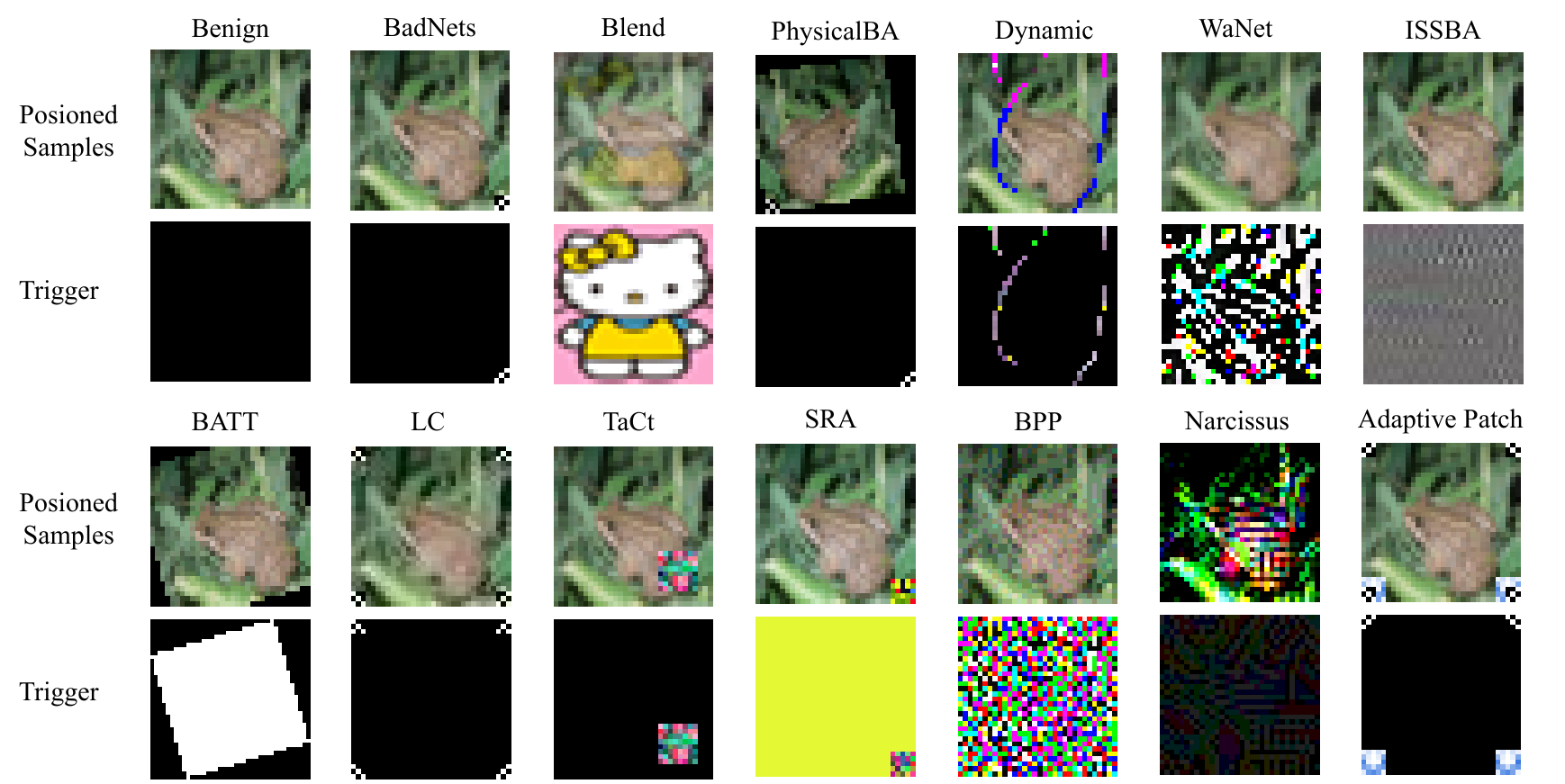}
    \vspace{-1mm}
    \caption{The examples of various triggers in the attacks adopted in our study and the corresponding poisoned samples.}
    \label{fig:trigger}
    \vspace{-1.5em}
\end{figure}

The poisoning rate $\rho$ for data-poisoning-based backdoor attacks is set to 0.1. The target class label is set to 0. In particular, the BATT attack consists of two attack modes, utilizing spatial rotation and translation transformations as triggers, respectively. In our study, we specifically employ spatial rotation as our triggers. The examples of both triggers and the corresponding poisoned samples are depicted in~\cref{fig:trigger}.

\subsection{Additional Details of Training Backdoored Models}

We adopt the standard training pipeline for developing backdoor models. This involves an SGD optimizer with a momentum of 0.9 and a weight decay of $10^{-4}$. The initial learning rate is set at 0.1, which is reduced to 10\% of its previous value at the 50th and 75th epochs. The training comprises 200 epochs with a batch size of 128. For data augmentation on the CIFAR-10 dataset, we apply RandomHorizontalFlip and RandomCrop32 (randomly cropping images to a size of 3 $\times$ 32 $\times$ 32). Additionally, RandomRotation15 is used to randomly rotate images within a range of [-15, 15] degrees. 

For data augmentation on the CIFAR-10 dataset, we utilize RandomHorizontalFlip with a probability of 0.5 and RandomCrop32, which randomly crops images to a size of 3 $\times$ 32 $\times$ 32. For the GTSRB dataset, we employ the RandomRotation15 augmentation technique, where images are randomly rotated within a range of [-15, 15] degrees. For the GTSRB dataset, we apply RandomCrop224, RandomHorizontalFlip, and RandomRotation20 to enhance the accuracy of the backdoored model on the benign samples.

All experiments are performed on a server with the Ubuntu 16.04.6 LTS operating system,  a 3.20GHz CPU, 2 NVIDIA's GeForce GTX3090 GPUs with 62G RAM, and an 8TB hard disk.

\subsection{Effectiveness of the Backdoored Attacks}

Following the settings in existing backdoor attacks, we use two metrics to measure the effectiveness of the backdoor attacks: attack success rate (ASR) and benign accuracy (BA). ASR indicates the success rate of classifying the poisoned samples into the corresponding target classes. BA measures the accuracy of a backdoored model on the benign testing dataset.

BA and ASR for different backdoor attacks are included in~\cref{tab:mainba} and~\cref{tab:otherba}.

\begin{table}[!t]
\centering
\setlength{\tabcolsep}{3.2pt}
\small
\caption{The performance (BA, ASR) on different attacks and datasets with ResNet18 model.}
\label{tab:mainba}
\begin{tabular}{lccccccccccccccccc} 
\toprule
\multirow{2}{*}{Datasets}  &\multicolumn{2}{c}{BadNets} &\multicolumn{2}{c}{Blend} & \multicolumn{2}{c}{PhysicalBA} & \multicolumn{2}{c}{Dynamic} & \multicolumn{2}{c}{WaNet} & \multicolumn{2}{c}{ISSBA} & \multicolumn{2}{c}{BATT} \\
  & BA & ASR &  BA & ASR& BA & ASR&   BA & ASR&   BA  & ASR  &  BA & ASR  &  BA & ASR &  \\
\midrule
CIFAR10 & 0.929 & 1 & 0.931 & 0.999 & 0.937 & 0.966 & 0.938 & 1 & 0.948 & 0.997 & 0.936 & 1 & 0.939 & 1\\
 \hline
GTSRB & 0.976 & 0.998 & 0.966  & 1 & 0.976 & 0.968 & 0.971 & 1 & 0.994 & 0.997 &  0.968 & 1 & 0.979 & 0.998\\
 \hline
SubImageNet-200 & 0.808&0.998 & 0.823 & 0.998 & 0.796& 0.994& 0.793& 1& 0.768& 0.967& 0.803 & 0.990 &0.695 & 0.997\\
\bottomrule
\end{tabular}
\vspace{-1.5em}
\end{table}

\begin{table}[!t]
\centering
\small
\caption{The performance (BA, ASR) on other backdoor attacks with the ResNet18 model on the CIFAR-10 dataset.}
\label{tab:otherba}
\begin{tabular}{lccccccccccc} 
\toprule
Metrics & LC & TaCT & SRA & BPP & NARCISSUS  & Adap-Patch\\
\midrule
BA &   92.28& 93.78  & 88.98 & 89.68& 89.80 & 93.54\\
ASR & 100&  99.00 & 99.90 & 99.70&  96.92& 99.89\\
\bottomrule
\end{tabular}
\vspace{-1.5em}
\end{table}

\begin{figure*}[!t]
\centering
\subfigure[BadNets\label{fig:badnets_roc}]{\includegraphics[width=0.23\textwidth]{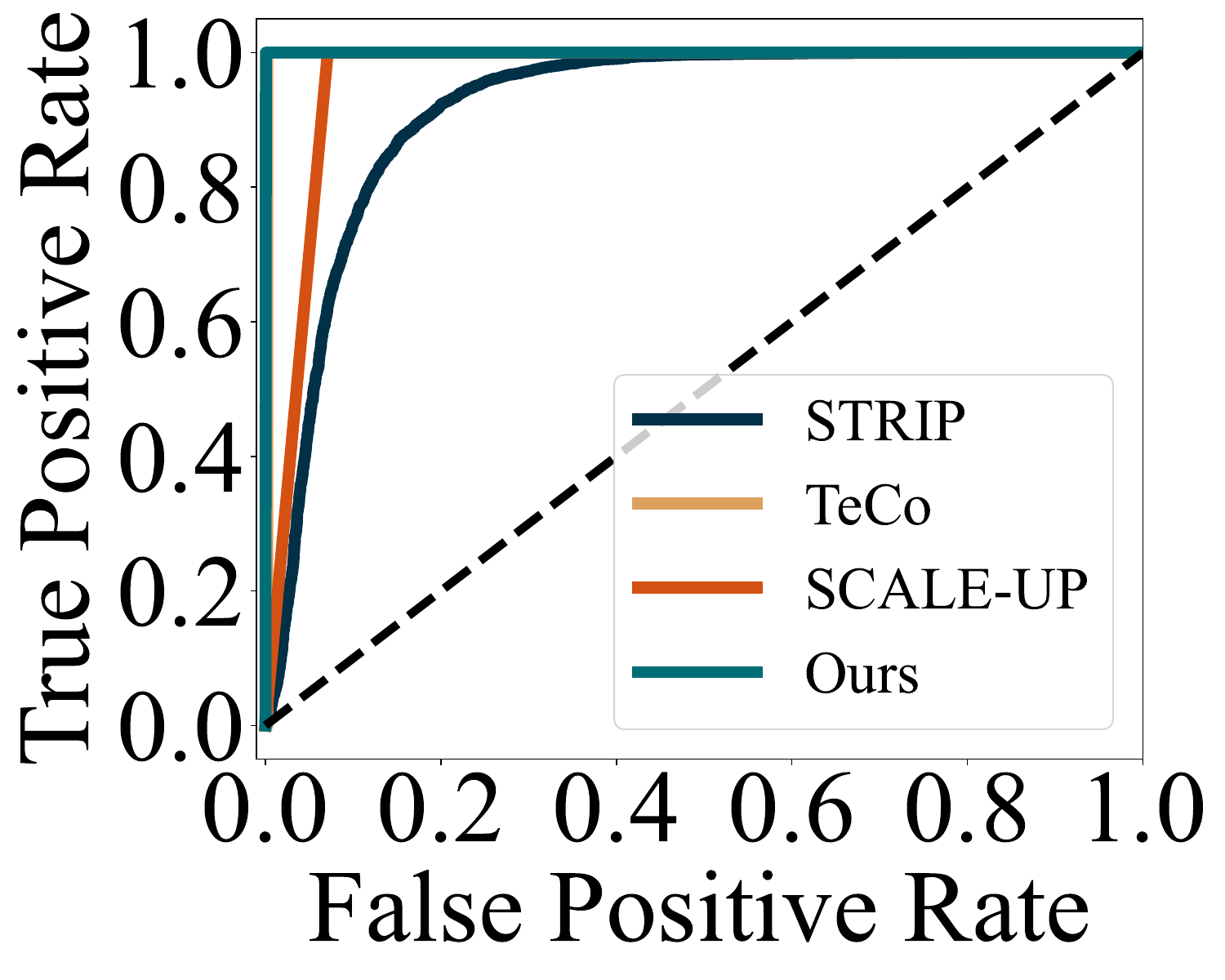}}
\subfigure[Blend\label{fig:blend_roc}]{\includegraphics[width=0.23\textwidth]{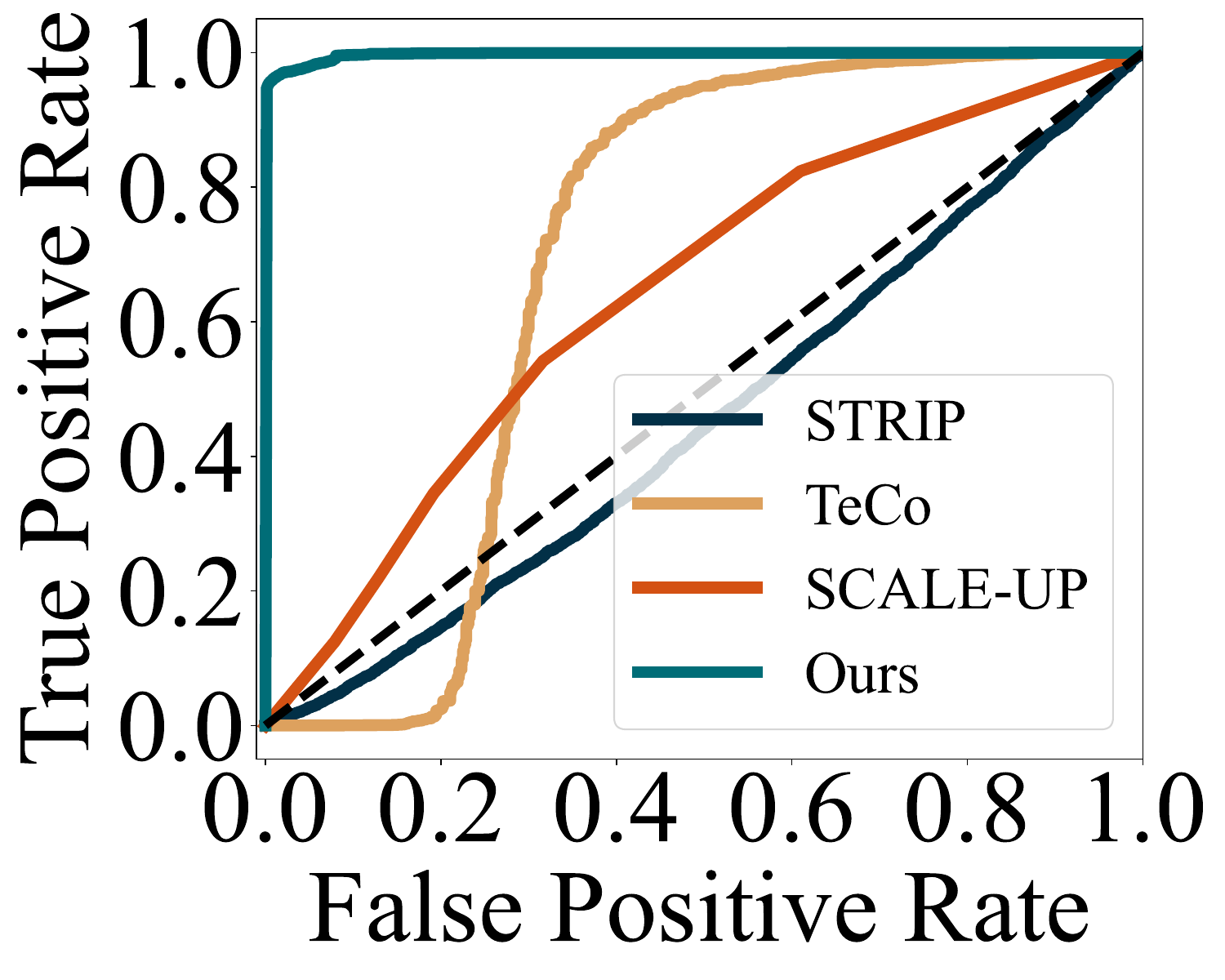}}
\subfigure[WaNet\label{fig:wanet_roc}]{\includegraphics[width=0.23\textwidth]{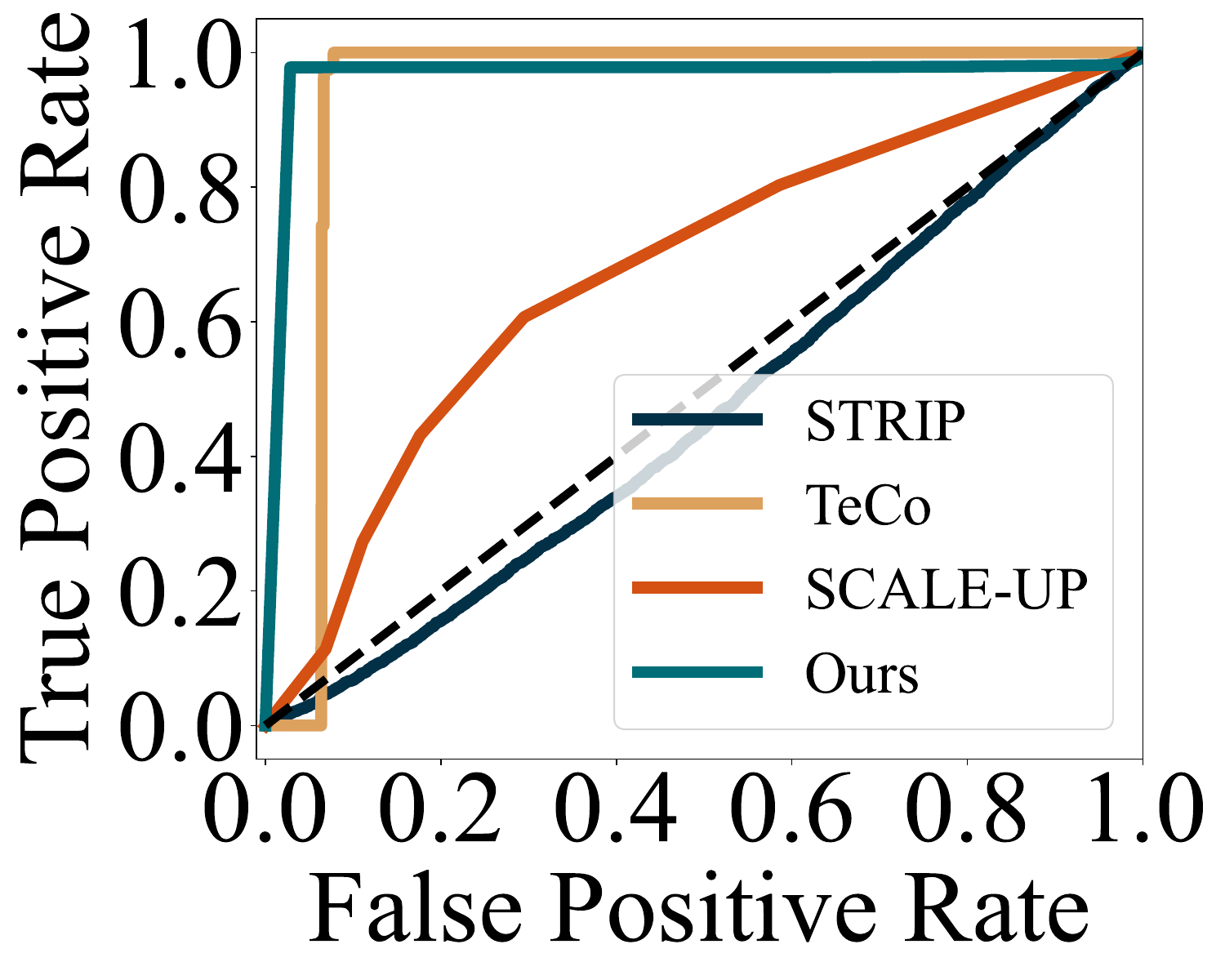}}
\subfigure[BATT\label{fig:batt_roc}]{\includegraphics[width=0.23\textwidth]{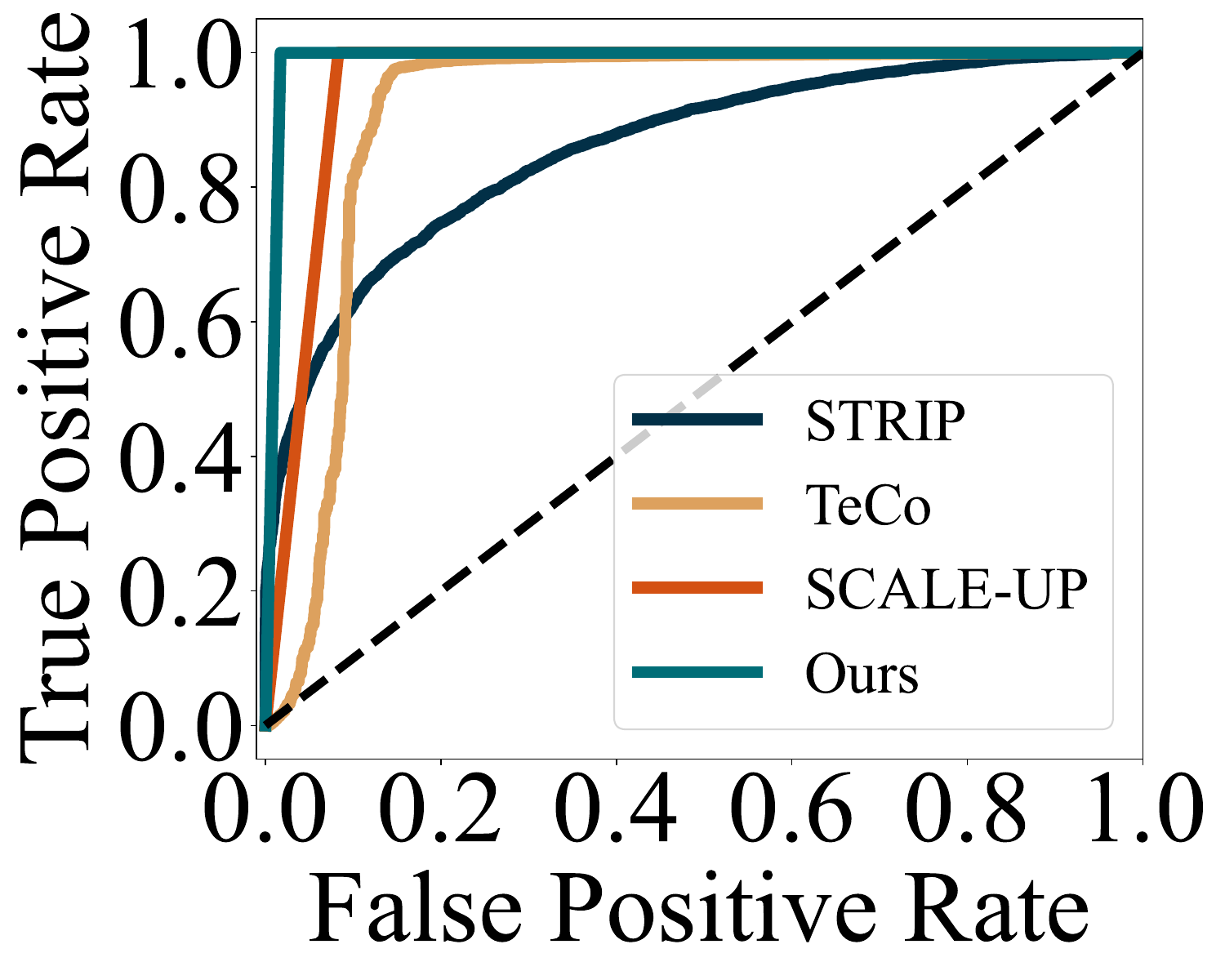}}
\vspace{-1em}
\caption{ROC curves of defenses against different attacks on the CIFAR-10 dataset.}
\label{fig:roc}
\vspace{-1em}
\end{figure*}

\section{Implementation of the Baseline Defenses}
\label{appendix:baselinedefenses}
\textbf{(1) STRIP:} We implement STRIP following their official open-sourced codes\footnote{\url{https://github.com/garrisongys/STRIP}}. STRIP detects backdoor attacks by observing the prediction behaviors of an input sample when superimposing benign features on it.

\textbf{(2) TeCo:}  We implement TeCo following their official open-sourced codes\footnote{\url{https://github.com/CGCL-codes/TeCo}}.

\textbf{(3) SCALE-UP:} We implement SCALE-UP (data-limited) following the most commonly used open-sourced toolbox codes\footnote{\url{https://github.com/vtu81/backdoor-toolbox}}.

\section{Generalizability to Other Model Architectures}
\label{appendix:moremodels}
We evaluate the effectiveness of our defense on additional model architectures including PreActResNet18~\cite{he2016identity}, and MobileNet~\cite{krizhevsky2009learning}. 
The defense performance is presented in~\cref{tab:modelarch}. 
As shown, most of the average AUROC and F1 scores on both architectures are above 0.96, with a few slightly lower scores (still above 0.93). This result indicates that our defense has general applicability across different model architectures.

\section{ROC Curve Comparison with Baseline Defenses}
\label{appendix_auccurve}
In addition to AUROC and F1 scire metrics, 
we also visually compare the ROC curves of competing defense methods against attacks. ROC curves for the CIFAR-10 experiments can be found in~\cref{fig:roc}.

\section{Performance of Our IBD-PSC Against Additional Backdoor Scenarios}
\label{appendix:moreattacks}
\cref{tab:mainother} presents the performance (AUROC and F1 scores) of our IBD-PSC against some other types of backdoor attacks, including clean-label attacks (LC~\cite{turner2019label}, NARCISSUS~\cite{zeng2023narcissus}), source-specific attack (TaCT~\cite{tang2021demon}), training-controlled attack (BPP~\cite{Wangbpp}),  model-controlled attack (SRA~\cite{qi2022towards}), and adaptive attack (Adap-Patch~\cite{qi2023revisiting}). The results demonstrate that IBD-PSC consistently outperforms other defense strategies across almost all types of backdoor attacks. It achieves the highest average scores in both AUROC and F1 metrics, marked in bold, underscoring its superior detection capabilities. This comprehensive evaluation affirms the robustness of IBD-PSC as a formidable defense mechanism in the ever-evolving landscape of backdoor attacks in cybersecurity.

\section{Settings for the Inference Time Comparison}
\label{appendix:time}
 The inference time is critical for this task (\ie, detecting poisoned testing images) because the detection is usually deployed as the `firewall' for online inference. In the case of STRIP, TeCo, SCALE-UP, and our defense, defenders utilize the target model's prediction for defense purposes. This means that both detection and prediction can be carried out simultaneously.

We calculate the inference time of all defense methods under identical and ideal conditions to evaluate efficiency. For example, we assume that defenders will load all required models and images simultaneously, which demands more memory requirements compared to the standard model inference. This comparison is fair and reasonable due to the significant differences in mechanisms and requirements among the various defenses. More precisely, before inference, we engage in preparatory steps such as selecting the BN layers to be amplified and preparing the parameter-amplified models. These models are subsequently deployed across different machines, enabling simultaneous processing of input samples. While this approach requires additional storage space to accommodate the various model versions, it considerably accelerates the detection process. For SCALE-UP, we calculate the inference time needed to obtain predictions for multiple augmented images associated with a given input. This is achieved by concurrently feeding all the images into the deployed model as a batch instead of predicting them individually.



\begin{table}[!t]
\centering
\small
\caption{The performance (AUROC, F1) of our defense on other model architectures.}
\label{tab:modelarch}
\begin{tabular}{lcccccccccc} 
\toprule
\multirow{2}{*}{Datasets}& Models$\rightarrow$ & \multicolumn{2}{c}{PreactResNet18} & \multicolumn{2}{c}{MobileNet} & \multicolumn{2}{c}{Avg.}\\
&Attacks$\downarrow$& AUROC & F1 & AUROC & F1 & AUROC & F1 \\
\hline
\multirow{4}{*}{CIFAR10}&BadNets &  0.978 & 0.931 & 0.970 & 0.943 & 0.974 & 0.937\\
&IAD &0.989  & 0.965 & 0.969 & 0.951 & 0.979 & 0.958\\
&WaNet & 0.977&  0.949 & 0.937& 0.940 & 0.957 & 0.945\\
&BATT & 0.972&0.958 & 0.951 & 0.953 & 0.962 & 0.956\\
\hline
\hline
\multirow{2}{*}{Datasets}& Models$\rightarrow$ & \multicolumn{2}{c}{PreactResNet18} & \multicolumn{2}{c}{MobileNet}  & \multicolumn{2}{c}{Avg.}\\
&Attacks$\downarrow$& AUROC & F1 & AUROC & F1 & AUROC & F1 \\
\hline
\multirow{4}{*}{GTSRB}&BadNets& 0.970&0.971 & 0.969 &0.971 & 0.970 & 0.971\\
&IAD & 0.970 & 0.970& 0.966 & 0.966 & 0.968 & 0.968\\
&WaNet& 0.964 & 0.933 & 0.986 &0.977 & 0.975 & 0.955\\
&BATT& 0.968 & 0.970& 0.970 & 0.957 & 0.969 & 0.964\\
\bottomrule
\end{tabular}
\vspace{-1.2em}
\end{table}

\begin{table}[!t]
\centering
\small
\caption{Performance (AUROC, F1) of our IBD-PSC against various backdoor attacks including clean-label, source-specific, training-controlled, model-controlled, and adaptive attacks. We mark the best result in boldface and failed cases ($<0.7$) in red.}
\label{tab:mainother}
\setlength{\tabcolsep}{2pt}
\begin{tabular}{lcccccccccccccc} 
\toprule
 Attacks$\rightarrow$ & \multicolumn{2}{c}{LC} & \multicolumn{2}{c}{TaCT} & \multicolumn{2}{c}{SRA} & \multicolumn{2}{c}{BPP}  & \multicolumn{2}{c}{NARCISSUS} & \multicolumn{2}{c}{Adap-Patch} &\multicolumn{2}{c}{Avg.}\\
 Defenses$\downarrow$ & AUROC & F1 & AUROC & F1 & AUROC & F1 & AUROC & F1  & AUROC & F1  & AUROC & F1 & AUROC & F1\\ 
\midrule
STRIP & \red{0.668} & \red{0.541} & \red{0.431} & \red{0.106} & \red{0.550} & \red{0.213}&  \red{0.331} & \red{0.081} & \bf{0.952} & \bf{0.949} & 0.858 & 0.715 & \red{0.632} & \red{0.434} \\ 
TeCo & 0.818 & \red{0.685} & \bf{1.000} & 0.946 & 0.933& 0.919&  \bf{0.992} &0.926  & 0.927 & 0.864& 0.947 & 0.948 & 0.940 & 0.908\\
SCALE-UP  & 0.943 & \bf{0.912}  & \red{0.614} & \red{0.234}& \red{0.580} & \red{0.453} & 
 0.860 &0.832& \red{0.673} & \red{0.000} & 0.941 & 0.913 &0.754 & \red{0.496}\\
IBD-PSC & \bf{0.980} & 0.834  & 0.986 & \bf{0.974} &  \bf{0.976} & \bf{0.943} & 0.990& \bf{0.968}  & 0.939 & 0.924 & \bf{0.999} &  \bf{0.961} & \bf{0.978}  
& \bf{0.944}\\
\bottomrule
\end{tabular}
\vspace{-1.2em}
\end{table}

\begin{figure}[!t]
	\centering
 \begin{minipage}{0.90\linewidth}
    \begin{minipage}{0.325\linewidth}
            \includegraphics[width=1\linewidth]{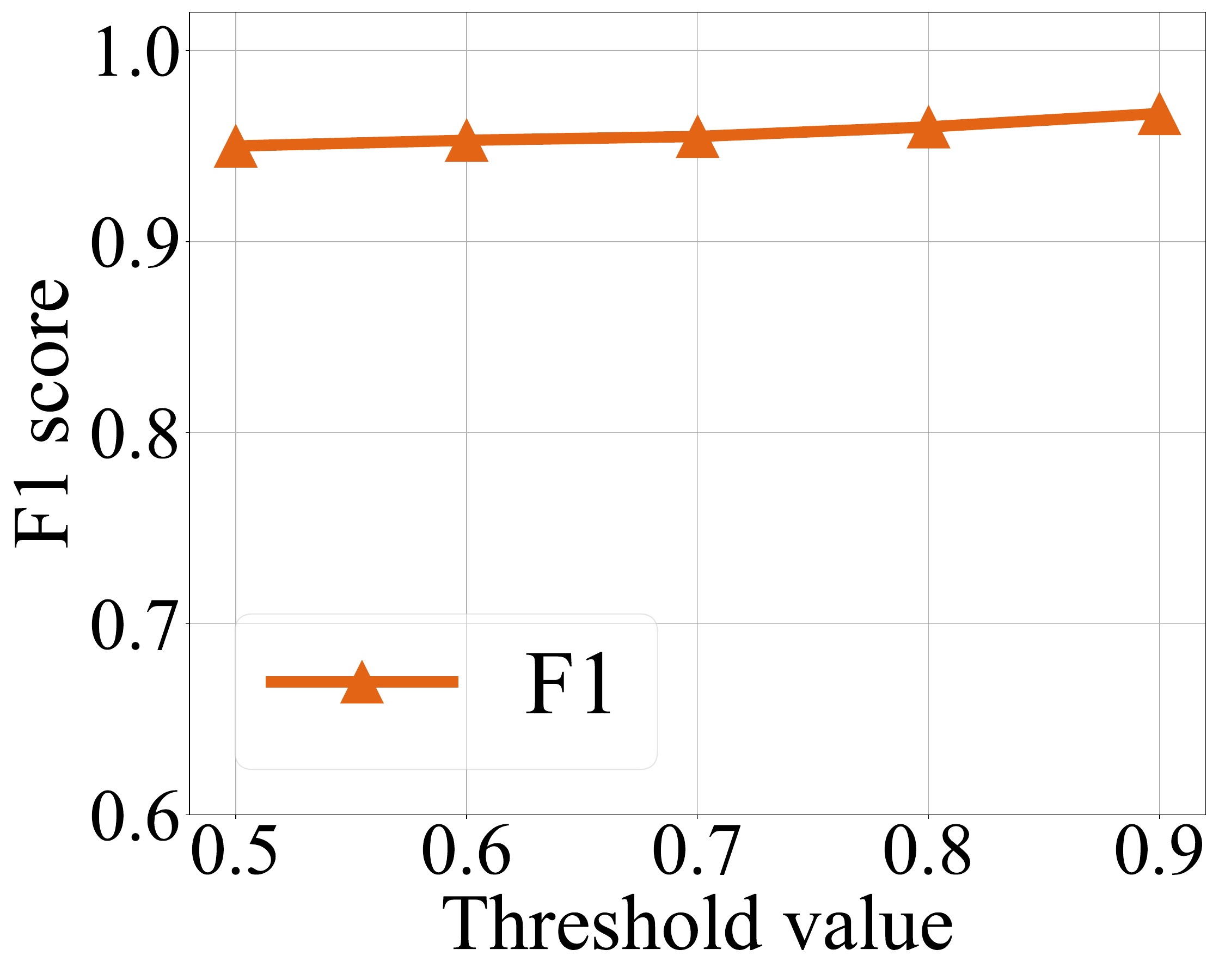}
            \centerline{(a) BadNets}
    \end{minipage}
    \begin{minipage}{0.325\linewidth}
            \includegraphics[width=1\linewidth]{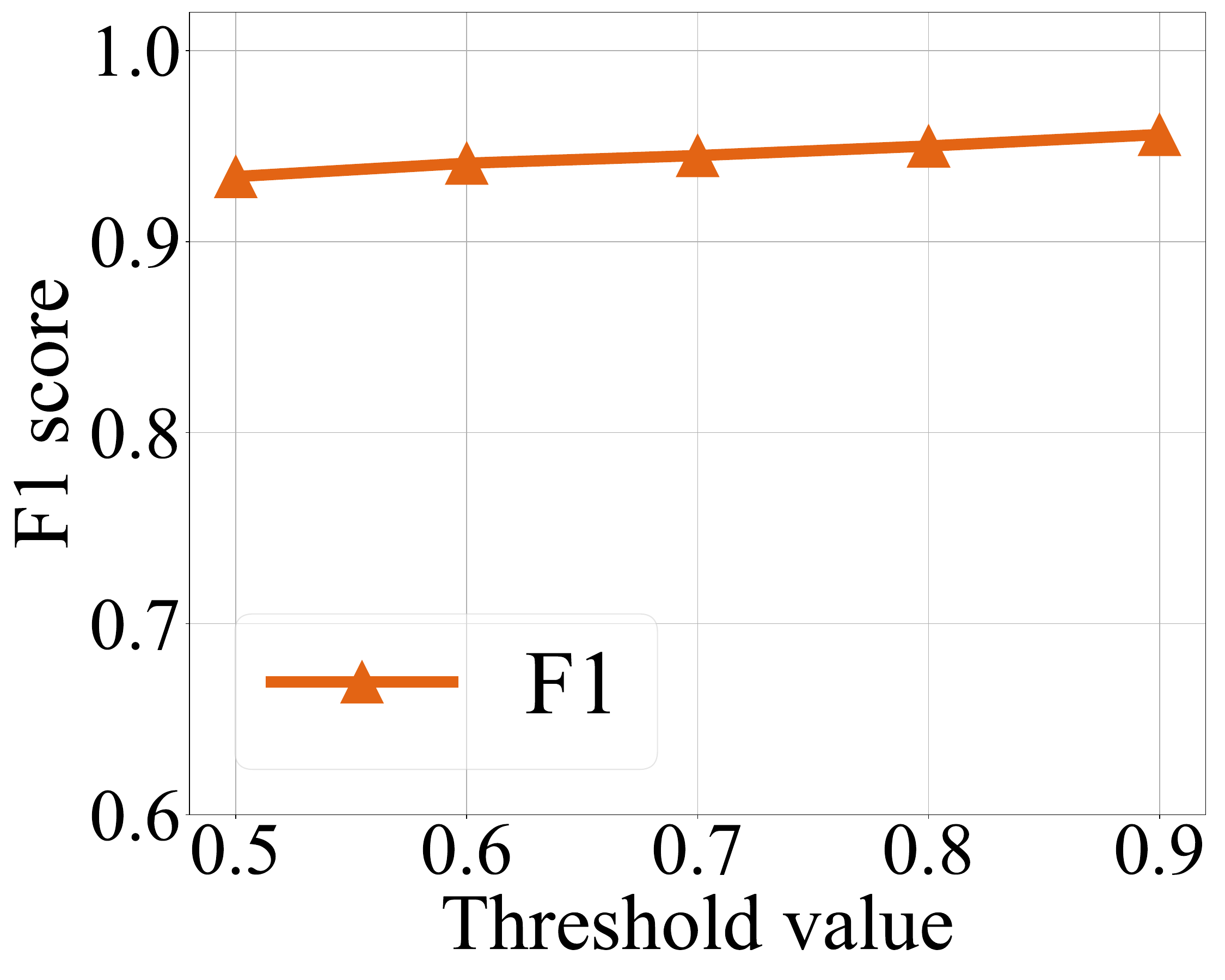}
            \centerline{(b) WaNet}
    \end{minipage}
    \begin{minipage}{0.325\linewidth}
            \includegraphics[width=1\linewidth]{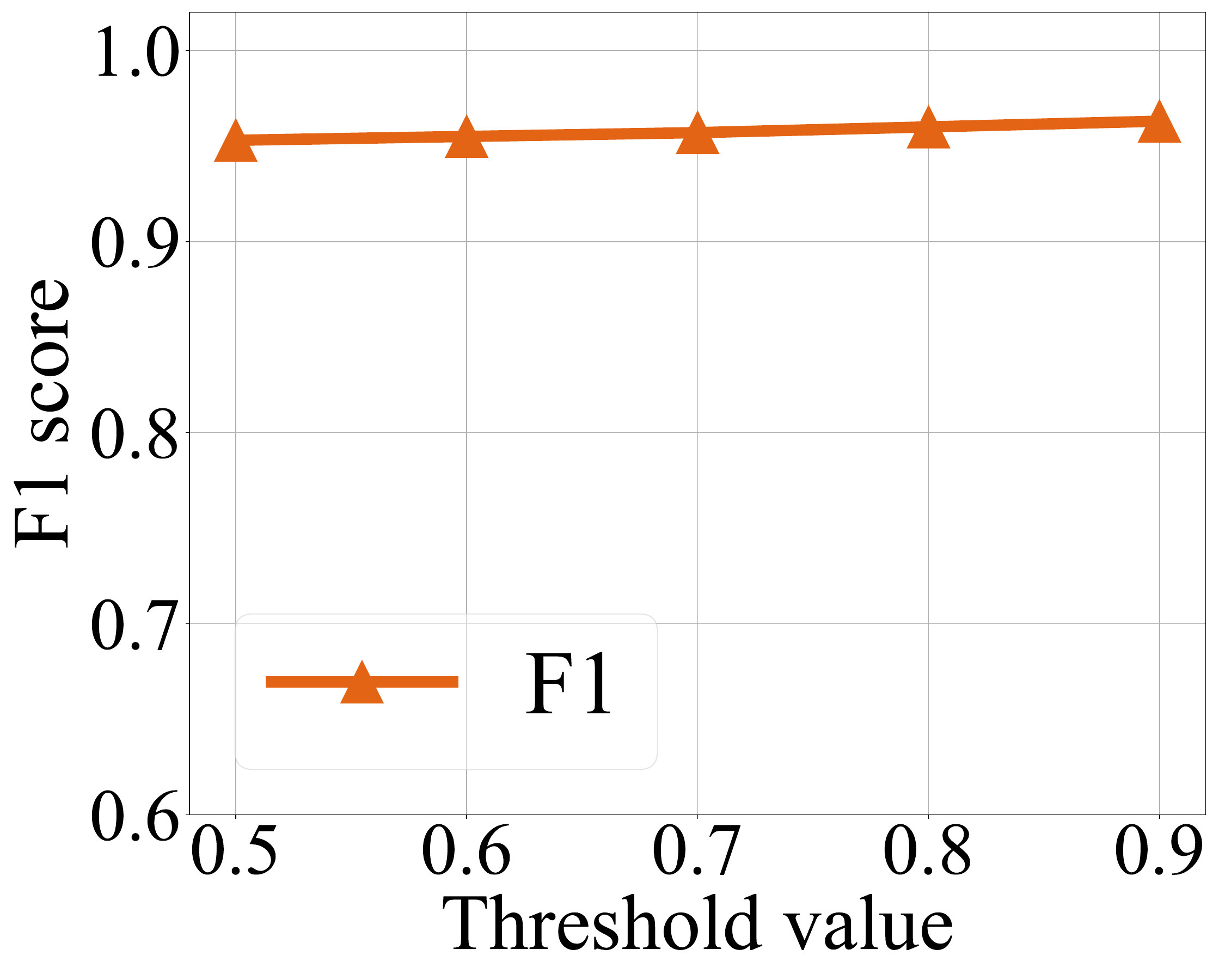}
            \centerline{(c) BATT}
    \end{minipage}
\end{minipage}
 \caption{Impact of the value of threshold $T$ on defense effectiveness.} 
  \label{fig:threshold_t}
  \vspace{-1.5em}
\end{figure}

\begin{figure}[!t]
	\centering
 \begin{minipage}{0.90\linewidth}
    \begin{minipage}{0.325\linewidth}
            \includegraphics[width=1\linewidth]{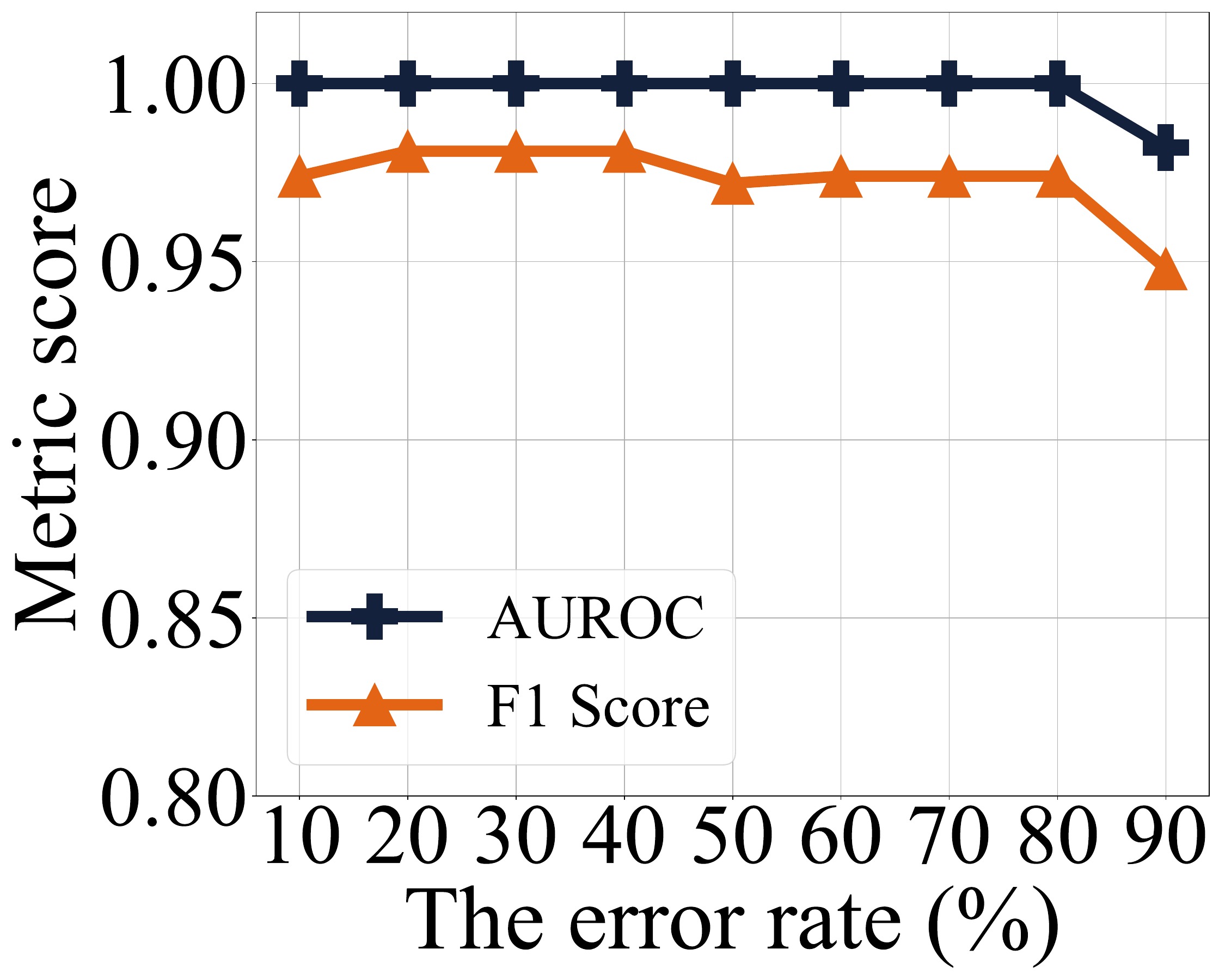}
            \centerline{(a) BadNets}
    \end{minipage}
    \begin{minipage}{0.325\linewidth}
            \includegraphics[width=1\linewidth]{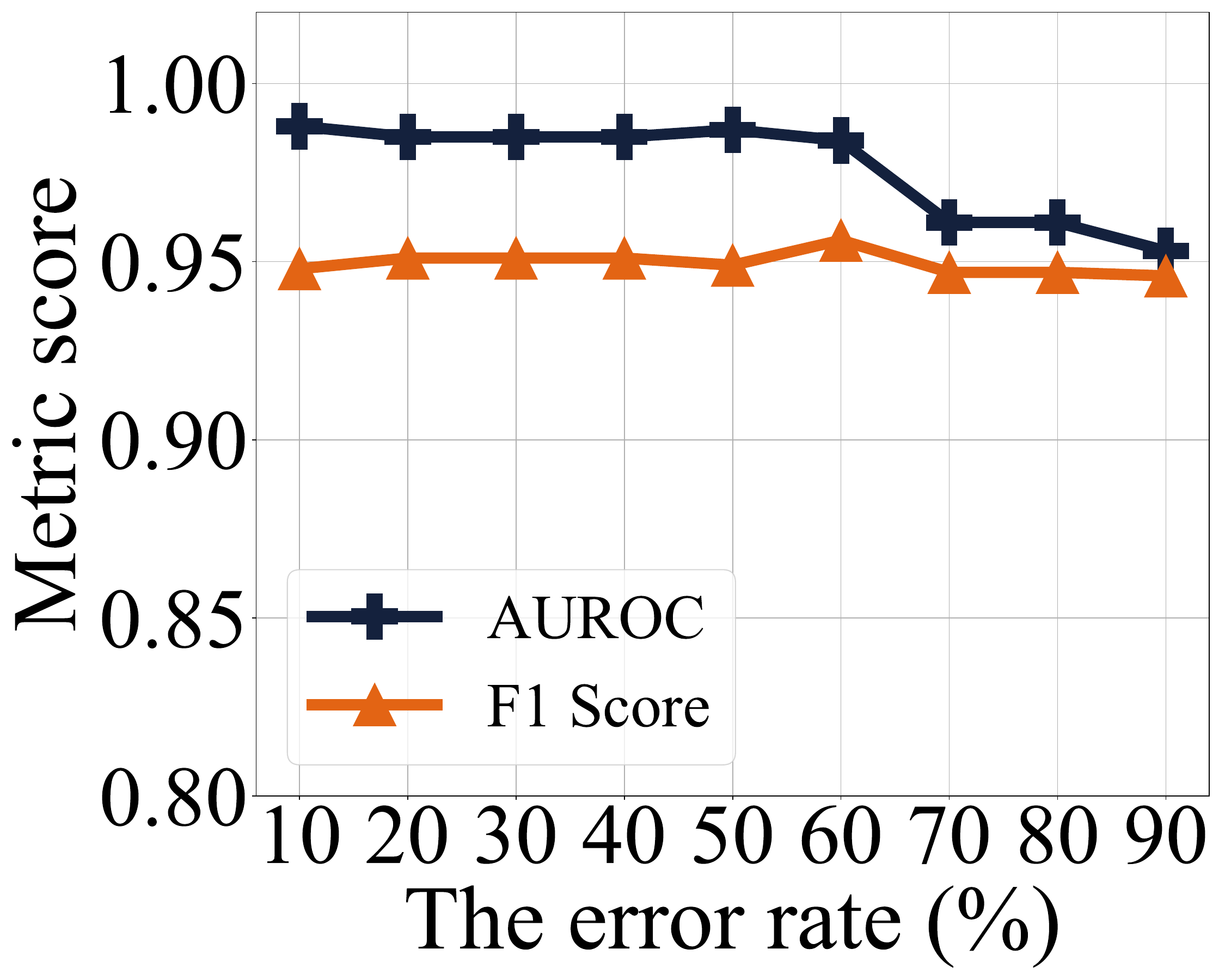}
            \centerline{(b) WaNet}
    \end{minipage}
    \begin{minipage}{0.325\linewidth}
            \includegraphics[width=1\linewidth]{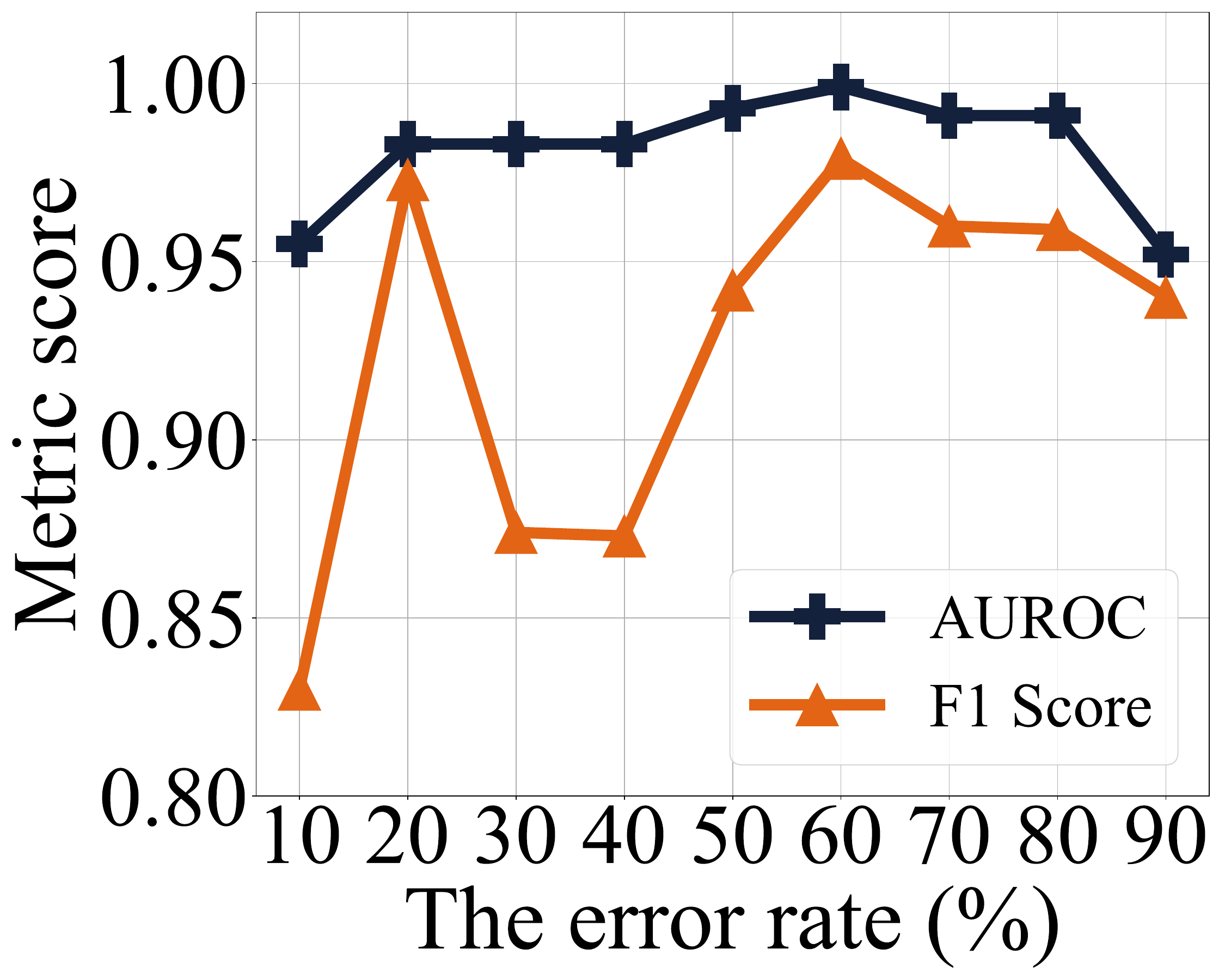}
            \centerline{(c) BATT}
    \end{minipage}
\end{minipage}
 \caption{The impact of error rate $\xi$ on defense effectiveness.} 
 \label{fig:errorrate}
 \vspace{-1.5em}
\end{figure}

\section{Ablation Studies}
\label{appendix:ablation}
\subsection{Impact of the Threshold $T$}
\label{appendix:spcvaluethres}

In our defense, we assess whether an input sample is malicious by comparing its PSC value to a predefined threshold $T$. 
Following the other experiments, we conduct an ablation study of $T$  on three representative attacks: BadNets, WaNet, and BATT on the CIFAR-10 dataset, by adjusting $T$ from 0.5 to 0.9. The results are shown in~\cref{fig:threshold_t}. As we can see, a wide range of values of $T$ can lead to a high F1 score. In our experiment, we set $T$ to 0.9.

\subsection{Impact of the Hyperparameter $\xi$}
\label{appendix:errorrate}
\begin{figure}[!t]
	\centering
 \begin{minipage}{0.90\linewidth}
    \begin{minipage}{0.325\linewidth}
            \includegraphics[width=1\linewidth]{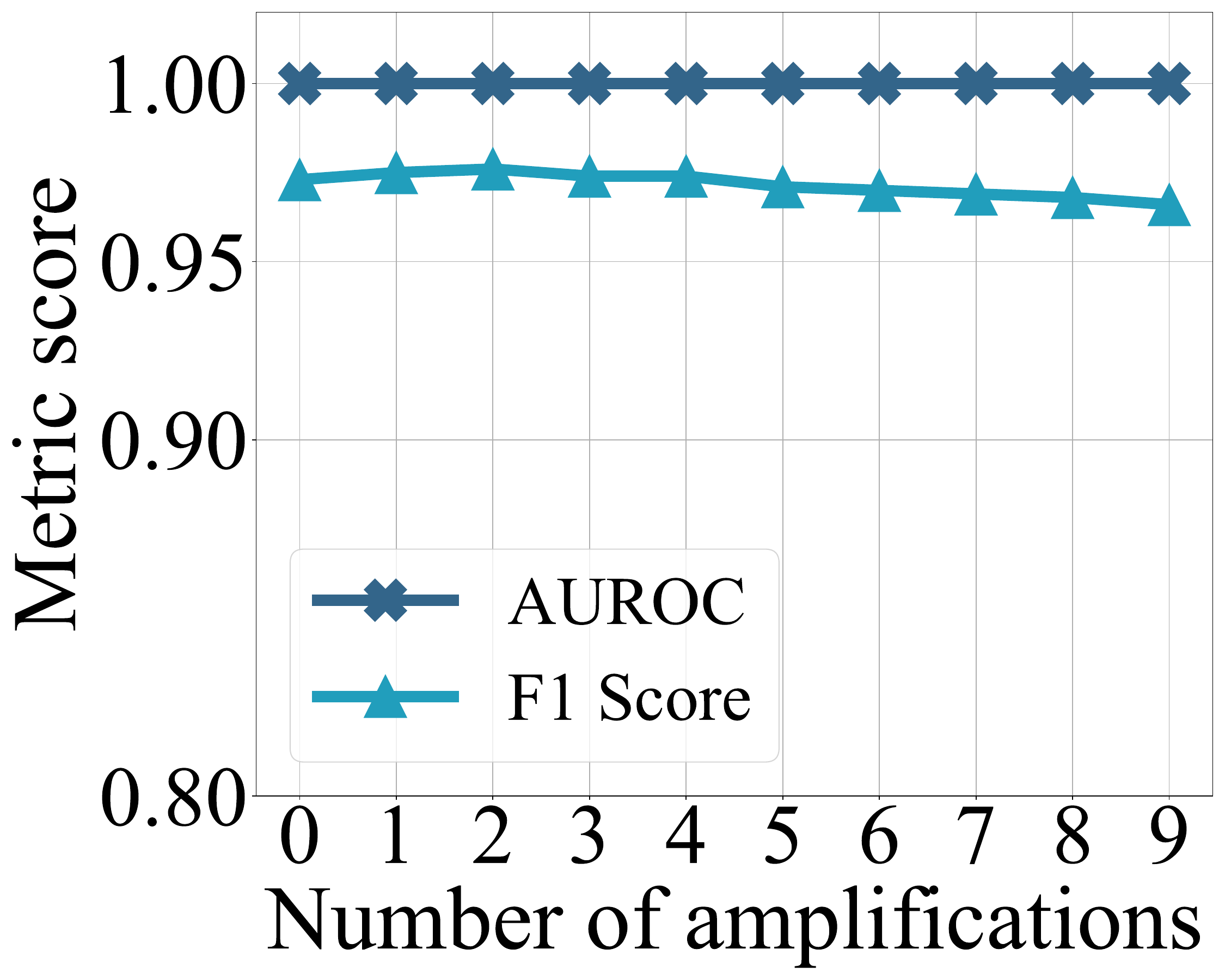}
            \centerline{(a) BadNets}
    \end{minipage}
    \begin{minipage}{0.325\linewidth}
            \includegraphics[width=1\linewidth]{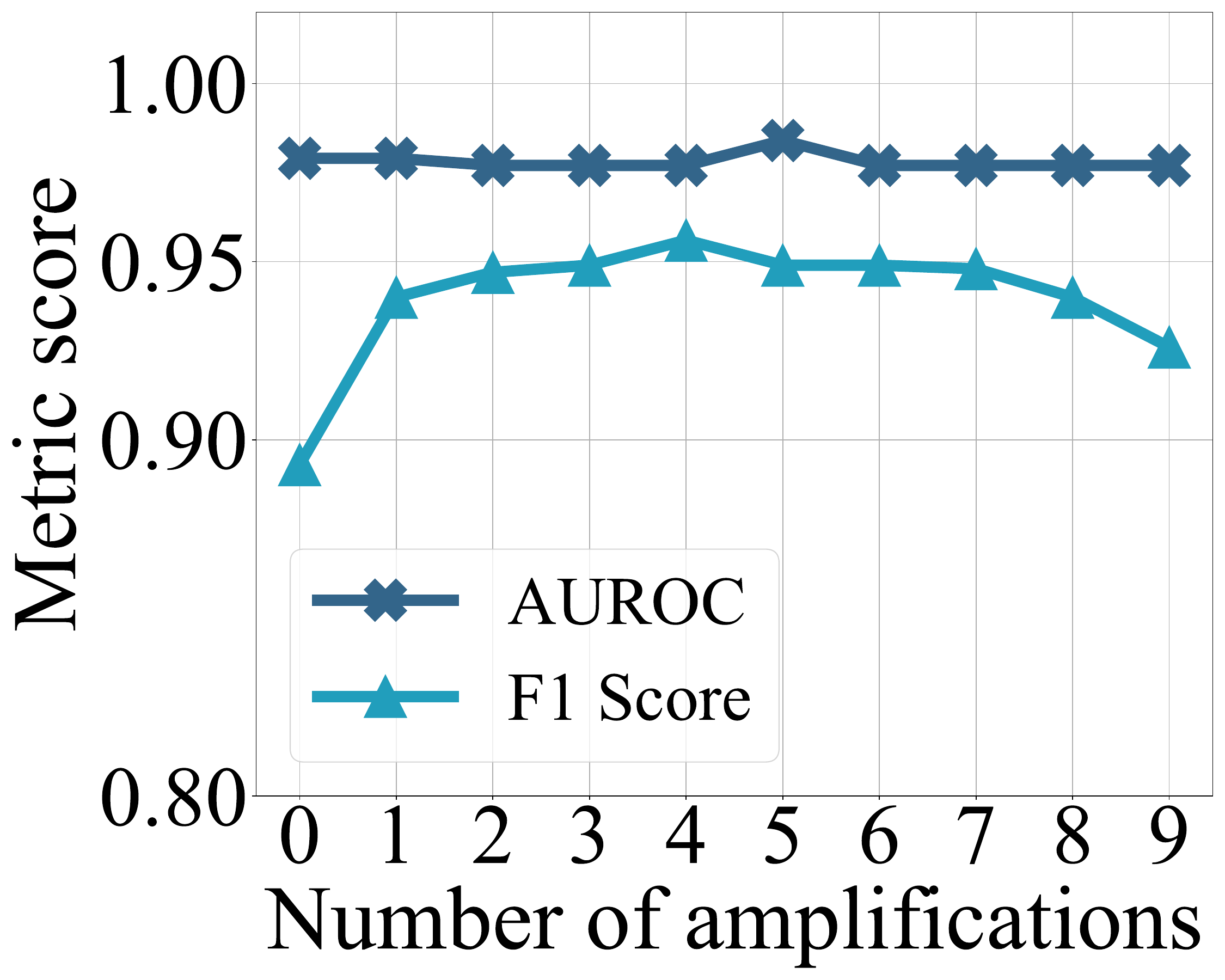}
            \centerline{(b) WaNet}
    \end{minipage}
    \begin{minipage}{0.325\linewidth}
            \includegraphics[width=1\linewidth]{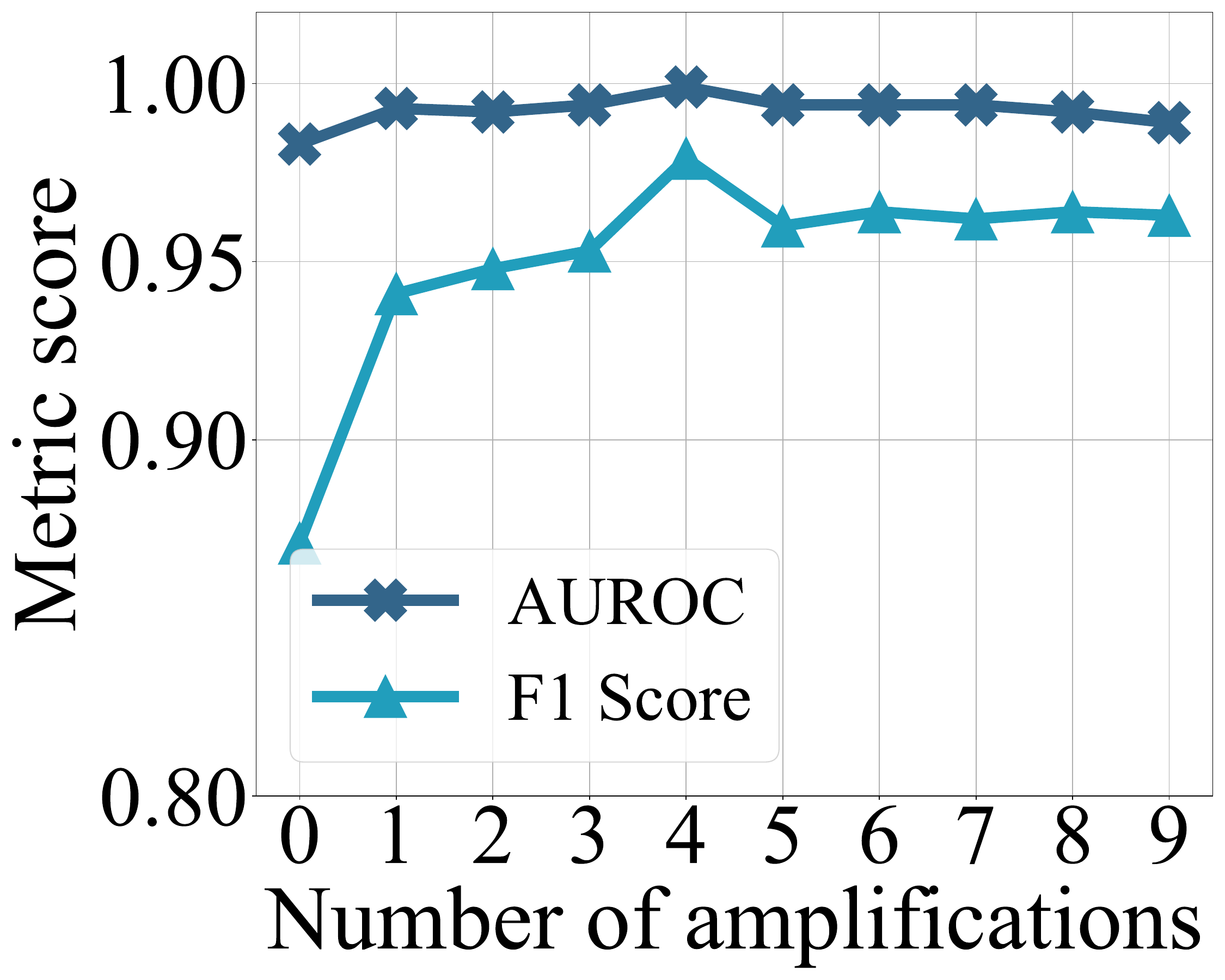}
            \centerline{(c) BATT}
    \end{minipage}
\end{minipage}
    \vspace{-2mm}
 \caption{Impact of the number of amplifications ($n$) on defense effectiveness.} 
 \label{fig:multiplicationtime}
     \vspace{-0.5em}
\end{figure}

In our defense, we design an adaptive algorithm to dynamically select a suitable number of the BN layers to be amplified. The algorithm uses a predefined hyperparameter error rate threshold $\xi$. Here we empirically show that our defense is insensitive to changes in $\xi$.
Again, this is demonstrated on three representative attacks: BadNets, WaNet, and BATT, with varying values of $\xi$ from 10\% to 90\%. ~\cref{fig:multiplicationtime} shows that the defense performance against BadNets and WaNet attacks exhibits remarkable resilience to variations in $\xi$.
While the BATT attack does manifest a more pronounced response to changes in $\xi$, with the F1 score experiencing fluctuations, the metric is eventually higher when the error rate reaches approximately 60\%. This observation signals that the overall influence of the error rate on the defense efficacy remains limited. Consequently, we advocate for an error rate of around 60\%, as it appears to strike a judicious balance, ensuring adequate detection accuracy without unduly compromising the defense strategy against the assessed backdoor threats.

\subsection{Impact of the Number of Amplified Models $n$}
\label{appendix:timesn}

In our defense, we build a profile of $n$ progressively amplified models,
to capture the model's dynamic response to such interventions.
In practice, the number of amplifications $n$ is a defender-assigned hyper-parameter. 
As illustrated in~\cref{fig:multiplicationtime}, the detection performance under the BadNets attack exhibits consistency across various values of \( n \), suggesting a relative insensitivity to the number of amplifications. In contrast, for the WaNet and BATT attacks, there is an improvement in detection effectiveness as $n$ increases, which plateaus when $n$ reaches five. This stabilization suggests an optimal defense performance, and thus we establish $n = 5$ as the optimal value for our defense, ensuring stable detection performance.

\subsection{Impact of the Target Class}
\label{appendix:labels}
\begin{figure}[!t]
	\centering
 \begin{minipage}{0.90\linewidth}
    \begin{minipage}{0.32\linewidth}
            \includegraphics[width=1\linewidth]{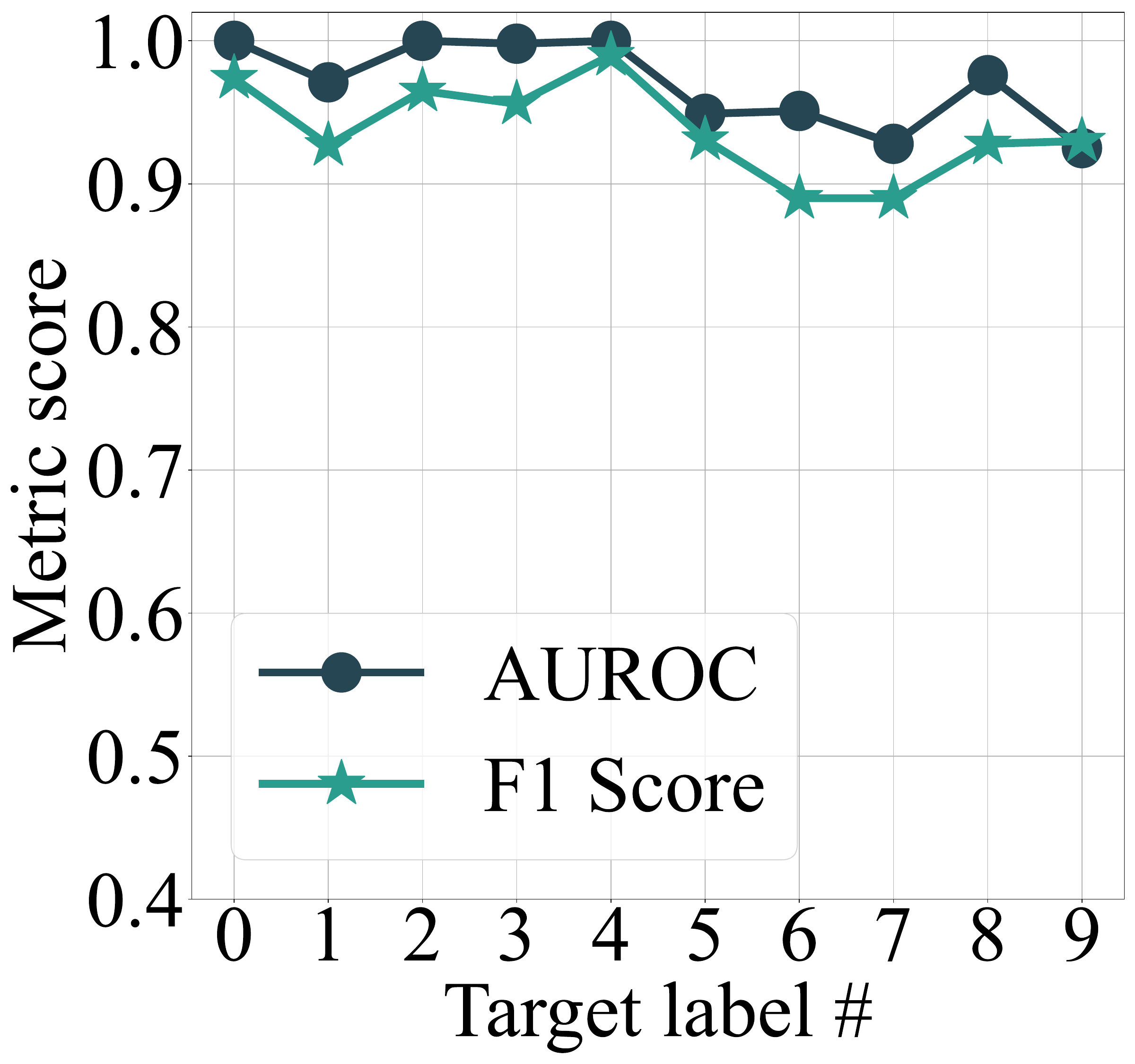}
            \centerline{(a) BadNets}
    \end{minipage}
    \begin{minipage}{0.32\linewidth}
            \includegraphics[width=1\linewidth]{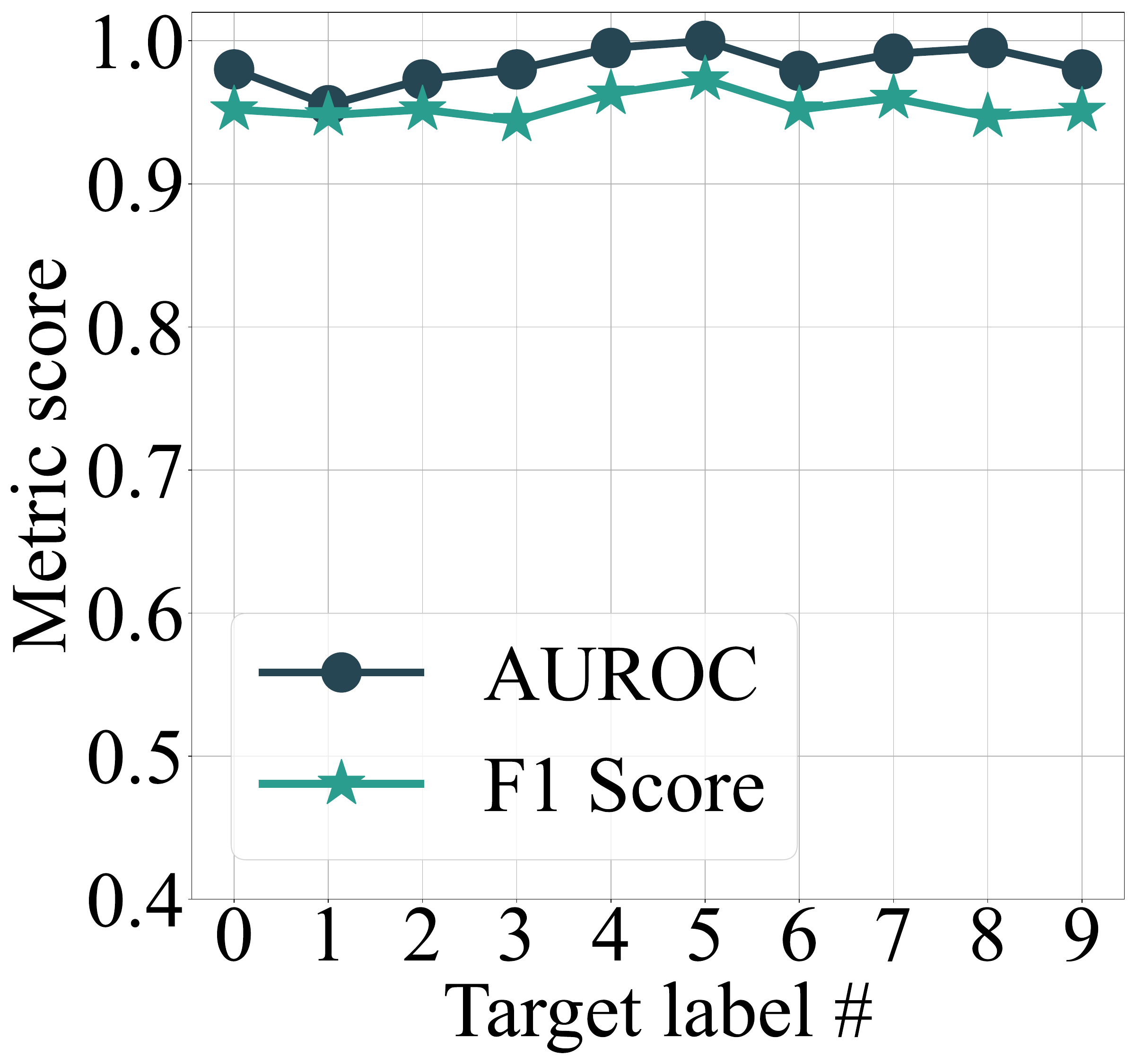}
            \centerline{(b) IAD}
    \end{minipage}
    \begin{minipage}{0.32\linewidth}
            \includegraphics[width=1\linewidth]{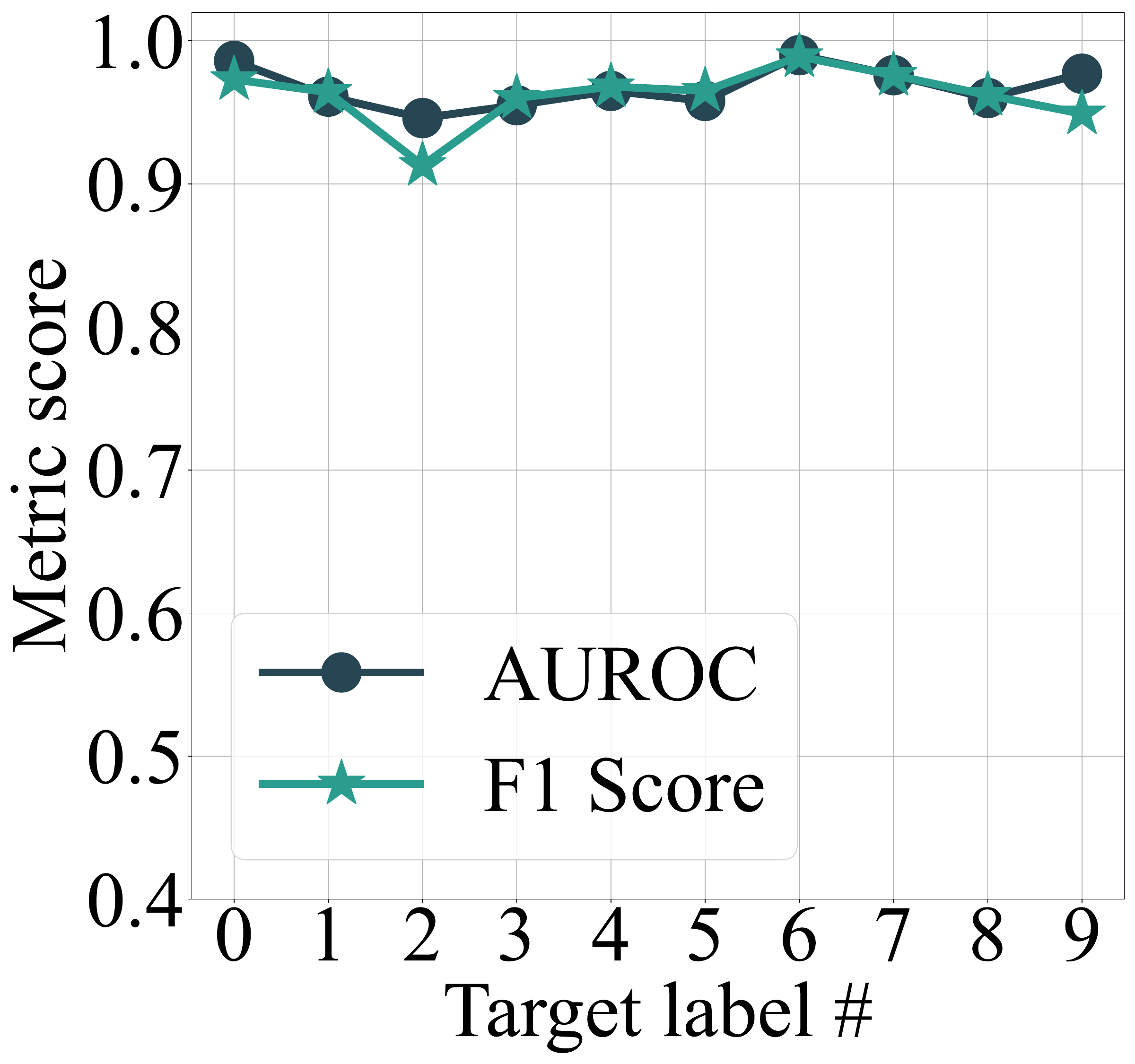}
            \centerline{(c) WaNet}
    \end{minipage}
\end{minipage}
 \vspace{-0.5em}
 \caption{Performance of our defense across 10 target labels of CIFAR10.} 
 \label{fig:targetlabel}
 \vspace{-0.8em}
\end{figure}

We further evaluate the robustness of our defense to the changes of the target class. We select three attacks, including the patch-based, dynamic, and physical backdoor attacks mentioned above, and apply them to target each of the ten labels of CIFAR-10. We display the AUROC and F1 scores of our defense against these backdoored models in~\cref{fig:targetlabel}. As shown, our defense demonstrates consistent performance against different attacks and target labels. Specifically, the AUROC and F1 scores are consistently close to 1, with the average AUROC and F1 scores of each attack all exceeding 0.96 and 0.94, respectively. This indicates that our defense maintains strong performance against different types of attacks and target labels. Additionally, the standard deviations of AUROC and F1 scores across different cases are generally below 0.02.

\begin{figure}[!t]
	\centering
 \begin{minipage}{0.90\linewidth}
    \begin{minipage}{0.32\linewidth}
            \includegraphics[width=1\linewidth]{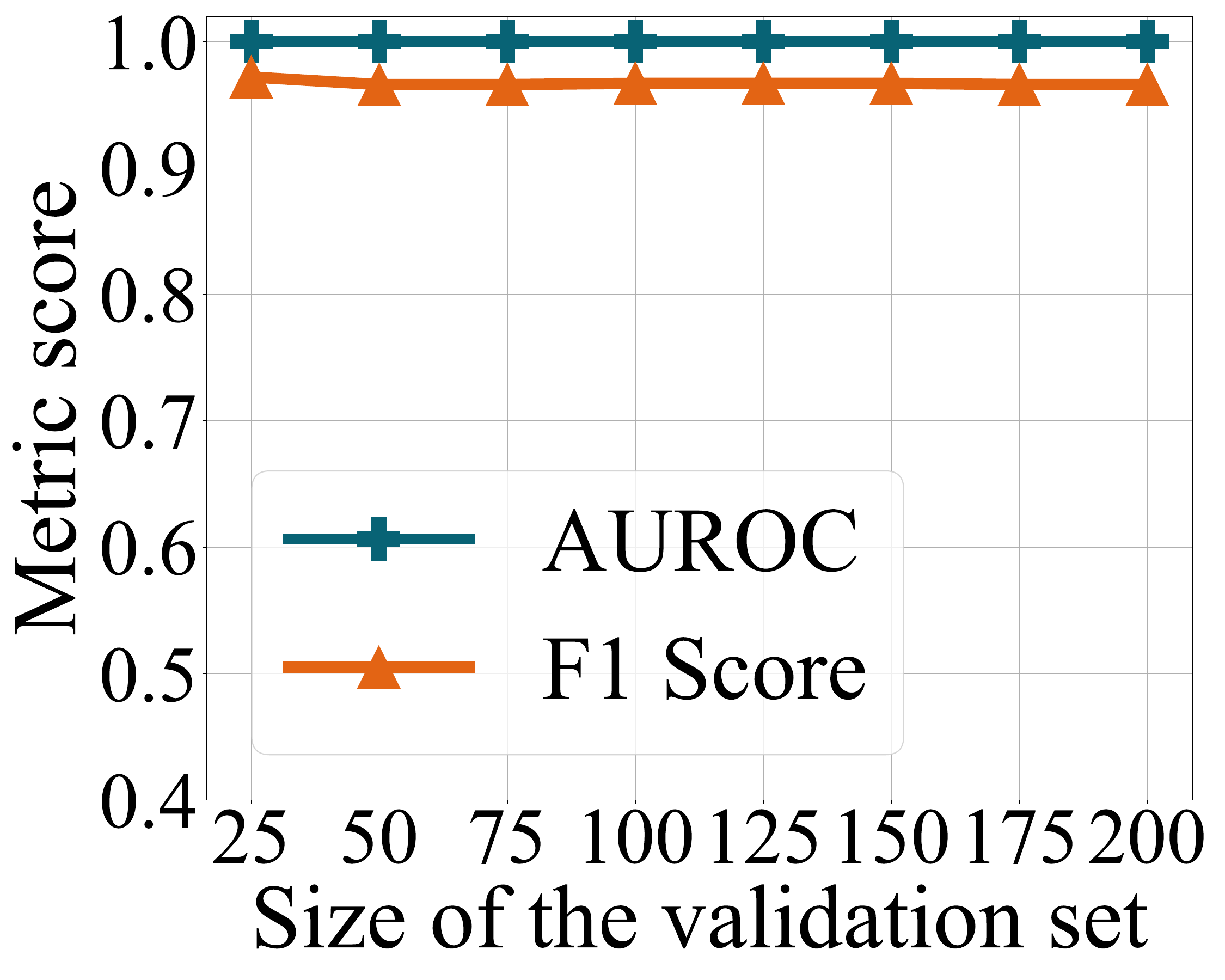}
            \centerline{(a) BadNets}
            \includegraphics[width=1\linewidth]{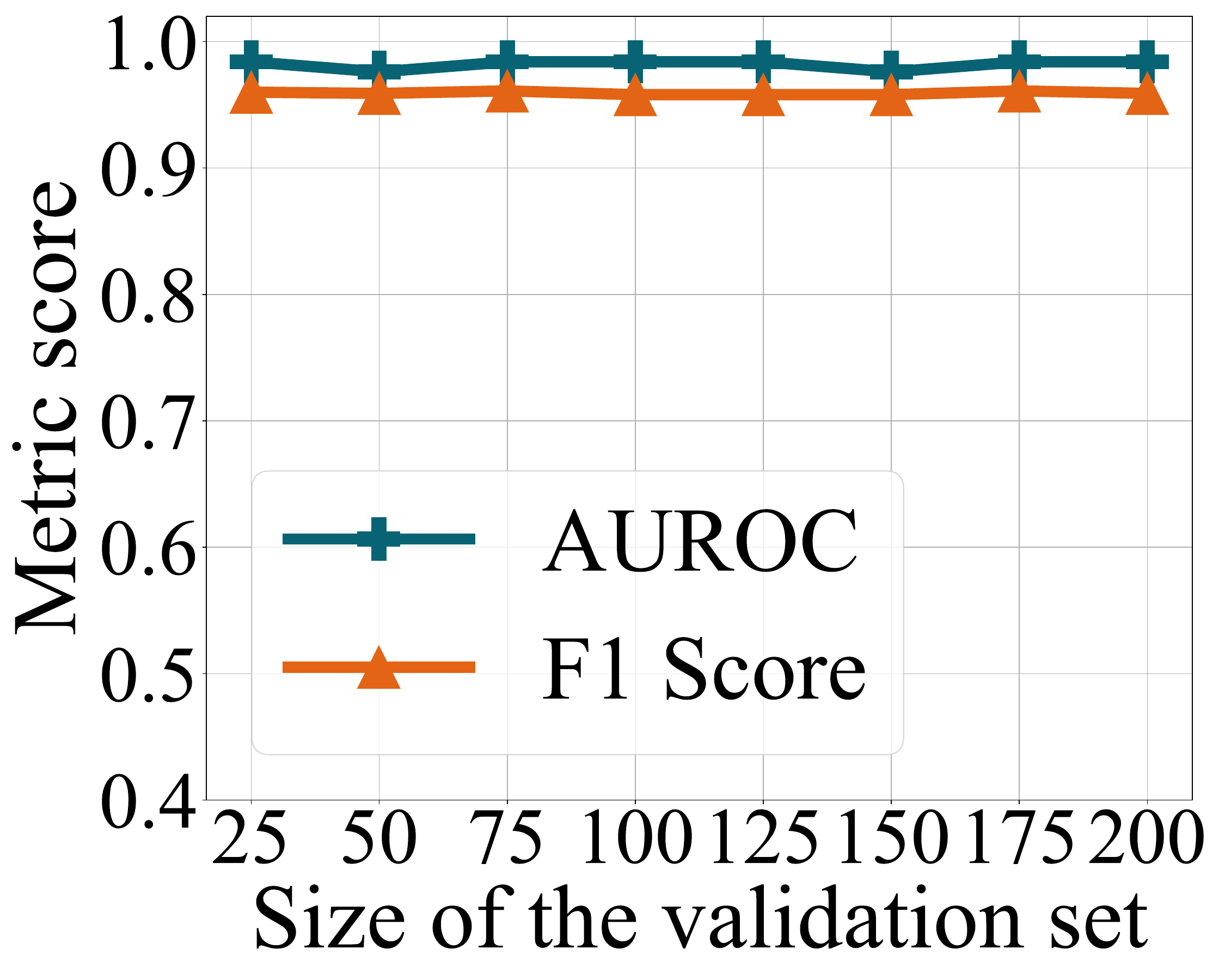}
            \centerline{(d) BATT}
    \end{minipage}
    \begin{minipage}{0.32\linewidth}
            \includegraphics[width=1\linewidth]{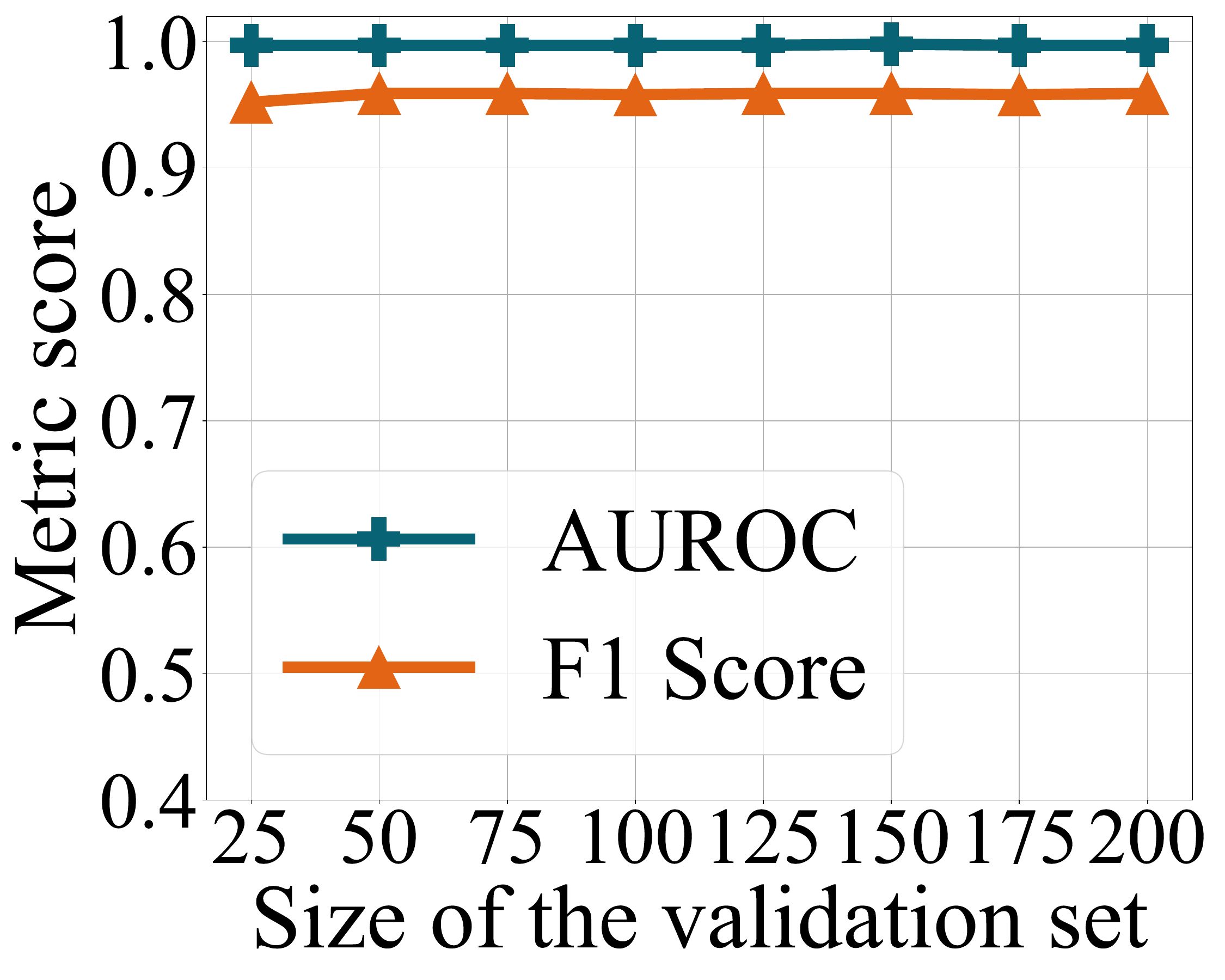}
            \centerline{(b) IAD}
            \includegraphics[width=1\linewidth]{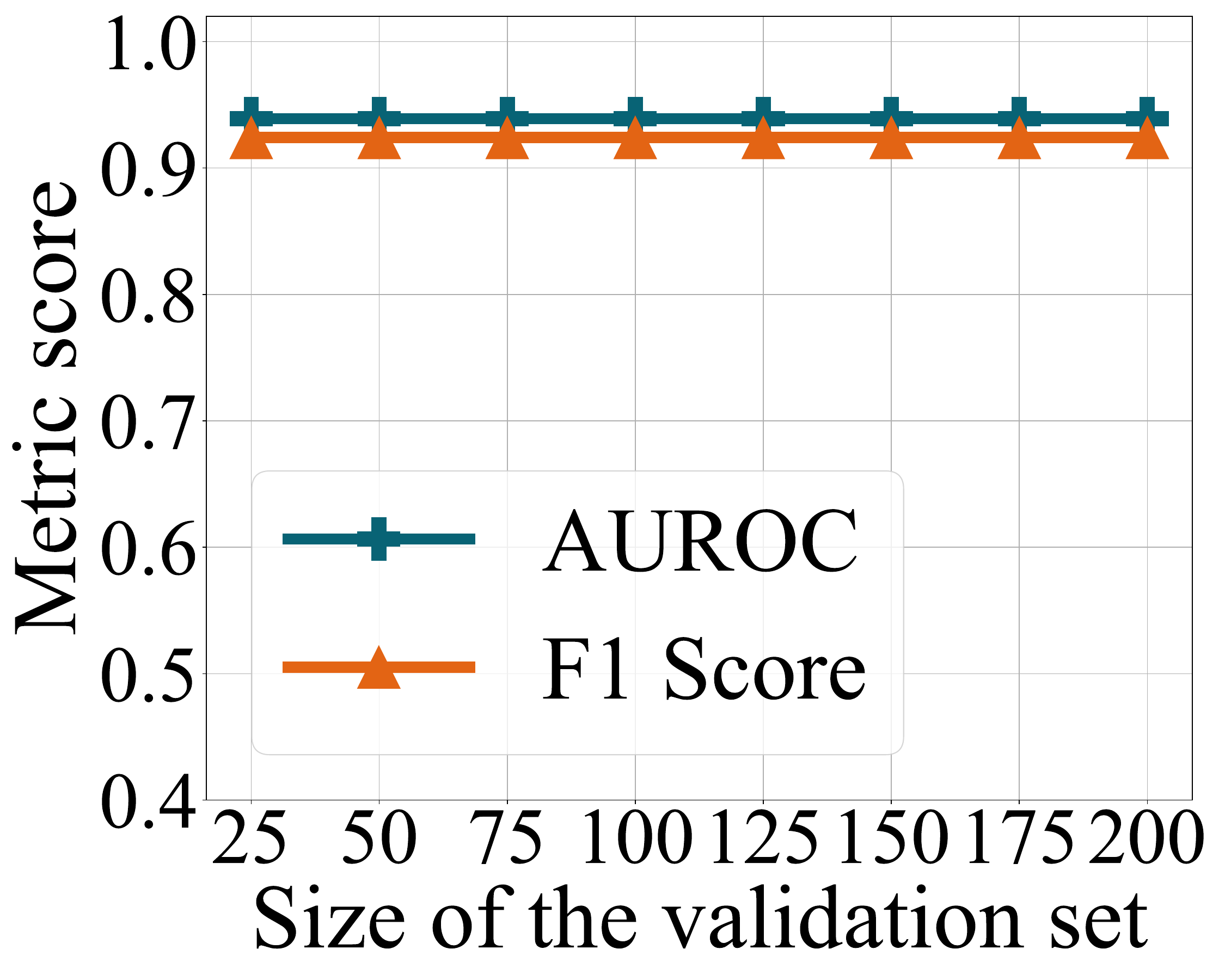}
            \centerline{(e) NARCISSUS}
    \end{minipage}
    \begin{minipage}{0.32\linewidth}
            \includegraphics[width=1\linewidth]{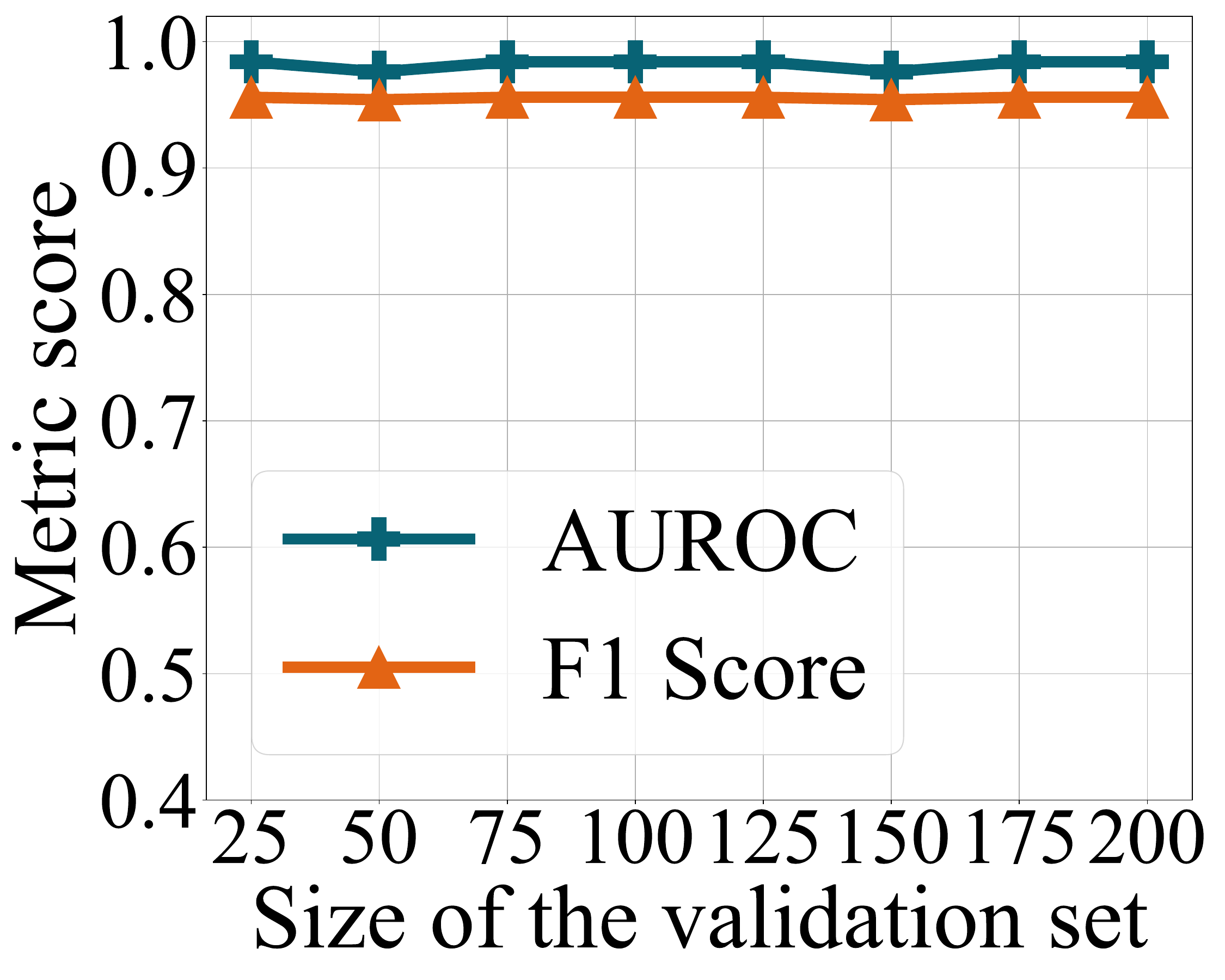}
            \centerline{(c) WaNet}
            \includegraphics[width=1\linewidth]{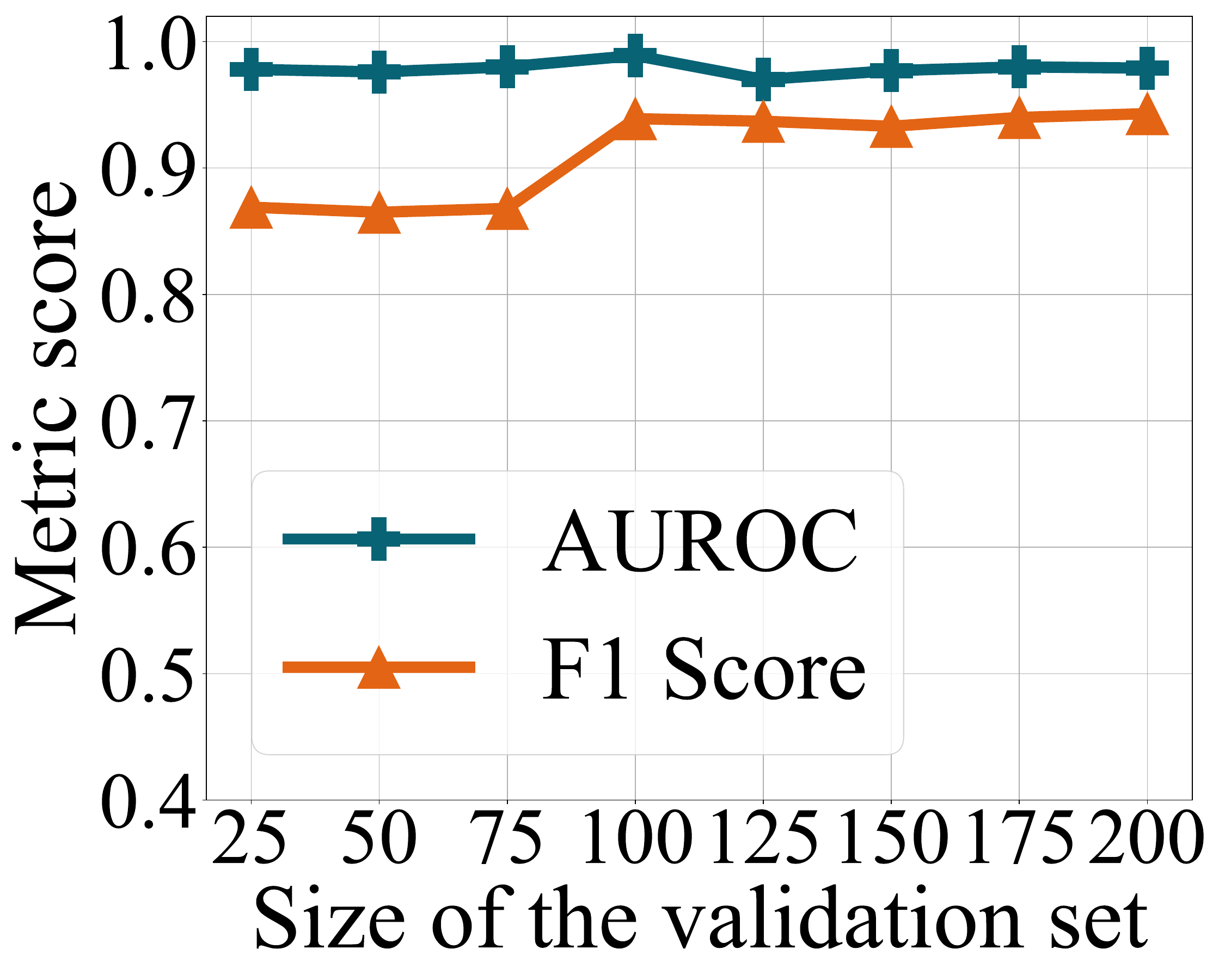}
            \centerline{(f) SRA}
    \end{minipage}
\end{minipage}
 \caption{The impact for the size of the validation set.} 
 \label{fig:impact_size}
 \vspace{-1.5em}
\end{figure}

\begin{table}[!t]
\centering
\small
\caption{The false positive rate (\%) of our defense on benign models on the CIFAR-10 dataset.}
\label{tab:fpr}
\begin{tabular}{lccccccccccc} 
\toprule
Dataset & Model 1 & Model 2 & Model 3 & Model 4 & Model 5\\
\midrule
CIFAR-10& 2.990 & 3.00 & 2.540 & 2.240 &2.100 \\
GTSRB & 6.050& 6.270 & 6.950 & 6.310 & 6.150 \\
SubImageNet-200& 0.290 & 0.370& 0.370 & 0.320 & 0.320 \\
\bottomrule
\end{tabular}
\vspace{-1em}
\end{table}

\subsection{Impact of the Size of Local Benign Samples $\mathcal{D}_r$}
\label{appendix:valsize}

Following similar studies in the literature~\cite{wang2019neural,guo2023scale}, we assume that defenders possess a small benign dataset $\mathcal{D}_r$ to calibrate
different model parameters.
By default, $\mathcal{D}_r$ contains merely 100 samples. 
We evaluate the robustness of our defense to the change in the size of $\mathcal{D}_r$. The results on the CIFAR-10 dataset using ResNet18 against six attacks on the CIFAR-10 dataset are shown in~\cref{fig:impact_size}. It is evident that for most attacks, including BadNets, IAD, WaNet, BATT, and NARCISSUS, the detection performance remains consistently high and relatively stable across varying sizes of $\mathcal{D}_r$. For the SRA attack, the F1 score increases and stabilizes when the size of $\mathcal{D}_r$ reaches 100. Overall, this demonstrates that our defense is effective with as few as 100 benign samples.

\subsection{False Positive Rates on Benign Models}
\label{appendix:benignmodel}
In this study, we investigate a scenario in which a defender obtains a third-party DNN model but cannot determine whether the model is compromised with backdoors. To ensure security, it is common to deploy an input-level backdoor detection system, similar to a network firewall, to filter potentially poisoned samples. In such a context, evaluating the impact of the deployed defenses on benign models is crucial. 

To achieve this objective, we train five benign models using different random seeds. Subsequently, we conduct tests on these models to calculate the false positive rate of our defense, which represents the proportion of benign samples incorrectly identified as backdoor samples. These benign samples are incorrectly rejected by the defense during inference. \cref{tab:fpr} presents the false positive rates of our defense on different benign models trained on various datasets. We can observe variations in the false positive rate among different models, but overall, it remains relatively low (below 3\% on the CIFAR-10 and 1\% on the SubImageNet-200 dataset, and around 6\% on the GTSRB dataset). 

We attribute the higher false positive rate on GTSRB to the relative simplicity of image features in this dataset, making the models more prone to overfitting. Consequently, when amplifying model weights, some benign samples may not decrease their prediction confidence due to overfitting. This overfitting phenomenon on GTSRB has also been reported in the SCALE-UP defense~\cite{guo2023scale}. However, in the real world, datasets are often more similar to the ImageNet dataset, characterized by its comprehensive and rich feature information. Our defense performs best on SubImageNet-200, achieving an error rate of less than 1\%.


\begin{figure}[!t]
\centering
\subfigure[BadNets]{\includegraphics[width=0.27\textwidth]{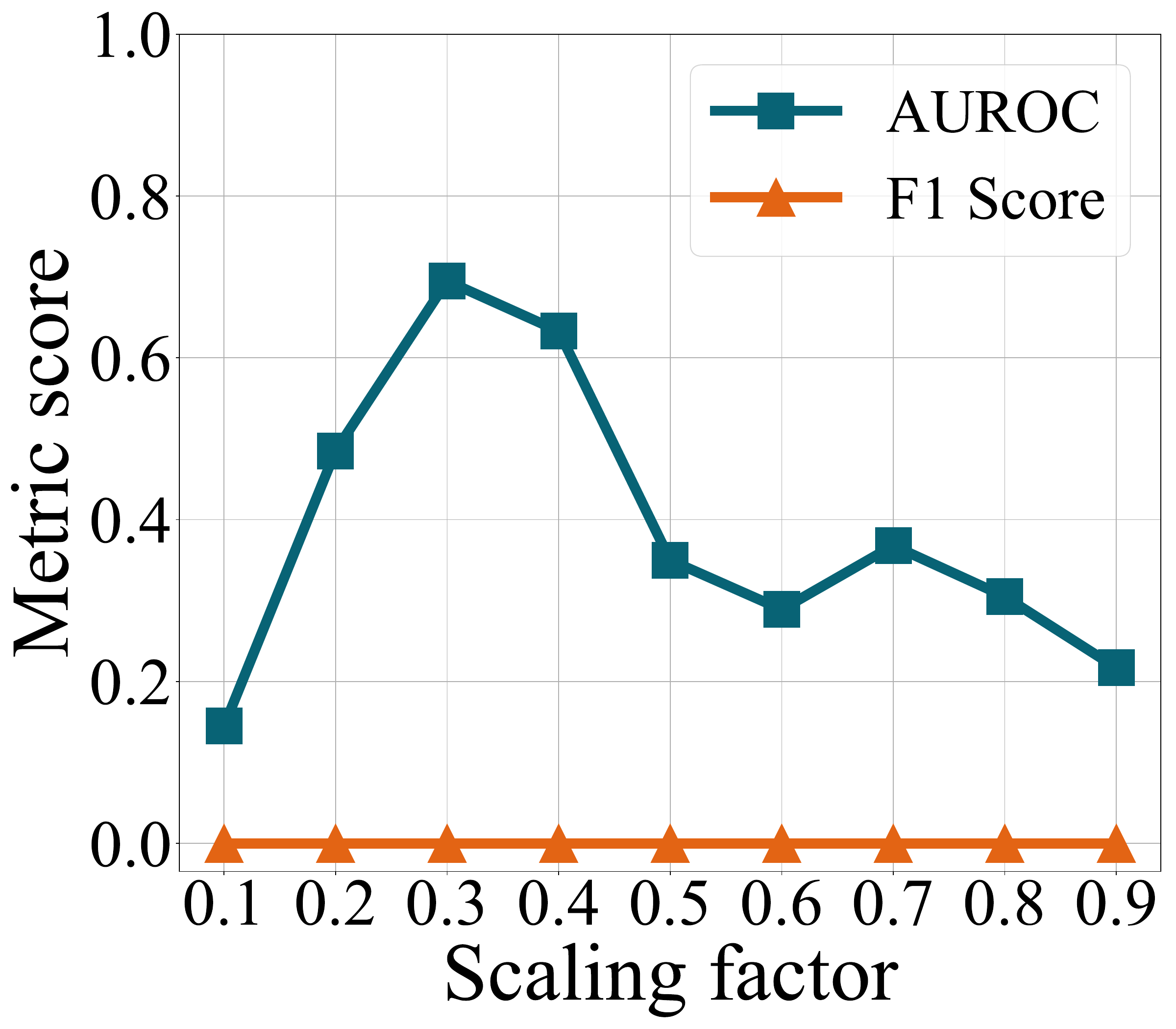}}
\subfigure[WaNet]{\includegraphics[width=0.27\textwidth]{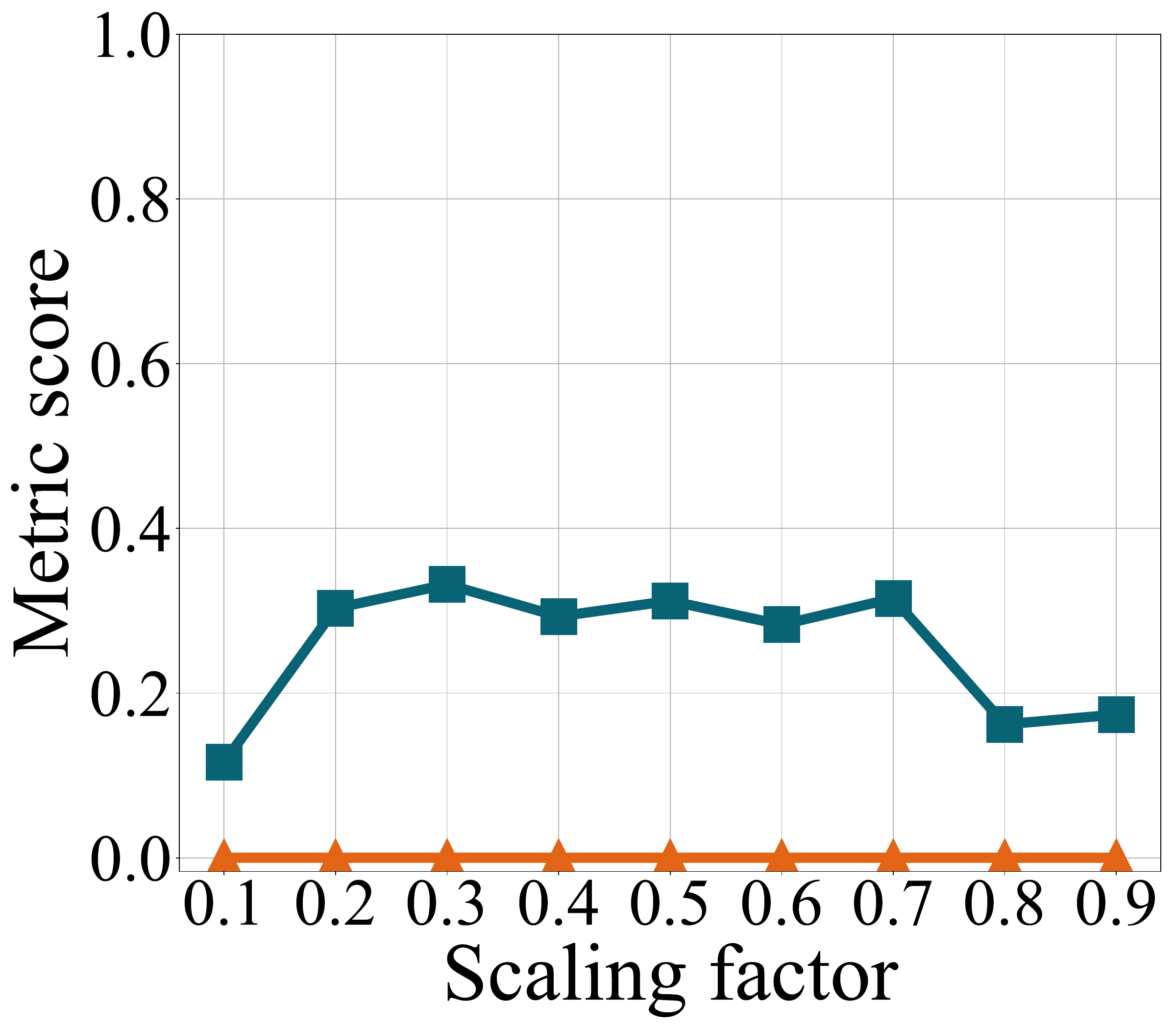}}
\subfigure[BATT]{\includegraphics[width=0.27\textwidth]{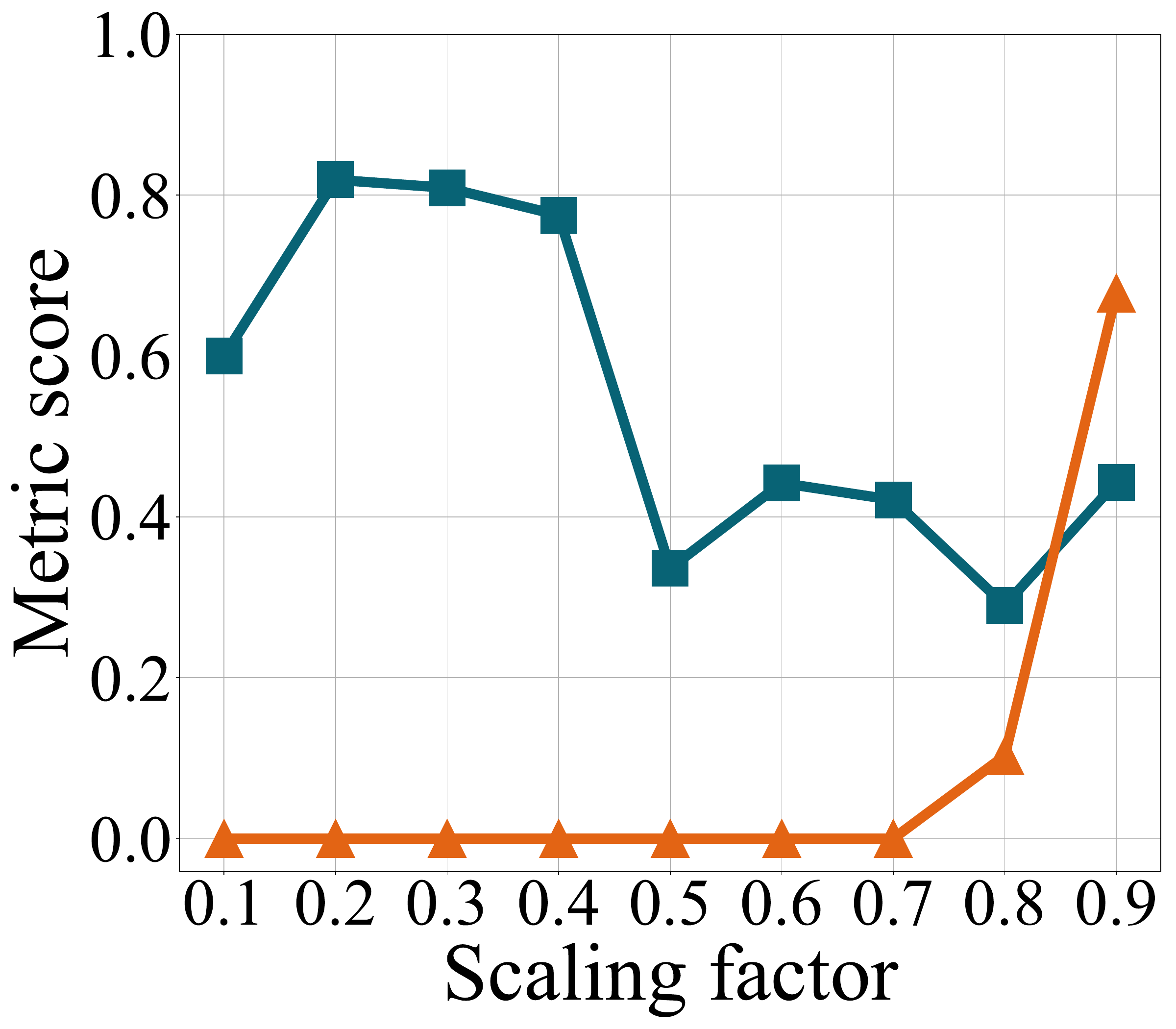}}
\vspace{-1.2em}
 \caption{The impact of scaling factor (smaller than 1.0) on defense effectiveness.} 
  \label{fig:magsmall}
  \vspace{-1em}
\end{figure}

\begin{figure}[!t]
\centering
\subfigure[Benign Model\label{fig:benign_l2_small1}]{\includegraphics[width=0.24\textwidth]{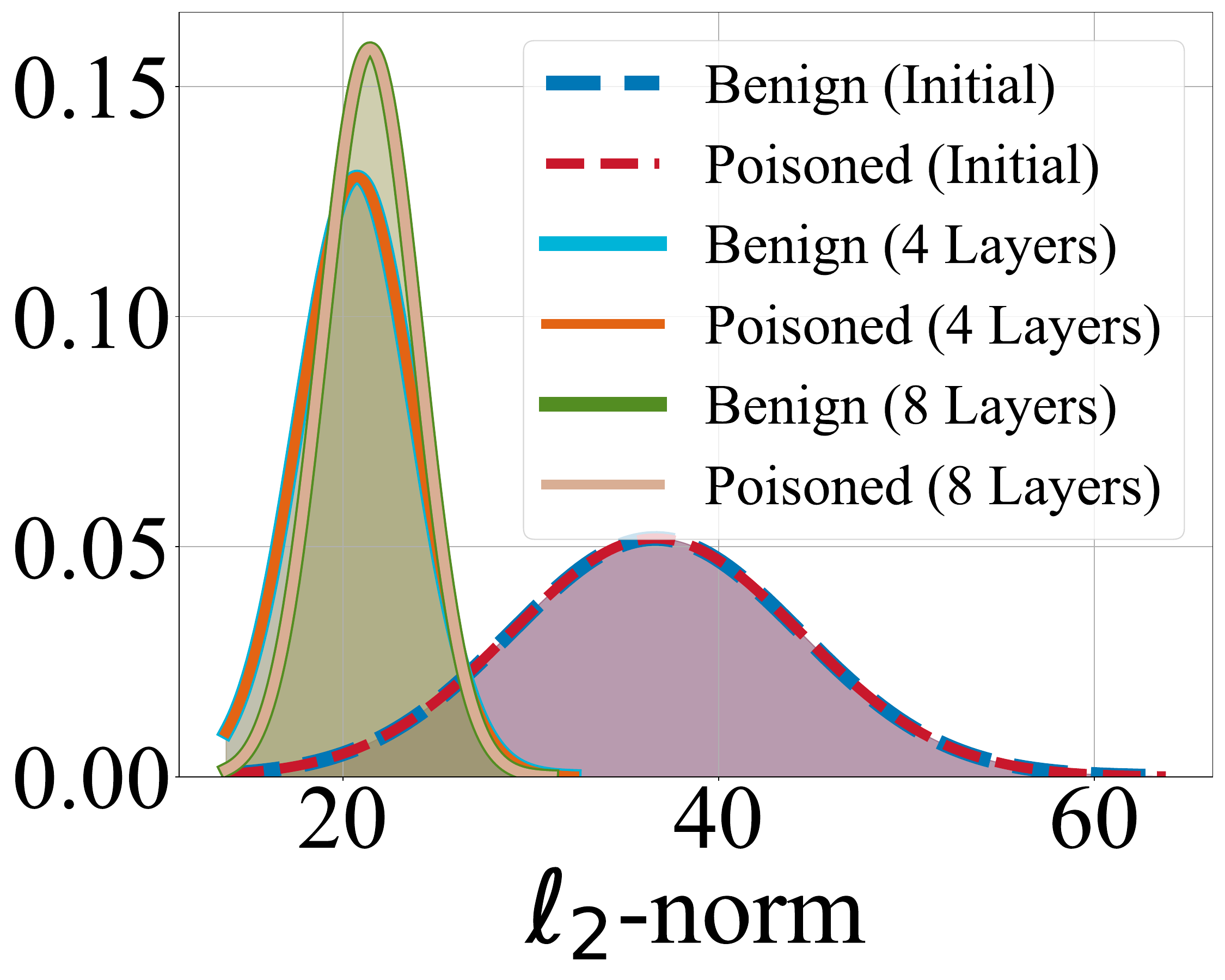}}
\subfigure[BadNets\label{fig:badnets_l2_small1}]{\includegraphics[width=0.24\textwidth]{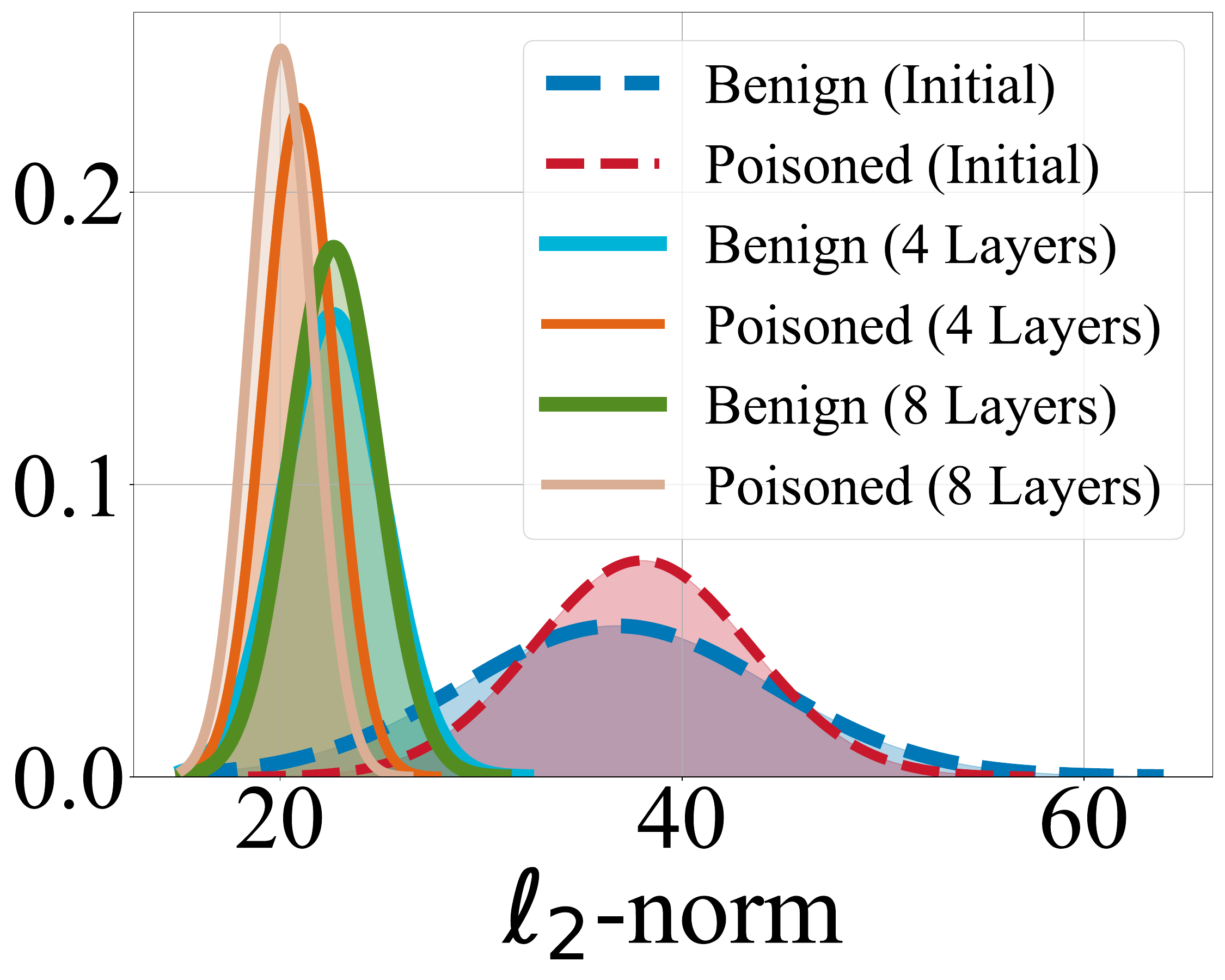}}
\subfigure[WaNet\label{fig:wanet_l2_small1}]{\includegraphics[width=0.24\textwidth]{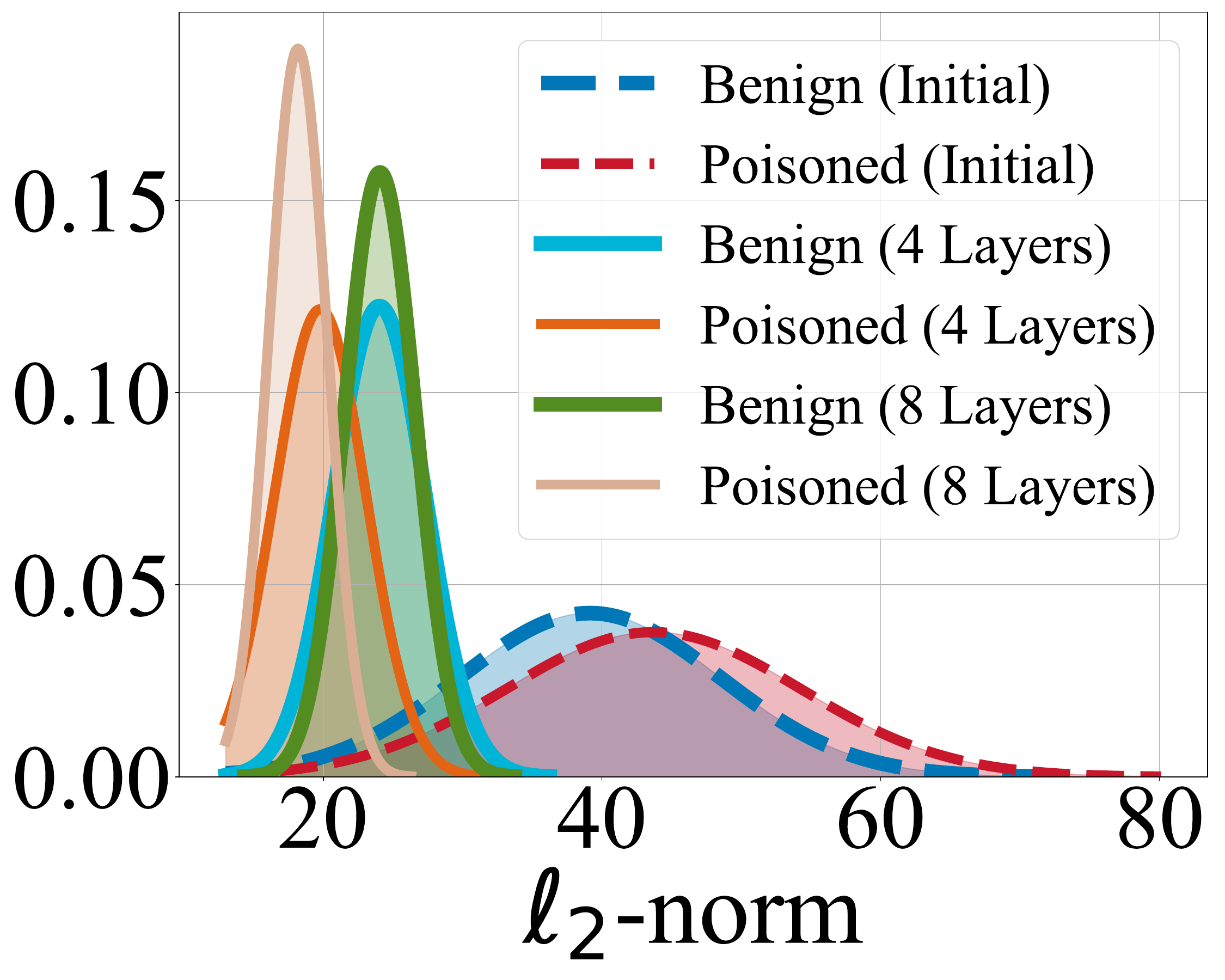}}
\subfigure[BATT\label{fig:batt_l2_small1}]{\includegraphics[width=0.24\textwidth]{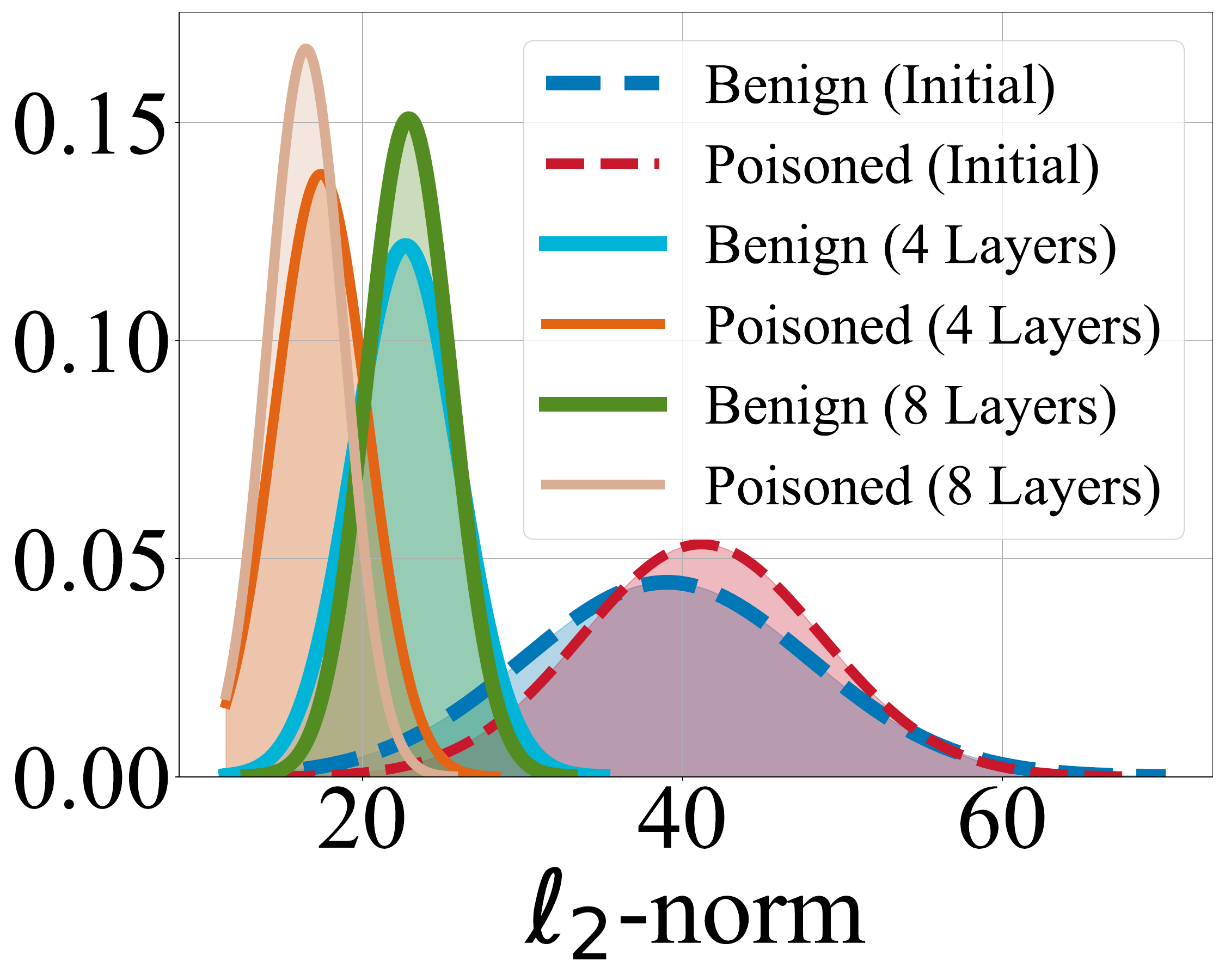}}
\caption{Distribution of L2 Norm Values for the last-hidden-layer activations under reduced model parameters (\ie, magnification factors less than 1), where ``$n$ Layers" represents the number of scaled layers, counting backward from the model's final layer.}
\vspace{-1.2em}
\label{fig:intuition_l2_small1}
\end{figure}

\subsection{Is Shrinking as Effective as Amplifying?}
\label{appendix:factorless1}
In this study, we focus on detecting poisoned samples by amplifying model parameters with a scaling factor greater than one. To complement this approach, we conducted ablation experiments in this section involving shrinking model parameters using a scaling factor smaller than one. 


According to~\cref{theorem1}, larger enough feature norms can induce a decrease in confidence for the original predicted class, if the inputs are benign samples (and
certain classical assumptions in learning theory are adopted). Poisoned samples, instead, will stay fine. Therefore, by inversely reducing the values of parameters, we expect to observe a degradation in detection performance. The experiments are conducted across a range of reduced magnification factors (0.1 to 0.9) against BadNet, WaNet, and BATT attacks on the CIFAR-10 dataset using the ResNet18 model. The results displayed in~\cref{fig:magsmall}, clearly indicate a reduction in detection performance, as evidenced by lowered AUROC values and F1 scores approaching zero. This trend remains consistent across the various attack methods examined.

The reduction in detection performance with decreased parameter values reveals the effectiveness of parameter amplification as a defensive strategy, offering a reason to adopt this approach in safeguarding against backdoor threats.

We also examine the L2 Norm for the last-hidden-layer activations under reduced model parameters. We set the parameter reduction factor to 0.9 and reduced the values of parameters of the model's last four and eight hidden layers, respectively. The L2 norm is calculated on both benign models and backdoored models under BadNet, WaNet, and BATT attacks. As shown in~\cref{fig:intuition_l2_small1}, a greater reduction in parameters led to a smaller L2 norm in both benign and backdoored models. 

This observation provides empirical justification for the necessity of parameter amplification, thereby reinforcing our insights for our proposed defense.


\section{Robustness Against Adaptive Attacks}

\subsection{Existing Attacks with Small Poisoning Rates}
\label{appendix:small_rate_defense}

We evaluate the robustness of our defense against backdoor attacks with low poisoning rates. Specifically, we focus on three representative attacks (BadNets, WaNet, and BATT) on the CIFAR-10 dataset with poisoning rates ($\rho$) ranging from 0.5\% to 10\%, and on the SubImageNet-200 dataset with $\rho$ from 2\% to 10\%, ensuring that most ASRs exceed 80\%. The results presented in~\cref{tab:poison_rate_small} and~\cref{tab:rate} demonstrate that our defense remains effective under low poisoning rates, achieving AUROC and F1 scores well above 0.9, even at a poisoning rate as low as 0.5\%. We also compare the performance of our defense with baseline defenses under these conditions. As shown in \cref{tab:smallrate_com}, our defense outperforms the baseline defenses in most scenarios across different poisoning rates.

\begin{table}[!t]
\centering
\small
\caption{The performance of our defense in defending against backdoor attacks with different poisoning rates on CIFAR-10.}
\label{tab:poison_rate_small}
\begin{tabular}{lcccccccccccccc} 
\toprule
Attacks $\rightarrow$& 	\multicolumn{3}{c}{BadNets}	&	\multicolumn{3}{c}{WaNet}& \multicolumn{3}{c}{BATT}\\
\cmidrule(lr){2-4}  \cmidrule(lr){5-7} \cmidrule(lr){8-10}
$\rho$  $\downarrow$ & ASR & AUROC	& F1 & ASR & AUROC	& F1 & ASR & AUROC	& F1\\
\midrule
0.5\% & 1.000 &0.955 & 0.950& 0.039&0.936 & 0.927& 0.961 &0.913&0.927 \\
1\% & 1.000	& 0.950 & 0.951 & 0.286& 0.953& 0.935& 0.993 & 0.948 & 0.948\\
2\% & 1.000 & 0.999& 0.928 & 0.968 & 0.966 & 0.944&0.999 & 0.999 & 0.972\\
4\%	&1.000	& 0.998 & 0.912 & 0.972 & 0.977 & 0.959 & 1.000 & 1.000 & 0.967\\
6\%	&1.000	&0.999 & 0.961 & 0.994 & 0.978 & 0.955 & 1.000 &0.998 & 0.942\\
8\%	&1.000	& 1.000	& 0.981 & 0.996 & 0.983 & 0.960 & 1.000 & 0.985 & 0.958\\
10\%	&1.000	& 1.000 & 0.967 & 0.997 & 0.984 & 0.956	&1.000 & 0.999 & 0.979\\
\bottomrule
\vspace{-2em}
\end{tabular}
\end{table}

\begin{table}[!t]
\small
\centering
\caption{The performance of our defense in defending against attacks with different poisoning rates ($\rho$) on SubImageNet-200.}
\label{tab:rate}
\begin{tabular}{lcccccccccccccc} 
\toprule
Attacks $\rightarrow$& 	\multicolumn{3}{c}{BadNets}	&	\multicolumn{3}{c}{WaNet}& \multicolumn{3}{c}{BATT}\\
\cmidrule(lr){2-4}  \cmidrule(lr){5-7} \cmidrule(lr){8-10}
$\rho$  $\downarrow$ & ASR & AUROC	& F1 & ASR & AUROC	& F1 & ASR & AUROC	& F1\\
\midrule
2\% & 0.955	&0.999&	0.905&	0.004	&-	&-&	0.981	&0.999	&0.998\\
4\%	&0.960	&1.000&	0.996	&0.182&	0.944&	-	&0.967	&1.000	&0.999\\
6\%	&0.972	&0.991	&0.989&	0.786	&1.000	&0.996&	0.981	&0.999	&0.997\\
8\%	&0.974	&0.997	&0.995	&0.818	&0.986&	0.976&	0.980	&1.000	&0.999 \\
10\%	&0.998	&1.000&	0.992	&0.967&	0.967&	0.981&	0.997	&0.998&	0.998\\
\bottomrule
\vspace{-2em}
\end{tabular}
\end{table}

\begin{table}[!t]
\centering
\small
\caption{Comparision of the performance in defending against backdoor attacks with different poisoning rates ($\rho$) on CIFAR-10.}
\label{tab:smallrate_com}
\scalebox{0.93}{
\begin{tabular}{lcccccccccccccc} 
\toprule
$\rho$  $\rightarrow$ & & \multicolumn{2}{c}{2\%} & \multicolumn{2}{c}{4\%} &	\multicolumn{2}{c}{6\%} & 	
\multicolumn{2}{c}{8\%}	 & \multicolumn{2}{c}{10\%} \\
\cmidrule(lr){3-4}   \cmidrule(lr){5-6} \cmidrule(lr){7-8} \cmidrule(lr){9-10} \cmidrule(lr){11-12} 
Defenses $\downarrow$ & Attacks $\downarrow$& 	AUROC & F1 & 	AUROC & F1 & 	AUROC & F1 & 	AUROC & F1  & 	AUROC & F1 \\
\midrule
\multirow{3}{*}{STRIP} & BadNets & 0.881&  0.679	& 0.895 & 0.752	& 0.868 & 0.657& 	0.769 & 0.429& 	0.931 & 0.824 \\
& WaNets & 0.493 &  0.137	&0.485& 0.138	&0.479& 0.131&	0.466	&0.116& 0.469& 0.125\\
& BATT & 0.779	& 0.579 &0.650& 0.364&	0.656 &0.385&	0.808	&0.639 &0.449 &0.258\\
\multirow{3}{*}{TeCo} & BadNets &1.000&	0.916& 0.997& 0.952	&0.981&	0.929& 0.994&	0.937 &0.998& 0.970\\
& WaNets & 0.992& 0.891&	0.976 &0.905&	0.999 &0.944&	0.906	&0.945& 0.923&  0.915\\
& BATT & 0.803&0.683&	0.814& 0.684&	0.809&0.685&	0.871&0.685&	0.914&0.673\\
\multirow{3}{*}{SCALE-UP} & BadNets &0.959	& 0.918& 0.964&	0.914& 0.959& 	0.910 &0.971 &0.915	&0.962 &0.913\\
& WaNets& 0.746 &0.698&	0.766&	0.726& 0.730& 0.624&	0.689& 0.646&	0.672& 0.529\\
& BATT &0.944 &0.880&	0.968&	 0.868& 0.957	&0.907& 0.968&	0.871& 0.959& 0.911\\
\multirow{3}{*}{Ours} & BadNets & 0.999&	0.928& 0.998 &0.912 &	0.999&	0.961& 1.000&	0.981& 1.000 &0.967\\
& WaNets& 0.966	&0.944 &0.977&	0.959 &0.978&	0.955& 0.983	&0.960& 0.984 &0.956\\
& BATT &0.999& 0.972&	1.000	&0.967 &0.998 &0.942 &	0.985&	0.958&0.999& 0.979\\
\bottomrule
\vspace{-2em}
\end{tabular}
}
\end{table} 

We observe that the ASRs (excluding testing samples from the target class) for WaNet are 3.9\% and 28.6\% at poisoning rates of 0.5\% and 1\%, respectively. This indicates that the attack nearly fails at a 0.5\% poisoning rate. Following the suggestion from the Backdoor-Toolbox, we remove samples containing trigger patterns that still cannot be correctly predicted as the target label by backdoored DNNs. As shown in~\cref{tab:com_poison_rate_small}, our defense remains highly effective in these cases, although its performance is slightly lower than that of TeCo, which requires significantly more inference time. In contrast, both STRIP and SCALE-UP fail.

\begin{table}[!t]
\centering
\small
\caption{The comparison of our defense and the baseline defenses in defending against the WaNet attack under poisoning rates of 0.5\% and 1\% on CIFAR-10.}
\label{tab:com_poison_rate_small}
\begin{tabular}{lcccccccccc} 
\toprule
 $\rho$ (\%) $\rightarrow$ & \multicolumn{3}{c}{0.5} & \multicolumn{3}{c}{1} \\
 \cmidrule(lr){2-4}  \cmidrule(lr){5-7}
Defenses $\downarrow$  & AUROC &TPR&	FPR & AUROC &TPR&	FPR\\
\midrule
STRIP	&0.403&	0.039&	0.100 &	0.421&	0.033&	0.100\\
TeCo	&1.000&	1.000&	0.156 	&1.000	&1.000&0.068\\
SCALE-UP &	0.461&	0.440	&0.389 &	0.467	&0.424&	0.345\\
Ours&	0.936	&0.791	&0.864	&0.953&	1.000&	0.129\\
\bottomrule
\vspace{-2em}
\end{tabular}
\end{table}

\subsection{Adaptive Attacks in the Worst-case Scenario}
\label{appendix:ada}
In addition to the existing attacks,
we further consider the worst-case scenario where potential adaptive attacks are tailored for our defense. 

\noindent\textbf{Design 1.} A natural assumption is that the adversary would design an adaptive loss $\mathcal{L}_{\mathrm{ada}}=\sum_{i=1}^{|\mathcal{D}_b|} \mathcal{L}(\hat{\mathcal{F}_k^{\omega}}(\vx_{i};\hat{\vtheta}),y_i)$ to ensure that the benign samples are correctly predicted when subjected to model parameter amplification, hence breaking our consistency assumption. This adaptive loss is then integrated into the overall loss function as $\mathcal{L} = \alpha \mathcal{L}_{\mathrm{bd}} + (1-\alpha) \mathcal{L}_{\mathrm{ada}}$, where $\alpha$ represents the weighting factor.

The adversary would aim to find a $\alpha$ value that best balances
the ASRs and the BAs.
\cref{tab:adaba} presents the performance (BA, ASR) of the adaptive attacks under various $\alpha$ settings. As evident from the results, all three attacks (BadNets, WaNet, BATT) on the CIFAR-10 dataset employed in the experiments consistently exhibit high ASRs and BA across different values of $\alpha$ on the CIFAR-10 dataset,
underscoring the effectiveness of the adaptive attacks.

On the other hand, we have shown in~\cref{tab:ada} such as adaptative attacks can still be effectively defended by our method. 
We conducted further investigation. 
We observed that 
the adaptive loss indeed induced a model $\mathcal{F}'$ substantially different from the nonadaptive version $\mathcal{F}$.
However, our defense, particular Algorithm~\ref{alg:identifylayers}, can readjust for the modified backdoor model (Note that model $\mathcal{F}$ is an input in Algorithm~\ref{alg:identifylayers}).
In particular, we observed that on a nonadaptive backdoor model $\mathcal{F}$, Algorithm~\ref{alg:identifylayers} returns $k=10$. On the adaptive model $\mathcal{F}'$, Algorithm~\ref{alg:identifylayers} returns $k=15$.
In other words, our algorithm learned to exploit the earlier layers not touched by the adaptive attack.
This ability to counter adaptive attacks is a key advantage of our method compared with the input-based SCALEUP method.


\begin{table}[!t]
\centering
\small
\caption{The attack performance (BA, ASR) with the adaptive attack settings in ``Design 1''.}
\label{tab:adaba}
\setlength{\tabcolsep}{5pt}
\begin{tabular}{lcccccccccccccc} 
\toprule
 Weight$\rightarrow$ & \multicolumn{2}{c}{0.2} & \multicolumn{2}{c}{0.5} & \multicolumn{2}{c}{0.9} & \multicolumn{2}{c}{0.99} \\
 Attacks$\downarrow$ & BA & ASR & BA & ASR& BA & ASR & BA & ASR \\ 
\midrule
BadNets &0.775 &0.992 &0.858 &0.985 & 0.881& 0.995 &0.891 &0.996 \\
WaNet & 0.906& 0.948& 0.891& 0.977& 0.877&0.935  &0.879 & 0.813\\ 
BATT & 0.851&0.986 &0.846 &0.994 &0.840 &0.983  &0.831 & 0.981\\
\bottomrule
\end{tabular}
\vspace{-1em}
\end{table}

\begin{table}[!t]
\centering
\small
\caption{The attack performance (BA, ASR) of the adaptive attack in ``Design 2'' and the detection performance (AUROC, F1) of IBD-PSC against the adaptive attack on CIFAR-10. We mark the failed cases (where $BA<70\%$) in red, given that the accuracy of models unaffected by backdoor attacks on clean samples is 94.40\%.}
\label{tab:adaba2}
\begin{tabular}{lcccccccccccccc} 
\toprule
$\alpha'$ $\rightarrow$ & 	\multicolumn{2}{c}{0.01}	&	\multicolumn{2}{c}{0.1}& \multicolumn{2}{c}{0.5}\\
\cmidrule(lr){2-3}  \cmidrule(lr){4-5} \cmidrule(lr){6-7}
Attacks$\downarrow$ & BA / ASR	& AUROC /F1	&  BA / ASR	&  AUROC / F1	& BA / ASR	& AUROC/ F1\\
\midrule
BadNets	&0.832 / 0.887	&0.877 / 0.924	&0.802 / 0.874	&0.874 / 0.861	&0.101/ 0.997	&- / - \\
WaNet	&90.88 / 99.87	&0.999 / 0.956	&87.07 / 99.15&	0.985 / 0.934	&85.16 / 89.10&	0.887 / 0.895\\
BATT&	0.745 / 0.997	&0.996 / 0.982	&0.648 / 0.998	&- / -	&0.463 / 0/994	&- / -\\
\bottomrule
\vspace{-2em}
\end{tabular}
\end{table}

\begin{table}[!t]
\centering
\small
\caption{The proportion (\%) of misclassified benign samples classified by the model on each category. In our cases, the target label is 0.}
\label{tab:ba_reduction_design2}
\setlength{\tabcolsep}{5pt}
\begin{tabular}{lcccccccccccccc} 
\toprule
Attacks $\downarrow$, Labels $\rightarrow$&	0	&1	&2	&3&	4	&5&	6	&7	&8	&9\\
\midrule
BadNets (Original)	&14.68	&5.71&	11.09	&22.02&	10.77&	14.19&	6.36	&3.92	&5.55	&5.71\\
BadNets (Adaptive)	&99.57	&0.42	&0.00	&0.00	&0.00	&0.00&	0.00	&0.00&	0.00	&0.01\\
BATT (Original)	&10.81&	5.50&	13.18&	22.27&	10.62&	13.18	&6.23	&5.88&	5.78&	6.54\\
BATT (Adaptive)	&94.67&	0.02	&0.30&	3.84	&0.91&	0.09&	0.06	&0.00	&0.02&	0.09\\
\bottomrule
\end{tabular}
\vspace{-1em}
\end{table}

\noindent\textbf{Design 2.} We can also design another form of adaptive attack to mitigate the impact of parameter amplification. Specifically, we aim to reduce the confidence with which parameter-amplified models predict poisoned samples as belonging to the target class. Inspired by label smoothing, we design an adaptive loss term $\mathcal{L}'_{\textrm{ada}}$ to decrease the confidence of poisoned samples when model parameter amplification occurs. The adaptive loss term $\mathcal{L}'_{\textrm{ada}}$ is defined as:
\begin{equation}
    \mathcal{L}'_{\textrm{ada}} = \sum_{j=1}^{|\mathcal{D}_p|} \mathcal{L}(\hat{\mathcal{F}_k^{\omega}} (\pmb{x}_{j};\hat{\pmb{\theta}}),\hat{y}_i),
\end{equation} 
where $\hat{y}_i$ represents the label-smoothing form of $t$, and $\mathcal{D}_p$ denotes the set of poisoned samples. $\hat{y}_i$ is defined as: 
\begin{equation}
\hat{y}_{i,c} = \begin{cases} 1 - \zeta  & \text{if } c = t \\ \frac{\zeta}{C-1} & \text{otherwise}. \end{cases}
\end{equation}

Here, $\zeta$ is set to 0.2, specifically chosen to reduce the confidence with which poisoned samples are classified into the target class. The term $C$ denotes the total number of classes, $|\mathcal{D}_p|$ represents the number of poisoned samples in the training set, and $\pmb{x}_{j}$ denotes a poisoned sample. 

We integrate the adaptive loss term $\mathcal{L}'_{\textrm{ada}}$ with the vanilla backdoor loss $\mathcal{L}_{\textrm{bd}}$ to formulate the overall loss function as $\mathcal{L}' = \alpha \mathcal{L}_{\textrm{bd}} + (1-\alpha') \mathcal{L}_{ada}'$, where $\alpha'$ is a weighting factor. We evaluate the robustness of our defense under the same settings as described in Section~\ref{sec: ada}. As shown in~\cref{tab:adaba2}, reducing the confidence of poisoned samples significantly decreases BA, making the attack more noticeable. Moreover, our defense remains effective even against this new adaptive attack. The effectiveness of IBD-SPC largely stems from our adaptive layer selection strategy, which dynamically identifies BN layers for amplification. This approach ensures the robustness of our defense mechanism across various scenarios, whether the model is vanilla or adaptively backdoored.

\begin{figure}[!t]
	\centering
 \begin{minipage}{0.90\linewidth}
    \begin{minipage}{0.49\linewidth}
            \includegraphics[width=1\linewidth]{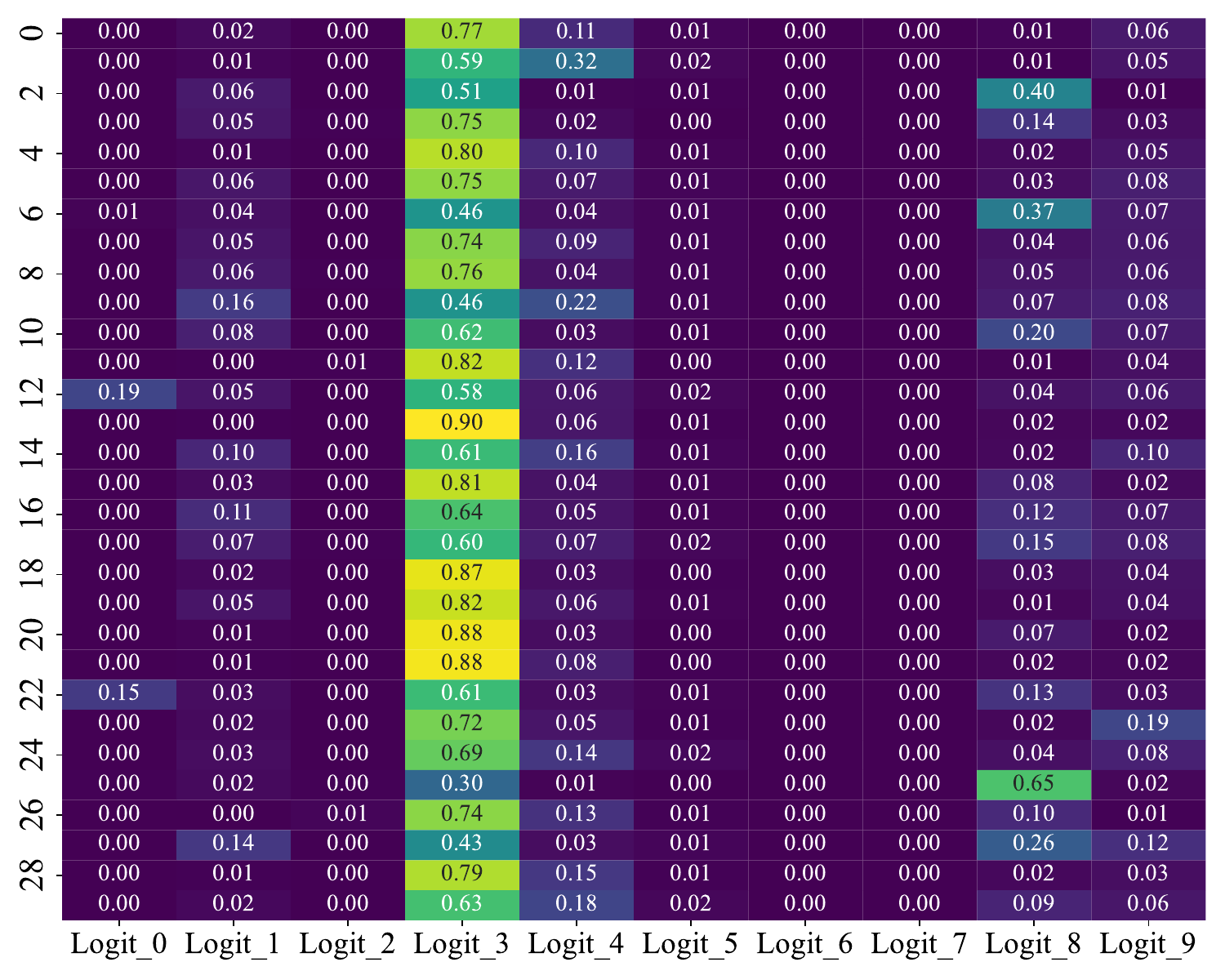}
            \centerline{(a) BadNets}
    \end{minipage}
    \begin{minipage}{0.49\linewidth}
            \includegraphics[width=1\linewidth]{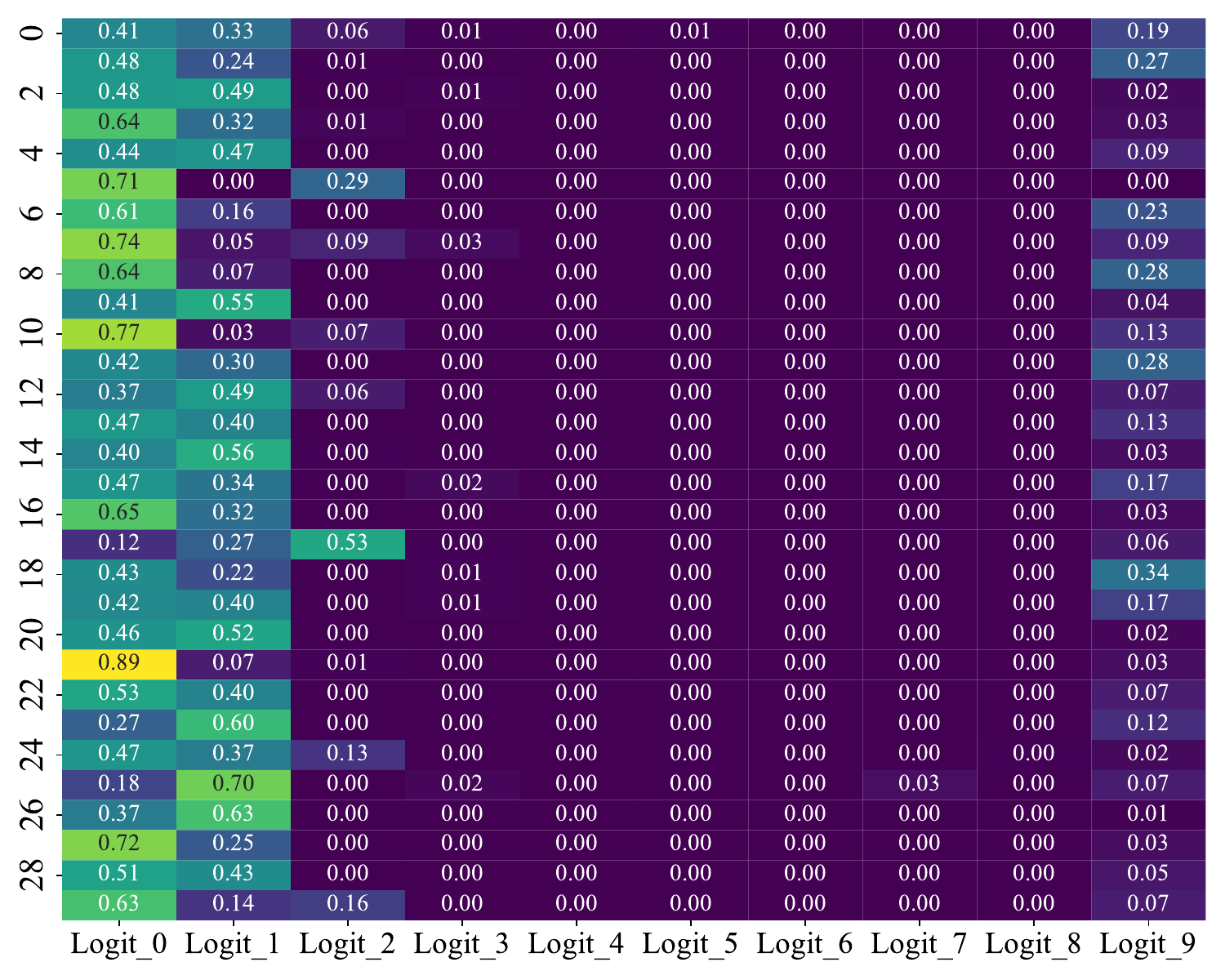}
            \centerline{(b) Blend}
    \end{minipage}
\end{minipage}\vspace{0.8em}
 \begin{minipage}{0.90\linewidth}
    \begin{minipage}{0.49\linewidth}
            \includegraphics[width=1\linewidth]{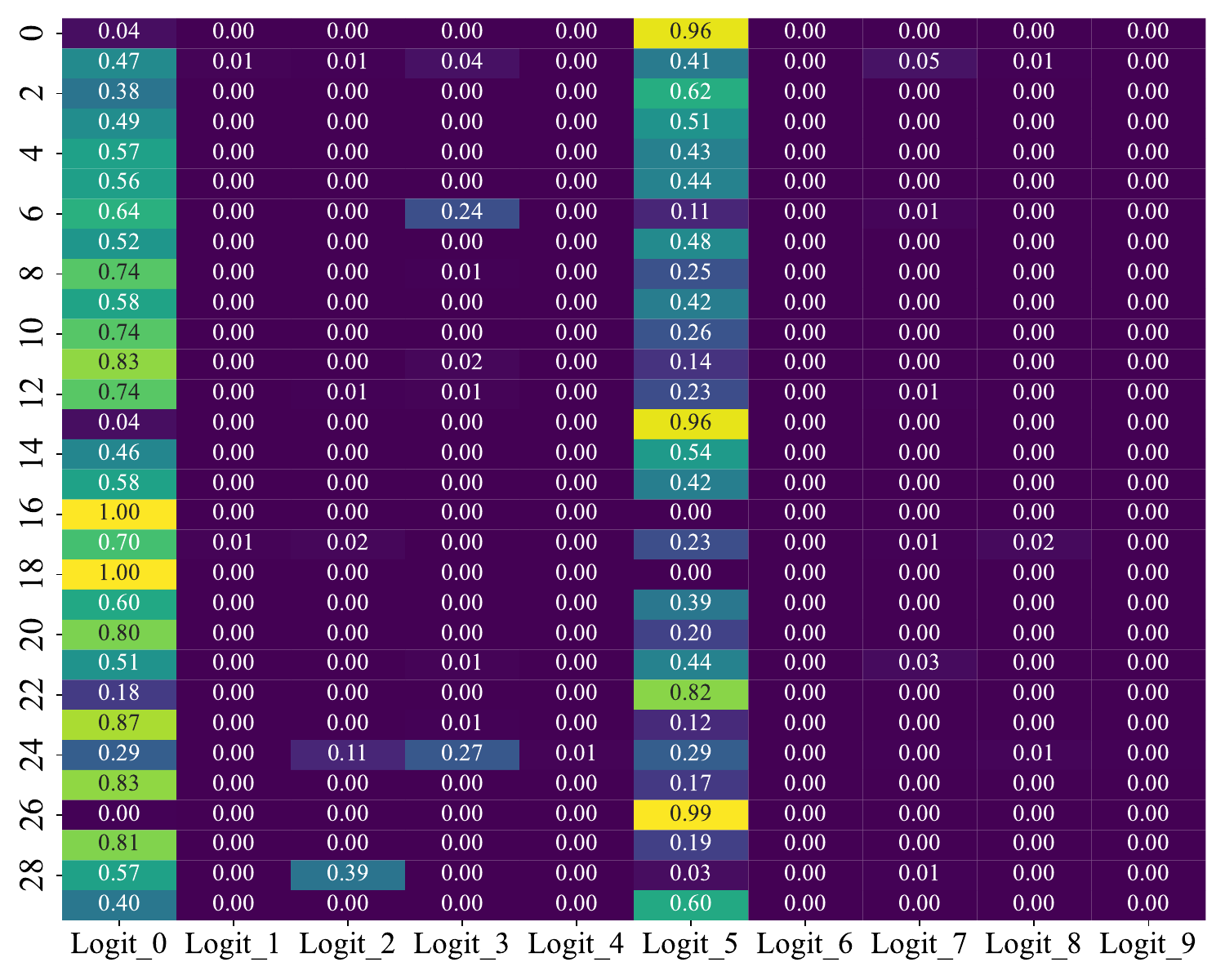}
            \centerline{(c) BATT}
    \end{minipage}
    \begin{minipage}{0.49\linewidth}
            \includegraphics[width=1\linewidth]{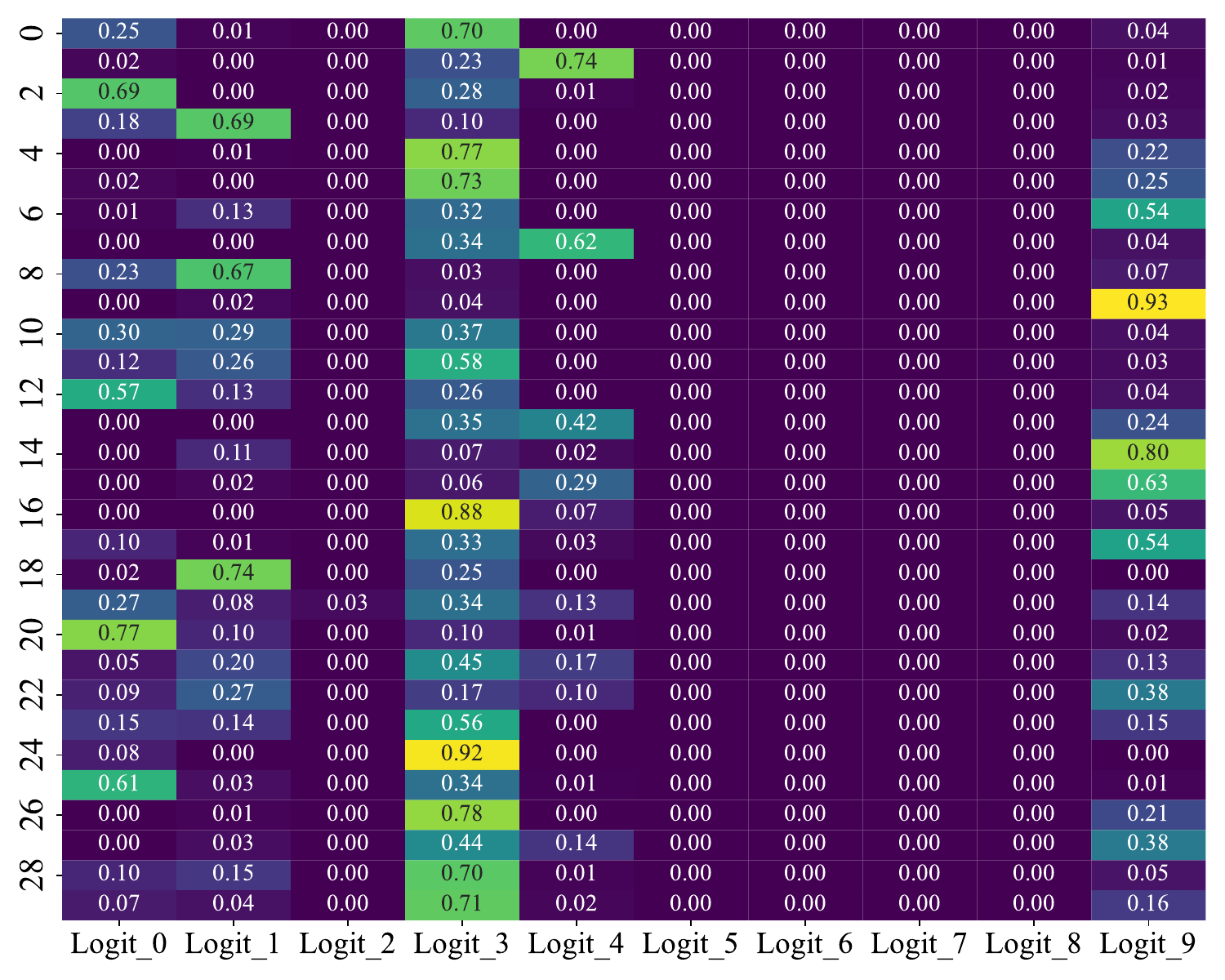}
            \centerline{(d) Ada-patch}
    \end{minipage}
\end{minipage}
\vspace{-0.6em}
 \caption{The heatmap of the average prediction confidences of benign samples within the target class predicted by the $n$ scaled models on the CIFAR-10 dataset. The threshold of our defense is set at 0.9; samples with PSC values above this threshold are classified as poisoned. In our cases, the target label is 0.}
 \label{fig:heatmap}
\end{figure}

We claim that the reduction of BA caused by the adaptive attack is mainly because the DNNs link both benign features and trigger features to the target class when using label smoothing, although the connection between trigger features and the target class is stronger. Specifically, attacked models tend to overfit trigger features when high confidence (\eg, 1) is applied on poisoned samples, as seen in vanilla backdoor attacks. However, after label smoothing, the attacked models rely on both trigger and other features, as the task becomes more complicated and harder to fit accurately. Consequently, the adaptive-attacked model is more likely to predict benign samples as the target class, resulting in relatively low benign accuracy. To further verify this, we calculate the distribution of misclassified benign samples. As shown in~\cref{tab:ba_reduction_design2}, almost all misclassified benign samples are predicted as the target label (\ie, 0) instead of other classes.

We also design an adaptive attack variant that directly classifies poisoned samples to their ground-truth labels under parameter amplification. However, this approach imposes excessive regularization, preventing the learning of backdoors. This phenomenon aligns with our findings in~\cref{theorem1}. Additionally, we design another adaptive loss term by making parameter-amplified models categorize poisoned images as a different class (rather than the target class). However, even with a very small trade-off hyper-parameter for the adaptive loss term, it significantly decreases the BA of the attacked model by more than $30\%$.

\section{The Impact of Parameter Amplification on the Benign Samples from Target class}
\label{appendix:target_benign}


To better understand the reduction in confidence scores of benign samples, we randomly select 30 benign samples from the target class and generate a heatmap to display the distribution of average logits across $n$ scaled models. As shown in~\cref{fig:heatmap}, the heatmaps for various attacks highlight a clear clustering phenomenon: scaled models consistently increase prediction confidence for a specific non-original label, thereby reducing confidence scores for the original labels of these benign samples. Moreover, these non-original labels correspond to categories that are more challenging to classify in CIFAR-10 dataset, typically having lower accuracy. For example, the category associated with Logit\_3 is ``Cat,"  which has an accuracy of less than 90\%, noticeably below the dataset's average accuracy of approximately 94\%.

\begin{figure}[!t]
    \centering
    \begin{minipage}[b]{0.99\linewidth}
        \centering
        \begin{minipage}[b]{0.02\linewidth}
            \rotatebox{90}{~~~~~~~~~~~~~~~~~~~~~~~~~~~~~~~~~~~~SCALE-UP}
            \rotatebox{90}{~~~~~~~~~~~~~~~~~~~~~~~~ Ours}
        \end{minipage}
        \begin{minipage}[b]{0.32\linewidth}
            \includegraphics[width=1\textwidth]{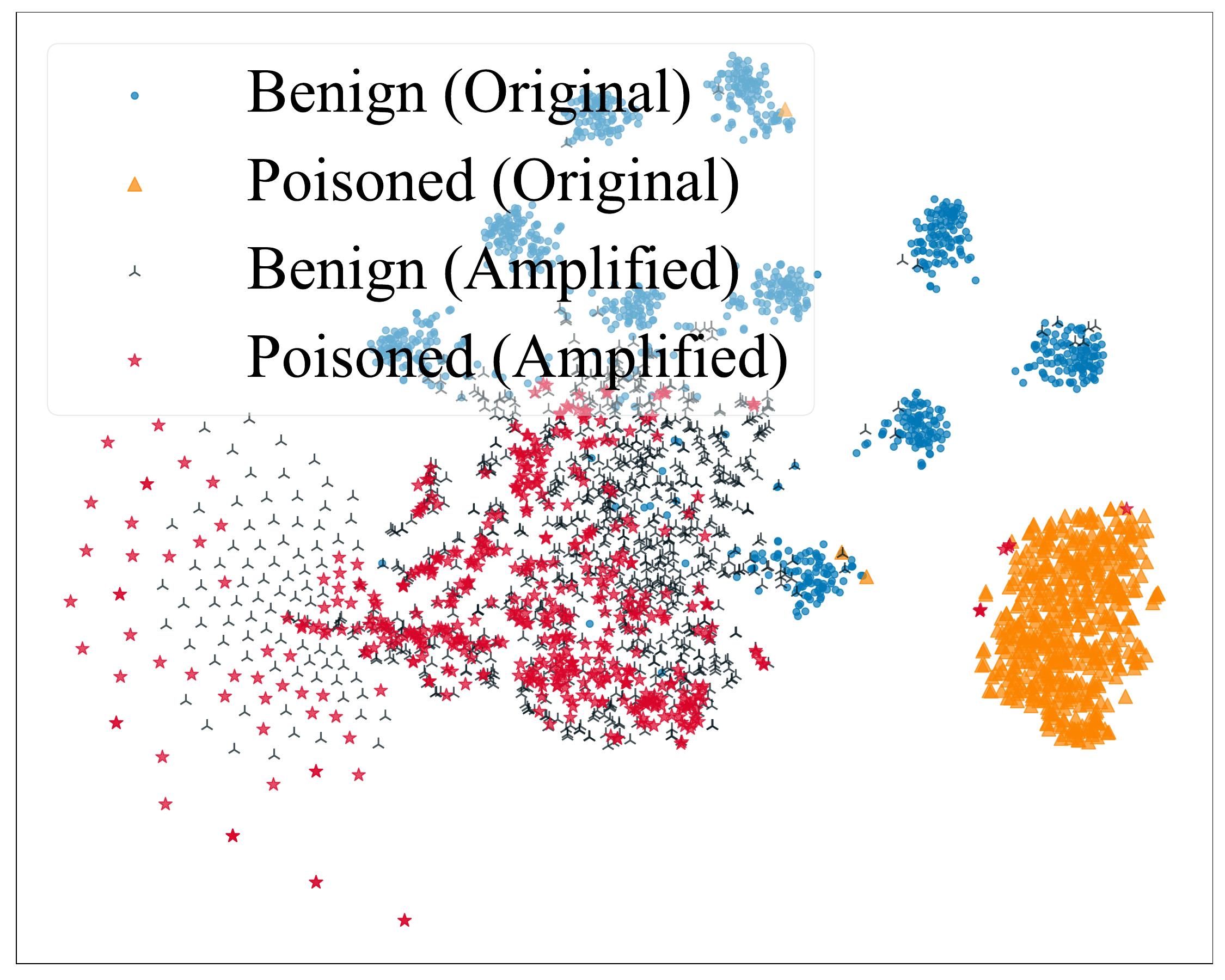}
            \includegraphics[width=1\textwidth]{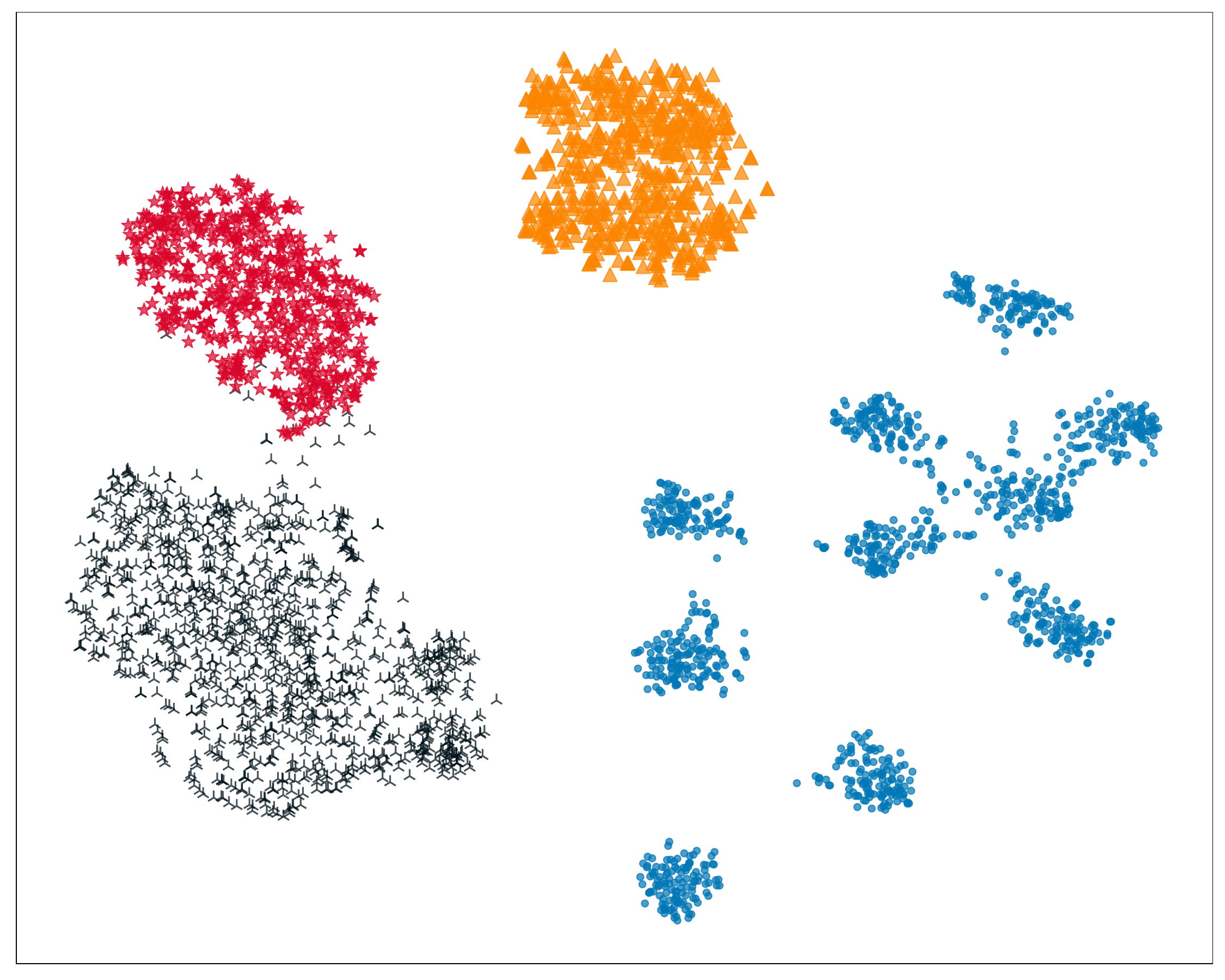}
            \centerline{(a) Blend}
        \end{minipage}
        \begin{minipage}[b]{0.32\linewidth}
            \includegraphics[width=1\textwidth]{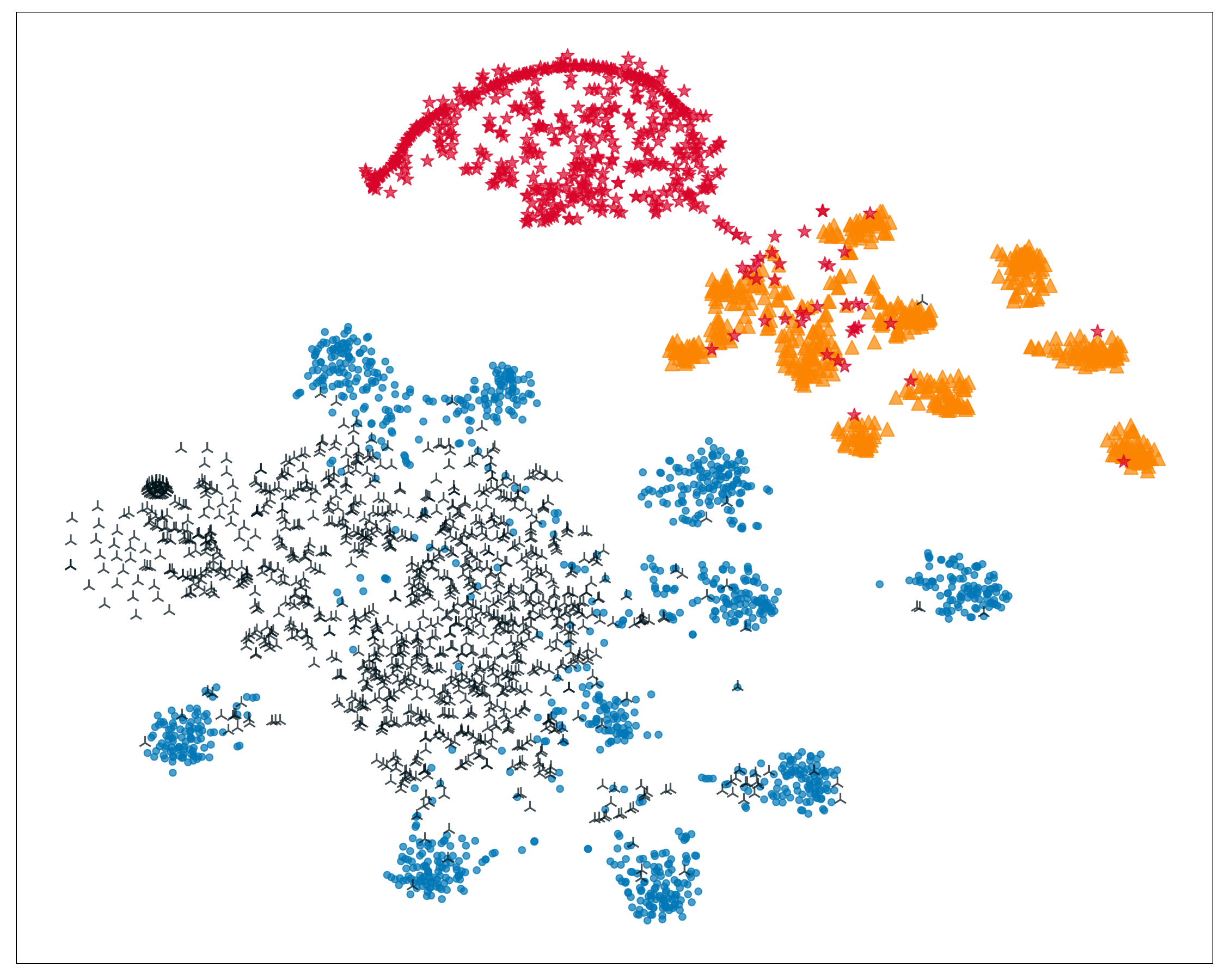}
            \includegraphics[width=1\textwidth]{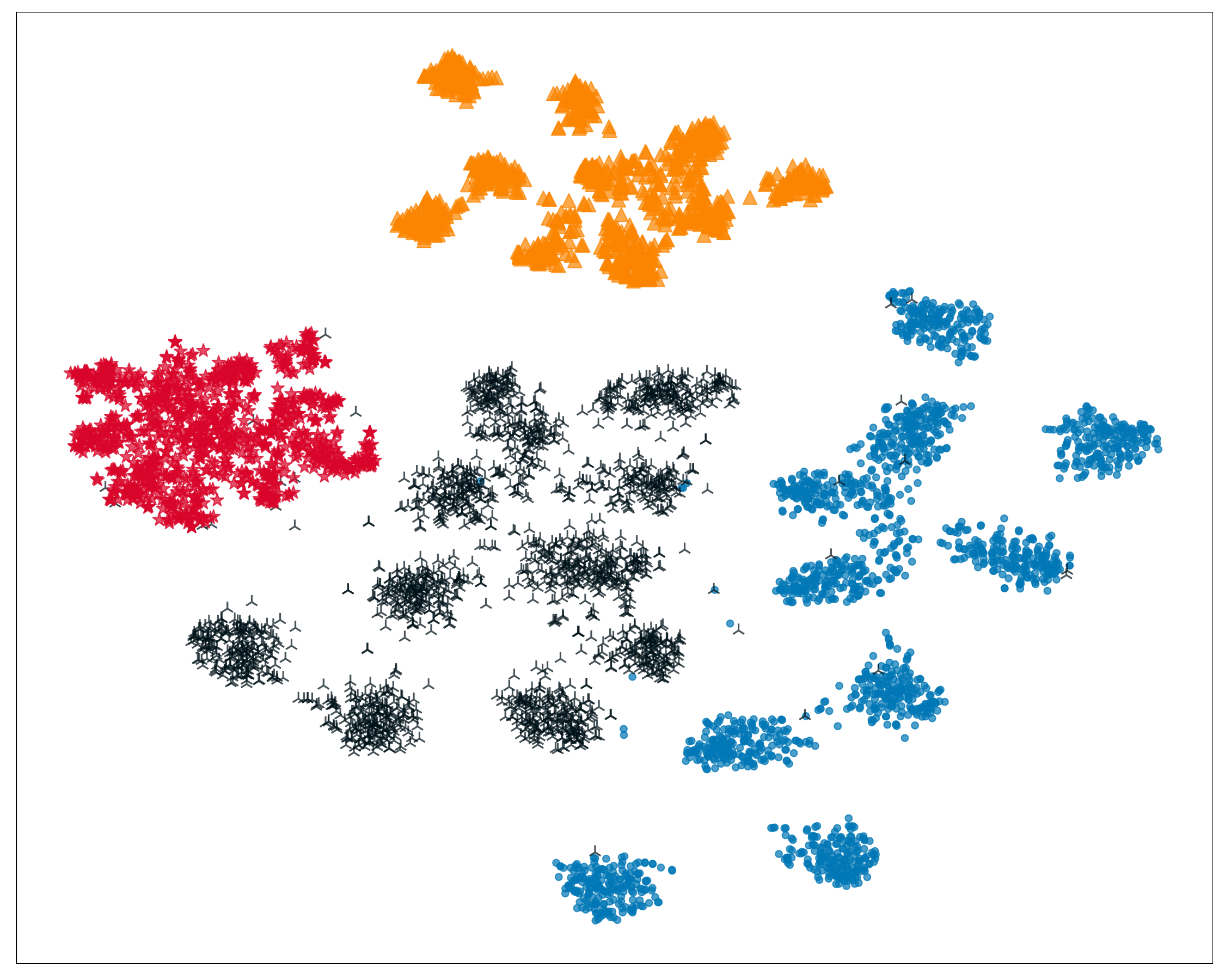}
            \centerline{(b) PhysicalBA}
        \end{minipage}
        \begin{minipage}[b]{0.32\linewidth}
            \includegraphics[width=1\textwidth]{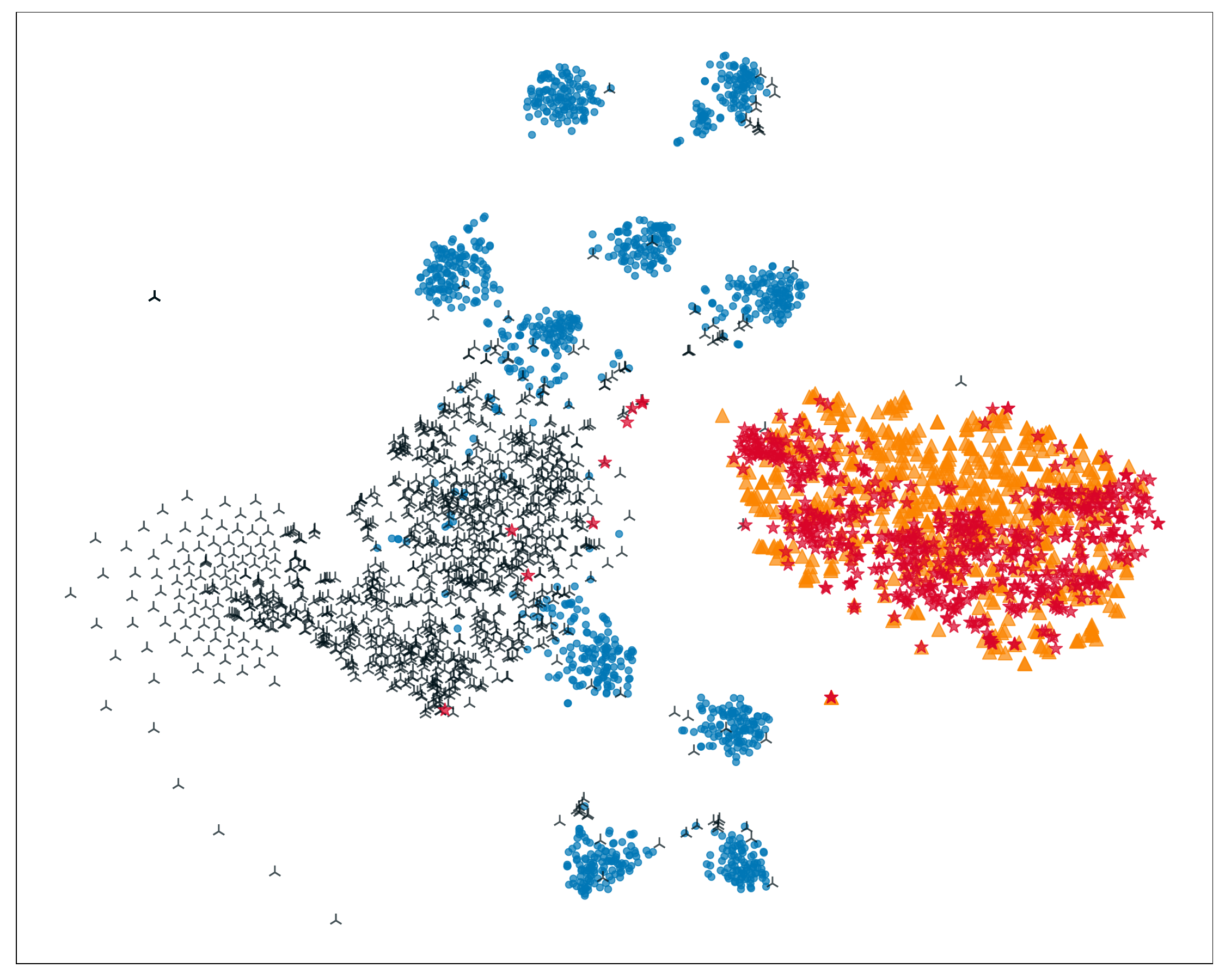}
            \includegraphics[width=1\textwidth]{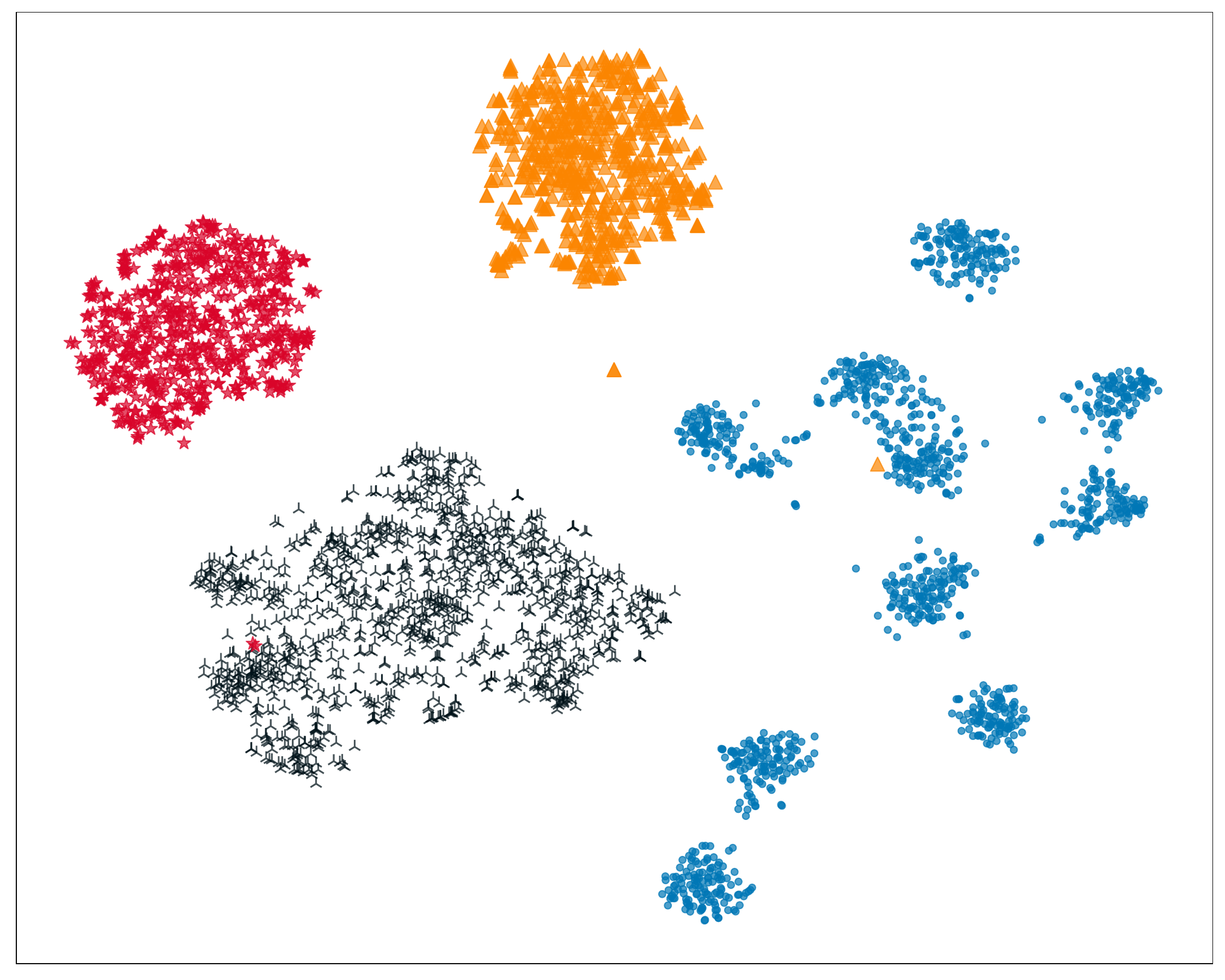}
            \centerline{(c) IAD}
        \end{minipage}
    \end{minipage}
    \begin{minipage}[b]{0.99\linewidth}
        \centering
        \begin{minipage}[b]{0.02\linewidth}
            \rotatebox{90}{~~~~~~~~~~~~~~~~~~~~~~~~~~~~~~~~~~~~SCALE-UP}
            \rotatebox{90}{~~~~~~~~~~~~~~~~~~~~~~~~ Ours}
        \end{minipage}
        \begin{minipage}[b]{0.32\linewidth}
            \includegraphics[width=1\textwidth]{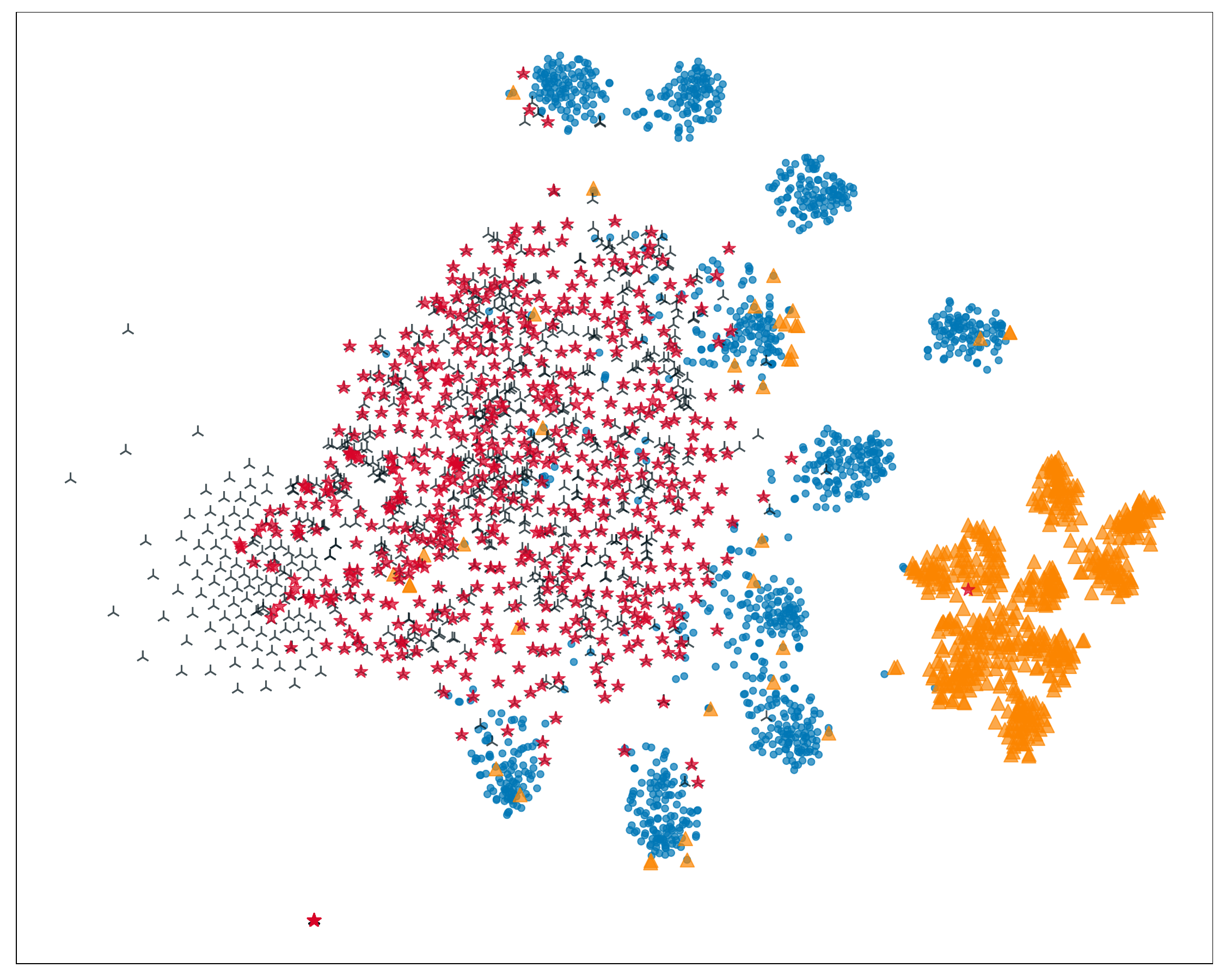}
            \includegraphics[width=1\textwidth]{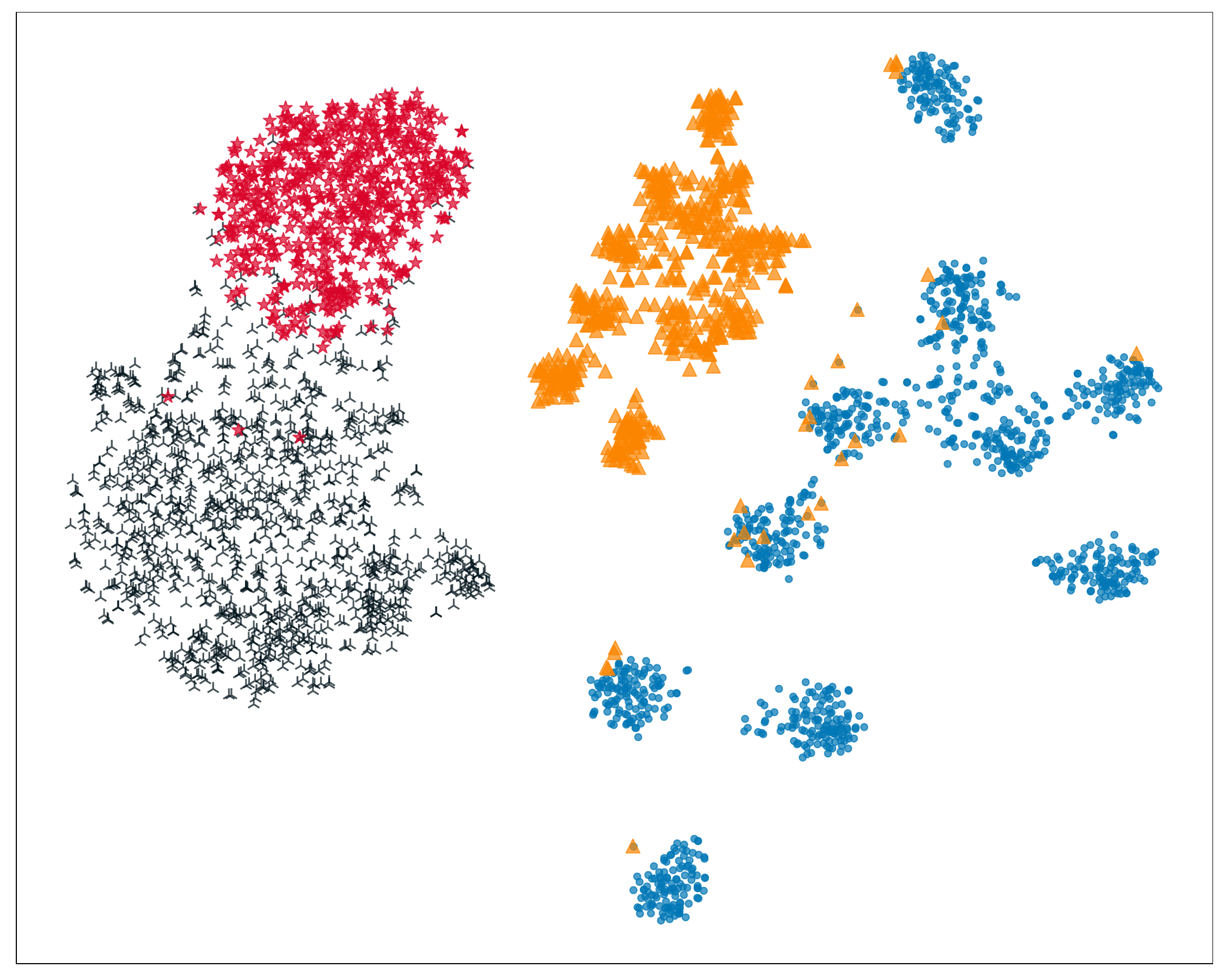}
            \centerline{(d) WaNet}
        \end{minipage}
        \begin{minipage}[b]{0.32\linewidth}
            \includegraphics[width=1\textwidth]{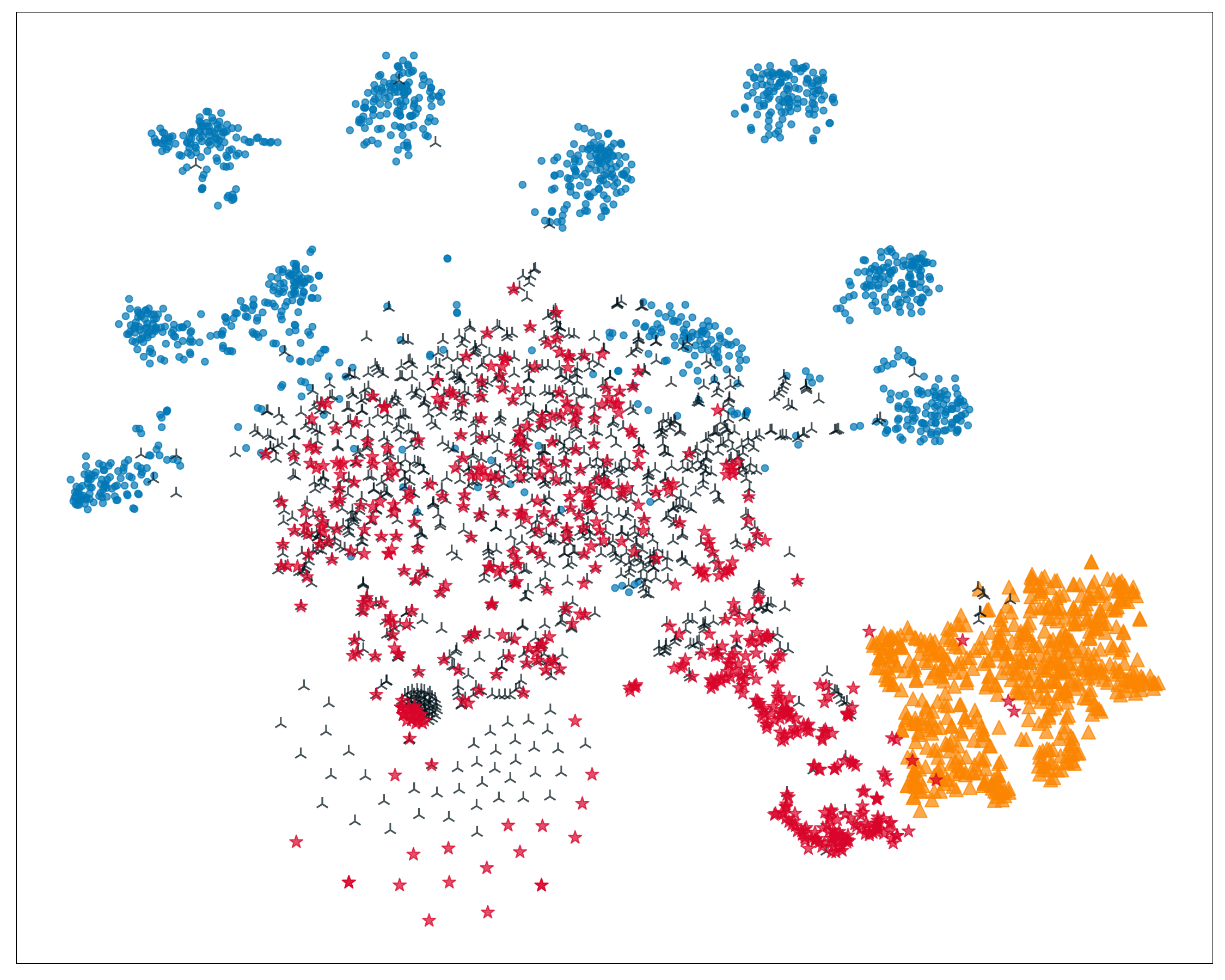}
            \includegraphics[width=1\textwidth]{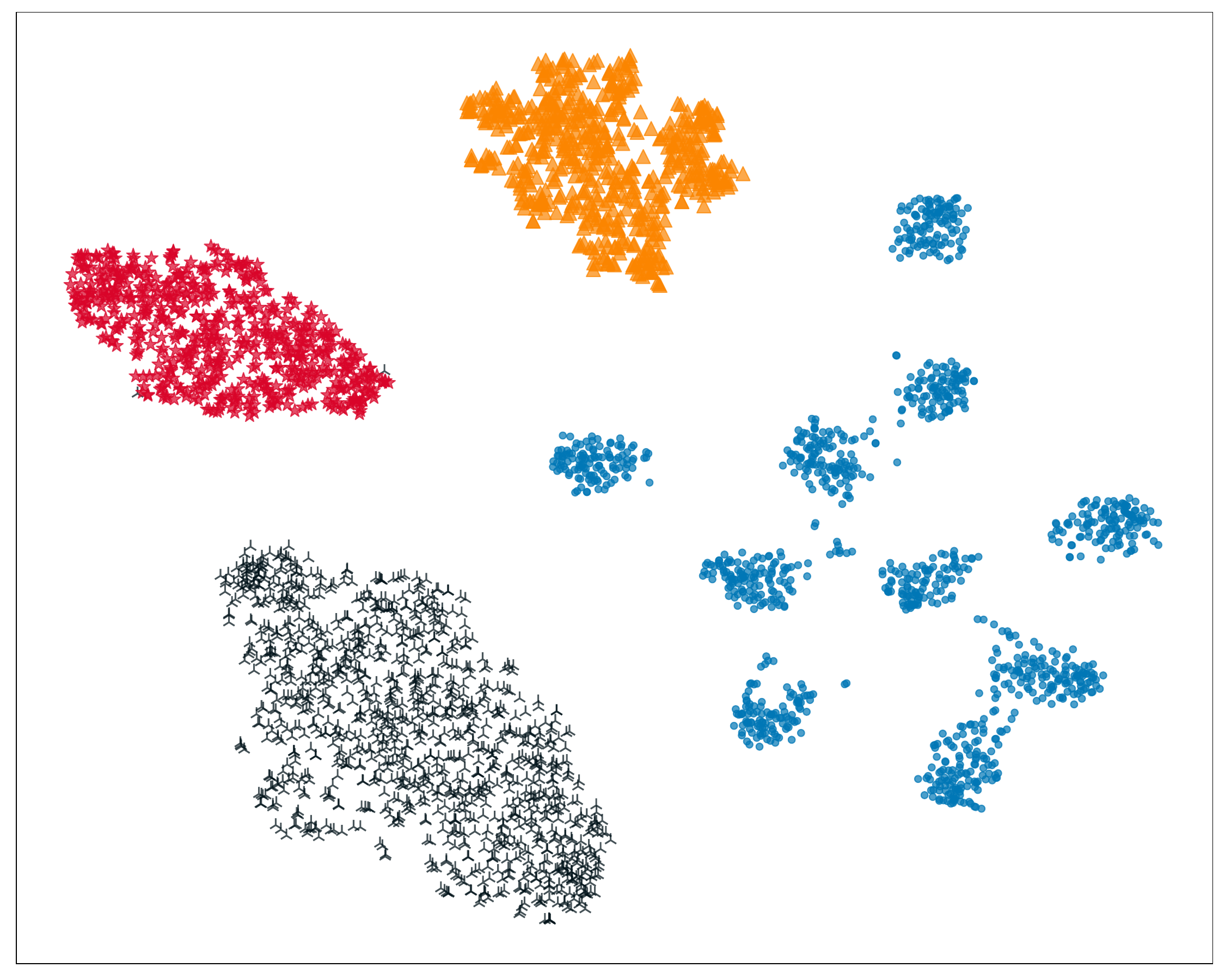}
            \centerline{(e) ISSBA}
        \end{minipage}
        \begin{minipage}[b]{0.32\linewidth}
            \includegraphics[width=1\textwidth]{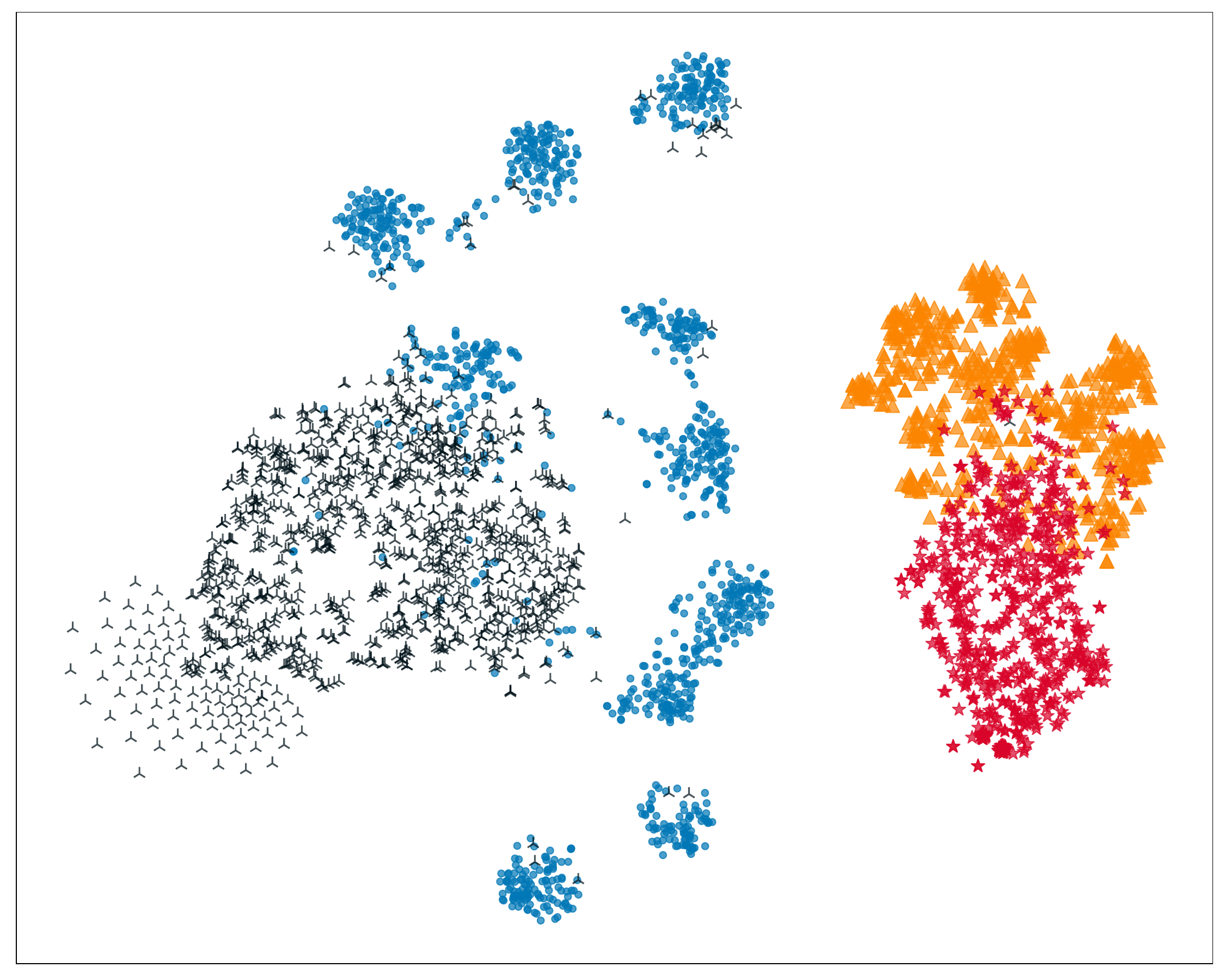}
            \includegraphics[width=1\textwidth]{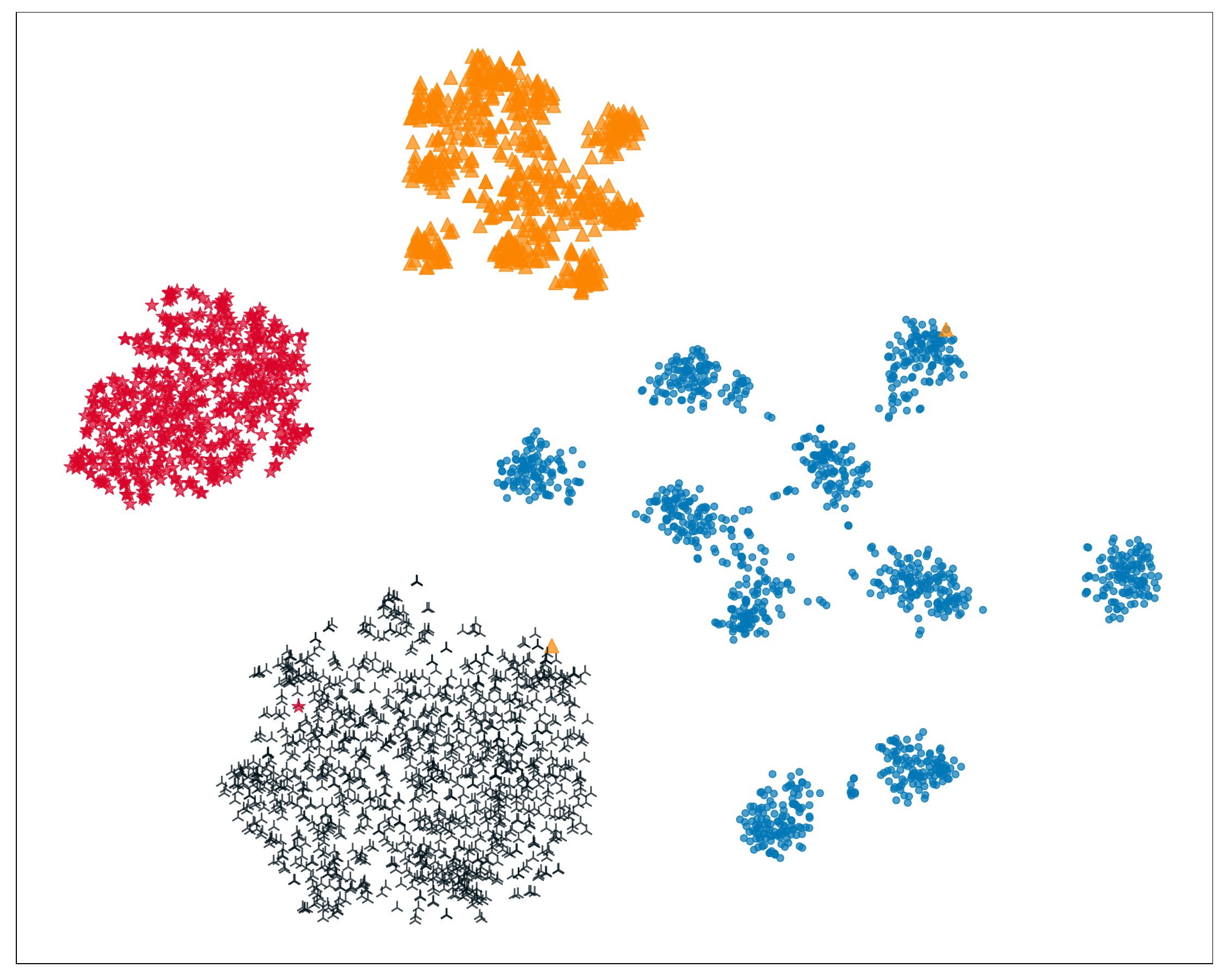}
            \centerline{(f) BATT}
        \end{minipage}
    \end{minipage}
       \caption{The t-SNE of feature representations of benign and poisoned samples on CIFAR-10 dataset against different backdoor attacks.} 
    \label{fig:cluster_other1}
    \vspace{-2em}
\end{figure}

\begin{figure}[!t]
    \centering
    \begin{minipage}[b]{0.99\textwidth}
    \centering
        \begin{minipage}[b]{0.02\linewidth}
            \rotatebox{90}{~~~~~~~~~~~~~~~~~~~~~~~~~~~~~~~~~~~~SCALE-UP}
            \rotatebox{90}{~~~~~~~~~~~~~~~~~~~~~~~~ Ours}
        \end{minipage}
        \begin{minipage}[b]{0.32\linewidth}
            \includegraphics[width=1\textwidth]{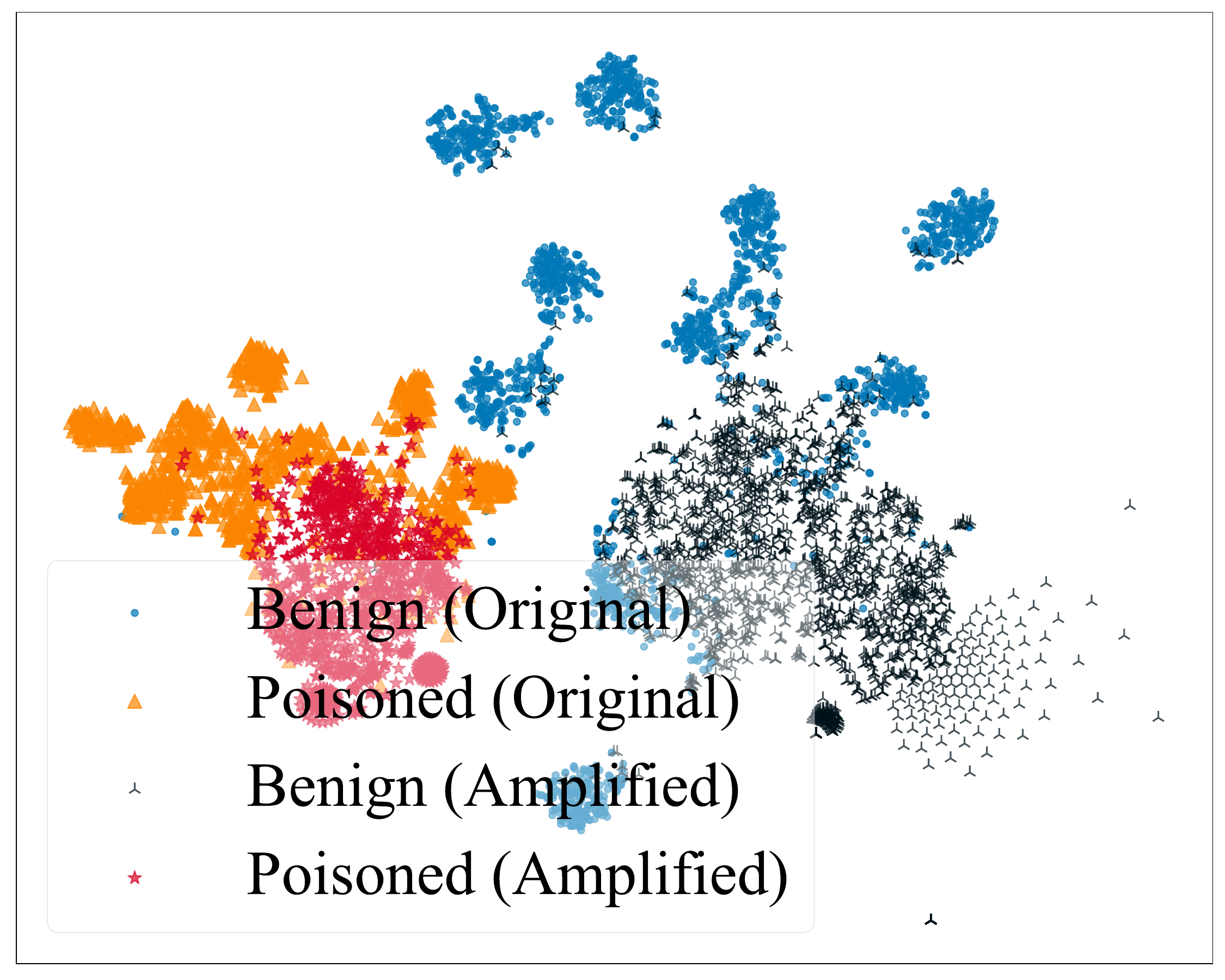}
            \includegraphics[width=1\textwidth]{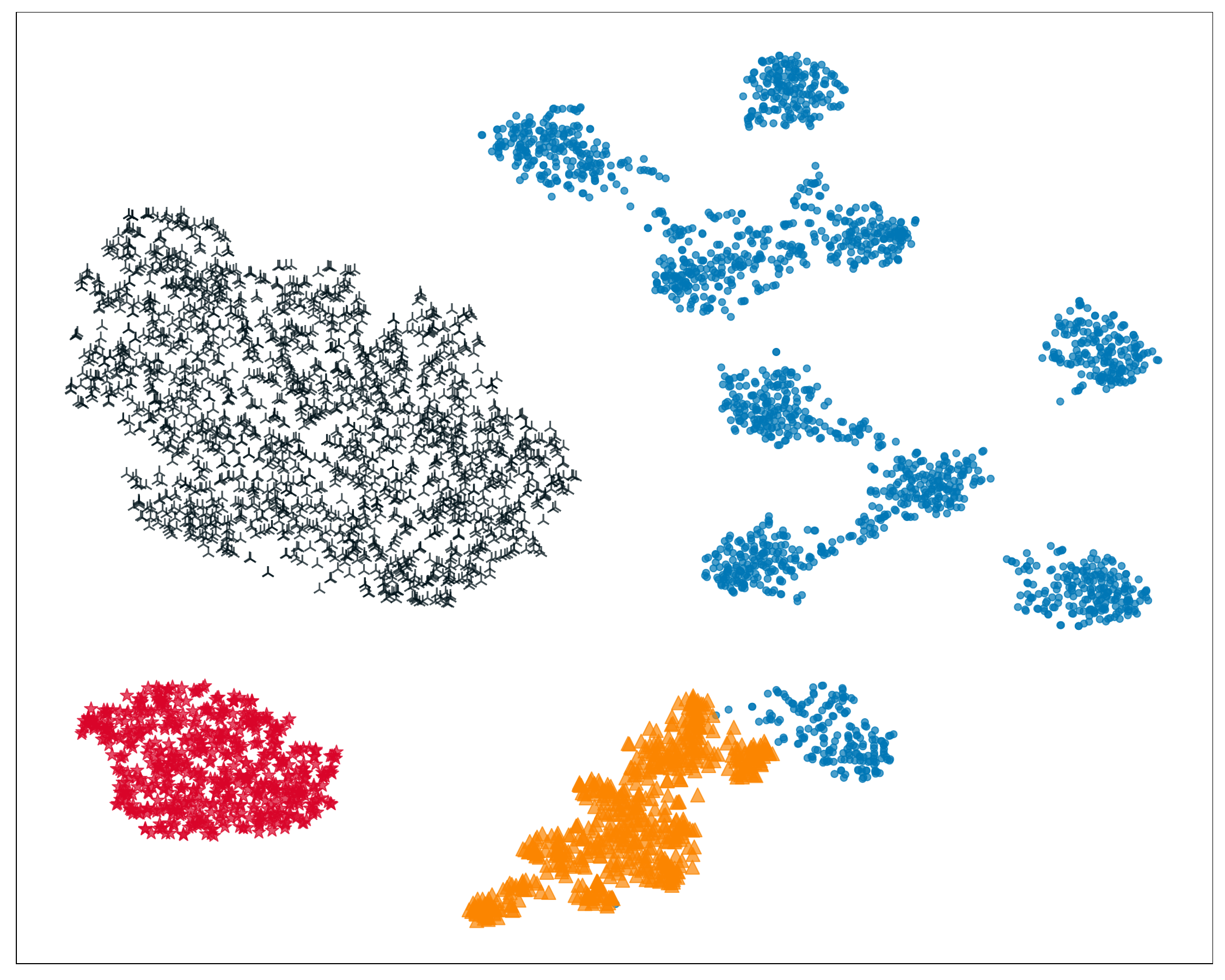}
            \centerline{(a) LC}
        \end{minipage}
        \begin{minipage}[b]{0.32\linewidth}
            \includegraphics[width=1\textwidth]{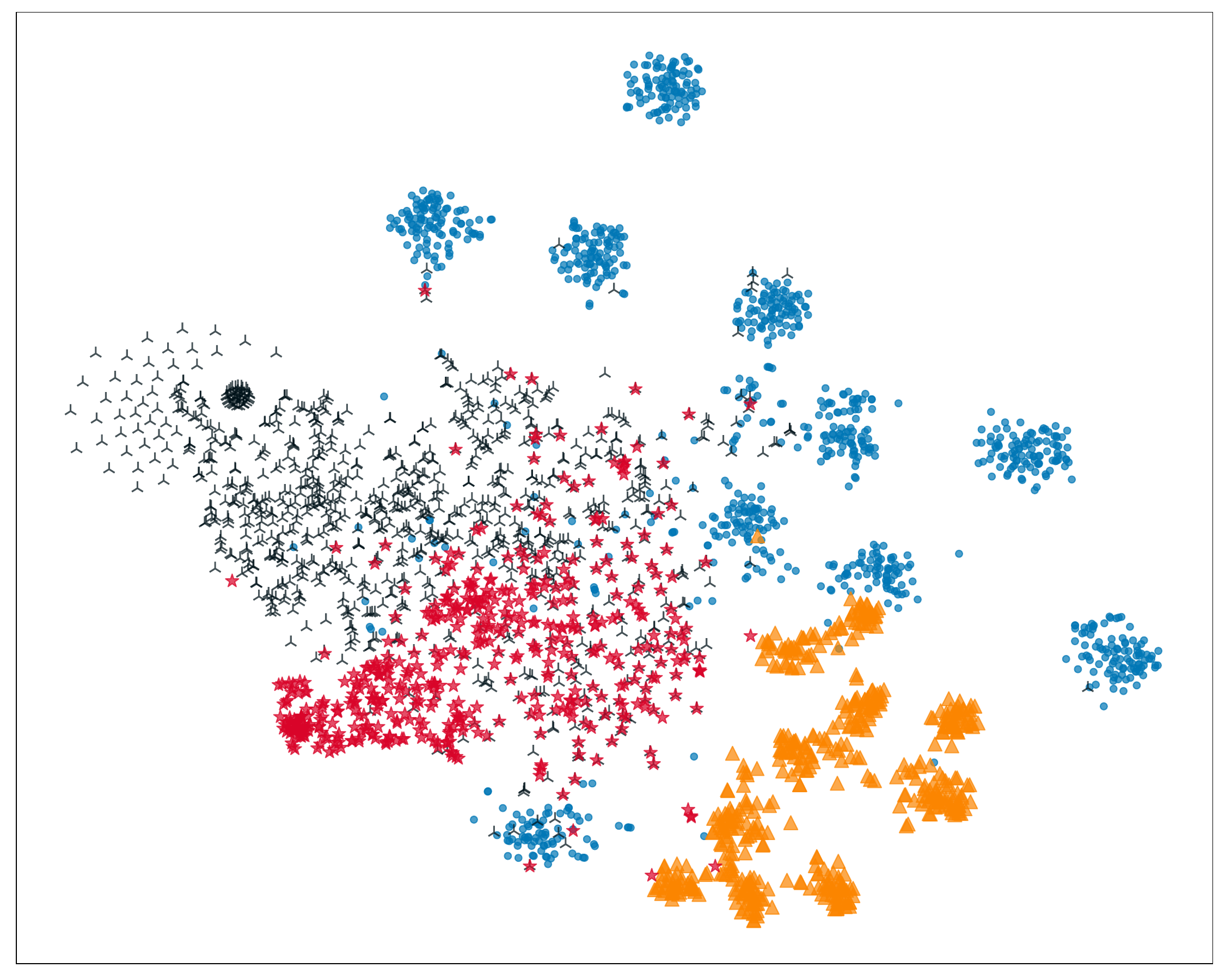}
            \includegraphics[width=1\textwidth]{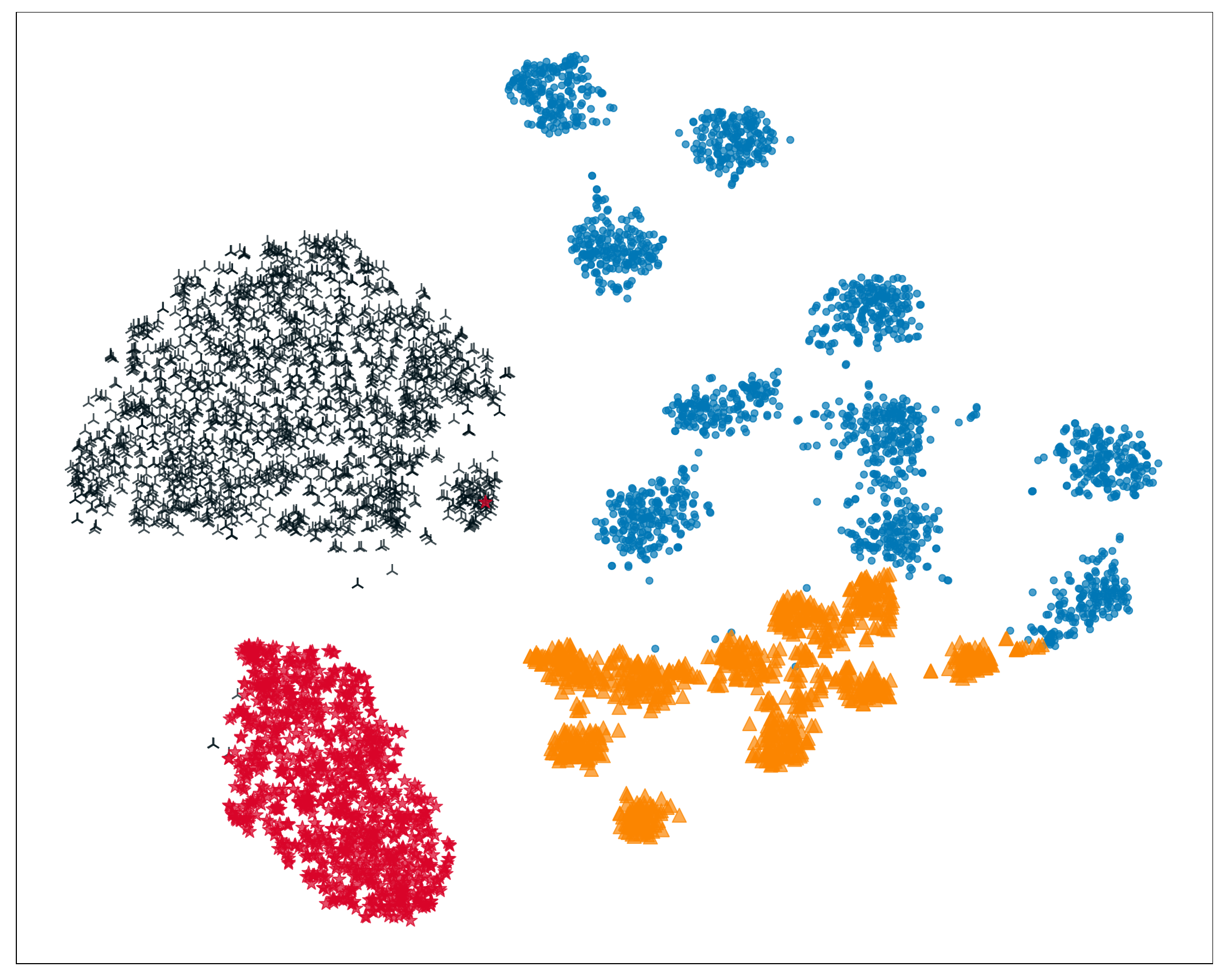}
            \centerline{(b) TaCT}
        \end{minipage}
        \begin{minipage}[b]{0.32\linewidth}
            \includegraphics[width=1\textwidth]{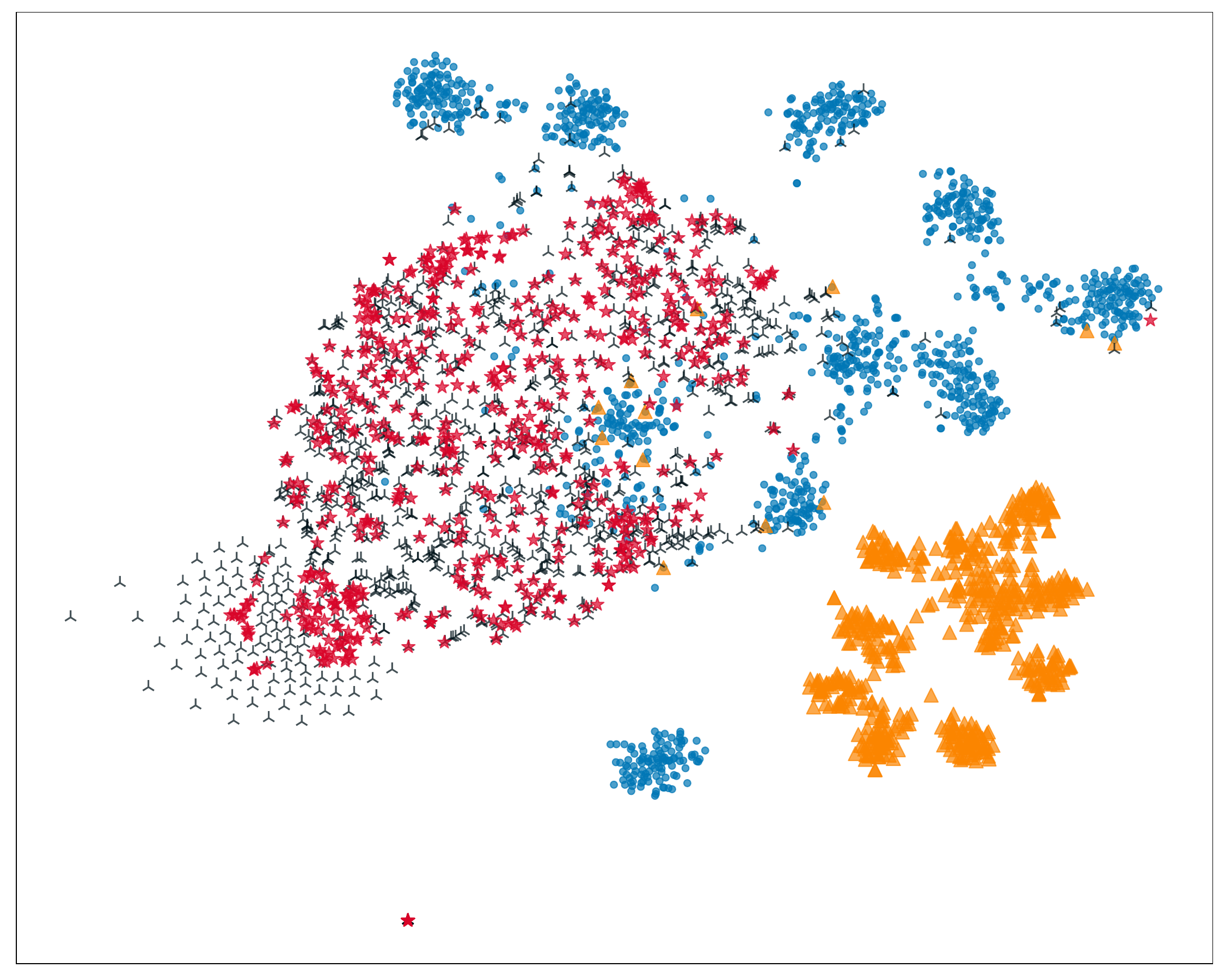}
            \includegraphics[width=1\textwidth]{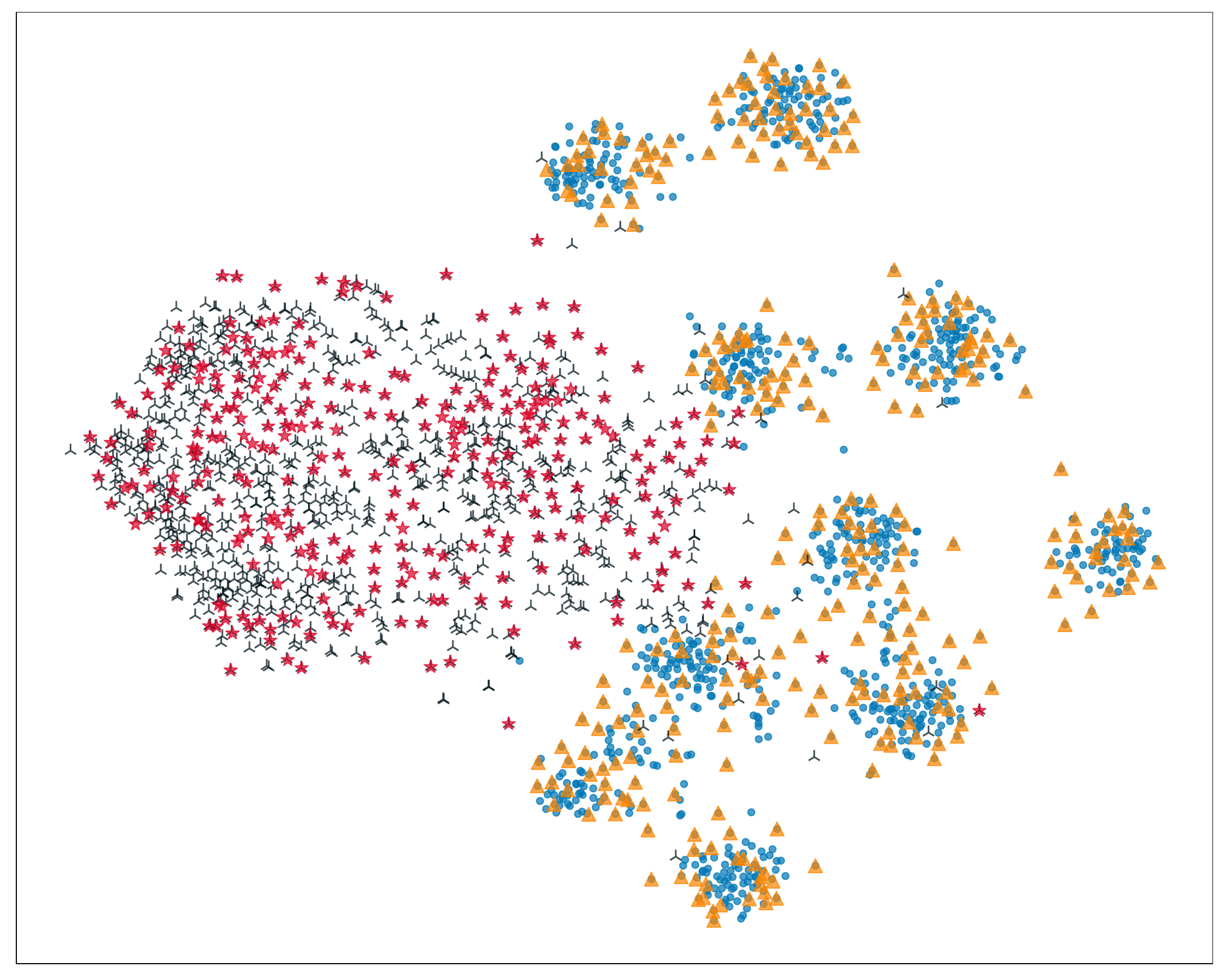}
            \centerline{(c) SRA}
        \end{minipage}
    \end{minipage}
    \begin{minipage}[b]{0.99\linewidth}
        \centering
        \begin{minipage}[b]{0.02\linewidth}
            \rotatebox{90}{~~~~~~~~~~~~~~~~~~~~~~~~~~~~~~~~~~~~SCALE-UP}
            \rotatebox{90}{~~~~~~~~~~~~~~~~~~~~~~~~ Ours}
        \end{minipage}
        \begin{minipage}[b]{0.32\linewidth}
            \includegraphics[width=1\textwidth]{scaleup_tsne_7_bpp.pdf}
            \includegraphics[width=1\textwidth]{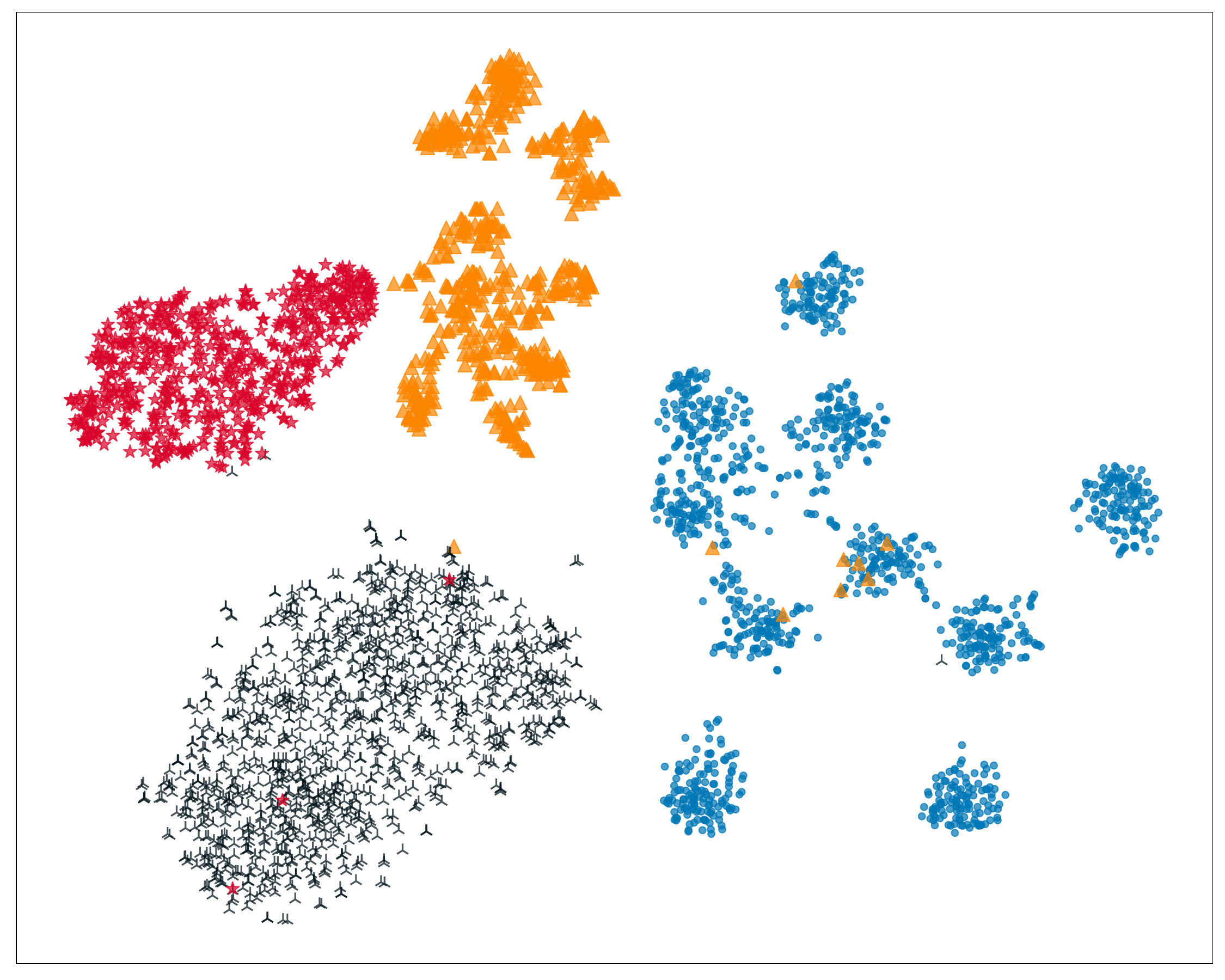}
            \centerline{(d) BPP}
        \end{minipage}
        \begin{minipage}[b]{0.32\linewidth}
            \includegraphics[width=1\textwidth]{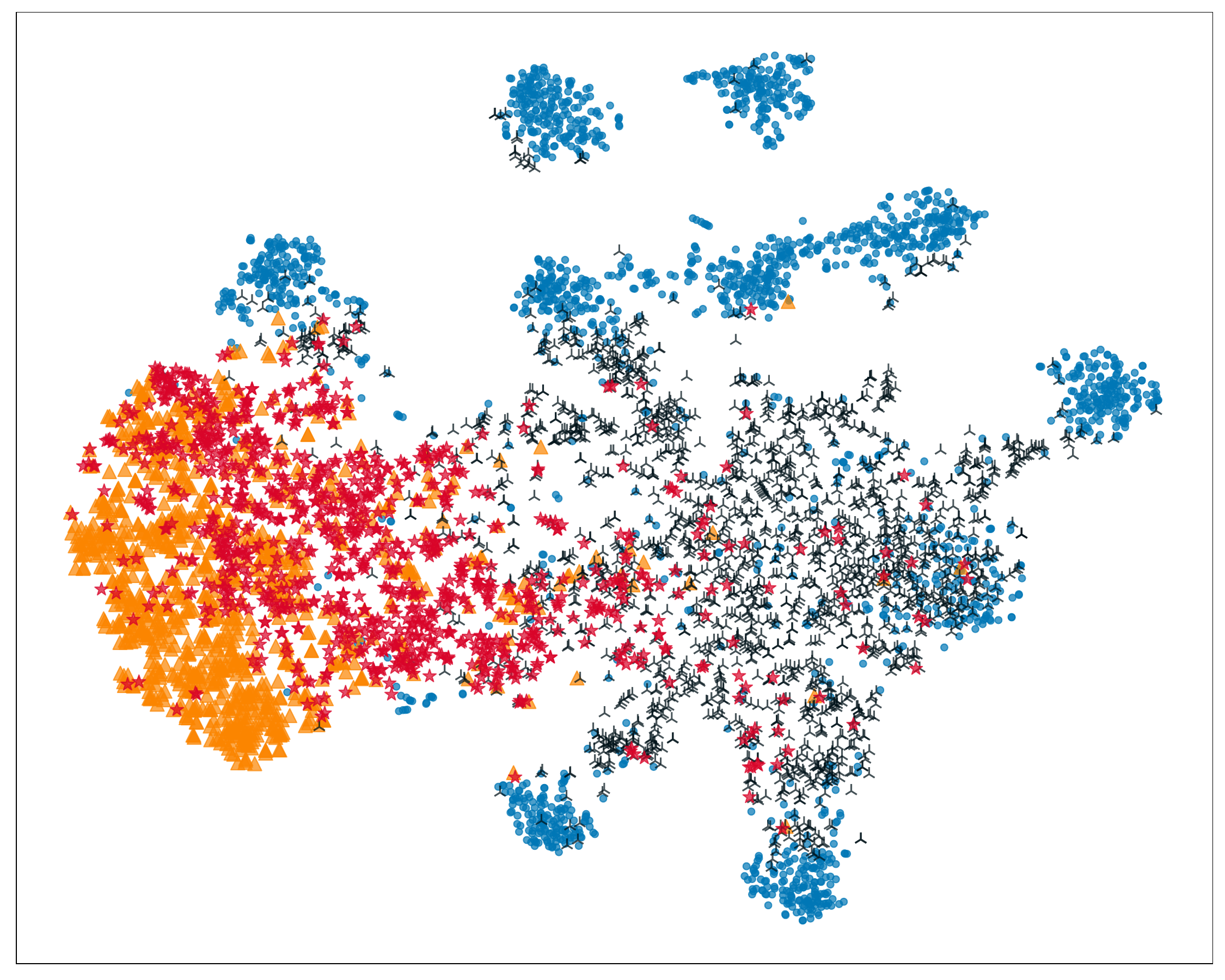}
            \includegraphics[width=1\textwidth]{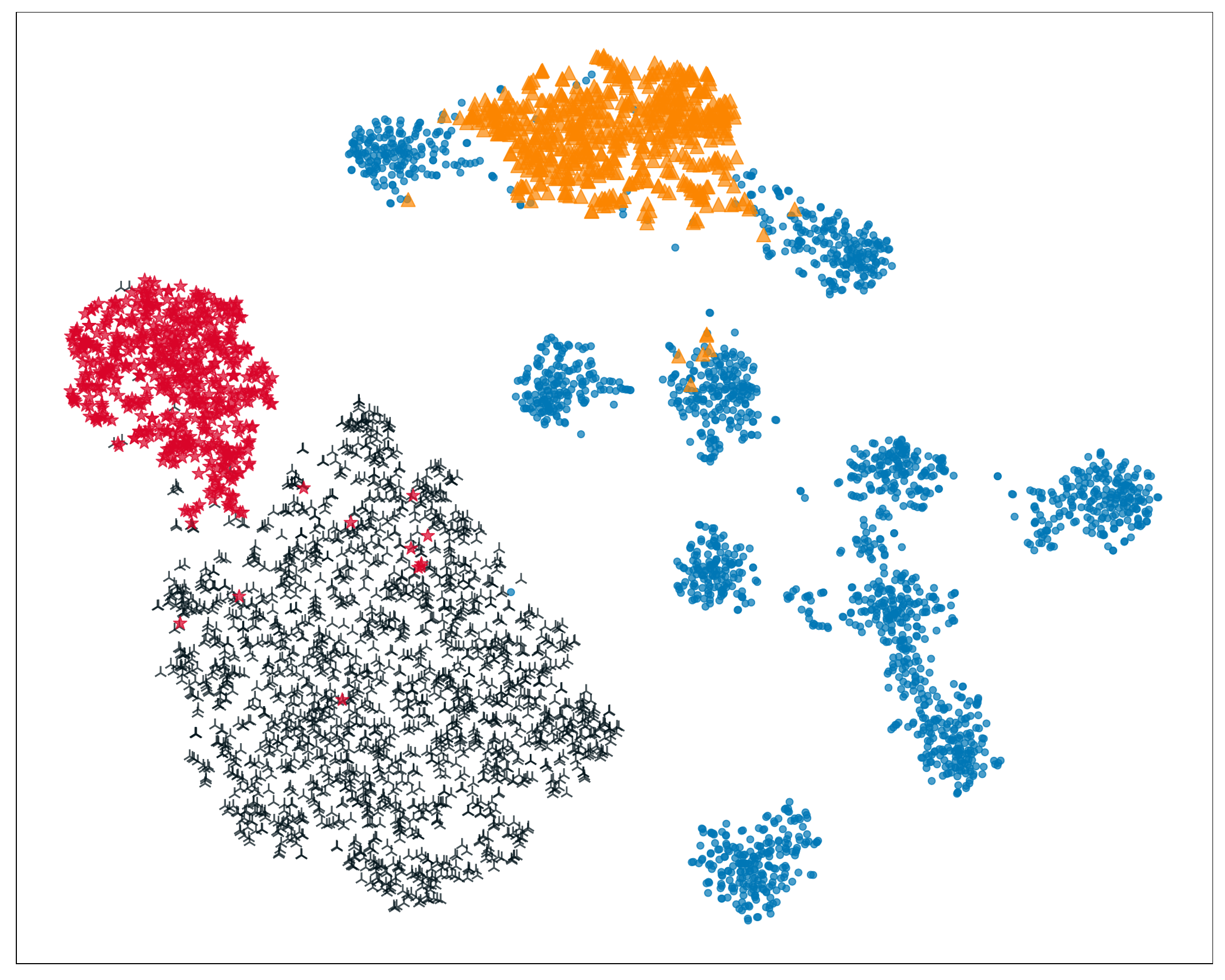}
            \centerline{(e) NARCISSUS}
        \end{minipage}
        \begin{minipage}[b]{0.32\linewidth}
            \includegraphics[width=1\textwidth]{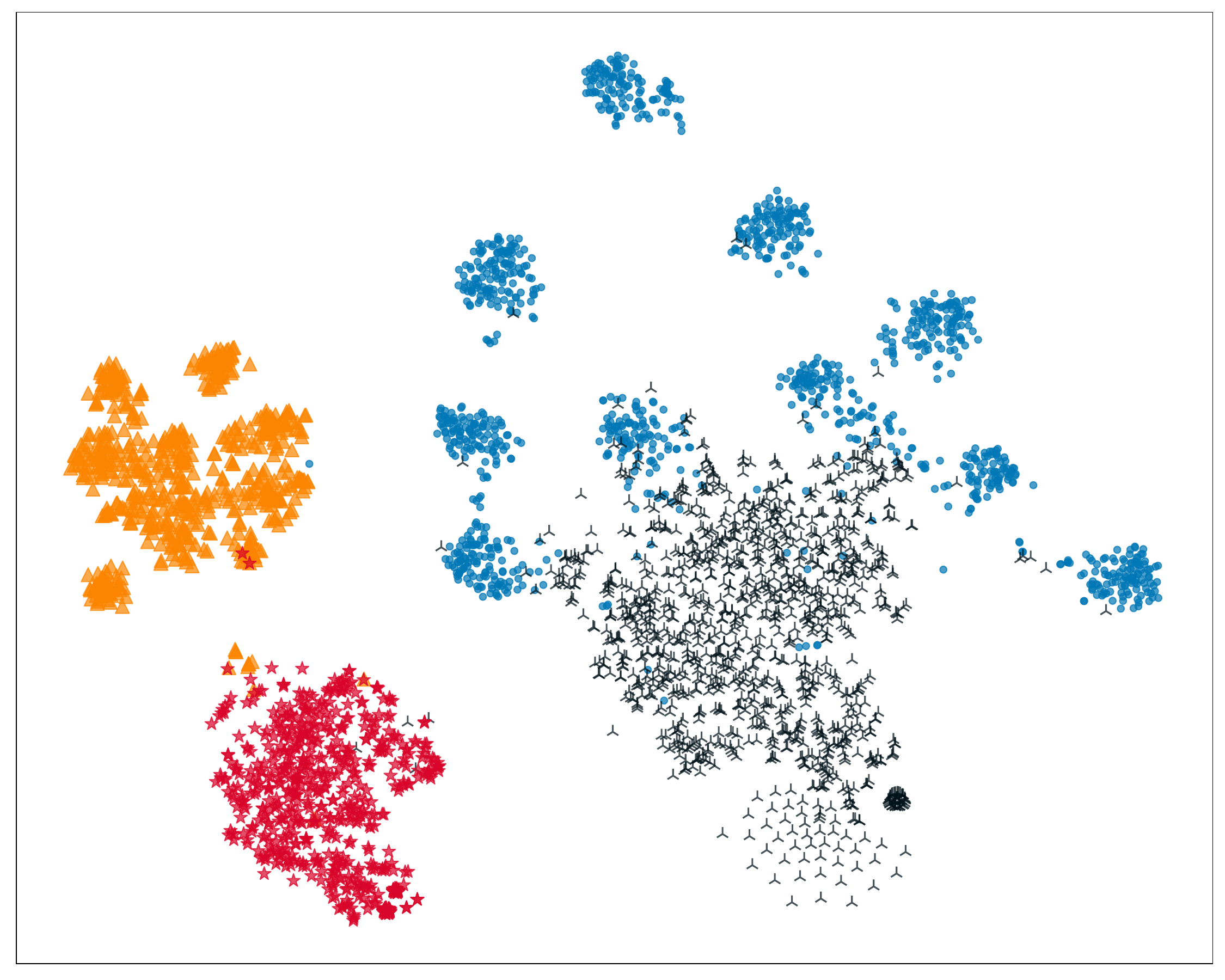}
            \includegraphics[width=1\textwidth]{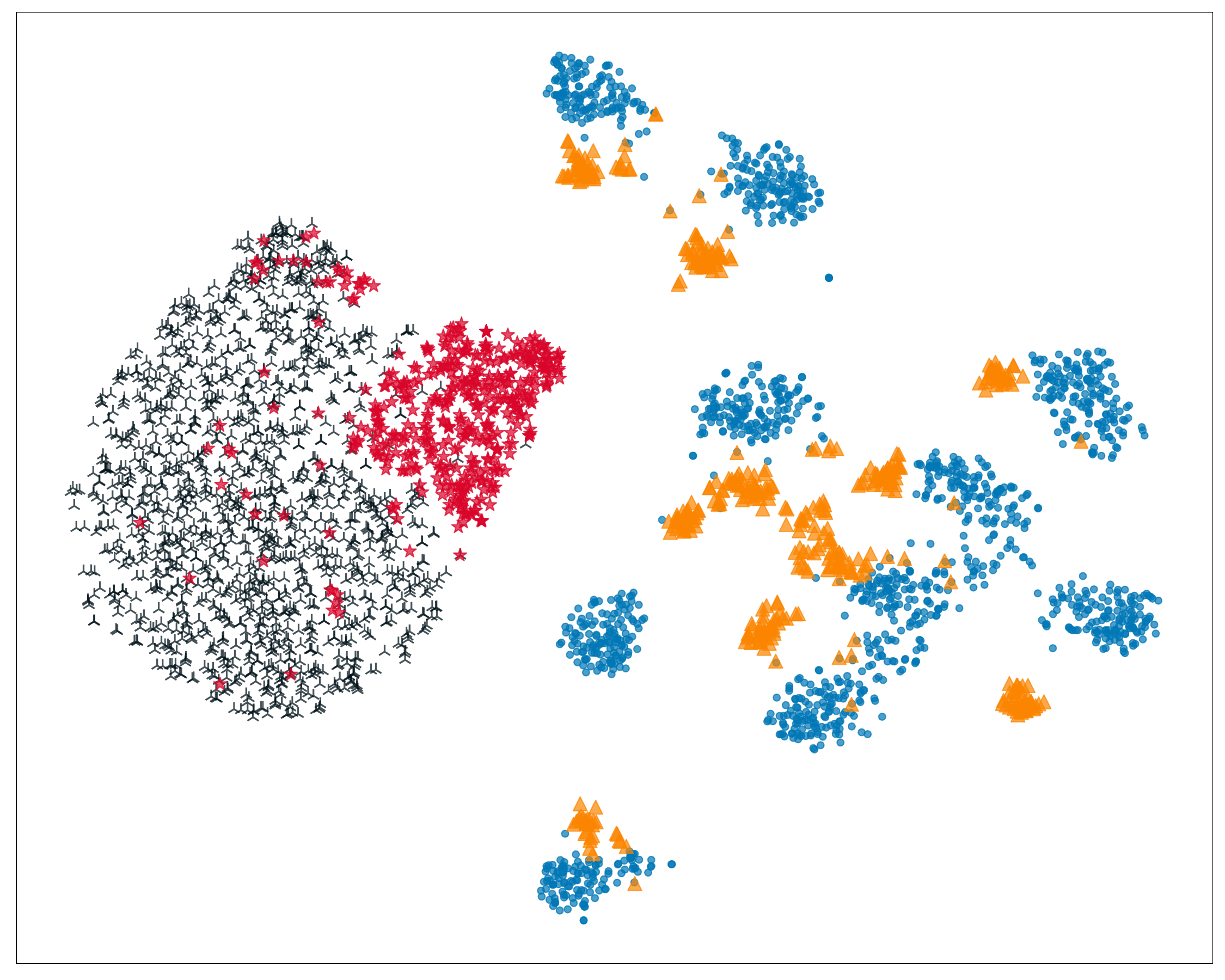}
            \centerline{(f) Adap-Patch}
        \end{minipage}
    \end{minipage}
       \caption{The t-SNE of feature representations of benign and poisoned samples on the CIFAR-10 dataset against other backdoor attacks (clean-label, specific-class, training-controlled, model-controlled, and adaptive attacks). }
    \label{fig:cluster_other2}
    \vspace{-1em}
\end{figure}

\section{How Model Amplification Changes the Latent Representation}
\label{appendix:closer}
In this section, we provide a comprehensive set of t-SNE visualizations for all the attacks considered in our study. These visualizations show how the hidden layer features of benign and poisoned samples change under the modifications by SCALE-UP and our defense strategy. As indicated in~\cref{fig:cluster_other1} and~\cref{fig:cluster_other2}, the amplification of pixel values by SCALE-UP results in a limited change within the feature space. In contrast, our defense achieves a more pronounced shift by modifying the model parameters, providing a more discernible differentiation between benign and poisoned samples.
This is intuitively why our method achieves better performance in backdoor attack detection. 

\begin{table}[!t]
\centering
\small
\caption{The performance (AUROC, TPR, FPR) of our defense on identifying the potential training poisoned samples.}
\label{tab:detection_training}
\begin{tabular}{lccccccccccccccccc} 
\toprule
Attacks $\rightarrow$ &\multicolumn{3}{c}{BadNets}	& \multicolumn{3}{c}{WaNet} &	\multicolumn{3}{c}{BATT}\\
\cmidrule(lr){2-4} \cmidrule(lr){5-7} \cmidrule(lr){8-10}
Defenses $\downarrow$	& AUROC	& TPR	&FPR	&AUROC	& TPR	&FPR &AUROC	& TPR	&FPR\\
\midrule
MSPC&	0.980	&1.000&	0.144	&0.747&	0.551&	0.186&	0.986&	0.991&	0.131\\
CD & 0.980	&0.895	&0.052&	0.710&	0.303&	0.121&	0.767&	0.403	&0.117\\
Ours&	1.000&	1.000	&0.066	&0.998&	1.000	&0.081	&0.994	&1.000&	0.079\\
\bottomrule
\vspace{-2em}
\end{tabular}
\end{table}

\begin{table}[!t]
\centering
\small
\caption{Effect of retraining models without poisoned samples identified by our defense.}
\label{tab:asr_training}
\begin{tabular}{lccccc} 
\toprule
Attacks &	ASR	&BA & \# Removed Samples\\
\midrule
BadNets&	0.005&	0.893&	7985\\
WaNet&	0.002&	0.878&	9119\\
BATT&	0.009&	0.860&	8536\\
\bottomrule
\vspace{-2em}
\end{tabular}
\end{table}

\section{The Extension to Training Set Purification}
\label{appendix:training_set}

\subsection{Related Work}
\label{appendix:related_pur}
Training set purification aims to filter out potentially poisoned samples from a contaminated training set, ensuring that the model trained on the purified dataset is free from backdoors. Many existing studies assume that backdoored models will develop abnormal latent representations for poisoned samples, which are significantly different from those of benign samples, allowing for the identification of poisoned samples. Chen~\etal~\cite{chen2018detecting} firstly observed that samples within the target class form two separate clusters in the feature space of the penultimate layer. They employed cluster analysis techniques, such as K-means, to segregate these clusters. Samples from the smaller cluster are classified as poisoned, based on the assumption that the number of poisoned samples is significantly lower than that of benign samples. Subsequent works generally utilized different cluster analysis methods, such as singular value decomposition (SVD)~\cite{tran2018spectral,hayase2021spectre}, Gram matrix~\cite{ma2022beatrix}, K-Nearest-Neighbors~\cite{peri2020deep}, and feature decomposition~\cite{tang2021demon}, to detect poisoned samples. Another line of research focuses on identifying poisoned samples based on differentiating characteristics, such as the faster speed of model fitting~\cite{li2021anti}, the presence of high-frequency artifacts~\cite{zeng2021rethinking}, and the sensitivity of poisoned samples to transformations~\cite{chen2022effective}.

More recently, Huang~\etal~\cite{huang2023distilling} hypothesized that poisoned samples require less input information to be predicted correctly. They introduced the cognitive pattern signature technique, which distills a minimal pattern (given by a mask) for an input sample to retain its original prediction. This technique reveals that poisoned samples typically exhibit a significantly smaller L1 norm in the cognitive pattern compared to benign samples. Pan~\etal~\cite{pan2023asset} proposed a proactive training set purification method called ASSET, which maximizes the loss difference between poisoned and benign samples by optimizing opposite objectives on the base and poisoned sets. Pal~\etal~\cite{pal2024backdoor} presented intriguing observations regarding the limitations of SCALE-UP~\cite{guo2023scale}, leading to the proposal of the masked scaled prediction consistency (MSPC) technique. This method selectively amplifies specific pixels in input samples, thereby more effectively exposing the prediction invariance of poisoned data under an input scaling factor.

\subsection{Comparing Our IBD-PSC with MSPC}

MSPC~\cite{pal2024backdoor} presents observations that are similar to ours regarding SCALE-UP. However, it is important to clarify that our findings on SCALE-UP constitute only a minor component of our research. While our study shares some similarities with the SCALE-UP framework, it significantly diverges by exploring parameter scaling, highlighting a substantial difference from MSPC. Additionally, our work focuses on different application scenarios compared to MSPC, which is primarily concerned with training set purification.

\noindent\textbf{Our Method Requires Fewer Assumptions about Potential Adversaries.} We explore scenarios where the users employ third-party models and need real-time detection of poisoned samples during the inference phase, similar to a firewall. This setup aligns with the framework proposed in SCALE-UP. Notably, we do not limit our adversaries to using poison-only attack methods, which is required by MSPC.

\noindent\textbf{Differences in Detection Focus.} Our defense operates during the inference stage, requiring the capability of real-time detection. In contrast, MSPC is less constrained in detection time as it operates during the data collection phase.

\subsection{Identifying and Filtering Potentially Poisoned Samples within Training set}
\label{appendix:filter_training}
Following the methodology in references~\cite{pal2024backdoor,huang2023distilling,pan2023asset}, we first train a model on a potentially compromised training set and then apply our detection method to identify and filter potentially poisoned samples within that dataset. The detection performance, presented in~\cref{tab:detection_training}, demonstrates the effectiveness of our method in filtering malicious training samples across various attacks, achieving a 100\% TPR and nearly 100\% AUROC while maintaining an FPR close to 0\%. Note that we reproduce MSPC using its open-source codes with default settings. However, it performs relatively poorly in defending against WaNet compared to the results reported in its original paper. We speculate this is because we test WaNet in noise mode, whereas MSPC is tested on the vanilla WaNet (as mentioned in their Appendix E). After removing suspected poisoned samples from the training set, we retrain the model on this purified training set to evaluate both its  BA and ASR. We conduct experiments on the CIFAR-10 dataset against three representative attacks, and the results, presented in~\cref{tab:asr_training}, show that the ASR scores of these retrained models are less than 0.5\%, rendering these backdoor attacks ineffective.

 \section{Potential Limitations and Future Directions}
\label{appendix:limitations}

In this section, we analyze the potential limitations and future directions of this work.

Firstly, our defense requires more memory and inference times than the standard model inference without any defense. Specifically, let $M_s$ and $M_d$ denote the memory (for loading models) required by the standard model inference and by that of our defense, respectively. Let $T_s$ and $T_d$ denote the inference time required by the standard model inference and by that of our defense. Assuming that we adopt $n$ (\eg, $n=5$) parameter-amplified models for our defense. We have the following equation: $M_d \cdot T_d = n \times M_s \cdot T_s.$ Accordingly, the users may need more GPUs to load all/some amplified models simultaneously to ensure efficiency or require more time for prediction by loading those models one by one when the memory is limited. In particular, the storage costs of our defense are similar to those without defense since we can easily obtain amplified models based on the standard one and, therefore, only need to save one model copy (\eg, vanilla model). We will explore how to reduce those costs in our future work.

Secondly, our IBD-PSC requires a few local benign samples, although their number could be small (\eg, 25, as shown in Figure \ref{fig:impact_size}). We will explore how to extend our method to the `data-free' scenarios in our future works.

Thirdly, our method can only detect whether a suspicious testing image is malicious. Currently, our defense cannot recover the correct label of malicious samples or their trigger patterns. As such, the users can only mark and refuse to predict those samples. We will explore how to incorporate those additional functionalities in our future works.

Fourthly, our work currently focuses only on image classification tasks. We will explore its performance on other modalities (\eg, text and audio) and tasks (\eg, detection and tracking) in our future work.


\section{Reproducibility Statement}

We have provided detailed descriptions encompassing the datasets utilized, training and evaluation settings, along the computational resources involved. To facilitate the replication of our experimental results, the corresponding codes and model checkpoints have been provided in the supplementary materials.

\section{Discussions about the Adopted Data}

In this paper, all the samples we used are from publicly available datasets, including CIFAR-10, GTSRB, and ImageNet. It's worth noting that our defense method is implemented by modifying the pre-trained model parameters, without making any alterations to the input samples themselves. Therefore, our study doesn't raise any concerns regarding the privacy of human-related images within the dataset.




\onecolumn

\end{document}